SECOND EDITION

Takaaki Fujita
Florentin Smarandache

# A Dynamic Survey of Soft Set Theory and Its Extensions

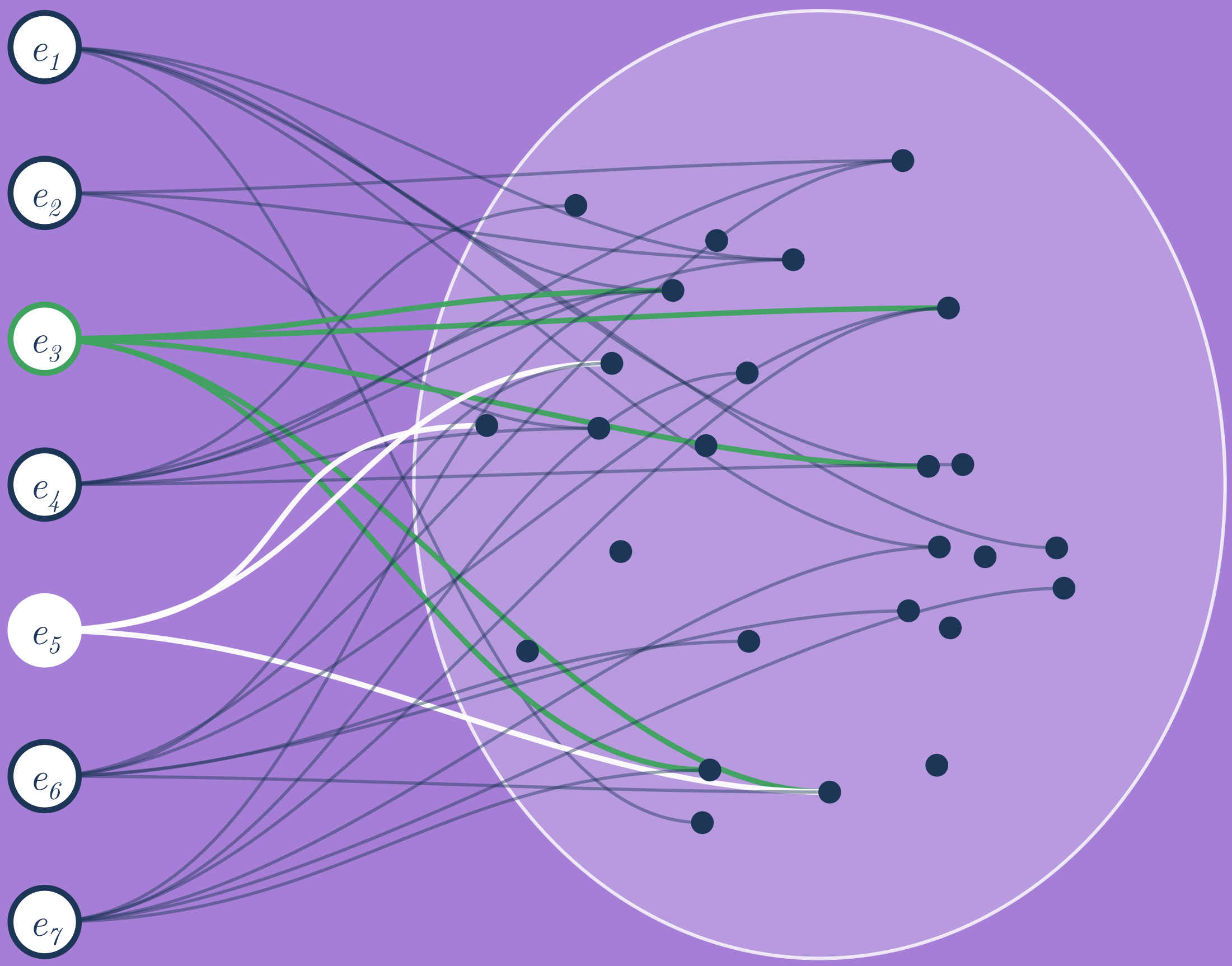

# A Dynamic Survey of Soft Set Theory and Its Extensions (2nd Edition)

**Takaaki Fujita** [1] * **and Florentin Smarandache**[2]
[1] Independent Researcher, Tokyo, Japan.
Email: Takaaki.fujita060@gmail.com
[2] University of New Mexico, Gallup Campus, NM 87301, USA.
Email: fsmarandache@gmail.com

## Abstract

Soft set theory provides a direct framework for parameterized decision modeling by assigning to each attribute (parameter) a subset of a given universe, thereby representing uncertainty in a structured way [1, 2]. Over the past decades, the theory has expanded into numerous variants—including hypersoft sets, superhypersoft sets, TreeSoft sets, bipolar soft sets, and dynamic soft sets—and has been connected to diverse areas such as topology and matroid theory. In this book, we survey soft sets and their major extensions, emphasizing parameter spaces, value spaces, representative constructions, and the assumptions required for valid mathematical comparisons. We distinguish literature-based definitions from exploratory formulations and discuss limitations of selected representation and reduction claims. This second edition adds further examples and concepts, corrects identified mathematical and typographical issues, and improves the organization.

PEER REVIEWERS

**Ranulfo Paiva Barbosa**
Web3 Blockchain Entrepreneur, San José, Costa Rica
*Email: ranulfo17@gmail.com*

**Jesus Estupinan Ricardo**
Centro de Estudios para la Calidad Educativa y la Investigación Científica, Toluca, Mexico
*Email: jestupinan2728@gmail.com*

# Contents of this book

The remainder of this book is organized as follows.

# Chapter 1

# Introduction

## 1.1 Soft Set Theory

Classical (crisp) set theory provides a precise and widely used language for formal reasoning and mathematical modeling [3]. Over the past decades, many generalized set frameworks have been introduced to represent uncertainty and vagueness, including fuzzy sets [4], intuitionistic fuzzy sets [5], hesitant fuzzy sets [6], picture fuzzy sets [7], single-valued neutrosophic sets [8, 9], quadripartitioned neutrosophic sets [10], pentapartitioned neutrosophic sets [11], double-valued neutrosophic sets [12], hesitant neutrosophic sets [13], plithogenic sets [14,15], and soft sets [2,16].

A fuzzy set assigns to each element $x$ a single membership grade $\mu(x) \in [0, 1]$, thereby capturing gradual inclusion rather than a sharp yes/no decision [4,17]. Neutrosophic sets extend this viewpoint by associating three (generally independent) degrees $T(x), I(x), F(x) \in [0, 1]$, interpreted as truth, indeterminacy, and falsity, respectively [8,18]. Because these models encode uncertainty more flexibly than crisp sets, they have been applied widely, for example in decision-making [19], robotics and system integration [20], artificial intelligence [21], and neural networks [22, 23].

A soft set offers a direct framework for parameterized decision modeling by associating each attribute (or parameter) with a subset of a universe, thereby handling uncertainty in a structured manner [1, 2]. Like fuzzy and neutrosophic frameworks, soft set theory has developed many extensions and variants, and its applications have been studied across a wide range of areas, including decision support and related fields.

## 1.2 Our Contributions

In light of these developments, research on soft set theory remains important. Moreover, because a large number of papers on soft sets and their extensions continue to appear, survey-style works play an increasingly valuable role in organizing and clarifying the landscape. Motivated by this need, this book organizes soft set theory and its extensions by parameter domain, value domain, structural compatibility, and decision-making mechanism. Besides recalling established definitions, it discusses representative examples and carefully specified extensions. Where a definition or comparison theorem relies on additional assumptions, these assumptions are stated explicitly.

Table 1.1: Structural distinctions among representative soft-set families.

| **Family** | **Parameter domain** | **Value / compatibility requirement** |
|---|---|---|
| Soft set | $A$ | $F : A \to \mathcal{P}(U)$ |
| HyperSoft set | $\prod_i A_i$ | $F : \prod_i A_i \to \mathcal{P}(U)$ |
| SuperHyperSoft set | $\prod_i \mathcal{P}(A_i)$ | Set-valued attribute choices, with outputs in $\mathcal{P}(U)$ |
| FiltrationSoft set | $A$ | Nested $(k+1)$-tuples of subsets of $U$ |
| Uncertainty-valued soft set | Model-dependent | Degree-domain admissibility and any additional constraints |

Structural embeddings preserve the properties demonstrated by their proofs; they do not, by themselves, establish numerical or computational superiority over existing methods. The second edition adds further constructions and explanatory material, corrects selected mathematical formulations, and improves notation and organization. Proposals and adaptations developed within this work are distinguished from literature-based formulations.

These representations need not be equivalent merely because they share a set-valued output. The admissibility, semantics, and compatibility of their parameter and value spaces must be compared separately.

# Chapter 2

# Types of Soft Sets

As types of soft sets, a wide variety of extended soft-set models have been proposed. In this chapter, we provide a survey-style introduction and brief discussion of these extensions.

## 2.1 Soft Set

A Soft Set offers a straightforward approach to parameterized decision modeling by associating attributes (or parameters) with subsets of a universal set, effectively addressing uncertainty in a structured manner [1, 2].

**Definition 2.1.1** (Soft Set)**.** [1,2] Let $U$ be a universal set and $A$ be a set of attributes. A soft set over $U$ is a pair $(\mathcal{F}, S)$, where $S \subseteq A$ and $\mathcal{F} : S \to \mathcal{P}(U)$. Here, $\mathcal{P}(U)$ denotes the power set of $U$. Mathematically, a soft set is represented as:

$$(\mathcal{F}, S) = \{(\alpha, \mathcal{F}(\alpha)) \mid \alpha \in S, \mathcal{F}(\alpha) \in \mathcal{P}(U)\}.$$

Each $\alpha \in S$ is called a parameter, and $\mathcal{F}(\alpha)$ is the set of elements in $U$ associated with $\alpha$.

## 2.2 HyperSoft Set

A HyperSoft set maps each multi-attribute value tuple to a subset of the universe, capturing combined parameter interactions [24–26]. HyperSoft Sets have been extensively studied in various studies in recent years [27–34].

**Definition 2.2.1** (Hypersoft Set)**.** [26] Let $U$ be a universal set, and let $\mathcal{A}_1, \mathcal{A}_2, \ldots, \mathcal{A}_m$ be attribute domains. Define $\mathcal{C} = \mathcal{A}_1 \times \mathcal{A}_2 \times \cdots \times \mathcal{A}_m$, the Cartesian product of these domains. A hypersoft set over $U$ is a pair $(G, \mathcal{C})$, where $G : \mathcal{C} \to \mathcal{P}(U)$. The hypersoft set is expressed as:

$$(G, \mathcal{C}) = \{(\gamma, G(\gamma)) \mid \gamma \in \mathcal{C}, G(\gamma) \in \mathcal{P}(U)\}.$$

For an $m$-tuple $\gamma = (\gamma_1, \gamma_2, \ldots, \gamma_m) \in \mathcal{C}$, where $\gamma_i \in \mathcal{A}_i$ for $i = 1, 2, \ldots, m$, $G(\gamma)$ represents the subset of $U$ corresponding to the combination of attribute values $\gamma_1, \gamma_2, \ldots, \gamma_m$.

**Example 2.2.2** (Example of a HyperSoft Set: laptop recommendation by exact attribute tuples)**.** Let $U$ be a finite set of laptop models:

$$U = \{\ell_1, \ell_2, \ell_3, \ell_4, \ell_5\},$$

where $\ell_1$ = Model A, $\ell_2$ = Model B, $\ell_3$ = Model C, $\ell_4$ = Model D, $\ell_5$ = Model E.

Consider $m = 3$ attribute domains:

$$\mathcal{A}_1 = \{\text{Low}, \text{Mid}, \text{High}\} \quad \text{(price tier)},$$

$$\mathcal{A}_2 = \{\text{Light}, \text{Standard}\} \quad \text{(weight class)},$$

$$\mathcal{A}_3 = \{\text{Long}, \text{Normal}\} \quad \text{(battery life)}.$$

Set the hypersoft parameter domain

$$\mathcal{C} = \mathcal{A}_1 \times \mathcal{A}_2 \times \mathcal{A}_3.$$

Define a mapping $G : \mathcal{C} \to \mathcal{P}(U)$ by assigning, to each attribute tuple $\gamma = (\gamma_1, \gamma_2, \gamma_3) \in \mathcal{C}$, the subset $G(\gamma) \subseteq U$ of laptops matching that exact combination. For instance, suppose the models have the following tags:

| model | price | weight | battery |
|---|---|---|---|
| $\ell_1$ | Low | Standard | Normal |
| $\ell_2$ | Mid | Light | Long |
| $\ell_3$ | Mid | Standard | Long |
| $\ell_4$ | High | Light | Long |
| $\ell_5$ | High | Standard | Normal |

As a concrete evaluation, take the tuple

$$\gamma^* = (\text{High}, \text{Light}, \text{Long}) \in \mathcal{C}.$$

Then

$$G(\gamma^*) = \{\ell_4\}.$$

Similarly,

$$G(\text{Mid}, \text{Standard}, \text{Long}) = \{\ell_3\}, \qquad G(\text{Mid}, \text{Light}, \text{Long}) = \{\ell_2\}.$$

Thus $(G, \mathcal{C})$ is a HyperSoft Set over $U$: each parameter is a *tuple* of attribute values, and $G(\gamma)$ returns the subset of objects in $U$ that satisfy exactly that combined tuple.

## 2.3 SuperHyperSoft Set

A SuperHyperSoft set maps tuples of subsets of attribute-value sets to universe subsets, modeling set-valued multi-attribute constraints [35–38]. SuperHyperSoft Sets have been studied in various studies [39–43].

**Definition 2.3.1** (SuperHyperSoft Set)**.** [38] Let $U$ be a universal set, and let $\mathcal{P}(U)$ denote the power set of $U$. Consider $n$ distinct attributes $a_1, a_2, \ldots, a_n$, where $n \geq 1$. Each attribute $a_i$ is associated with a set of attribute values $A_i$, satisfying the property $A_i \cap A_j = \emptyset$ for all $i \neq j$.

Define $\mathcal{P}(A_i)$ as the power set of $A_i$ for each $i = 1, 2, \ldots, n$. Then, the Cartesian product of the power sets of attribute values is given by:

$$\mathcal{C} = \mathcal{P}(A_1) \times \mathcal{P}(A_2) \times \cdots \times \mathcal{P}(A_n).$$

A SuperHyperSoft Set over $U$ is a pair $(F, \mathcal{C})$, where:

$$F : \mathcal{C} \to \mathcal{P}(U),$$

and $F$ maps each element $(\alpha_1, \alpha_2, \ldots, \alpha_n) \in \mathcal{C}$ (with $\alpha_i \in \mathcal{P}(A_i)$) to a subset $F(\alpha_1, \alpha_2, \ldots, \alpha_n) \subseteq U$. Mathematically, the SuperHyperSoft Set is represented as:

$$(F, \mathcal{C}) = \{(\gamma, F(\gamma)) \mid \gamma \in \mathcal{C}, F(\gamma) \in \mathcal{P}(U)\}.$$

Here, $\gamma = (\alpha_1, \alpha_2, \ldots, \alpha_n) \in \mathcal{C}$, where $\alpha_i \in \mathcal{P}(A_i)$ for $i = 1, 2, \ldots, n$, and $F(\gamma)$ corresponds to the subset of $U$ defined by the combined attribute values $\alpha_1, \alpha_2, \ldots, \alpha_n$.

**Example 2.3.2** (Example of a SuperHyperSoft Set: meal planning with set-valued attribute choices)**.** Let $U$ be a set of dinner recipes:

$$U = \{r_1, r_2, r_3, r_4, r_5, r_6\},$$

where $r_1$ = tofu salad, $r_2$ = chicken stir-fry, $r_3$ = lentil soup, $r_4$ = salmon bowl, $r_5$ = gluten-free pasta, $r_6$ = vegetable curry.

Consider $n = 3$ distinct attributes:

$$a_1 = \text{Diet type}, \qquad a_2 = \text{Main protein}, \qquad a_3 = \text{Cooking time}.$$

Let the corresponding attribute-value sets be

$$A_1 = \{\text{Vegan}, \text{Omnivore}, \text{Pescatarian}\},$$

$$A_2 = \{\text{Tofu}, \text{Chicken}, \text{Fish}, \text{Legumes}\},$$

$$A_3 = \{\text{Quick}, \text{Medium}, \text{Long}\},$$

so $A_i \cap A_j = \varnothing$ for $i \neq j$. Define the super-parameter domain

$$\mathcal{C} = \mathcal{P}(A_1) \times \mathcal{P}(A_2) \times \mathcal{P}(A_3).$$

For each $\gamma = (\alpha_1, \alpha_2, \alpha_3) \in \mathcal{C}$, interpret $\alpha_i \subseteq A_i$ as a *set of acceptable values* for attribute $a_i$ (rather than a single value). Define $F : \mathcal{C} \to \mathcal{P}(U)$ by selecting recipes compatible with the acceptable sets. For instance, suppose the recipe tags are:

| recipe | diet | protein | time |
|---|---|---|---|
| $r_1$ | Vegan | Tofu | Quick |
| $r_2$ | Omnivore | Chicken | Medium |
| $r_3$ | Vegan | Legumes | Long |
| $r_4$ | Pescatarian | Fish | Quick |
| $r_5$ | Omnivore | Legumes | Medium |
| $r_6$ | Vegan | Legumes | Medium |

As a concrete evaluation, take

$$\gamma^* = (\alpha_1, \alpha_2, \alpha_3) = (\{\text{Vegan}, \text{Pescatarian}\},\ \{\text{Tofu}, \text{Fish}\},\ \{\text{Quick}\}).$$

Then $F(\gamma^*)$ is the subset of recipes whose diet is in $\alpha_1$, protein is in $\alpha_2$, and cooking time is in $\alpha_3$:

$$F(\gamma^*) = \{r_1, r_4\}.$$

Thus $(F, \mathcal{C})$ is a SuperHyperSoft Set: each parameter is a *tuple of subsets* $(\alpha_1, \alpha_2, \alpha_3)$ specifying acceptable attribute values at each coordinate, and $F(\alpha_1, \alpha_2, \alpha_3)$ returns the recipes satisfying the combined set-valued constraints.

A comparison of soft sets, hypersoft sets, and superhypersoft sets is presented in Table 2.1.

Table 2.1: Concise comparison of Soft sets, HyperSoft sets, and SuperHyperSoft sets.

| **Aspect** | **Soft set** | **HyperSoft set (Hypersoft set)** | **SuperHyperSoft set** |
|---|---|---|---|
| Parameter domain | A subset $A \subseteq E$ of parameters. | A Cartesian product $\mathcal{C} = \mathcal{A}_1 \times \cdots \times \mathcal{A}_m$ of attribute-value domains. | A product of *powersets* $\mathcal{C} = \mathcal{P}(A_1) \times \cdots \times \mathcal{P}(A_n)$ (subset-valued attribute choices). |
| Evaluation map (codomain) | $F : A \to \mathcal{P}(U)$. | $G : \mathcal{C} \to \mathcal{P}(U)$. | $F : \mathcal{C} \to \mathcal{P}(U)$ with $\mathcal{C}$ as above. |
| Meaning of one parameter | A single attribute/criterion $e \in A$ selects a subset $F(e) \subseteq U$. | A full attribute-value tuple $\gamma \in \mathcal{C}$ selects $G(\gamma) \subseteq U$. | A tuple of *value-subsets* $(\alpha_1, \dots, \alpha_n) \in \mathcal{C}$ selects $F(\alpha_1, \dots, \alpha_n) \subseteq U$. |
| Typical use | Parameterized selection under independent criteria. | Multi-attribute selection under simultaneous value assignments. | Multi-attribute selection under *set-valued* (possibly multi-choice) constraints per attribute. |

## 2.4 $(m, n)$-SuperHyperSoft Set

An $(m, n)$-SuperHyperSoft set parameterizes objects by $m$ attribute groups across $n$ hierarchical subset-levels, mapping each tuple to a universe subset. Let $U$ be a nonempty universe, and let

$$A_1, A_2, \dots, A_m$$

be $m$ pairwise-disjoint attribute domains. We write $\mathcal{P}(A_i)$ for the power set of $A_i$. Introduce

$$\mathcal{C} = \prod_{i=1}^{m} \mathcal{P}(A_i) \ = \ \mathcal{P}(A_1) \times \cdots \times \mathcal{P}(A_m),$$

whose elements are tuples $\boldsymbol{\alpha} = (\alpha_1, \dots, \alpha_m)$ with $\alpha_i \subseteq A_i$.

Similarly, fix a single universal codomain $U$ and consider

$$\mathcal{D} = \prod_{j=1}^{n} \mathcal{P}(U) \;=\; \underbrace{\mathcal{P}(U) \times \cdots \times \mathcal{P}(U)}_{n \text{ factors}}.$$

An element of $\mathcal{D}$ is an $n$-tuple $\mathbf{X} = (X_1, \ldots, X_n)$ with each $X_j \subseteq U$.

**Definition 2.4.1** ($(m,n)$-SuperHyperSoft Set)**.** An $(m,n)$-*SuperHyperSoft Set* on $U$ (with attribute domains $A_1, \ldots, A_m$) is a function

$$F : \mathcal{C} \longrightarrow \mathcal{D}.$$

Equivalently, one may write

$$F(\alpha_1, \ldots, \alpha_m) = \big(F_1(\alpha_1, \ldots, \alpha_m),\ \ldots,\ F_n(\alpha_1, \ldots, \alpha_m)\big),$$

where each coordinate

$$F_j : \mathcal{C} \longrightarrow \mathcal{P}(U) \quad (j = 1, \ldots, n)$$

is itself a "classical" $m$-SuperHyperSoft Set[1].

**Remark 2.4.2.** Thus an $(m,n)$-SuperHyperSoft Set encodes $n$ different $m$-SuperHyperSoft evaluations in parallel, one per coordinate.

**Example 2.4.3** (Example of an $(m,n)$-SuperHyperSoft Set: course recommendation with two-level outputs)**.** Let $U$ be a set of university courses:

$$U = \{c_1, c_2, c_3, c_4, c_5, c_6\},$$

where $c_1$ = Linear Algebra, $c_2$ = Discrete Mathematics, $c_3$ = Machine Learning, $c_4$ = Databases, $c_5$ = Algorithms, $c_6$ = Statistics.

Take $m = 3$ pairwise-disjoint attribute domains describing a student's preferences:

$$A_1 = \{\text{Math}, \text{CS}, \text{Data}\} \quad \text{(interest area)},$$

$$A_2 = \{\text{Beginner}, \text{Intermediate}, \text{Advanced}\} \quad \text{(difficulty tolerance)},$$

$$A_3 = \{\text{Short}, \text{Normal}\} \quad \text{(weekly workload)}.$$

Define the input domain

$$\mathcal{C} = \mathcal{P}(A_1) \times \mathcal{P}(A_2) \times \mathcal{P}(A_3).$$

Fix $n = 2$ output levels and define

$$\mathcal{D} = \mathcal{P}(U) \times \mathcal{P}(U).$$

Interpret the first component as *recommended* courses and the second as *optional* courses.

Define $F : \mathcal{C} \to \mathcal{D}$ by

$$F(\alpha_1, \alpha_2, \alpha_3) = \big(F_1(\alpha_1, \alpha_2, \alpha_3),\ F_2(\alpha_1, \alpha_2, \alpha_3)\big),$$

[1] I.e. a mapping from $\mathcal{C}$ into $\mathcal{P}(U)$.

where $F_1, F_2 : \mathcal{C} \to \mathcal{P}(U)$ are given by a simple rule-based advisor.

For a concrete parameter tuple, take

$$(\alpha_1, \alpha_2, \alpha_3) = (\{\mathrm{CS}, \mathrm{Data}\},\ \{\mathrm{Intermediate}, \mathrm{Advanced}\},\ \{\mathrm{Normal}\}).$$

Suppose the advisor outputs

$$F_1(\alpha_1, \alpha_2, \alpha_3) = \{c_3, c_5, c_4\} \quad \text{and} \quad F_2(\alpha_1, \alpha_2, \alpha_3) = \{c_6, c_2\}.$$

Thus

$$F(\alpha_1, \alpha_2, \alpha_3) = \big(\{c_3, c_5, c_4\},\ \{c_6, c_2\}\big) \in \mathcal{D}.$$

Interpretation: the input $(\alpha_1, \alpha_2, \alpha_3)$ specifies *sets* of acceptable values for each attribute group (interest area, difficulty, workload), and the output is an $n$-tuple of subsets of $U$ (recommended and optional course lists). Hence $F$ is an $(m, n)$-SuperHyperSoft Set on $U$ with $(m, n) = (3, 2)$.

For reference, the comparison between a SuperHyperSoft set and an $(m, n)$-SuperHyperSoft set is summarized in Table 2.2.

Table 2.2: Concise comparison between a SuperHyperSoft set and an $(m, n)$-SuperHyperSoft set on a universe $U$.

| **Aspect** | **SuperHyperSoft set (single-output)** | **$(m, n)$-SuperHyperSoft set (multi-output)** |
|---|---|---|
| Attribute domains | Pairwise-disjoint domains $A_1, \dots, A_m$; each input component is a subset $\alpha_i \subseteq A_i$. | Same domains $A_1, \dots, A_m$ and the same subset-valued input components $\alpha_i \subseteq A_i$. |
| Input (parameter) space | $\mathcal{C} = \prod_{i=1}^{m} \mathcal{P}(A_i)$. | Same $\mathcal{C} = \prod_{i=1}^{m} \mathcal{P}(A_i)$. |
| Mapping (codomain) | $F : \mathcal{C} \to \mathcal{P}(U)$. | $F : \mathcal{C} \to \mathcal{D}$, where $\mathcal{D} = \prod_{j=1}^{n} \mathcal{P}(U)$. |
| Output semantics | For each $\boldsymbol{\alpha} \in \mathcal{C}$, a single selected subset $F(\boldsymbol{\alpha}) \subseteq U$. | For each $\boldsymbol{\alpha} \in \mathcal{C}$, an $n$-tuple $F(\boldsymbol{\alpha}) = (F_1(\boldsymbol{\alpha}), \dots, F_n(\boldsymbol{\alpha}))$ with $F_j(\boldsymbol{\alpha}) \subseteq U$ (e.g., recommended/optional/rejected tiers). |
| Equivalent viewpoint | One $m$-attribute, subset-valued selector on $U$. | An ordered family of $n$ parallel selectors $(F_1, \dots, F_n)$, each an $m$-SuperHyperSoft-type map $\mathcal{C} \to \mathcal{P}(U)$. |
| Reduction / relation | Special case of $(m, n)$ with $n = 1$ (identify $\mathcal{D} = \mathcal{P}(U)$). | Strict extension of the single-output model by allowing $n$ coordinated outputs for each input tuple. |
| Typical use | Set-valued multi-attribute constraints/selection with set-valued attribute inputs. | Multi-stage screening, multi-tier reporting, or hierarchical decision outputs under the same set-valued multi-attribute inputs. |

## 2.5 TreeSoft Set

A TreeSoft set maps subsets of a hierarchical attribute tree to universe subsets, modeling refined parameters across multiple levels [44–47]. Related concepts include *PolyTree-soft sets* [48] and *Tree-to-Tree soft sets* [49].

**Definition 2.5.1** (TreeSoft Set)**.** [50] Let $U$ be a universe of discourse and let $H$ be a nonempty subset of $U$. Write $\mathcal{P}(H)$ for the power set of $H$. Let $A = \{A_1, A_2, \dots, A_n\}$ be a set of attributes (parameters, factors, etc.), where $n \geq 1$ and each $A_i$ is regarded as a *first-level* attribute.

Each first-level attribute $A_i$ may be refined into a set of *second-level* sub-attributes

$$A_i = \{A_{i,1}, A_{i,2}, \dots\}.$$

Likewise, each second-level sub-attribute $A_{i,j}$ may be further refined into *third-level* sub-sub-attributes,

$$A_{i,j} = \{A_{i,j,1}, A_{i,j,2}, \dots\},$$

and so on. In general, one may consider sub-attributes at the $m$-th level, indexed by $A_{i_1,i_2,\dots,i_m}$, where each index $i_k$ specifies the position at level $k$.

This hierarchical attribute organization determines a rooted tree, denoted by Tree($A$), whose root is $A$ (level 0) and whose nodes consist of all attributes and sub-attributes across levels 1 through $m$. The terminal nodes (nodes without descendants) are called the *leaves* of Tree($A$).

A *TreeSoft Set* on $H$ (with attribute-tree Tree($A$)) is a mapping

$$F : \mathcal{P}\big(\mathrm{Tree}(A)\big) \longrightarrow \mathcal{P}(H),$$

where $\mathcal{P}(\mathrm{Tree}(A))$ denotes the power set of the node set of Tree($A$).

**Example 2.5.2** (Example of a TreeSoft Set: medical triage rules organized by a symptom tree)**.** Let $U$ be a universe of patients and let

$$H = \{p_1, p_2, p_3, p_4, p_5, p_6\} \subseteq U$$

be a finite set of patients currently in a clinic.

Consider a hierarchical attribute system with two first-level attributes:

$$A = \{A_1, A_2\}, \qquad A_1 = \text{“Respiratory”}, \quad A_2 = \text{“Cardiovascular”}.$$

Refine each first-level attribute into second-level sub-attributes:

$$A_1 = \{A_{1,1}, A_{1,2}\}, \qquad A_{1,1} = \text{“Cough”}, \quad A_{1,2} = \text{“Shortness of breath”},$$

$$A_2 = \{A_{2,1}, A_{2,2}\}, \qquad A_{2,1} = \text{“Chest pain”}, \quad A_{2,2} = \text{“Palpitations”}.$$

Refine one second-level attribute further into third-level sub-sub-attributes:

$$A_{1,2} = \{A_{1,2,1}, A_{1,2,2}\}, \qquad A_{1,2,1} = \text{“Mild dyspnea”}, \quad A_{1,2,2} = \text{“Severe dyspnea”}.$$

Let Tree($A$) denote the rooted attribute tree whose nodes are

$$\mathrm{Tree}(A) = \{A, A_1, A_2, A_{1,1}, A_{1,2}, A_{2,1}, A_{2,2}, A_{1,2,1}, A_{1,2,2}\}.$$

Define a TreeSoft Set

$$F : \mathcal{P}(\mathrm{Tree}(A)) \longrightarrow \mathcal{P}(H)$$

by mapping any chosen set of nodes $X \subseteq \mathrm{Tree}(A)$ to the subset $F(X) \subseteq H$ of patients who satisfy *all* clinical features represented in $X$. Concretely, suppose the clinic records yield:

$$F(\{A_{1,1}\}) = \{p_1, p_2, p_5\} \quad \text{(patients with cough)},$$

$$F(\{A_{1,2,2}\}) = \{p_2, p_6\} \quad \text{(patients with severe dyspnea)},$$

$$F(\{A_{2,1}\}) = \{p_3, p_6\} \quad \text{(patients with chest pain)}.$$

For a combined node-set, define $F$ by intersection of the corresponding patient groups; for example,

$$F(\{A_{1,1}, A_{1,2,2}\}) = F(\{A_{1,1}\}) \cap F(\{A_{1,2,2}\}) = \{p_2\},$$

and

$$F(\{A_{2,1}, A_{1,2,2}\}) = F(\{A_{2,1}\}) \cap F(\{A_{1,2,2}\}) = \{p_6\}.$$

Thus $F$ assigns to each subset of attribute-tree nodes a subset of patients in $H$ matching the selected hierarchical symptom description, and therefore $(F, \mathrm{Tree}(A))$ constitutes a TreeSoft Set on $H$.

## 2.6 ForestSoft Set

A *ForestSoft Set* is formed by taking a collection of TreeSoft Sets and "gluing" (uniting) them together so as to obtain a single function whose domain is the union of all tree-nodes' power sets and whose values in $P(H)$ combine the images given by the individual TreeSoft Sets [51–54].

**Definition 2.6.1** (ForestSoft Set)**.** [53] Let $U$ be a universe of discourse, $H \subseteq U$ be a non-empty subset, and $P(H)$ be the power set of $H$. Suppose we have a finite (or countable) collection of TreeSoft Sets

$$\{\, F_t : P(\mathrm{Tree}(A^{(t)})) \to P(H)\}_{t \in T},$$

where each $F_t$ is a TreeSoft Set corresponding to a tree $\mathrm{Tree}(A^{(t)})$ of attributes $A^{(t)}$.

We construct a *forest* by taking the (disjoint) union of all these trees:

$$\mathrm{Forest}\big(\{A^{(t)}\}_{t \in T}\big) \;=\; \bigsqcup_{t \in T} \mathrm{Tree}\big(A^{(t)}\big).$$

A *ForestSoft Set*, denoted by

$$\mathbf{F} : P\big(\mathrm{Forest}(\{A^{(t)}\})\big) \;\longrightarrow\; P(H),$$

is defined as the *union* of all TreeSoft Set mappings $F_t$. Concretely, for any element $X \in P\big(\mathrm{Forest}(\{A^{(t)}\})\big)$, we set

$$\mathbf{F}(X) \;=\; \bigcup_{\substack{t \in T \\ X \cap \mathrm{Tree}(A^{(t)}) \neq \varnothing}} F_t\big(X \cap \mathrm{Tree}(A^{(t)})\big),$$

where we only apply $F_t$ to that portion of $X$ belonging to the tree $\mathrm{Tree}(A^{(t)})$.

**Example 2.6.2** (Example of a ForestSoft Set: hospital triage across multiple specialty trees)**.** Let $U$ be a universe of patients and let

$$H = \{p_1, p_2, p_3, p_4, p_5, p_6, p_7, p_8\} \subseteq U$$

be the set of patients currently under assessment.

Assume two medical specialties provide *separate* hierarchical attribute trees (a forest):

$$T = \{t_{\text{Resp}}, t_{\text{Card}}\}.$$

**(1) Respiratory TreeSoft Set.** Let $\mathrm{Tree}(A^{(t_{\text{Resp}})})$ be the respiratory attribute tree with nodes

$$\mathrm{Tree}(A^{(t_{\text{Resp}})}) = \{R,\ R_{\text{Cough}},\ R_{\text{Dyspnea}},\ R_{\text{SevDyspnea}}\},$$

interpreted as $R$ = Respiratory (root), $R_{\text{Cough}}$ = Cough, $R_{\text{Dyspnea}}$ = Dyspnea, $R_{\text{SevDyspnea}}$ = Severe dyspnea. Define a TreeSoft Set

$$F_{t_{\text{Resp}}} : \mathcal{P}(\mathrm{Tree}(A^{(t_{\text{Resp}})})) \to \mathcal{P}(H)$$

by patient groups:

$$F_{t_{\text{Resp}}}(\{R_{\text{Cough}}\}) = \{p_1, p_2, p_5\}, \qquad F_{t_{\text{Resp}}}(\{R_{\text{SevDyspnea}}\}) = \{p_2, p_6, p_8\},$$

and (as a typical rule) for combined node-sets use intersections, e.g.,

$$F_{t_{\text{Resp}}}(\{R_{\text{Cough}}, R_{\text{SevDyspnea}}\}) = \{p_1, p_2, p_5\} \cap \{p_2, p_6, p_8\} = \{p_2\}.$$

**(2) Cardiovascular TreeSoft Set.** Let $\mathrm{Tree}(A^{(t_{\text{Card}})})$ be the cardiovascular attribute tree with nodes

$$\mathrm{Tree}(A^{(t_{\text{Card}})}) = \{C,\ C_{\text{ChestPain}},\ C_{\text{Arrhythmia}}\},$$

interpreted as $C$ = Cardiovascular (root), $C_{\text{ChestPain}}$ = Chest pain, $C_{\text{Arrhythmia}}$ = Arrhythmia/palpitations. Define a TreeSoft Set

$$F_{t_{\text{Card}}} : \mathcal{P}(\mathrm{Tree}(A^{(t_{\text{Card}})})) \to \mathcal{P}(H)$$

by

$$F_{t_{\text{Card}}}(\{C_{\text{ChestPain}}\}) = \{p_3, p_6\}, \qquad F_{t_{\text{Card}}}(\{C_{\text{Arrhythmia}}\}) = \{p_4, p_7\},$$

and for a combined node-set,

$$F_{t_{\text{Card}}}(\{C_{\text{ChestPain}}, C_{\text{Arrhythmia}}\}) = \{p_3, p_6\} \cap \{p_4, p_7\} = \varnothing.$$

**(3) Forest and ForestSoft Set aggregation.** Form the forest by disjoint union:

$$\text{Forest} = \mathrm{Tree}(A^{(t_{\text{Resp}})}) \ \sqcup\ \mathrm{Tree}(A^{(t_{\text{Card}})}).$$

Define the ForestSoft Set $\mathbf{F} : \mathcal{P}(\text{Forest}) \to \mathcal{P}(H)$ by

$$\mathbf{F}(X) = \bigcup_{\substack{t \in T \\ X \cap \mathrm{Tree}(A^{(t)}) \neq \varnothing}} F_t\big(X \cap \mathrm{Tree}(A^{(t)})\big).$$

For example, take the mixed selection

$$X = \{R_{\text{SevDyspnea}},\ C_{\text{ChestPain}}\} \subseteq \text{Forest}.$$

Then

$$\mathbf{F}(X) = F_{t_{\text{Resp}}}(\{R_{\text{SevDyspnea}}\}) \cup F_{t_{\text{Card}}}(\{C_{\text{ChestPain}}\}) = \{p_2, p_6, p_8\} \cup \{p_3, p_6\} = \{p_2, p_3, p_6, p_8\}.$$

Interpretation: the ForestSoft Set aggregates the (possibly different) specialty-specific TreeSoft Set outputs, enabling a unified view across multiple hierarchical symptom trees.

## 2.7 IndetermSoft Set

An IndetermSoft Set extends the classical soft-set framework by allowing uncertainty, ambiguity, or nonuniqueness in parameter-dependent approximations [55–59].

Various formulations of IndetermSoft Sets have been investigated in the literature [24,50,60,61]. Here, we consider a mathematically precise set-valued formulation in which each parameter may determine multiple admissible subsets of the universe.

**Definition 2.7.1** (IndetermSoft Set)**.** Let $U$ be a nonempty universe, let $\varnothing \neq H \subseteq U$, and let $A$ be a nonempty set of parameters.

Define
$$\mathcal{P}^*(\mathcal{P}(H)) = \mathcal{P}(\mathcal{P}(H)) \setminus \{\varnothing\}.$$

An *IndetermSoft Set* over $H$ is a pair
$$(F, A),$$
where
$$F : A \longrightarrow \mathcal{P}^*(\mathcal{P}(H)).$$

For each $a \in A$, the value
$$F(a) \subseteq \mathcal{P}(H)$$
is a nonempty family of admissible approximations.

The IndetermSoft Set is called *genuinely indeterminate* if
$$\exists a \in A \quad |F(a)| > 1.$$

If
$$|F(a)| = 1 \qquad \text{for every } a \in A,$$
then the structure reduces to an ordinary Soft Set by identifying each singleton family with its unique member.

**Remark 2.7.2.** The mapping $F$ is single-valued as a function into $\mathcal{P}^*(\mathcal{P}(H))$, but its images may contain multiple admissible subsets of $H$.

Thus, the formulation distinguishes a uniquely specified family of possible approximations from a uniquely determined ordinary soft approximation.

More general IndetermSoft models may also allow indeterminacy in the parameter domain or the universe itself. Such extensions require additional specifications.

**Example 2.7.3** (IndetermSoft Set: Recruiting with Incomplete Evidence)**.** Let

$$U = \{u_1, u_2, u_3, u_4, u_5\}, \qquad H = U,$$

represent five applicants for a data-engineering position.

Consider the parameter set

$$A = \{\text{Pass}, \text{Borderline}, \text{Fail}\}.$$

Suppose that $u_1$ and $u_3$ have confirmed passing results, while $u_4$ has a confirmed failing result.

However, the screening outcomes of $u_2$ and $u_5$ are uncertain because of missing records and conflicting reports.

Define

$$F : A \longrightarrow \mathcal{P}^*(\mathcal{P}(H))$$

by

$$F(\text{Pass}) = \big\{\{u_1, u_3\}\big\},$$
$$F(\text{Fail}) = \big\{\{u_4\}\big\},$$

and

$$F(\text{Borderline}) = \big\{\{u_2\}, \{u_5\}\big\}.$$

Thus, the parameter Borderline has two admissible approximations.

In particular,

$$|F(\text{Borderline})| = 2 > 1.$$

Moreover, for every $a \in A$,

$$F(a) \in \mathcal{P}^*(\mathcal{P}(H)).$$

Therefore, $F$ is well-defined, and

$$(F, A)$$

is a genuinely indeterminate IndetermSoft Set over $H$.

The example illustrates how a parameter may admit several possible approximations when the available information is incomplete.

Related concepts include the following notions.

- **IndetermHyperSoft Set** [56, 62–65]: A hypersoft-set extension allowing indeterminacy in parameter tuples, attribute values, or corresponding approximations.
- **IndetermSuperHyperSoft Set** [66]: A superhypersoft-set extension incorporating indeterminacy in higher-order parameter structures and their associated approximations.
- **Bipolar IndetermSoft Set** [67]: An IndetermSoft Set incorporating positive and negative evaluations, potentially allowing indeterminacy in both components.
- **Weighted IndetermSoft Set** [57]: An IndetermSoft Set equipped with parameter weights to represent relative importance in decision-making under incomplete information.

## 2.8 ContraSoft Set

A ContraSoft Set extends the classical soft-set framework by incorporating contradiction degrees between attribute values and using threshold-based selection to obtain parameter-dependent approximations [68]. This framework enables contradiction-aware information filtering and decision-making.

**Definition 2.8.1** (Contradiction on Attribute Values)**.** [68] Let $V$ be a nonempty finite set of attribute values.

A *contradiction function* on $V$ is a mapping

$$c : V \times V \longrightarrow [0, 1]$$

satisfying

$$c(v, v) = 0, \qquad c(v, w) = c(w, v)$$

for all $v, w \in V$.

The value $c(v, w)$ represents the degree of contradiction between $v$ and $w$. Larger values indicate stronger contradiction.

**Definition 2.8.2** (ContraSoft Structure)**.** Let $U$ be a nonempty universe and $E$ a nonempty parameter set.

For each $e \in E$, let

$$V_e$$

be a nonempty finite set of attribute values, let

$$c_e : V_e \times V_e \longrightarrow [0, 1]$$

be a contradiction function, and let

$$v_e^\star \in V_e$$

be a designated reference value.

Define the disjoint union

$$\mathcal{V} = \bigsqcup_{e \in E} (\{e\} \times V_e).$$

A *ContraSoft Structure* consists of these data together with a mapping

$$G : \mathcal{V} \longrightarrow \mathcal{P}(U).$$

For every $e \in E$ and $v \in V_e$, the set

$$G(e, v) \subseteq U$$

represents the objects associated with attribute value $v$ under parameter $e$.

**Definition 2.8.3** (ContraSoft Set)**.** Let

$$\mathsf{CS} = (U, E, \{(V_e, c_e, v_e^\star)\}_{e\in E}, G)$$

be a ContraSoft Structure.

For a fixed threshold

$$\tau \in [0,1],$$

define

$$F^{(\tau)} : E \longrightarrow \mathcal{P}(U)$$

by

$$F^{(\tau)}(e) = \bigcup_{\substack{v\in V_e \\ c_e(v,v_e^\star)\le\tau}} G(e,v).$$

The pair

$$(F^{(\tau)}, E)$$

is called the *threshold-induced ContraSoft Set* associated with $\mathsf{CS}$.

Thus, only attribute values whose contradiction degrees relative to the reference value do not exceed $\tau$ contribute to the resulting soft approximation.

**Theorem 2.8.4** (Well-Definedness of ContraSoft Sets)**.** *Let* $\mathsf{CS}$ *be a ContraSoft Structure and let* $\tau \in [0,1]$*.*

*Then*

$$F^{(\tau)} : E \longrightarrow \mathcal{P}(U)$$

*is well-defined.*

*Moreover, if*

$$0 \le \tau_1 \le \tau_2 \le 1,$$

*then*

$$F^{(\tau_1)}(e) \subseteq F^{(\tau_2)}(e)$$

*for every* $e \in E$*.*

*Proof.* For each $e \in E$, the family of selected attribute values is nonempty, because

$$c_e(v_e^\star, v_e^\star) = 0 \le \tau.$$

Since

$$G(e,v) \subseteq U$$

for every $v \in V_e$, their union is a uniquely determined subset of $U$.

Hence,

$$F^{(\tau)}(e) \in \mathcal{P}(U),$$

so $F^{(\tau)}$ is well-defined.

Furthermore, if $\tau_1 \leq \tau_2$, then

$$\{v : c_e(v, v_e^\star) \leq \tau_1\} \subseteq \{v : c_e(v, v_e^\star) \leq \tau_2\}.$$

Taking unions of the corresponding sets $G(e, v)$ proves the inclusion. □

**Remark 2.8.5** (Reduction to Classical Soft Sets)**.** Suppose that

$$V_e = \{v_e^\star\} \qquad \text{for every } e \in E.$$

Then, for every $\tau \in [0, 1]$,

$$F^{(\tau)}(e) = G(e, v_e^\star).$$

Consequently, the ContraSoft Set reduces to the classical soft set

$$(F^\circ, E),$$

where

$$F^\circ(e) = G(e, v_e^\star).$$

More generally, if every contradiction function is identically zero, then

$$F^{(\tau)}(e) = \bigcup_{v \in V_e} G(e, v).$$

Thus, the resulting mapping is an ordinary soft-set mapping without contradiction-based exclusion.

Unlike neutrosophic indeterminacy, contradiction measures incompatibility between attribute values rather than an independent degree of uncertainty.

In addition to contradiction between attribute values, one may consider contradiction between parameters.

**Definition 2.8.6** (Parameter-Level ContraSoft Extension)**.** Let $\mathsf{CS}$ be a ContraSoft Structure.

A *parameter-level contradiction function* is a mapping

$$c_P : E \times E \longrightarrow [0, 1]$$

satisfying

$$c_P(e, e) = 0, \qquad c_P(e, f) = c_P(f, e)$$

for all $e, f \in E$.

For $\eta \in [0, 1]$, a subset $B \subseteq E$ is called *$\eta$-compatible* if

$$c_P(e, f) \leq \eta \qquad \text{for all } e, f \in B.$$

The family of compatible parameter subsets is

$$\mathcal{C}_\eta = \{B \subseteq E : c_P(e, f) \leq \eta \text{ for all } e, f \in B\}.$$

A parameter-level ContraSoft extension is the pair

$$(\mathsf{CS}, c_P).$$

**Example 2.8.7** (ContraSoft Set: Hiring with Contradictory Criteria)**.** Let

$$U = \{u_1, u_2, u_3, u_4, u_5, u_6\}$$

be a finite set of job applicants.

Consider the parameter set

$$E = \{e_{\text{Exp}}, e_{\text{Salary}}, e_{\text{Remote}}, e_{\text{Onsite}}\},$$

representing experience, salary expectations, remote-work preference, and on-site work preference, respectively.

For every $e \in E$, let

$$V_e = \{\text{Yes}, \text{No}\}, \qquad v_e^\star = \text{Yes}.$$

Define

$$c_e(\text{Yes}, \text{Yes}) = c_e(\text{No}, \text{No}) = 0,$$

and

$$c_e(\text{Yes}, \text{No}) = c_e(\text{No}, \text{Yes}) = 1.$$

Let

$$G(e, \text{Yes}) = S_e, \qquad G(e, \text{No}) = U \setminus S_e,$$

where

$$S_{e_{\text{Exp}}} = \{u_1, u_3, u_5\},$$
$$S_{e_{\text{Salary}}} = \{u_2, u_4, u_6\},$$
$$S_{e_{\text{Remote}}} = \{u_1, u_2, u_6\},$$

and

$$S_{e_{\text{Onsite}}} = \{u_3, u_4, u_5\}.$$

Choose the threshold

$$\tau = 0.2.$$

Since

$$c_e(\text{Yes}, \text{Yes}) = 0$$

and

$$c_e(\text{No}, \text{Yes}) = 1,$$

we obtain

$$F^{(0.2)}(e) = G(e, \text{Yes}) = S_e.$$

Consequently,

$$F^{(0.2)}(e_{\text{Exp}}) = \{u_1, u_3, u_5\},$$
$$F^{(0.2)}(e_{\text{Salary}}) = \{u_2, u_4, u_6\},$$
$$F^{(0.2)}(e_{\text{Remote}}) = \{u_1, u_2, u_6\},$$

and

$$F^{(0.2)}(e_{\text{Onsite}}) = \{u_3, u_4, u_5\}.$$

Now introduce parameter-level contradiction through

$$c_P : E \times E \to [0, 1].$$

Set

$$c_P(e_{\text{Remote}}, e_{\text{Onsite}}) = c_P(e_{\text{Onsite}}, e_{\text{Remote}}) = 0.95,$$

and

$$c_P(e_{\text{Exp}}, e_{\text{Salary}}) = c_P(e_{\text{Salary}}, e_{\text{Exp}}) = 0.60.$$

For all $e \in E$, let

$$c_P(e, e) = 0,$$

and assign contradiction degree 0.20 to every other pair of distinct parameters.

Choose the compatibility threshold

$$\eta = 0.70.$$

Then
$$\{e_{\text{Exp}}, e_{\text{Salary}}\} \in \mathcal{C}_{0.70},$$
because
$$c_P(e_{\text{Exp}}, e_{\text{Salary}}) = 0.60 \leq 0.70.$$

However,
$$\{e_{\text{Remote}}, e_{\text{Onsite}}\} \notin \mathcal{C}_{0.70},$$
because
$$c_P(e_{\text{Remote}}, e_{\text{Onsite}}) = 0.95 > 0.70.$$

Thus, the model distinguishes attribute-value selection from parameter-level contradiction.

The former determines the accepted applicants under each criterion, whereas the latter identifies combinations of criteria that are considered mutually compatible.

A comparison between classical Soft Sets and ContraSoft Sets is presented in Table 2.3.

Table 2.3: Comparison of Soft Sets and ContraSoft Sets.

| **Aspect** | **Soft Set** | **ContraSoft Set** |
|---|---|---|
| Basic structure | A parameterized family of subsets. | A soft-set structure equipped with attribute-value contradiction degrees. |
| Mapping | $F : E \to \mathcal{P}(U)$. | $G : \mathcal{V} \to \mathcal{P}(U)$ with induced mapping $F^{(\tau)} : E \to \mathcal{P}(U)$. |
| Additional data | No contradiction function required. | Value sets $V_e$, contradiction maps $c_e$, and reference values $v_e^\star$. |
| Selection rule | Direct parameter-based selection. | Threshold-based selection using $c_e(v, v_e^\star) \leq \tau$. |
| Parameter conflict | Not explicitly modeled. | Can be incorporated through an additional map $c_P : E \times E \to [0, 1]$. |
| Main application | Parameterized selection and uncertainty modeling. | Contradiction-aware filtering and decision-making. |
| Reduction | Classical model. | Reduces to a classical Soft Set when all $V_e$ are singletons. |

## 2.9 HesiSoft Set

A *HesiSoft Set* is a soft set $F : E \to \mathcal{P}(U)$ together with a symmetric hesitancy map $h : E \times E \to \mathcal{P}_{\text{fin}}([0, 1])$. Related concepts include *hesitant fuzzy sets* [6,69] and *hesitant neutrosophic sets* [13, 70].

**Definition 2.9.1** (HesiSoft Set)**.** Let $U$ be a finite universe of objects and $E$ a finite set of parameters. A *HesiSoft Set* is a quadruple
$$\mathsf{HSS} := (U,\ E,\ F,\ h),$$
where

- $F : E \to \mathcal{P}(U)$ is the (crisp) soft mapping; $F(e) \subseteq U$ is the set of objects *accepted* (or classified as positive) under parameter $e$;

- $h : E \times E \to \mathcal{P}_{\text{fin}}([0,1])$ is a *hesitancy map* on parameters, symmetric and normalized on the diagonal:
$$h(e,e) = \{0\}, \qquad h(e,f) = h(f,e) \quad (\forall\, e,f \in E),$$
where $\mathcal{P}_{\text{fin}}([0,1])$ denotes the family of all finite subsets of $[0,1]$.

For $x \in U$ and $e \in E$, the atomic statement "$x$ is accepted by $e$" is represented by
$$A(x,e) : \quad x \in F(e),$$
with truth value **T** if $x \in F(e)$ and **F** otherwise.

**Example 2.9.2** (Hiring shortlisting with parameter-wise hesitancy)**.** Let $U$ be the set of applicants
$$U = \{u_1, u_2, u_3, u_4, u_5\},$$
and let $E$ be the set of evaluation parameters
$$E = \{\mathsf{Tech}, \mathsf{Comm}, \mathsf{Lead}, \mathsf{Culture}\},$$
standing for technical skills, communication, leadership, and culture fit, respectively.

**(1) Crisp acceptance map.** Define $F : E \to \mathcal{P}(U)$ by
$$\begin{aligned} F(\mathsf{Tech}) &= \{u_1, u_2, u_4\}, \\ F(\mathsf{Comm}) &= \{u_2, u_3, u_5\}, \\ F(\mathsf{Lead}) &= \{u_1, u_3\}, \\ F(\mathsf{Culture}) &= \{u_2, u_4, u_5\}. \end{aligned}$$
Thus, for example, $A(u_4, \mathsf{Tech})$ is true (since $u_4 \in F(\mathsf{Tech})$), while $A(u_4, \mathsf{Comm})$ is false (since $u_4 \notin F(\mathsf{Comm})$).

**(2) Hesitancy map on parameters.** Define $h : E \times E \to \mathcal{P}_{\text{fin}}([0,1])$ by setting $h(e,e) = \{0\}$ for all $e \in E$, and for distinct parameters specify (symmetrically):
$$\begin{aligned} h(\mathsf{Tech}, \mathsf{Comm}) &= \{0.2, 0.4\}, & h(\mathsf{Tech}, \mathsf{Lead}) &= \{0.1, 0.3\}, \\ h(\mathsf{Tech}, \mathsf{Culture}) &= \{0.4, 0.6\}, & h(\mathsf{Comm}, \mathsf{Lead}) &= \{0.3, 0.5\}, \\ h(\mathsf{Comm}, \mathsf{Culture}) &= \{0.1, 0.2\}, & h(\mathsf{Lead}, \mathsf{Culture}) &= \{0.5, 0.7\}, \end{aligned}$$
and $h(e,f) = h(f,e)$ for all pairs. Here each finite set $h(e,f)$ records a *committee hesitancy profile* about how strongly the two criteria $e$ and $f$ should co-influence a final hiring decision (e.g., disagreements or multiple plausible weights coming from different interviewers).

Then
$$\mathsf{HSS} = (U, E, F, h)$$
is a HesiSoft Set modeling a real hiring shortlist: $F$ captures crisp accept/reject outcomes per criterion, while $h$ captures finite-valued hesitancy between criteria induced by mixed expert opinions.

## 2.10 MultiPolar Soft Set

A multipolar soft set assigns to each parameter multiple "polar" subsets, thereby capturing several perspectives or evaluations over the same universe. Related notions include multipolar fuzzy sets [71, 72] and multipolar neutrosophic sets [73–75].

**Definition 2.10.1** (Multipolar Soft Set)**.** Let $U$ be a nonempty universe and let $E$ be a nonempty set of parameters. Fix an integer $m \geq 2$ (the number of poles). An *$m$-polar (multipolar) soft set* over $U$ with respect to $E$ is an ordered pair $(F, E)$, where

$$F : E \longrightarrow \big(\mathcal{P}(U)\big)^m$$

is a mapping. For each parameter $e \in E$, we write

$$F(e) = \big(F_1(e), F_2(e), \ldots, F_m(e)\big),$$

where each component satisfies $F_i(e) \subseteq U$ for $i = 1, 2, \ldots, m$. The subset $F_i(e)$ is called the *$i$-th polar approximation* of $U$ under the parameter $e$, and it represents the evaluation of $e$ from the $i$-th perspective (pole). Equivalently, an $m$-polar soft set can be viewed as an ordered family of ordinary soft sets $(F_1, E), \ldots, (F_m, E)$ on the same universe and the same parameter set.

**Remark 2.10.2.** If $m = 1$, then $F : E \to \mathcal{P}(U)$ and $(F, E)$ reduces to an ordinary (crisp) soft set. For $m = 2$, the model yields a two-pole soft representation (often studied as a bipolar-type framework, possibly with additional constraints depending on the chosen bipolar definition).

**Example 2.10.3** (Real-life example of a multipolar soft set: multi-stakeholder project risk screening)**.** Let $U$ be a set of candidate IT projects in a company:

$$U = \{p_1, p_2, p_3, p_4, p_5\}.$$

Let $E$ be a set of risk-related parameters and take

$$E = \{e_{\text{Sec}}, e_{\text{Cost}}, e_{\text{Sched}}\},$$

where $e_{\text{Sec}}$ = "security risk", $e_{\text{Cost}}$ = "budget risk", $e_{\text{Sched}}$ = "schedule risk".

Suppose three different stakeholder groups evaluate each risk parameter:

$$m = 3, \qquad \text{(1) Security team, (2) Finance team, (3) PMO.}$$

Define a mapping $F : E \to \big(\mathcal{P}(U)\big)^3$ by letting, for each parameter $e \in E$,

$$F(e) = \big(F_1(e), F_2(e), F_3(e)\big),$$

where $F_i(e) \subseteq U$ is the set of projects judged by stakeholder $i$ to have *high* risk under $e$.

For instance, assume the following assessments:

$$F(e_{\text{Sec}}) = \big(\{p_2, p_4\},\ \{p_4\},\ \{p_2, p_3, p_4\}\big),$$
$$F(e_{\text{Cost}}) = \big(\{p_3\},\ \{p_1, p_3, p_5\},\ \{p_1, p_5\}\big),$$
$$F(e_{\text{Sched}}) = \big(\{p_2, p_5\},\ \{p_5\},\ \{p_1, p_2, p_5\}\big).$$

Then $(F, E)$ is a 3-polar (multipolar) soft set over $U$: for each risk parameter $e$, the three components represent the "high-risk" project subsets identified from three distinct perspectives.

Interpretation: this structure supports decisions such as *consensus high risk* for $e$ (intersection $\bigcap_{i=1}^{3} F_i(e)$), or *any-stakeholder high risk* (union $\bigcup_{i=1}^{3} F_i(e)$), depending on the organization's risk policy.

Related notions include the following concepts.

- *Bipolar Soft Sets* [76–78]: model parameters by paired positive/negative approximations, enabling simultaneous support and opposition assessments for each object.
- *Bipolar HyperSoft Sets* [79–81]: extend bipolar soft sets to multi-attribute tuple parameters, assigning positive and negative approximations to each tuple.
- *HyperPolar Soft Sets* [82]: A hyperpolar soft set assigns each parameter $n$ polarity-based subsets of the universe, capturing qualitative evaluations from multiple agents simultaneously.

## 2.11 Dynamic Soft Set

A dynamic soft set is a time-indexed family of soft sets, modeling parameterized approximations that evolve across time or contexts [83–85].

**Definition 2.11.1** (Dynamic soft set)**.** [83–85] Let $U$ be a nonempty universe of discourse, let $E$ be a (nonempty) set of parameters, and let $T$ be a nonempty index set (e.g., time points, system states, or contexts).

A *dynamic soft set* over $(U, E)$ indexed by $T$ is a family

$$\mathcal{S} \;=\; \{\, (t, F_t, A_t) \mid t \in T \,\},$$

such that for each $t \in T$:

$$A_t \subseteq E \quad \text{and} \quad F_t : \; A_t \longrightarrow \mathcal{P}(U).$$

Equivalently, one may represent $\mathcal{S}$ as the set of triples

$$\mathcal{S} \;=\; \{\, (t, e, F_t(e)) \mid t \in T,\; e \in A_t \,\},$$

where $F_t(e) \subseteq U$ is the (crisp) approximation of $U$ with respect to parameter $e$ at index $t$.

For each $t \in T$, the pair $(F_t, A_t)$ is the *time-/context-slice soft set* of $\mathcal{S}$ at $t$.

**Remark 2.11.2** (Reduction to a classical soft set)**.** If there exists a fixed $A \subseteq E$ and a fixed mapping $F : A \to \mathcal{P}(U)$ such that $A_t = A$ and $F_t = F$ for all $t \in T$, then the dynamic soft set $\mathcal{S}$ reduces to the classical (static) soft set $(F, A)$.

**Example 2.11.3** (Real-life example of a dynamic soft set: daily product availability in a grocery store)**.** Let $U$ be a set of products sold by a grocery store:

$$U = \{p_1, p_2, p_3, p_4, p_5, p_6\}.$$

Let $E$ be a set of availability-related parameters:

$$E = \{e_{\text{In}}, e_{\text{Sale}}, e_{\text{Out}}\},$$

where $e_{\text{In}}$ = "in stock", $e_{\text{Sale}}$ = "on sale", and $e_{\text{Out}}$ = "out of stock". Let the time index set be three consecutive days,

$$T = \{t_1, t_2, t_3\}.$$

For each day $t \in T$, define a time-slice soft set $(F_t, A_t)$ describing the store status. Take $A_t = \{e_{\text{In}}, e_{\text{Sale}}, e_{\text{Out}}\} \subseteq E$ for all $t \in T$.

**Day $t_1$:**

$$F_{t_1}(e_{\text{In}}) = \{p_1, p_2, p_3, p_5\}, \quad F_{t_1}(e_{\text{Sale}}) = \{p_2, p_5\}, \quad F_{t_1}(e_{\text{Out}}) = \{p_4, p_6\}.$$

**Day $t_2$:**

$$F_{t_2}(e_{\text{In}}) = \{p_1, p_3, p_4, p_5\}, \quad F_{t_2}(e_{\text{Sale}}) = \{p_1, p_4\}, \quad F_{t_2}(e_{\text{Out}}) = \{p_2, p_6\}.$$

**Day $t_3$:**

$$F_{t_3}(e_{\text{In}}) = \{p_1, p_2, p_4, p_6\}, \quad F_{t_3}(e_{\text{Sale}}) = \{p_2, p_6\}, \quad F_{t_3}(e_{\text{Out}}) = \{p_3, p_5\}.$$

Define

$$\mathcal{S} = \{(t, F_t, A_t) \mid t \in T\}.$$

Then $\mathcal{S}$ is a dynamic soft set over $(U, E)$ indexed by $T$: each time point $t$ has its own soft mapping $F_t : A_t \to \mathcal{P}(U)$ describing which products are in stock, on sale, or out of stock on that day.

Interpretation: the same parameter (e.g., "in stock") may select different subsets of products as inventory changes over time, and the dynamic soft set records these time-dependent approximations.

## 2.12 Type-$n$ Soft Set

A Type-$n$ soft set iteratively parameterizes soft sets, assigning each parameter a Type-$(n-1)$ soft set over the universe. Related concepts include Type-2 fuzzy sets [86, 87], Type-2 neutrosophic sets [88–90], and Type-2 soft sets [91].

**Definition 2.12.1** (Type-$n$ soft set)**.** Let $U$ be a nonempty universe and let $E$ be a nonempty set of parameters. Define, recursively, the classes $\Sigma^{(n)}(U, E)$ of Type-$n$ soft sets over $(U, E)$ as follows.

(1) **Type-1 (classical) soft sets.**

$$\Sigma^{(1)}(U,E) \;:=\; \{\,(F,A) \;\big|\; A\subseteq E,\;\; F:A\to\mathcal{P}(U)\,\}.$$

Elements of $\Sigma^{(1)}(U,E)$ are (crisp) soft sets in the sense of Molodtsov.

(2) **Inductive step.** For each integer $n\geq 2$, define

$$\Sigma^{(n)}(U,E) \;:=\; \{\,(F,A) \;\big|\; A\subseteq E,\;\; F:A\to\Sigma^{(n-1)}(U,E)\,\}.$$

An element $(F,A)\in\Sigma^{(n)}(U,E)$ is called a *Type-n soft set* (briefly, $T_nSS$) over $(U,E)$. The set $A$ is the *primary parameter set*. For each $a\in A$ we may write

$$F(a)=(F_a,A_a)\in\Sigma^{(n-1)}(U,E),$$

so that $A_a\subseteq E$ is a (possibly $a$-dependent) *underlying parameter set at level* 2. Iterating this decomposition, for any chain

$$a_1\in A,\quad a_2\in A_{a_1},\quad \ldots,\quad a_n\in A_{a_1\cdots a_{n-1}},$$

the terminal evaluation is a subset of the universe,

$$F_{a_1\cdots a_{n-1}}(a_n)\;\subseteq\;U.$$

**Remark 2.12.2.** (i) For $n=1$, Definition 2.12.1 reduces to the usual (crisp) soft set $F:A\to\mathcal{P}(U)$.

(ii) For $n=2$, $F:A\to\Sigma^{(1)}(U,E)$, i.e., each primary parameter $a\in A$ is assigned a Type-1 soft set; this is the usual Type-2 soft set [91–94] viewpoint (parameterization over an already parameterized family).

**Example 2.12.3** (Real-life example of a Type-3 soft set: company selection by department $\to$ criterion $\to$ strictness)**.** Let $U$ be a set of job candidates:

$$U=\{u_1,u_2,u_3,u_4,u_5,u_6\}.$$

Let $E$ be a pool of evaluation parameters (used at all levels), including:

$$E=\{e_{\mathrm{Alg}},e_{\mathrm{Comm}},e_{\mathrm{Exp}},e_{\mathrm{Strong}},e_{\mathrm{Moderate}}\}.$$

Interpret $e_{\mathrm{Alg}}=$ "good algorithms", $e_{\mathrm{Comm}}=$ "good communication", $e_{\mathrm{Exp}}=$ "relevant experience", and $e_{\mathrm{Strong}},e_{\mathrm{Moderate}}$ as *strictness levels* (meta-criteria) for acceptance.

We construct a Type-3 soft set $(F,A)\in\Sigma^{(3)}(U,E)$ that models the following hierarchy:

$$\text{Department}\;\longrightarrow\;\text{Criterion}\;\longrightarrow\;\text{Strictness}.$$

**Level 1 (primary parameters): departments.** Let

$$A=\{a_{\mathrm{Eng}},a_{\mathrm{PM}}\}\subseteq E$$

be the primary parameter set, where $a_{\mathrm{Eng}}$ = "Engineering department" and $a_{\mathrm{PM}}$ = "Product management department". For each department $a \in A$, we define a Type-2 soft set

$$F(a) = (F_a, A_a) \in \Sigma^{(2)}(U, E).$$

**Level 2: criteria used by each department.** Let

$$A_{a_{\mathrm{Eng}}} = \{e_{\mathrm{Alg}}, e_{\mathrm{Exp}}\} \subseteq E, \qquad A_{a_{\mathrm{PM}}} = \{e_{\mathrm{Comm}}, e_{\mathrm{Exp}}\} \subseteq E.$$

Thus Engineering focuses on $(\mathrm{Alg}, \mathrm{Exp})$, while PM focuses on $(\mathrm{Comm}, \mathrm{Exp})$. For each criterion $c \in A_a$, we define a Type-1 soft set

$$F_a(c) = (F_{a,c}, A_{a,c}) \in \Sigma^{(1)}(U, E),$$

where $A_{a,c}$ is a strictness-parameter set and $F_{a,c} : A_{a,c} \to \mathcal{P}(U)$ returns the candidate subset accepted under that strictness.

**Level 3: strictness parameters and final accepted subsets.** For each $(a, c)$ above, take

$$A_{a,c} = \{e_{\mathrm{Strong}}, e_{\mathrm{Moderate}}\} \subseteq E.$$

Define the terminal (Type-1) acceptance mappings as follows.

*Engineering → Algorithms:*

$$F_{a_{\mathrm{Eng}},e_{\mathrm{Alg}}}(e_{\mathrm{Strong}}) = \{u_1, u_4\}, \qquad F_{a_{\mathrm{Eng}},e_{\mathrm{Alg}}}(e_{\mathrm{Moderate}}) = \{u_1, u_3, u_4, u_6\}.$$

*Engineering → Experience:*

$$F_{a_{\mathrm{Eng}},e_{\mathrm{Exp}}}(e_{\mathrm{Strong}}) = \{u_2, u_4\}, \qquad F_{a_{\mathrm{Eng}},e_{\mathrm{Exp}}}(e_{\mathrm{Moderate}}) = \{u_1, u_2, u_4, u_5\}.$$

*PM → Communication:*

$$F_{a_{\mathrm{PM}},e_{\mathrm{Comm}}}(e_{\mathrm{Strong}}) = \{u_3, u_5\}, \qquad F_{a_{\mathrm{PM}},e_{\mathrm{Comm}}}(e_{\mathrm{Moderate}}) = \{u_1, u_3, u_5, u_6\}.$$

*PM → Experience:*

$$F_{a_{\mathrm{PM}},e_{\mathrm{Exp}}}(e_{\mathrm{Strong}}) = \{u_2, u_5\}, \qquad F_{a_{\mathrm{PM}},e_{\mathrm{Exp}}}(e_{\mathrm{Moderate}}) = \{u_2, u_4, u_5, u_6\}.$$

**Verification of the Type-3 structure.** By construction, for each $a \in A$, the pair $(F_a, A_a)$ is a Type-2 soft set because $F_a : A_a \to \Sigma^{(1)}(U, E)$. Moreover, for each $c \in A_a$, the pair $(F_{a,c}, A_{a,c})$ is a Type-1 soft set because $F_{a,c} : A_{a,c} \to \mathcal{P}(U)$. Hence $(F, A) \in \Sigma^{(3)}(U, E)$ is a Type-3 soft set.

**Interpretation.** A chain $(a, c, s)$ with $a \in \{\mathrm{Eng}, \mathrm{PM}\}$, $c$ a criterion used by $a$, and $s \in \{\mathrm{Strong}, \mathrm{Moderate}\}$ yields the final accepted subset $F_{a,c}(s) \subseteq U$. For example, the chain

$$a_{\mathrm{Eng}} \;\to\; e_{\mathrm{Alg}} \;\to\; e_{\mathrm{Strong}}$$

selects the candidates $\{u_1, u_4\}$, whereas

$$a_{\mathrm{PM}} \;\to\; e_{\mathrm{Comm}} \;\to\; e_{\mathrm{Moderate}}$$

selects $\{u_1, u_3, u_5, u_6\}$.

## 2.13 L-Soft Set

An $L$-soft set maps each parameter to an $L$-valued set on the universe, enabling lattice-graded, parameterized membership evaluations [95, 96].

**Definition 2.13.1** ($L$-set)**.** [95] Let $X$ be a nonempty universe and let $(L, \leq)$ be a bounded lattice (or, more generally, a poset of truth degrees with designated $0_L, 1_L$). An *$L$-set* (also called an *$L$-fuzzy set*) on $X$ is a mapping

$$A : X \longrightarrow L.$$

For $x \in X$, the value $A(x) \in L$ is interpreted as the $L$-valued grade of membership (truth degree) of the statement "$x \in A$". We denote by $L^X$ the class of all $L$-sets on $X$.

**Definition 2.13.2** ($L$-soft set)**.** [95] Let $X$ be a nonempty universe, let $E$ be a nonempty set of parameters, and fix a bounded lattice $(L, \leq, 0_L, 1_L)$. Let $A \subseteq E$ be nonempty. An *$L$-soft set* over $X$ (with parameter set $A$) is a pair

$$\Theta = (F, A),$$

where

$$F : \ A \longrightarrow L^X.$$

Equivalently, $\Theta$ can be identified with a single mapping

$$\mu_\Theta : \ A \times X \longrightarrow L, \qquad \mu_\Theta(a, x) := F(a)(x),$$

so that for each fixed $a \in A$ the section $x \mapsto \mu_\Theta(a, x)$ is an $L$-set on $X$. For $a \in A$, the $L$-set $F(a) \in L^X$ is called the *$a$-approximation* (or $a$-evaluation) of $\Theta$.

**Example 2.13.3** (An $L$-soft set for qualitative product-rating in e-commerce)**.** Let $X$ be a finite set of smartphones

$$X = \{p_1, p_2, p_3, p_4\},$$

and let the parameter set be

$$A = \{\mathsf{Battery}, \mathsf{Camera}, \mathsf{Price}\}.$$

We use a bounded lattice of linguistic grades

$$L = \{\mathbf{L}, \mathbf{M}, \mathbf{H}\}, \qquad \mathbf{L} \leq \mathbf{M} \leq \mathbf{H}, \qquad 0_L = \mathbf{L}, \ \ 1_L = \mathbf{H},$$

interpreted as *Low*, *Medium*, and *High* satisfaction, respectively.

Define an $L$-soft set $\Theta = (F, A)$ by specifying, for each parameter $a \in A$, an $L$-set $F(a) \in L^X$ (i.e., a map $X \to L$) that records the qualitative evaluation of each phone:

$$\begin{aligned} &F(\mathsf{Battery}) : X \to L, \\ &F(\mathsf{Battery})(p_1) = \mathbf{H}, \qquad F(\mathsf{Battery})(p_2) = \mathbf{M}, \\ &F(\mathsf{Battery})(p_3) = \mathbf{H}, \qquad F(\mathsf{Battery})(p_4) = \mathbf{L}, \end{aligned}$$

$$\begin{aligned} &F(\mathsf{Camera}) : X \to L, \\ &F(\mathsf{Camera})(p_1) = \mathbf{M}, \qquad F(\mathsf{Camera})(p_2) = \mathbf{H}, \\ &F(\mathsf{Camera})(p_3) = \mathbf{M}, \qquad F(\mathsf{Camera})(p_4) = \mathbf{H}, \end{aligned}$$

$$\begin{aligned} &F(\mathsf{Price}) : X \to L, \\ &F(\mathsf{Price})(p_1) = \mathbf{M}, \qquad F(\mathsf{Price})(p_2) = \mathbf{L}, \\ &F(\mathsf{Price})(p_3) = \mathbf{H}, \qquad F(\mathsf{Price})(p_4) = \mathbf{M}. \end{aligned}$$

Equivalently, the associated membership map $\mu_\Theta : A \times X \to L$ given by $\mu_\Theta(a,p) = F(a)(p)$ encodes, for each criterion $a$, the lattice-graded satisfaction level of each product $p \in X$.

**Interpretation.** For example, $\mu_\Theta(\mathsf{Battery}, p_3) = \mathbf{H}$ means that $p_3$ is evaluated as *high* on battery life, while $\mu_\Theta(\mathsf{Price}, p_2) = \mathbf{L}$ means that $p_2$ is judged *low* (unfavorable) in price attractiveness. Thus $\Theta$ models parameterized, qualitative (lattice-valued) assessments without using real-valued scores.

## 2.14 PosetSoft set (monotone soft set)

A PosetSoft set is a soft set whose parameter order enforces monotonic inclusion: stronger parameters yield larger object subsets.

**Definition 2.14.1** (PosetSoft set (monotone soft set))**.** Let $U$ be a nonempty universe and let $(A, \preceq)$ be a nonempty partially ordered set of parameters. A *PosetSoft set* over $U$ is a soft set $(F, A)$ with

$$F : A \to \mathcal{P}(U)$$

satisfying the monotonicity constraint

$$a \preceq b \implies F(a) \subseteq F(b) \qquad (a, b \in A).$$

**Example 2.14.2** (Example of a PosetSoft set: apartment shortlisting under increasing budget)**.** Let the universe be a finite set of apartments

$$U = \{u_1, u_2, u_3, u_4, u_5, u_6\}.$$

Assume their monthly rents (in thousand JPY) are:

$$\mathrm{rent}(u_1) = 75,\ \mathrm{rent}(u_2) = 80,\ \mathrm{rent}(u_3) = 92,\ \mathrm{rent}(u_4) = 100,\ \mathrm{rent}(u_5) = 108,\ \mathrm{rent}(u_6) = 125.$$

Let the parameter set be the set of budget thresholds

$$A = \{80, 100, 120\},$$

equipped with the usual order $\preceq := \leq$ (so $80 \preceq 100 \preceq 120$). Define a soft set $(F, A)$ over $U$ by

$$F(b) = \{\, u \in U \mid \mathrm{rent}(u) \leq b \,\} \qquad (b \in A).$$

Concretely,

$$F(80) = \{u_1, u_2\}, \qquad F(100) = \{u_1, u_2, u_3, u_4\}, \qquad F(120) = \{u_1, u_2, u_3, u_4, u_5\}.$$

Then for any $a, b \in A$,

$$a \preceq b \implies F(a) \subseteq F(b),$$

because increasing the allowable budget can only add (and never remove) feasible apartments. Hence $(F, A)$ is a PosetSoft set over $U$.

**Theorem 2.14.3** (Soft-set structure and well-definedness of PosetSoft sets)**.** *Let $U$ be a nonempty universe and let $(A, \preceq)$ be a nonempty poset of parameters. Let $F : A \to \mathcal{P}(U)$ be a mapping satisfying*

$$a \preceq b \implies F(a) \subseteq F(b) \qquad (a, b \in A).$$

*Then:*

(i) ***(Soft-set structure).*** *$(F, A)$ is a (classical) soft set over $U$.*

(ii) ***(Well-defined monotonicity predicate).*** *The monotonicity constraint is a well-defined property of $(F, A)$, i.e., for each comparable pair $(a, b) \in A \times A$ with $a \preceq b$, the inclusion statement $F(a) \subseteq F(b)$ is unambiguous.*

(iii) ***(Induced order-preserving operator).*** *Define $\iota : \mathcal{P}(U)^A \to \{0, 1\}$ by $\iota(F) = 1$ iff $F$ satisfies $a \preceq b \Rightarrow F(a) \subseteq F(b)$. Then $\iota$ is well-defined, and the class*

$$\mathsf{PosetSoft}(U; A, \preceq) := \{(F, A) \mid F : A \to \mathcal{P}(U),\ \iota(F) = 1\}$$

*of all PosetSoft sets over $U$ (with parameter poset $(A, \preceq)$) is well-defined.*

(iv) ***(Canonical associated relation on $U$).*** *Define a binary relation $\preceq_F$ on $U$ by*

$$x \preceq_F y \quad :\Longleftrightarrow \quad (\forall a \in A)\ (x \in F(a) \Rightarrow y \in F(a)).$$

*Then $\preceq_F$ is well-defined and is a preorder on $U$ (reflexive and transitive).*

*Proof.* (i) By assumption $A \neq \varnothing$ and $F : A \to \mathcal{P}(U)$, hence $(F, A)$ is a soft set over $U$ by the standard definition.

(ii) For any $a, b \in A$ with $a \preceq b$, the sets $F(a)$ and $F(b)$ are uniquely determined subsets of $U$ because $F$ is a function. Therefore the statement $F(a) \subseteq F(b)$ is unambiguous, and the implication $a \preceq b \Rightarrow F(a) \subseteq F(b)$ is a well-defined predicate on $(F, A)$.

(iii) The map $\iota$ assigns to each function $F \in \mathcal{P}(U)^A$ a unique truth value in $\{0, 1\}$, because the defining condition is a (well-defined) universal statement over the set of comparable pairs in $A$. Hence $\iota$ is well-defined, and consequently the subset of soft sets satisfying $\iota(F) = 1$ is well-defined; this is exactly $\mathsf{PosetSoft}(U; A, \preceq)$.

(iv) We check reflexivity and transitivity.

*Reflexive.* Fix $x \in U$. For every $a \in A$, the implication $x \in F(a) \Rightarrow x \in F(a)$ holds, hence $x \preceq_F x$.

*Transitive.* Assume $x \preceq_F y$ and $y \preceq_F z$. Let $a \in A$ and suppose $x \in F(a)$. From $x \preceq_F y$ we obtain $y \in F(a)$, and then from $y \preceq_F z$ we obtain $z \in F(a)$. Thus $(\forall a \in A)(x \in F(a) \Rightarrow z \in F(a))$, i.e., $x \preceq_F z$.

Therefore $\preceq_F$ is a well-defined preorder on $U$. □

## 2.15 Random soft set

A random soft set is a measurable mapping from outcomes to soft sets, yielding parameter-indexed random subsets under uncertainty.

**Definition 2.15.1** (Measurable space of soft sets)**.** Fix $U$ and $A$ as above and write

$$\mathsf{SS}(U,A) := \big(\mathcal{P}(U)\big)^{A} = \{\, F : A \to \mathcal{P}(U) \,\},$$

the set of all soft mappings on $(U, A)$ (equivalently, all soft sets $(F, A)$ over $U$ with this fixed parameter set).

For each $a \in A$ and $u \in U$, define the *membership cylinder*

$$\mathcal{C}_{a,u} := \{\, F \in \mathsf{SS}(U,A) \mid u \in F(a) \,\} \subseteq \mathsf{SS}(U,A).$$

Let

$$\Sigma_{U,A} := \sigma\big(\{\mathcal{C}_{a,u} : a \in A,\ u \in U\}\big)$$

be the $\sigma$-algebra generated by all such cylinders.

**Definition 2.15.2** (Random soft set)**.** Let $(\Omega, \mathcal{F}, \mathbb{P})$ be a probability space and fix a universe $U$ and parameter set $A$. A *random soft set* (on $(U, A)$) is an $(\mathcal{F}, \Sigma_{U,A})$-measurable mapping

$$\mathbf{F} :\ (\Omega, \mathcal{F}) \longrightarrow \big(\mathsf{SS}(U,A), \Sigma_{U,A}\big).$$

For $\omega \in \Omega$, write $\mathbf{F}(\omega) = F_\omega \in \mathsf{SS}(U,A)$; then each outcome $\omega$ induces a (deterministic) soft set $(F_\omega, A)$ over $U$.

**Example 2.15.3** (Random soft set: commuting routes under random traffic)**.** Let $U = \{r_1, r_2, r_3, r_4\}$ be a finite set of commuting routes, such as train lines or driving routes, and let

$$A = \{a_{\text{Fast}}, a_{\text{Cheap}}, a_{\text{Safe}}\}$$

be a parameter set, where $a_{\text{Fast}}$ = "fast", $a_{\text{Cheap}}$ = "cheap", and $a_{\text{Safe}}$ = "low incident risk".

Model the (uncertain) morning condition by a finite probability space

$$\Omega = \{\omega_L, \omega_N, \omega_H\}, \qquad \mathcal{F} = 2^{\Omega}, \qquad \mathbb{P}(\omega_L) = 0.3,\ \mathbb{P}(\omega_N) = 0.5,\ \mathbb{P}(\omega_H) = 0.2,$$

where $\omega_L, \omega_N, \omega_H$ represent *light/normal/heavy* congestion (or disruption) states.

For each outcome $\omega \in \Omega$, define a soft set $(F_\omega, A)$ over $U$ by specifying $F_\omega(a) \subseteq U$ as the set of routes satisfying criterion $a$ under state $\omega$. For instance, let:

| | $F_\omega(a_{\text{Fast}})$ | $F_\omega(a_{\text{Cheap}})$ | $F_\omega(a_{\text{Safe}})$ |
|---|---|---|---|
| $\omega = \omega_L$ | $\{r_1, r_2\}$ | $\{r_3, r_4\}$ | $\{r_2, r_4\}$ |
| $\omega = \omega_N$ | $\{r_2\}$ | $\{r_3, r_4\}$ | $\{r_2, r_3\}$ |
| $\omega = \omega_H$ | $\{r_4\}$ | $\{r_4\}$ | $\{r_3, r_4\}$ |

(Interpretation: under heavy congestion, only route $r_4$ remains "fast" and also "cheap", while the "safe" set depends on disruption patterns.)

Define the mapping
$$\mathbf{F} : \Omega \longrightarrow \mathsf{SS}(U, A), \qquad \mathbf{F}(\omega) := (F_\omega, A).$$
Since $U$ and $A$ are finite, the soft-set space $\mathsf{SS}(U, A)$ is finite; taking $\Sigma_{U,A} = 2^{\mathsf{SS}(U,A)}$, the map $\mathbf{F}$ is automatically $(\mathcal{F}, \Sigma_{U,A})$-measurable. Hence $\mathbf{F}$ is a *random soft set* in the sense of Definition 2.15.2: each realized traffic state $\omega$ induces a deterministic soft set $(F_\omega, A)$ describing, parameterwise, which routes are acceptable that day.

**Proposition 2.15.4** (Pointwise measurability criterion)**.** *A mapping* $\mathbf{F} : \Omega \to \mathsf{SS}(U, A)$ *is a random soft set if and only if for every* $a \in A$ *and* $u \in U$ *the event*
$$\{\omega \in \Omega : \ u \in F_\omega(a)\} \in \mathcal{F}.$$

*Proof.* By Definition 2.15.2, $\mathbf{F}$ is measurable iff $\mathbf{F}^{-1}(B) \in \mathcal{F}$ for all $B \in \Sigma_{U,A}$. Since $\Sigma_{U,A}$ is generated by the cylinders $\mathcal{C}_{a,u}$ (Definition 2.15.1), this is equivalent to requiring $\mathbf{F}^{-1}(\mathcal{C}_{a,u}) \in \mathcal{F}$ for all $(a, u)$. But
$$\mathbf{F}^{-1}(\mathcal{C}_{a,u}) = \{\omega \in \Omega : \ \mathbf{F}(\omega) \in \mathcal{C}_{a,u}\} = \{\omega \in \Omega : \ u \in F_\omega(a)\},$$
which yields the claim. □

**Remark 2.15.5** (Relation to random set theory)**.** For each fixed parameter $a \in A$, the coordinate map
$$\mathbf{F}_a : \Omega \to \mathcal{P}(U), \qquad \mathbf{F}_a(\omega) := F_\omega(a),$$
is a *random subset* of $U$ in the sense that all membership events $\{\omega : \ u \in \mathbf{F}_a(\omega)\}$ are measurable. Thus a random soft set is precisely a *family of random subsets indexed by parameters*, packaged as a single measurable map into the soft-set space; this parallels the standard viewpoint of random sets as measurable set-valued mappings.

## 2.16 Capacitary soft set

A capacitary soft set assigns each parameter a normalized monotone capacity on $U$, representing nonadditive, uncertainty-aware set evaluations.

**Definition 2.16.1** (Capacitary soft set (nonadditive set-function valued))**.** Let $U$ be a nonempty finite universe and let
$$\mathrm{Cap}(U) := \Big\{\nu : \mathcal{P}(U) \to [0, 1] \ \Big| \ \nu(\varnothing) = 0, \ \nu(U) = 1, \ A \subseteq B \Rightarrow \nu(A) \leq \nu(B)\Big\}$$
be the family of (normalized) capacities on $U$. Let $A$ be a nonempty parameter set. A *capacitary soft set* over $U$ is a pair $(F, A)$ with
$$F : A \to \mathrm{Cap}(U).$$

**Example 2.16.2** (Real-life example of a capacitary soft set: evaluating cybersecurity control bundles under different threat contexts)**.** Let $U$ be a finite set of candidate cybersecurity controls:

$$U = \{c_1, c_2, c_3\},$$

where $c_1$ = multi-factor authentication (MFA), $c_2$ = endpoint protection (EDR), and $c_3$ = network monitoring (NDR). Let $A = \{a_{\text{Low}}, a_{\text{High}}\}$ be a parameter set of threat contexts (low-threat vs. high-threat season).

A capacitary soft set $(F, A)$ assigns to each $a \in A$ a normalized capacity $\nu_a : \mathcal{P}(U) \to [0, 1]$, which quantifies the (possibly nonadditive) overall *risk-reduction effectiveness* of any bundle $S \subseteq U$ under context $a$.

Define $\nu_{a_{\text{Low}}}$ by

$$\nu_{a_{\text{Low}}}(\varnothing) = 0, \qquad \nu_{a_{\text{Low}}}(U) = 1,$$

and

$$\begin{aligned}\nu_{a_{\text{Low}}}(\{c_1\}) &= 0.40, & \nu_{a_{\text{Low}}}(\{c_2\}) &= 0.30, & \nu_{a_{\text{Low}}}(\{c_3\}) &= 0.25,\\ \nu_{a_{\text{Low}}}(\{c_1, c_2\}) &= 0.70, & \nu_{a_{\text{Low}}}(\{c_1, c_3\}) &= 0.65, & \nu_{a_{\text{Low}}}(\{c_2, c_3\}) &= 0.55.\end{aligned}$$

Define $\nu_{a_{\text{High}}}$ by

$$\nu_{a_{\text{High}}}(\varnothing) = 0, \qquad \nu_{a_{\text{High}}}(U) = 1,$$

and

$$\begin{aligned}\nu_{a_{\text{High}}}(\{c_1\}) &= 0.45, & \nu_{a_{\text{High}}}(\{c_2\}) &= 0.35, & \nu_{a_{\text{High}}}(\{c_3\}) &= 0.35,\\ \nu_{a_{\text{High}}}(\{c_1, c_2\}) &= 0.85, & \nu_{a_{\text{High}}}(\{c_1, c_3\}) &= 0.80, & \nu_{a_{\text{High}}}(\{c_2, c_3\}) &= 0.75.\end{aligned}$$

Each $\nu_a$ is a normalized capacity (monotone w.r.t. $\subseteq$, with $\nu_a(\varnothing) = 0$ and $\nu_a(U) = 1$), and it is typically *nonadditive*; for instance, under high threat,

$$\nu_{a_{\text{High}}}(\{c_1, c_2\}) = 0.85 \quad \text{need not equal} \quad \nu_{a_{\text{High}}}(\{c_1\}) + \nu_{a_{\text{High}}}(\{c_2\}) = 0.80,$$

reflecting synergy/overlap effects among controls.

Now define $F : A \to \text{Cap}(U)$ by

$$F(a_{\text{Low}}) = \nu_{a_{\text{Low}}}, \qquad F(a_{\text{High}}) = \nu_{a_{\text{High}}}.$$

Then $(F, A)$ is a capacitary soft set over $U$ in the sense of Definition 2.16.1.

**Theorem 2.16.3** (Soft-set structure and well-definedness of capacitary soft sets)**.** *Let $U$ be a nonempty* finite *universe, let $A$ be a nonempty parameter set, and let*

$$\text{Cap}(U) := \Big\{\nu : \mathcal{P}(U) \to [0, 1] \;\Big|\; \nu(\varnothing) = 0,\ \nu(U) = 1,\ X \subseteq Y \Rightarrow \nu(X) \le \nu(Y)\Big\}.$$

*If $(F, A)$ satisfies $F : A \to \text{Cap}(U)$, then:*

(i) ***Soft-set structure.*** *$(F, A)$ is a $\mathcal{T}$-valued soft set over $U$ with codomain $\mathcal{T} = \text{Cap}(U)$; equivalently, it is a mapping $F : A \to \mathcal{T}^{\mathcal{P}(U)}$ whose values are monotone set functions normalized by $\nu(\varnothing) = 0$ and $\nu(U) = 1$.*

(ii) ***Well-defined parameter evaluations.*** *For each $a \in A$, the object $F(a)$ is a uniquely determined capacity on $U$, i.e., a uniquely determined function*

$$F(a) : \mathcal{P}(U) \to [0,1] \quad \textit{with} \quad F(a)(\varnothing) = 0,\ F(a)(U) = 1,\ X \subseteq Y \Rightarrow F(a)(X) \le F(a)(Y).$$

(iii) ***Well-defined induced evaluation map.*** *The map*

$$\Phi_{(F,A)} :\ A \times \mathcal{P}(U) \longrightarrow [0,1], \qquad \Phi_{(F,A)}(a, X) := F(a)(X),$$

*is well-defined (single-valued).*

*Proof.* (i) By assumption, $A$ is nonempty and $F$ is a mapping $A \to \mathrm{Cap}(U)$. A (classical) soft set over a universe $Z$ with parameter set $A$ is any map $A \to \mathcal{P}(Z)$. Here the "universe" being evaluated is $\mathcal{P}(U)$ and the codomain is not $\mathcal{P}(\cdot)$ but the prescribed codomain $\mathcal{T} = \mathrm{Cap}(U)$; thus $(F, A)$ is precisely a $\mathcal{T}$-valued soft set in the sense of Definition 2.19.1. Moreover, each $F(a) \in \mathrm{Cap}(U)$ is, by definition of $\mathrm{Cap}(U)$, a normalized monotone set function on $\mathcal{P}(U)$.

(ii) Fix $a \in A$. Since $F$ is a function, the value $F(a)$ is uniquely determined. Because the codomain of $F$ is $\mathrm{Cap}(U)$, we have $F(a) \in \mathrm{Cap}(U)$, hence $F(a) : \mathcal{P}(U) \to [0,1]$ and the three axioms $F(a)(\varnothing) = 0$, $F(a)(U) = 1$, and $X \subseteq Y \Rightarrow F(a)(X) \le F(a)(Y)$ hold for all $X, Y \subseteq U$. Therefore each parameter $a$ determines a uniquely defined capacity.

(iii) Define $\Phi_{(F,A)}(a, X) := F(a)(X)$ for $(a, X) \in A \times \mathcal{P}(U)$. This is meaningful because, by (ii), for each fixed $a \in A$ the expression $F(a)(X)$ is defined for every $X \in \mathcal{P}(U)$ and belongs to $[0, 1]$. Uniqueness follows from the single-valuedness of $F(a)$ as a function. Hence $\Phi_{(F,A)} : A \times \mathcal{P}(U) \to [0, 1]$ is well-defined. □

## 2.17 CoverSoft set

A CoverSoft set assigns each parameter a cover of $U$ by nonempty subsets, encoding parameter-dependent decompositions or granularizations.

**Definition 2.17.1** (CoverSoft set)**.** Let $U$ be a nonempty universe. Write

$$\mathrm{Cov}(U) := \Big\{ \mathcal{C} \subseteq \mathcal{P}(U) \setminus \{\varnothing\} \ \Big|\ \bigcup_{C \in \mathcal{C}} C = U \Big\}$$

for the family of all covers of $U$ by nonempty subsets. Let $A$ be a nonempty parameter set. A *CoverSoft set* over $U$ is a pair $(F, A)$ with

$$F : A \to \mathrm{Cov}(U).$$

**Example 2.17.2** (Real-life example of a CoverSoft set: decomposing a delivery region into service zones under different strategies)**.** Let $U$ be a finite universe of delivery addresses in a small region:

$$U = \{u_1, u_2, u_3, u_4, u_5, u_6\}.$$

Let $A = \{a_{\mathrm{Geo}}, a_{\mathrm{Time}}\}$ be a set of operational parameters, where $a_{\mathrm{Geo}}$ denotes a *geographic zoning strategy* and $a_{\mathrm{Time}}$ denotes a *time-window zoning strategy.*

A CoverSoft set $(F, A)$ assigns to each $a \in A$ a cover $F(a) \in \mathrm{Cov}(U)$, i.e., a family of nonempty subsets of $U$ whose union is $U$.

**(1) Geographic zones.** Define

$$F(a_{\mathrm{Geo}}) = \Big\{C_1,\ C_2,\ C_3\Big\},$$

where

$$C_1 = \{u_1, u_2\}, \qquad C_2 = \{u_3, u_4\}, \qquad C_3 = \{u_5, u_6\}.$$

Then each $C_i \neq \varnothing$ and

$$C_1 \cup C_2 \cup C_3 = U,$$

so $F(a_{\mathrm{Geo}}) \in \mathrm{Cov}(U)$.

**(2) Time-window zones (overlapping cover).** Define

$$F(a_{\mathrm{Time}}) = \Big\{D_1,\ D_2,\ D_3\Big\},$$

where

$$D_1 = \{u_1, u_3, u_5\} \quad \text{(morning-feasible)},$$
$$D_2 = \{u_2, u_3, u_4\} \quad \text{(afternoon-feasible)},$$
$$D_3 = \{u_4, u_6\} \quad \text{(evening-feasible)}.$$

Again, each $D_i \neq \varnothing$ and

$$D_1 \cup D_2 \cup D_3 = \{u_1, u_2, u_3, u_4, u_5, u_6\} = U,$$

hence $F(a_{\mathrm{Time}}) \in \mathrm{Cov}(U)$.

**(3) The CoverSoft set.** Therefore the mapping $F : A \to \mathrm{Cov}(U)$ given by

$$F(a_{\mathrm{Geo}}) = \{C_1, C_2, C_3\}, \qquad F(a_{\mathrm{Time}}) = \{D_1, D_2, D_3\},$$

defines a CoverSoft set $(F, A)$ over $U$ in the sense of Definition 2.17.1.

**Theorem 2.17.3** (Soft-set structure and well-definedness of CoverSoft sets)**.** *Let $U$ be a nonempty universe and let $A$ be a nonempty parameter set. Define*

$$\mathrm{Cov}(U) := \Big\{\mathcal{C} \subseteq \mathcal{P}(U) \setminus \{\varnothing\} \ \Big|\ \bigcup_{C \in \mathcal{C}} C = U\Big\}.$$

*Assume $(F, A)$ satisfies $F : A \to \mathrm{Cov}(U)$. Then:*

(i) ***(Soft-set structure).*** *$(F, A)$ is a $\mathcal{T}$-valued soft set over $U$ in the sense of Definition 2.19.1 with codomain $\mathcal{T} = \mathrm{Cov}(U)$.*

(ii) ***(Well-defined parameterwise covers).*** *For each $a \in A$, the value $F(a)$ is a uniquely determined cover of $U$ by nonempty subsets; in particular,*

$$F(a) \subseteq \mathcal{P}(U) \setminus \{\varnothing\} \quad \text{and} \quad \bigcup_{C \in F(a)} C = U.$$

(iii) ***(Well-defined induced granulation neighborhoods).*** *Define, for $(a, u) \in A \times U$,*

$$\mathcal{N}_{(F,A)}(a, u) \;:=\; \{\, C \in F(a) \mid u \in C \,\} \;\subseteq\; F(a).$$

*Then $\mathcal{N}_{(F,A)} : A \times U \to \mathcal{P}(\mathcal{P}(U))$ is well-defined and satisfies $\mathcal{N}_{(F,A)}(a, u) \neq \varnothing$ for all $(a, u) \in A \times U$.*

*Proof.* First note that $\mathrm{Cov}(U)$ is nonempty because $\{U\} \in \mathrm{Cov}(U)$ (as $U \neq \varnothing$).

(i) Since $A \neq \varnothing$ and $F : A \to \mathrm{Cov}(U)$, the pair $(F, A)$ is exactly a soft set whose values lie in the fixed codomain $\mathcal{T} = \mathrm{Cov}(U)$; this is precisely the notion of a $\mathcal{T}$-valued soft set.

(ii) Fix $a \in A$. Because $F$ is a function, $F(a)$ is uniquely determined. Moreover, $F(a) \in \mathrm{Cov}(U)$ implies by definition that $F(a) \subseteq \mathcal{P}(U) \setminus \{\varnothing\}$ and $\bigcup_{C \in F(a)} C = U$. Hence $F(a)$ is a well-defined cover of $U$ by nonempty subsets.

(iii) Fix $(a, u) \in A \times U$ and define $\mathcal{N}_{(F,A)}(a, u) = \{C \in F(a) : u \in C\}$. This set is well-defined because membership $u \in C$ is unambiguous for each $C \subseteq U$. To see nonemptiness, use $\bigcup_{C \in F(a)} C = U$ from (ii): since $u \in U$, there exists $C \in F(a)$ with $u \in C$, hence $\mathcal{N}_{(F,A)}(a, u) \neq \varnothing$. Therefore $\mathcal{N}_{(F,A)}$ is a well-defined (everywhere nonempty) neighborhood/granulation map. □

## 2.18 FiltrationSoft set

A FiltrationSoft set assigns each parameter a nested chain of subsets, representing multi-level selection stages from strict to relaxed.

**Definition 2.18.1** (FiltrationSoft set (multi-level subset output))**.** Let $U$ be a nonempty universe and fix an integer $k \geq 1$. Define

$$\mathrm{Fil}_k(U) := \Big\{ (S_0, \ldots, S_k) \in \mathcal{P}(U)^{k+1} \;\Big|\; S_0 \subseteq S_1 \subseteq \cdots \subseteq S_k \Big\}.$$

Let $A$ be a nonempty parameter set. A *FiltrationSoft set* over $U$ (of depth $k$) is a mapping

$$F : A \to \mathrm{Fil}_k(U).$$

**Example 2.18.2** (Real-life example of a FiltrationSoft set: multi-stage hiring shortlists under different job profiles)**.** Let $U$ be a finite set of job applicants:

$$U = \{u_1, u_2, u_3, u_4, u_5, u_6, u_7\}.$$

Fix depth $k = 3$, so that each output is a nested chain

$$(S_0, S_1, S_2, S_3) \in \mathrm{Fil}_3(U) \quad \text{with} \quad S_0 \subseteq S_1 \subseteq S_2 \subseteq S_3.$$

Let the parameter set be

$$A = \{a_{\mathrm{SE}}, a_{\mathrm{DS}}\},$$

where $a_{\mathrm{SE}}$ represents a *software engineer* profile and $a_{\mathrm{DS}}$ represents a *data scientist* profile.

Define $F : A \to \mathrm{Fil}_3(U)$ by specifying, for each profile $a \in A$, four successive shortlists:

- $S_0(a)$: candidates passing a strict initial screen (must-haves),
- $S_1(a)$: candidates passing a technical screen,
- $S_2(a)$: candidates passing interviews,
- $S_3(a)$: candidates considered at all for that profile.

**(1) Software engineer profile.** Let

$$F(a_{\mathrm{SE}}) = \big(S_0^{\mathrm{SE}}, S_1^{\mathrm{SE}}, S_2^{\mathrm{SE}}, S_3^{\mathrm{SE}}\big) := \big(\{u_1, u_2\},\ \{u_1, u_2, u_4\},\ \{u_1, u_2, u_4, u_6\},\ \{u_1, u_2, u_3, u_4, u_6\}\big).$$

Then

$$S_0^{\mathrm{SE}} \subseteq S_1^{\mathrm{SE}} \subseteq S_2^{\mathrm{SE}} \subseteq S_3^{\mathrm{SE}} \subseteq U,$$

so $F(a_{\mathrm{SE}}) \in \mathrm{Fil}_3(U)$.

**(2) Data scientist profile.** Let

$$F(a_{\mathrm{DS}}) = \big(S_0^{\mathrm{DS}}, S_1^{\mathrm{DS}}, S_2^{\mathrm{DS}}, S_3^{\mathrm{DS}}\big) := \big(\{u_2\},\ \{u_2, u_5\},\ \{u_2, u_5, u_7\},\ \{u_2, u_4, u_5, u_7\}\big).$$

Again,

$$S_0^{\mathrm{DS}} \subseteq S_1^{\mathrm{DS}} \subseteq S_2^{\mathrm{DS}} \subseteq S_3^{\mathrm{DS}} \subseteq U,$$

so $F(a_{\mathrm{DS}}) \in \mathrm{Fil}_3(U)$.

**(3) FiltrationSoft set interpretation.** Hence $F : A \to \mathrm{Fil}_3(U)$ is a FiltrationSoft set over $U$ in the sense of Definition 2.18.1. Each parameter $a \in A$ selects a *multi-level* (nested) sequence of acceptable applicants, representing progressively relaxed stages of the hiring pipeline for that job profile.

**Theorem 2.18.3** (Soft-set structure and well-definedness of FiltrationSoft sets)**.** *Let $U$ be a nonempty universe, fix an integer $k \geq 1$, and define*

$$\mathrm{Fil}_k(U) := \Big\{(S_0, \ldots, S_k) \in \mathcal{P}(U)^{k+1} \ \Big|\ S_0 \subseteq S_1 \subseteq \cdots \subseteq S_k\Big\}.$$

*Let $A$ be a nonempty parameter set, and let $F : A \to \mathrm{Fil}_k(U)$ be a mapping. Then:*

(i) ***(Nonemptiness of the codomain).*** $\mathrm{Fil}_k(U) \neq \varnothing$.

(ii) ***(Soft-set structure).*** *The filtration determines a $\mathcal{T}$-valued soft set in the sense of Definition 2.19.1, where*

$$\mathcal{T} = \{(b_0, \ldots, b_k) \in \{0,1\}^{k+1} : b_0 \leq \cdots \leq b_k\},$$

*via the map $\mu : A \times U \to \mathcal{T}$ defined by*

$$\mu(a,u) = \big(\mathbf{1}_{u \in S_0^{(a)}}, \ldots, \mathbf{1}_{u \in S_k^{(a)}}\big).$$

*The original filtration can be recovered from this map.*

(iii) ***(Well-defined filtration at each parameter).*** *For every $a \in A$ there exists a unique tuple*

$$F(a) = (S_0^{(a)}, \ldots, S_k^{(a)}) \in \mathcal{P}(U)^{k+1}$$

*satisfying the nesting chain*

$$S_0^{(a)} \subseteq S_1^{(a)} \subseteq \cdots \subseteq S_k^{(a)}.$$

(iv) ***(Well-defined stagewise soft sets).*** *For each $i \in \{0, 1, \ldots, k\}$ define the $i$-th stage map*

$$F_i : A \longrightarrow \mathcal{P}(U), \qquad F_i(a) := \pi_i(F(a)),$$

*where $\pi_i : \mathrm{Fil}_k(U) \to \mathcal{P}(U)$ is the projection $\pi_i(S_0, \ldots, S_k) = S_i$. Then each $(F_i, A)$ is a classical soft set over $U$, and moreover the family is pointwise monotone:*

$$F_i(a) \subseteq F_{i+1}(a) \qquad (\forall\, a \in A,\ \forall\, 0 \leq i < k).$$

*Proof.* (i) Since $U \neq \varnothing$, the tuple $(\varnothing, \ldots, \varnothing) \in \mathcal{P}(U)^{k+1}$ satisfies $\varnothing \subseteq \cdots \subseteq \varnothing$, hence $(\varnothing, \ldots, \varnothing) \in \mathrm{Fil}_k(U)$. Therefore $\mathrm{Fil}_k(U) \neq \varnothing$.

(ii) For $a \in A$, write $F(a) = (S_0^{(a)}, \ldots, S_k^{(a)})$. The pointwise nesting condition implies that $\mu(a,u) \in \mathcal{T}$ for all $(a,u) \in A \times U$. Thus $\mu : A \times U \to \mathcal{T}$ defines a $\mathcal{T}$-valued soft set by Definition 2.19.1. Conversely, the nested stage sets are uniquely recovered by

$$S_i^{(a)} = \{u \in U : (\mu(a,u))_i = 1\}.$$

Hence the representation is lossless.

(iii) Fix $a \in A$. Since $F$ is a function, $F(a)$ is uniquely determined. Also, $F(a) \in \mathrm{Fil}_k(U)$ means precisely that $F(a)$ is a $(k+1)$-tuple of subsets of $U$, say $F(a) = (S_0^{(a)}, \ldots, S_k^{(a)})$, satisfying $S_0^{(a)} \subseteq \cdots \subseteq S_k^{(a)}$. Hence the filtration chain at parameter $a$ is well-defined.

(iv) For each fixed $i$, the projection $\pi_i$ is a well-defined function from $\mathrm{Fil}_k(U)$ to $\mathcal{P}(U)$, so $F_i = \pi_i \circ F : A \to \mathcal{P}(U)$ is well-defined, and therefore $(F_i, A)$ is a classical soft set over $U$. Finally, for any $a \in A$, the tuple $F(a) = (S_0^{(a)}, \ldots, S_k^{(a)})$ lies in $\mathrm{Fil}_k(U)$, so $S_i^{(a)} \subseteq S_{i+1}^{(a)}$ for $0 \leq i < k$. But $F_i(a) = S_i^{(a)}$ and $F_{i+1}(a) = S_{i+1}^{(a)}$ by definition of $F_i$, hence $F_i(a) \subseteq F_{i+1}(a)$ for all $a$ and $i$. □

## 2.19 $\mathcal{T}$-valued soft set

A $\mathcal{T}$-valued soft set assigns each parameter a $\mathcal{T}$-valued function on the universe, encoding parameterized evaluations without requiring crisp subsets.

**Definition 2.19.1** ($\mathcal{T}$-valued soft set (general template))**.** Let $X$ be a nonempty universe, let $E$ be a nonempty set of parameters, and let $A \subseteq E$ be nonempty. Let $\mathcal{T}$ be a nonempty codomain set and write $\mathcal{T}^X := \{ f : X \to \mathcal{T} \}$. A *$\mathcal{T}$-valued soft set* over $X$ (with parameter set $A$) is a pair

$$\Theta = (F, A),$$

where

$$F : A \longrightarrow \mathcal{T}^X.$$

Equivalently, $\Theta$ can be identified with a single mapping

$$\mu_\Theta : A \times X \longrightarrow \mathcal{T}, \qquad \mu_\Theta(a, x) := F(a)(x),$$

so that for each fixed $a \in A$ the section $x \mapsto \mu_\Theta(a, x)$ is a $\mathcal{T}$-valued function on $X$.

**Theorem 2.19.2** (Well-definedness and canonical identification of $\mathcal{T}$-valued soft sets)**.** *In the setting of Definition 2.19.1, the two descriptions*

$$F : A \to \mathcal{T}^X \qquad \text{and} \qquad \mu_\Theta : A \times X \to \mathcal{T}, \quad \mu_\Theta(a, x) = F(a)(x),$$

*are equivalent in a canonical (bijection) sense. More precisely, the map*

$$\Phi : \mathcal{T}^{X\,A} := \{F : A \to \mathcal{T}^X\} \longrightarrow \mathcal{T}^{A\times X} := \{\mu : A \times X \to \mathcal{T}\}, \qquad \Phi(F)(a, x) := F(a)(x),$$

*is a well-defined bijection with inverse*

$$\Psi : \mathcal{T}^{A\times X} \longrightarrow \mathcal{T}^{X\,A}, \qquad \Psi(\mu)(a)(x) := \mu(a, x).$$

*Consequently, a $\mathcal{T}$-valued soft set $\Theta = (F, A)$ determines a unique map $\mu_\Theta : A \times X \to \mathcal{T}$, and conversely every such $\mu$ determines a unique $\mathcal{T}$-valued soft set $(\Psi(\mu), A)$.*

*Proof.* **Step 1 (Well-definedness of $\Phi$).** Let $F : A \to \mathcal{T}^X$. For each $a \in A$, $F(a) \in \mathcal{T}^X$ is (by definition of $\mathcal{T}^X$) a function $F(a) : X \to \mathcal{T}$. Hence for each $(a, x) \in A \times X$ the value $F(a)(x) \in \mathcal{T}$ is uniquely determined. Therefore $\Phi(F) : A \times X \to \mathcal{T}$ given by $(a, x) \mapsto F(a)(x)$ is a well-defined function.

**Step 2 (Well-definedness of $\Psi$).** Let $\mu : A \times X \to \mathcal{T}$. Fix $a \in A$. Define $\Psi(\mu)(a) : X \to \mathcal{T}$ by $x \mapsto \mu(a, x)$. This is a well-defined function in $\mathcal{T}^X$. Thus $a \mapsto \Psi(\mu)(a)$ defines a well-defined mapping $\Psi(\mu) : A \to \mathcal{T}^X$.

**Step 3 ($\Phi$ and $\Psi$ are inverses).** For $F : A \to \mathcal{T}^X$ and any $(a, x) \in A \times X$,

$$(\Phi \circ \Psi)(\Phi(F))(a, x) = \Phi(F)(a, x) = F(a)(x),$$

and more directly,

$$(\Psi \circ \Phi)(F)(a)(x) = \Phi(F)(a, x) = F(a)(x) \quad (\forall a \in A,\ \forall x \in X),$$

so $\Psi(\Phi(F)) = F$ as functions $A \to \mathcal{T}^X$.

Similarly, for $\mu : A \times X \to \mathcal{T}$ and any $(a, x) \in A \times X$,

$$(\Phi \circ \Psi)(\mu)(a, x) = \Psi(\mu)(a)(x) = \mu(a, x),$$

so $\Phi(\Psi(\mu)) = \mu$ as functions $A \times X \to \mathcal{T}$.

Thus $\Phi$ is bijective with inverse $\Psi$. The stated uniqueness claims follow immediately from the existence of this bijection. □

**Definition 2.19.3** (Vector-valued soft set)**.** Let $X$ be a nonempty universe, let $E$ be a nonempty set of parameters, and let $A \subseteq E$ be nonempty. Let $V$ be a vector space over a field $\mathbb{K}$. A *vector-valued soft set* over $X$ (with parameter set $A$) is a $V$-valued soft set

$$\Theta_V = (F, A), \qquad F : A \longrightarrow V^X.$$

Equivalently, it is a map $\mu_{\Theta_V} : A \times X \to V$ given by $\mu_{\Theta_V}(a, x) = F(a)(x)$.

**Example 2.19.4** (Vector-valued soft set: multi-criteria scoring of job candidates)**.** Let $X = \{x_1, x_2, x_3\}$ be three job candidates and let $E$ be a parameter pool. Choose the active parameter set

$$A = \{\mathsf{Tech}, \mathsf{Comm}\} \subseteq E$$

(technical skill and communication skill). Let $V = \mathbb{R}^2$.

Define a vector-valued soft set $\Theta_V = (F, A)$ by specifying, for each $a \in A$, a function $F(a) : X \to \mathbb{R}^2$ (so $F : A \to (\mathbb{R}^2)^X$):

$$F(\mathsf{Tech})(x_1) = (0.80, 0.70), \quad F(\mathsf{Tech})(x_2) = (0.60, 0.90), \quad F(\mathsf{Tech})(x_3) = (0.75, 0.65),$$
$$F(\mathsf{Comm})(x_1) = (0.60, 0.55), \quad F(\mathsf{Comm})(x_2) = (0.40, 0.60), \quad F(\mathsf{Comm})(x_3) = (0.85, 0.80).$$

Interpretation: under $\mathsf{Tech}$ the two coordinates represent (coding, algorithms) scores, and under $\mathsf{Comm}$ they represent (presentation, teamwork) scores. Equivalently, $\mu_{\Theta_V} : A \times X \to \mathbb{R}^2$ is given by $\mu_{\Theta_V}(a, x) = F(a)(x)$.

**Definition 2.19.5** (Matrix-valued soft set)**.** Let $X$ be a nonempty universe, let $E$ be a nonempty set of parameters, and let $A \subseteq E$ be nonempty. Fix integers $m, n \geq 1$ and a field $\mathbb{K}$, and denote by $\mathrm{Mat}_{m \times n}(\mathbb{K})$ the space of $m \times n$ matrices over $\mathbb{K}$. A *matrix-valued soft set* over $X$ (with parameter set $A$) is a $\mathrm{Mat}_{m \times n}(\mathbb{K})$-valued soft set

$$\Theta_M = (F, A), \qquad F : A \longrightarrow \mathrm{Mat}_{m \times n}(\mathbb{K})^X.$$

Equivalently, it is a map $\mu_{\Theta_M} : A \times X \to \mathrm{Mat}_{m \times n}(\mathbb{K})$ with $\mu_{\Theta_M}(a, x) = F(a)(x)$.

**Example 2.19.6** (Matrix-valued soft set: shift-dependent state-transition estimates of machines)**.** Let $X = \{M_1, M_2\}$ be two production machines and take

$$A = \{\mathsf{Day}, \mathsf{Night}\} \subseteq E$$

as operating-shift parameters. Fix $m = n = 2$ and $\mathbb{K} = \mathbb{R}$, so the codomain is $\mathrm{Mat}_{2\times 2}(\mathbb{R})$.

Define a matrix-valued soft set $\Theta_M = (F, A)$ by giving, for each $a \in A$, a function $F(a) : X \to \mathrm{Mat}_{2\times 2}(\mathbb{R})$:

$$F(\mathsf{Day})(M_1) = \begin{pmatrix} 0.97 & 0.03 \\ 0.40 & 0.60 \end{pmatrix}, \qquad F(\mathsf{Day})(M_2) = \begin{pmatrix} 0.95 & 0.05 \\ 0.30 & 0.70 \end{pmatrix},$$

$$F(\mathsf{Night})(M_1) = \begin{pmatrix} 0.94 & 0.06 \\ 0.50 & 0.50 \end{pmatrix}, \qquad F(\mathsf{Night})(M_2) = \begin{pmatrix} 0.92 & 0.08 \\ 0.45 & 0.55 \end{pmatrix}.$$

Interpretation: each 2×2 matrix is a simple estimated transition matrix between states $\{\mathsf{OK}, \mathsf{Fault}\}$ for the corresponding shift (rows = current state, columns = next state). Equivalently, $\mu_{\Theta_M}(a, x) = F(a)(x) \in \mathrm{Mat}_{2\times 2}(\mathbb{R})$.

**Definition 2.19.7** (Tensor-valued soft set)**.** Let $X$ be a nonempty universe, let $E$ be a nonempty set of parameters, and let $A \subseteq E$ be nonempty. Fix a field $\mathbb{K}$ and vector spaces $V_1, \ldots, V_r$ over $\mathbb{K}$ ($r \geq 1$), and put

$$\mathcal{T} := V_1 \otimes \cdots \otimes V_r$$

(the $r$-th order tensor space of type $(V_1, \ldots, V_r)$). A *tensor-valued soft set* over $X$ (with parameter set $A$) is a $\mathcal{T}$-valued soft set

$$\Theta_T = (F, A), \qquad F : A \longrightarrow \mathcal{T}^X.$$

Equivalently, it is a map $\mu_{\Theta_T} : A \times X \to \mathcal{T}$ with $\mu_{\Theta_T}(a, x) = F(a)(x)$.

**Example 2.19.8** (Tensor-valued soft set: store-dependent $2 \times 2 \times 2$ demand-context tensor)**.** Let $X = \{S_1, S_2\}$ be two retail stores and let

$$A = \{\mathsf{Weekday}, \mathsf{Weekend}\} \subseteq E$$

be context parameters. Fix $\mathbb{K} = \mathbb{R}$ and take

$$V_1 = \mathbb{R}^2 \text{ (demand: Low/High)}, \qquad V_2 = \mathbb{R}^2 \text{ (weather: Cool/Hot)}, \qquad V_3 = \mathbb{R}^2 \text{ (promotion: Off/On)},$$

so $\mathcal{T} = V_1 \otimes V_2 \otimes V_3$ can be represented as $2 \times 2 \times 2$ arrays.

Define a tensor-valued soft set $\Theta_T = (F, A)$ by specifying $F(a) : X \to \mathcal{T}$. For readability, we write $F(a)(x) = [t_{ijk}]_{i,j,k\in\{1,2\}}$ as two $2 \times 2$ slices (promotion Off/On):

**Store $S_1$.**

$$F(\mathsf{Weekday})(S_1): \quad k = 1\ (\mathsf{Off}) \Rightarrow \begin{pmatrix} 0.30 & 0.10 \\ 0.20 & 0.05 \end{pmatrix}, \quad k = 2\ (\mathsf{On}) \Rightarrow \begin{pmatrix} 0.15 & 0.05 \\ 0.10 & 0.05 \end{pmatrix},$$

$$F(\mathsf{Weekend})(S_1): \quad k = 1\ (\mathsf{Off}) \Rightarrow \begin{pmatrix} 0.20 & 0.10 \\ 0.25 & 0.10 \end{pmatrix}, \quad k = 2\ (\mathsf{On}) \Rightarrow \begin{pmatrix} 0.10 & 0.05 \\ 0.15 & 0.05 \end{pmatrix}.$$

**Store $S_2$.**

$$F(\mathsf{Weekday})(S_2): \quad k = 1\ (\mathsf{Off}) \Rightarrow \begin{pmatrix} 0.35 & 0.10 \\ 0.15 & 0.05 \end{pmatrix}, \quad k = 2\ (\mathsf{On}) \Rightarrow \begin{pmatrix} 0.20 & 0.05 \\ 0.08 & 0.02 \end{pmatrix}.$$

Interpretation: $t_{ijk}$ is a store- and context-dependent weight (e.g., empirical frequency) for the joint situation (demand $i$, weather $j$, promotion $k$). Equivalently, $\mu_{\Theta_T}(a, x) = F(a)(x) \in V_1 \otimes V_2 \otimes V_3$.

## 2.20 Cubic Soft Set

Cubic soft sets map each parameter to cubic sets: interval-valued membership degrees plus fuzzy membership, modeling uncertainty with dual information. The definition of the Cubic Soft Set is described as follows [97–99].

**Definition 2.20.1** (Cubic Soft Set)**.** [97] Let $X$ be a nonempty universe and let $E$ be a set of parameters. First, a *cubic set* in $X$ is defined as a mapping that assigns to each element $x \in X$ a pair

$$\langle [A^-(x), A^+(x)],\ \lambda(x) \rangle,$$

where:

- $[A^-(x), A^+(x)] \subseteq [0, 1]$ is an interval representing the degree of membership as given by an *interval-valued fuzzy set*, and
- $\lambda(x) \in [0, 1]$ is the membership degree provided by a *fuzzy set*.

Then, a *cubic soft set* over $X$ with respect to the parameter set $E$ is a mapping

$$F : E \to \{\text{cubic sets in } X\}.$$

Equivalently, a cubic soft set can be expressed as the collection

$$\widetilde{F} = \{(e, \widetilde{F}(e)) : e \in E\},$$

where for each parameter $e \in E$, the set $\widetilde{F}(e)$ is a cubic set in $X$; that is,

$$\widetilde{F}(e) = \{\langle x, [A_e^-(x), A_e^+(x)],\ \lambda_e(x) \rangle : x \in X\}.$$

**Example 2.20.2** (Real-life example of a cubic soft set: apartment screening with uncertain scores and point estimates)**.** Let $X$ be a set of apartments:

$$X = \{a_1, a_2, a_3, a_4\}.$$

Let $E$ be a set of decision parameters and consider

$$E = \{e_{\text{Safe}}, e_{\text{Comm}}\},$$

where $e_{\text{Safe}}$ = "neighborhood safety" and $e_{\text{Comm}}$ = "commute convenience".

A cubic soft set $F : E \to \{\text{cubic sets in } X\}$ assigns to each parameter $e \in E$ a cubic set on $X$ of the form

$$\widetilde{F}(e) = \Big\{ \langle x,\ [A_e^-(x), A_e^+(x)],\ \lambda_e(x) \rangle \ \Big|\ x \in X \Big\},$$

where $[A_e^-(x), A_e^+(x)]$ is an interval-valued membership (reflecting uncertainty in the score) and $\lambda_e(x)$ is a single membership degree (a point estimate).

For instance, based on crime statistics and expert judgment, suppose safety is assessed as:

$$\widetilde{F}(e_{\text{Safe}}) = \Big\{ \langle a_1, [0.70, 0.85], 0.80 \rangle,\ \langle a_2, [0.40, 0.60], 0.50 \rangle,\ \langle a_3, [0.55, 0.75], 0.65 \rangle,\ \langle a_4, [0.20, 0.35], 0.30 \rangle \Big\}.$$

Similarly, using travel-time variability data, suppose commute convenience is assessed as:

$$\widetilde{F}(e_{\text{Comm}}) = \Big\{ \langle a_1, [0.30, 0.50], 0.40 \rangle,\ \langle a_2, [0.75, 0.90], 0.85 \rangle,\ \langle a_3, [0.50, 0.70], 0.60 \rangle,\ \langle a_4, [0.60, 0.80], 0.70 \rangle \Big\}.$$

Then $F$ is a cubic soft set on $X$: each parameter $e$ is associated with a cubic evaluation of every apartment, combining an interval-valued degree (uncertainty band) with a representative point degree. In decision-making, one may aggregate $\widetilde{F}(e_{\text{Safe}})$ and $\widetilde{F}(e_{\text{Comm}})$ to rank apartments under both safety and commuting considerations.

## 2.21 Probabilistic Soft Set

A probabilistic soft set maps each parameter to a probability distribution over the universe, modeling uncertainty via normalized likelihoods [100, 101].

**Definition 2.21.1** (Probabilistic Soft Set)**.** [100,101] Let $U$ be a non-empty finite universe and $E$ be a set of parameters. Let $A \subseteq E$ be a subset of parameters. Denote by

$$D(U) = \left\{ \mu : U \to [0,1] \,\middle|\, \sum_{u \in U} \mu(u) = 1 \right\}$$

the set of all probability distributions on $U$. A **probabilistic soft set** over $U$ is a pair $(F, A)$ where

$$F : A \to D(U)$$

such that for each $e \in A$, the function $F(e) : U \to [0,1]$ satisfies

$$\sum_{u \in U} F(e)(u) = 1.$$

In other words, for every parameter $e \in A$, $F(e)$ is a probability distribution on $U$.

**Example 2.21.2** (Real-life example of a probabilistic soft set: choosing a commuting route under different criteria)**.** Let $U$ be a finite set of candidate commuting routes:

$$U = \{r_1, r_2, r_3, r_4\}.$$

Let $E$ be a set of decision parameters and take

$$A = \{e_{\text{Fast}}, e_{\text{Cheap}}, e_{\text{Comfort}}\} \subseteq E,$$

where $e_{\text{Fast}}$ = "fastest", $e_{\text{Cheap}}$ = "cheapest", and $e_{\text{Comfort}}$ = "most comfortable".

A probabilistic soft set $(F, A)$ assigns to each parameter $e \in A$ a probability distribution on $U$, interpreted as the likelihood that each route is the best choice under criterion $e$ (e.g., estimated from historical travel data and user preference models).

For instance, define $F : A \to D(U)$ by the following distributions:

| $F(e)(u)$ | $r_1$ | $r_2$ | $r_3$ | $r_4$ |
|---|---|---|---|---|
| $F(e_{\text{Fast}})(\cdot)$ | 0.55 | 0.25 | 0.15 | 0.05 |
| $F(e_{\text{Cheap}})(\cdot)$ | 0.10 | 0.20 | 0.60 | 0.10 |
| $F(e_{\text{Comfort}})(\cdot)$ | 0.15 | 0.50 | 0.10 | 0.25 |

Each row sums to 1, so each $F(e)$ is a probability distribution on $U$.

Thus $(F, A)$ is a probabilistic soft set over $U$. Interpretation: under "fastest" the model favors route $r_1$, under "cheapest" it favors $r_3$, and under "most comfortable" it favors $r_2$.

The comparison of classical soft sets and probabilistic soft sets is presented in Table 2.4.

Table 2.4: Concise comparison of classical soft sets and probabilistic soft sets over a finite universe $U$.

| **Aspect** | **Soft set** | **Probabilistic soft set** |
|---|---|---|
| Universe/parameters | Finite (or general) universe $U$ and parameter set $A \subseteq E$. | Finite universe $U$ and parameter set $A \subseteq E$ (typically finite to interpret distributions). |
| Value assigned to a parameter $a \in A$ | A crisp subset $F(a) \subseteq U$. | A probability distribution $F(a) \in D(U)$, i.e. $F(a) : U \to [0,1]$ with $\sum_{u \in U} F(a)(u) = 1$. |
| Mathematical type of $F$ | $F : A \to \mathcal{P}(U)$. | $F : A \to D(U) \subseteq [0,1]^U$. |
| Interpretation | "Accepted/feasible objects under parameter $a$" (yes/no membership). | "Likelihood/degree of preference of each object under parameter $a$" (normalized uncertainty). |
| Aggregation across parameters | Often via set operations (union/intersection) or counting scores. | Often via probabilistic combination (e.g., weighted mixtures, Bayesian updates, expected-utility rules). |
| Relation between the two | Baseline model with crisp information. | Generalizes soft sets: a soft set can be embedded by using point-mass distributions on $F(a)$ (or thresholding $F(a)$). |

## 2.22 D-soft set

A D-soft set maps each parameter to a D-number mass assignment on subsets, handling incompleteness and nonexclusive evidence.

**Definition 2.22.1** (D-number on $U$)**.** [102, 103] Let $U$ be a nonempty *finite* universe. A *D-number* on $U$ is a mapping

$$D : 2^U \longrightarrow [0,1]$$

satisfying

$$D(\varnothing) = 0, \qquad \sum_{B \subseteq U} D(B) \le 1.$$

(Unlike classical Dempster–Shafer basic probability assignments, D-number theory does not require the elements of $U$ to be mutually exclusive, and it allows incomplete information when the above sum is $< 1$.) Let

$$\mathsf{DNum}(U) := \left\{ D : 2^U \to [0,1] \;\middle|\; D(\varnothing) = 0, \sum_{B \subseteq U} D(B) \le 1 \right\}$$

be the set of all D-numbers on $U$.

**Definition 2.22.2** (D-soft set)**.** Let $U$ be a nonempty *finite* universe and let $E$ be a nonempty set of parameters. Let $A \subseteq E$ be nonempty. A *D-soft set* over $U$ with parameter set $A$ is a pair $(F, A)$ where

$$F : A \longrightarrow \mathsf{DNum}(U).$$

Thus, for each parameter $e \in A$, the value $F(e) = D_e$ is a D-number on $U$, i.e., a mass assignment $D_e : 2^U \to [0,1]$ with $\sum_{B \subseteq U} D_e(B) \le 1$.

**Example 2.22.3** (Real-life example of a D-soft set: supplier selection under incomplete and nonexclusive evidence)**.** Let $U$ be a finite set of candidate suppliers for a manufacturing order:

$$U = \{s_1, s_2, s_3, s_4\}.$$

Let $E$ be a set of evaluation parameters and take the nonempty subset

$$A = \{e_{\text{Qual}}, e_{\text{Deliv}}\} \subseteq E,$$

where $e_{\text{Qual}}$ = "high product quality" and $e_{\text{Deliv}}$ = "reliable delivery".

A D-soft set $(F, A)$ assigns to each parameter $e \in A$ a D-number $D_e \in \mathsf{DNum}(U)$, i.e., a map $D_e : 2^U \to [0, 1]$ with $D_e(\varnothing) = 0$ and $\sum_{B \subseteq U} D_e(B) \leq 1$.

**(1) Evidence for quality.** Suppose quality audits provide the following (possibly incomplete) evidence:

$$D_{e_{\text{Qual}}}(\{s_1\}) = 0.45, \qquad D_{e_{\text{Qual}}}(\{s_1, s_3\}) = 0.20, \qquad D_{e_{\text{Qual}}}(\{s_2\}) = 0.10,$$

$$D_{e_{\text{Qual}}}(\{s_2, s_4\}) = 0.05, \qquad D_{e_{\text{Qual}}}(B) = 0 \text{ for all other } B \subseteq U.$$

Then

$$\sum_{B \subseteq U} D_{e_{\text{Qual}}}(B) = 0.45 + 0.20 + 0.10 + 0.05 = 0.80 \leq 1,$$

so $D_{e_{\text{Qual}}} \in \mathsf{DNum}(U)$; the remaining mass $1 - 0.80 = 0.20$ represents unassigned (unknown) information.

**(2) Evidence for delivery.** From shipping records and logistics reports, suppose we obtain:

$$D_{e_{\text{Deliv}}}(\{s_3\}) = 0.30, \qquad D_{e_{\text{Deliv}}}(\{s_1, s_3\}) = 0.25, \qquad D_{e_{\text{Deliv}}}(\{s_1, s_2, s_3\}) = 0.10,$$

$$D_{e_{\text{Deliv}}}(B) = 0 \text{ for all other } B \subseteq U.$$

Hence

$$\sum_{B \subseteq U} D_{e_{\text{Deliv}}}(B) = 0.30 + 0.25 + 0.10 = 0.65 \leq 1,$$

so $D_{e_{\text{Deliv}}} \in \mathsf{DNum}(U)$. Note that the focal sets $\{s_3\} \subseteq \{s_1, s_3\} \subseteq \{s_1, s_2, s_3\}$ are *not* mutually exclusive, which is allowed in D-number modeling.

**(3) The D-soft set.** Define $F : A \to \mathsf{DNum}(U)$ by

$$F(e_{\text{Qual}}) = D_{e_{\text{Qual}}}, \qquad F(e_{\text{Deliv}}) = D_{e_{\text{Deliv}}}.$$

Then $(F, A)$ is a D-soft set over $U$: each parameter is associated with a D-number encoding subset-valued evidence, while permitting incompleteness ($\sum D_e < 1$) and nonexclusive focal sets.

A comparison between a classical soft set and a D-soft set is presented in Table 2.5.

Table 2.5: Concise comparison between a classical soft set and a D-soft set over a (finite) universe $U$.

| **Aspect** | **Classical soft set** $(F, A)$ | **D-soft set** $(F, A)$ |
|---|---|---|
| Universe requirement | $U$ any nonempty set (often finite in applications). | $U$ is assumed finite since each value is a D-number on $2^U$. |
| Parameter domain | Nonempty $A \subseteq E$. | Nonempty $A \subseteq E$. |
| Codomain / value type | $F(e) \in \mathcal{P}(U)$ (a crisp subset of alternatives). | $F(e) = D_e \in \mathsf{DNum}(U)$ where $D_e : 2^U \to [0, 1]$ (a subset-mass assignment). |
| Meaning of $F(e)$ | Objects *accepted* under parameter $e$ (yes/no selection). | Evidence about *which subset(s)* of $U$ are plausible under $e$, via focal sets with masses. |
| Granularity of information | Element-level inclusion only. | Subset-level evidence; can express ambiguity such as "either $u_1$ or $u_3$" by mass on $\{u_1, u_3\}$. |
| Normalization / completeness | No probability-like normalization constraint. | $\sum_{B \subseteq U} D_e(B) \leq 1$; the gap $1 - \sum_B D_e(B)$ represents unassigned/unknown information. |
| Mutual exclusivity requirement | Not an evidence model; typically treats alternatives crisply. | D-number framework does not require mutual exclusivity of focal sets (and may model nonexclusive evidence). |
| Typical decision extraction | Scoring/ranking by counting satisfied parameters (or other set-based aggregations). | Ranking via belief/plausibility/credibility transforms or other mass-to-score rules (application-dependent). |

## 2.23 Complex Soft Sets

A complex soft set assigns to each parameter a complex-valued membership function on the universe, thereby encoding both magnitude and phase information (cf. [104]). Related notions include *complex fuzzy sets* [105, 106] and *complex neutrosophic sets* [107–109].

**Definition 2.23.1** (Complex fuzzy set)**.** [110, 111] Let $U$ be a nonempty universe. Define the closed unit disk

$$\mathbb{D} := \{ z \in \mathbb{C} \mid |z| \leq 1 \}.$$

A *complex fuzzy set* (CFS) on $U$ is a mapping

$$\mu_A : U \longrightarrow \mathbb{D}.$$

Equivalently, for each $u \in U$ we may write

$$\mu_A(u) = r_A(u)\, e^{i\omega_A(u)}, \qquad r_A(u) \in [0, 1], \quad \omega_A(u) \in [0, 2\pi],$$

where $r_A(u)$ is the *amplitude* (membership magnitude) and $\omega_A(u)$ is the *phase.* We denote by $\mathsf{CFS}(U)$ the family of all complex fuzzy sets on $U$.

**Definition 2.23.2** (Complex soft set)**.** Let $U$ be a nonempty universe and let $E$ be a nonempty set of parameters. Let $A \subseteq E$ be nonempty. A *complex soft set* (CSS) over $U$ with parameter set $A$ is a pair $(F, A)$ where

$$F : A \longrightarrow \mathsf{CFS}(U).$$

Thus, for each parameter $e \in A$, the value $F(e)$ is a complex fuzzy set on $U$, i.e.,

$$F(e) = \mu_e : U \longrightarrow \mathbb{D}.$$

Equivalently, a complex soft set can be identified with a single mapping

$$\mu : \ A \times U \longrightarrow \mathbb{D}, \qquad \mu(e,u) := \mu_e(u),$$

such that for every fixed $e \in A$, the section $\mu_e(\cdot)$ is a complex fuzzy membership function on $U$.

**Remark 2.23.3** (Reductions)**.** 1. If $\mu(e,u) \in [0,1] \subset \mathbb{C}$ (equivalently, all phases are 0), then $(F,A)$ reduces to a fuzzy soft set.

2. If $\mu(e,u) \in \{0,1\}$ for all $(e,u)$, then $(F,A)$ reduces to a (crisp) soft set.

**Example 2.23.4** (Real-life example of a complex soft set: wearable-sensor sleep assessment with confidence phase)**.** Let $U$ be a set of nights (sleep records) for a person:

$$U = \{n_1, n_2, n_3, n_4\}.$$

Let $E$ be a set of assessment parameters and take

$$A = \{e_{\text{Deep}}, e_{\text{Stress}}\} \subseteq E,$$

where $e_{\text{Deep}}$ = "deep-sleep quality" and $e_{\text{Stress}}$ = "low night-time stress".

A complex soft set $(F,A)$ assigns to each parameter $e \in A$ a complex fuzzy set $\mu_e : U \to \mathbb{D} = \{z \in \mathbb{C} \mid |z| \le 1\}$. Interpret the *amplitude* $r(e,n) \in [0,1]$ as the *degree* to which night $n$ satisfies $e$, and interpret the *phase* $\omega(e,n) \in [0,2\pi]$ as a *confidence/regularity marker* derived from signal quality (e.g., stable sensors yield small phase, noisy sensors yield larger phase).

Define $\mu : A \times U \to \mathbb{D}$ by $\mu(e,n) = r(e,n)e^{i\omega(e,n)}$ with the following values:

$$\mu(e_{\text{Deep}}, n_1) = 0.80\, e^{\,i0.10\pi}, \quad \mu(e_{\text{Deep}}, n_2) = 0.45\, e^{\,i0.35\pi},$$

$$\mu(e_{\text{Deep}}, n_3) = 0.70\, e^{\,i0.15\pi}, \quad \mu(e_{\text{Deep}}, n_4) = 0.20\, e^{\,i0.60\pi},$$

$$\mu(e_{\text{Stress}}, n_1) = 0.60\, e^{\,i0.20\pi}, \quad \mu(e_{\text{Stress}}, n_2) = 0.30\, e^{\,i0.55\pi},$$

$$\mu(e_{\text{Stress}}, n_3) = 0.75\, e^{\,i0.10\pi}, \quad \mu(e_{\text{Stress}}, n_4) = 0.50\, e^{\,i0.40\pi}.$$

For each fixed $e \in A$, the section $\mu_e(\cdot) = \mu(e,\cdot) : U \to \mathbb{D}$ is a complex membership function, so it defines a complex fuzzy set $F(e) = \mu_e \in \mathsf{CFS}(U)$. Hence $(F,A)$ (equivalently $\mu$) is a complex soft set.

Interpretation: the magnitude $|\mu(e,n)| = r(e,n)$ expresses how well the night $n$ satisfies the criterion $e$, while the phase $\arg(\mu(e,n)) = \omega(e,n)$ encodes a secondary aspect such as confidence or signal stability, which is useful when sensor quality varies across nights.

## 2.24 Real Soft Set

A real soft set maps each parameter to a bounded nonempty subset of real numbers, representing parameter-dependent possible real values [112–114].

**Definition 2.24.1** (Soft real set (real soft set))**.** [114] Let $A$ be a nonempty set of parameters and let $\mathbb{R}$ be the set of real numbers. Define the collection of all nonempty bounded subsets of $\mathbb{R}$ by

$$\mathcal{P}_{\mathrm{b}}(\mathbb{R}) \ := \ \{\, B \subseteq \mathbb{R} \mid B \neq \varnothing \text{ and } B \text{ is bounded}\,\}.$$

A *soft real set* (also called a *real soft set*) over $\mathbb{R}$ with parameter set $A$ is a pair $(F, A)$, where

$$F : \ A \longrightarrow \mathcal{P}_{\mathrm{b}}(\mathbb{R})$$

is a mapping. For each $\lambda \in A$, the value $F(\lambda) \subseteq \mathbb{R}$ is interpreted as the (parameter-dependent) set of possible real values under the parameter $\lambda$.

**Remark 2.24.2** (Soft real numbers as singleton soft real sets)**.** If $(F, A)$ is a *singleton* soft real set, i.e., $F(\lambda) = \{r(\lambda)\}$ for every $\lambda \in A$, then it is naturally identified with the corresponding *soft real number*

$$\tilde{r} : \ A \longrightarrow \mathbb{R}, \qquad \tilde{r}(\lambda) = r(\lambda).$$

**Example 2.24.3** (Real-life example of a soft real set: delivery-time windows under different shipping options)**.** Let $A$ be a set of shipping options for an online store:

$$A = \{\lambda_{\mathrm{Std}}, \lambda_{\mathrm{Exp}}, \lambda_{\mathrm{Eco}}\},$$

where $\lambda_{\mathrm{Std}}$ = standard shipping, $\lambda_{\mathrm{Exp}}$ = express shipping, and $\lambda_{\mathrm{Eco}}$ = economy shipping.

Consider the real quantity "delivery time" measured in days, so the value domain is $\mathbb{R}$. Because delivery times are uncertain but bounded for each option, define

$$F : A \longrightarrow \mathcal{P}_{\mathrm{b}}(\mathbb{R})$$

by assigning to each option the (bounded, nonempty) set of plausible delivery times:

$$F(\lambda_{\mathrm{Std}}) = [2, 5], \qquad F(\lambda_{\mathrm{Exp}}) = [1, 2], \qquad F(\lambda_{\mathrm{Eco}}) = [4, 9].$$

Each $F(\lambda) \subseteq \mathbb{R}$ is nonempty and bounded, hence $F(\lambda) \in \mathcal{P}_{\mathrm{b}}(\mathbb{R})$. Therefore $(F, A)$ is a soft real set over $\mathbb{R}$.

Interpretation: the same order may have different feasible delivery-time windows depending on the chosen shipping parameter $\lambda \in A$.

## 2.25 Intersectional soft sets

An intersectional soft set satisfies $F(x) \cap F(y) \subseteq F(x*y)$, ensuring combined parameters include all jointly satisfying objects [115–118].

**Definition 2.25.1** (Intersectional soft set)**.** [115–118] Let $U$ be a universal set and let $E$ be a set of parameters equipped with a binary operation

$$* : \; E \times E \longrightarrow E.$$

Let $A \subseteq E$ be a nonempty subset and let $(F, A)$ be a soft set over $U$, i.e.,

$$F : \; A \longrightarrow \mathcal{P}(U).$$

Then $(F, A)$ is called an *intersectional soft set* over $U$ (with respect to $*$) if, for all $x, y \in A$ such that $x * y \in A$, one has

$$F(x) \cap F(y) \; \subseteq \; F(x * y).$$

**Remark 2.25.2.** The condition says that whenever the "combined" parameter $x*y$ is admissible (lies in $A$), it must contain all objects that satisfy both $x$ and $y$ simultaneously.

**Example 2.25.3** (Real-life example of an intersectional soft set: product filtering with a "bundle" operation)**.** Let $U$ be a set of products in an online store:

$$U = \{p_1, p_2, p_3, p_4, p_5, p_6\}.$$

Let $E$ be a set of product-tag parameters and take

$$A = \{e_{\text{Bio}}, e_{\text{GF}}, e_{\text{BioGF}}\} \subseteq E,$$

where

$$e_{\text{Bio}} = \text{“organic”}, \qquad e_{\text{GF}} = \text{“gluten-free”}, \qquad e_{\text{BioGF}} = \text{“organic and gluten-free”}.$$

Define a binary operation $* : E \times E \to E$ (a "bundle" of tags) on the relevant parameters by

$$e_{\text{Bio}} * e_{\text{GF}} \;=\; e_{\text{BioGF}}, \qquad e_{\text{GF}} * e_{\text{Bio}} \;=\; e_{\text{BioGF}},$$

and set $x * x = x$ for $x \in A$ (idempotence on $A$).

Define a soft mapping $F : A \to \mathcal{P}(U)$ by listing products that satisfy each tag:

$$F(e_{\text{Bio}}) = \{p_1, p_2, p_4\}, \qquad F(e_{\text{GF}}) = \{p_2, p_3, p_5\}, \qquad F(e_{\text{BioGF}}) = \{p_2\}.$$

Then

$$F(e_{\text{Bio}}) \cap F(e_{\text{GF}}) = \{p_2\} \subseteq F(e_{\text{BioGF}}),$$

so the intersectional condition $F(x) \cap F(y) \subseteq F(x * y)$ holds for $x = e_{\text{Bio}}$ and $y = e_{\text{GF}}$ (and similarly for the symmetric order).

Interpretation: the combined tag parameter $e_{\text{BioGF}}$ must include every product that is both organic and gluten-free, ensuring consistency of the tag-bundling rule.

## 2.26 $N$-soft Sets

An $N$-soft set assigns to each object, for each parameter, a discrete grade from $\{0, \ldots, N-1\}$, thereby yielding graded approximations [119–122]. Related notions include *$N$-HyperSoft sets* [123, 124].

**Definition 2.26.1** ($N$-soft set)**.** [119, 120] Let $U$ be a nonempty universe of discourse and let $E$ be a nonempty set of parameters. Fix an integer $N \geq 2$ and let

$$G_N := \{0, 1, \ldots, N-1\}$$

be a set of (ordered) grades. Let $A \subseteq E$ be nonempty.

A triple $(F, A, N)$ is called an *$N$-soft set* on $U$ (with parameter set $A$) if

$$F : A \longrightarrow \mathcal{P}(U \times G_N)$$

satisfies the *uniqueness-of-grade* condition: for every $a \in A$ and every $u \in U$, there exists a unique $r \in G_N$ such that

$$(u, r) \in F(a).$$

Equivalently, for each $a \in A$ the set $F(a) \subseteq U \times G_N$ is the graph of a unique function

$$g_a : U \longrightarrow G_N, \qquad g_a(u) = r \iff (u, r) \in F(a),$$

and hence the whole $N$-soft set can be identified with a single *information function*

$$g : A \times U \longrightarrow G_N, \qquad g(a, u) := g_a(u).$$

**Example 2.26.2** (Real-life example of an $N$-soft set: grading students by discrete performance levels)**.** Let $U$ be a set of students in a class:

$$U = \{s_1, s_2, s_3, s_4, s_5\}.$$

Let $E$ be a set of evaluation parameters and take

$$A = \{a_{\text{Math}}, a_{\text{Prog}}\} \subseteq E,$$

where $a_{\text{Math}}$ = "mathematics" and $a_{\text{Prog}}$ = "programming".

Fix $N = 5$ and hence

$$G_5 = \{0, 1, 2, 3, 4\},$$

interpreted as ordered grades

$$0 = \text{very poor} \prec 1 = \text{poor} \prec 2 = \text{average} \prec 3 = \text{good} \prec 4 = \text{excellent}.$$

Define the information function $g : A \times U \to G_5$ by discrete rubric-based assessments:

| $g(a, u)$ | $a_{\text{Math}}$ | $a_{\text{Prog}}$ |
|---|---|---|
| $s_1$ | 4 | 3 |
| $s_2$ | 2 | 4 |
| $s_3$ | 1 | 2 |
| $s_4$ | 3 | 1 |
| $s_5$ | 0 | 2 |

Equivalently, define $F : A \to \mathcal{P}(U \times G_5)$ by taking graphs of the grade functions $g_a(u) := g(a, u)$:

$$F(a_{\text{Math}}) = \{(s_1, 4), (s_2, 2), (s_3, 1), (s_4, 3), (s_5, 0)\},$$

$$F(a_{\text{Prog}}) = \{(s_1, 3), (s_2, 4), (s_3, 2), (s_4, 1), (s_5, 2)\}.$$

For each $a \in A$ and $u \in U$, there is a unique $r \in G_5$ such that $(u, r) \in F(a)$, so $(F, A, 5)$ is an $N$-soft set on $U$.

Interpretation: each student receives a *discrete* grade for each parameter (subject), and the resulting $N$-soft set supports graded decision rules such as selecting students with $g(a_{\text{Prog}}, u) \geq 3$.

## 2.27 $n$-ary soft set

An $n$-ary soft set maps each parameter to an $n$-tuple of subsets over multiple universes, enabling multi-attribute approximations.

**Definition 2.27.1** (Binary soft set)**.** [125, 126] Let $U_1$ and $U_2$ be two nonempty universe sets, and let $E$ be a set of parameters. Fix a nonempty parameter subset $A \subseteq E$. A *binary soft set* over $(U_1, U_2)$ (with parameter set $A$) is a pair $(F, A)$, where

$$F : A \longrightarrow \mathcal{P}(U_1) \times \mathcal{P}(U_2).$$

For each $e \in A$, we write

$$F(e) = (X_e, Y_e), \qquad X_e \subseteq U_1, \quad Y_e \subseteq U_2.$$

Equivalently, $(F, A)$ can be identified with two ordinary soft sets $(F_1, A)$ over $U_1$ and $(F_2, A)$ over $U_2$, where $F_1(e) := X_e$ and $F_2(e) := Y_e$ for all $e \in A$.

**Definition 2.27.2** ($n$-ary soft set)**.** Let $n \in \mathbb{N}$ with $n \geq 2$, let $U_1, \dots, U_n$ be nonempty universe sets, and let $E$ be a set of parameters. Fix a nonempty parameter subset $A \subseteq E$. An *$n$-ary soft set* over $(U_1, \dots, U_n)$ (with parameter set $A$) is a pair $(F, A)$, where

$$F : A \longrightarrow \prod_{i=1}^{n} \mathcal{P}(U_i).$$

Thus, for each parameter $e \in A$ we have an $n$-tuple

$$F(e) = (X_e^{(1)}, X_e^{(2)}, \dots, X_e^{(n)}), \qquad X_e^{(i)} \subseteq U_i \quad (i = 1, \dots, n).$$

Equivalently, $(F, A)$ is uniquely determined by an $n$-tuple of (ordinary) soft sets

$$(F_i, A) \ \text{ over } U_i \quad (i = 1, \dots, n),$$

via the component maps $F_i : A \to \mathcal{P}(U_i)$ defined by

$$F_i(e) := X_e^{(i)} \qquad (e \in A, \ i = 1, \dots, n),$$

so that $F(e) = (F_1(e), \dots, F_n(e))$ for all $e \in A$.

## 2.28 Linguistic Soft Set

A linguistic soft set maps parameters to subsets of the universe by using ordered linguistic terms, thereby enabling qualitative evaluations instead of numeric membership degrees. Linguistic hypersoft sets have been studied in the literature [127, 128]; here we rewrite the concept in the (classical) *linguistic soft set* form. Related models have also been investigated in various contexts [129–133].

**Definition 2.28.1** (Linguistic Soft Set)**.** [127, 128] Let $\Omega$ be a finite universe of objects (e.g., rural health service centers), and let $E$ be a nonempty set of parameters (criteria). Fix a nonempty subset $A \subseteq E$ of parameters to be used.

For each parameter $e \in A$, let $\Upsilon_e$ be a finite, strictly ordered set of linguistic values,

$$\Upsilon_e = \{\kappa_{e,1}, \kappa_{e,2}, \ldots, \kappa_{e,m_e}\}, \qquad \kappa_{e,1} \prec \kappa_{e,2} \prec \cdots \prec \kappa_{e,m_e},$$

(where, for example, $\kappa_{e,1}$ = "very low" and $\kappa_{e,m_e}$ = "very high"). Let

$$\Upsilon := \bigsqcup_{e \in A} \Upsilon_e$$

denote the disjoint union of the linguistic term sets.

A *linguistic soft set* (LSS) over $\Omega$ with parameter set $A$ is a pair $(\Gamma, A)$, where

$$\Gamma : A \longrightarrow \mathcal{P}(\Omega) \times \Upsilon$$

is a mapping such that, for each $e \in A$, there exists a linguistic term $\kappa_e \in \Upsilon_e$ with

$$\Gamma(e) = (\Gamma_e, \kappa_e), \qquad \Gamma_e \in \mathcal{P}(\Omega), \quad \kappa_e \in \Upsilon_e.$$

Equivalently, one may view an LSS as an annotated soft set

$$(\Gamma, A) = \Big\{ (e, \kappa_e, \Gamma_e) \;\Big|\; e \in A \Big\},$$

where $\Gamma_e$ is the subset of objects described (or selected) by the linguistic evaluation $\kappa_e$ under the criterion $e$.

In applications, the annotation $\kappa_e$ is determined by experts or data-driven rules, and $\Gamma_e$ collects the objects in $\Omega$ that satisfy the parameter $e$ at the linguistic level $\kappa_e$.

**Example 2.28.2** (Real-life example of a Linguistic Soft Set: rating restaurants by linguistic service/price levels)**.** Let $\Omega$ be a small set of restaurants:

$$\Omega = \{r_1, r_2, r_3, r_4, r_5\}.$$

Let $E$ be a set of evaluation criteria and take

$$A = \{e_{\mathrm{Srv}}, e_{\mathrm{Price}}\} \subseteq E,$$

where $e_{\mathrm{Srv}}$ = "service quality" and $e_{\mathrm{Price}}$ = "price level".

For each $e \in A$, fix an ordered set of linguistic values:

$$\Upsilon_{e_{\mathrm{Srv}}} = \{\text{poor} \prec \text{fair} \prec \text{good} \prec \text{excellent}\}, \qquad \Upsilon_{e_{\mathrm{Price}}} = \{\text{cheap} \prec \text{moderate} \prec \text{expensive}\}.$$

Let $\Upsilon = \Upsilon_{e_{\mathrm{Srv}}} \sqcup \Upsilon_{e_{\mathrm{Price}}}$.

Define $\Gamma : A \to \mathcal{P}(\Omega) \times \Upsilon$ by assigning to each criterion a linguistic label together with the subset of restaurants that match that label (according to aggregated reviews):

$$\Gamma(e_{\mathrm{Srv}}) = \big(\{r_1, r_3\},\ \text{excellent}\big), \qquad \Gamma(e_{\mathrm{Price}}) = \big(\{r_2, r_5\},\ \text{cheap}\big).$$

Equivalently, the linguistic soft set is the annotated collection

$$(\Gamma, A) = \Big\{\big(e_{\mathrm{Srv}}, \text{excellent}, \{r_1, r_3\}\big),\ \big(e_{\mathrm{Price}}, \text{cheap}, \{r_2, r_5\}\big)\Big\}.$$

Interpretation: under the criterion "service quality", restaurants $r_1$ and $r_3$ are assessed as *excellent*; under "price level", restaurants $r_2$ and $r_5$ are assessed as *cheap.* Such a linguistic soft set supports qualitative filtering (e.g., seeking restaurants with excellent service or cheap price) without introducing numeric membership degrees.

## 2.29 MetaSoft Set

A MetaSoft set is a soft set whose universe consists of soft sets, classifying families of soft sets by meta-parameters [134]. We first fix a general single-sorted, finitary *signature*

$$\Sigma \;=\; \big(\mathsf{Func},\ \mathsf{Rel},\ \mathrm{ar}_{\mathsf{Func}},\ \mathrm{ar}_{\mathsf{Rel}}\big),$$

where $\mathsf{Func}$ (resp. $\mathsf{Rel}$) is a set of function (resp. relation) symbols, and ar records arities. A (single-sorted) $\Sigma$-*structure* is

$$\mathbf{C} \;=\; \big(H,\ (f^{\mathbf{C}})_{f\in\mathsf{Func}},\ (R^{\mathbf{C}})_{R\in\mathsf{Rel}}\big),$$

with carrier $H \neq \emptyset$, interpretations $f^{\mathbf{C}} : H^m \to H$ for each $f \in \mathsf{Func}$ of arity $m$, and relations $R^{\mathbf{C}} \subseteq H^r$ for each $R \in \mathsf{Rel}$ of arity $r$. Let $\mathrm{Str}_\Sigma$ denote the class of all $\Sigma$-structures.

**Definition 2.29.1** (MetaStructure over a fixed signature)**.** (cf. [135]) Fix $\Sigma$ as above. A *MetaStructure* ("structure of structures") over $\Sigma$ is a pair

$$\mathbb{M} \;=\; \big(U,\ (\Phi_\ell)_{\ell\in\Lambda}\big),$$

where:

- $U$ is a nonempty set with $U \subseteq \mathrm{Str}_\Sigma$ (its elements are *objects* at level 0);
- for each label $\ell \in \Lambda$ of *meta-arity* $k_\ell \in \mathbb{N}$, the *meta-operation*

$$\Phi_\ell \;:\; U^{k_\ell} \longrightarrow U$$

is specified by uniform *carrier- and symbol-constructors*:

$$\Gamma_\ell : \ (\mathbf{C}_1, \ldots, \mathbf{C}_{k_\ell}) \mapsto H_\ell$$
(new carrier $H_\ell$ built functorially);
$$\forall f \in \mathsf{Func} :$$
$$f^{\Phi_\ell(\mathbf{C}_1,\ldots,\mathbf{C}_{k_\ell})} \ = \ \Lambda_\ell^f\big(f^{\mathbf{C}_1}, \ldots, f^{\mathbf{C}_{k_\ell}}\big);$$
$$\forall R \in \mathsf{Rel} :$$
$$R^{\Phi_\ell(\mathbf{C}_1,\ldots,\mathbf{C}_{k_\ell})} \ = \ \Xi_\ell^R\big(R^{\mathbf{C}_1}, \ldots, R^{\mathbf{C}_{k_\ell}}\big),$$

where $\Lambda_\ell^f$ and $\Xi_\ell^R$ are *uniform* recipes turning the symbols' interpretations on inputs into the symbol's interpretation on the output, over the new carrier $H_\ell$.

Moreover, each $\Phi_\ell$ is *isomorphism-invariant* (a.k.a. natural): if $\alpha_i : \mathbf{C}_i \cong \mathbf{D}_i$ for $1 \le i \le k_\ell$, then there is an induced isomorphism

$$\Phi_\ell(\alpha_1, \ldots, \alpha_{k_\ell}) \ : \ \Phi_\ell(\mathbf{C}_1, \ldots, \mathbf{C}_{k_\ell}) \xrightarrow{\cong} \Phi_\ell(\mathbf{D}_1, \ldots, \mathbf{D}_{k_\ell})$$

commuting with all interpretations of symbols of $\Sigma$.

**Definition 2.29.2** (MetaSoft Set)**.** Let $U$ be a nonempty universe of objects and let $S$ be a (possibly finite) set of parameters. A (crisp) soft set on $(U, S)$ is a mapping

$$\mathcal{F} : S \longrightarrow \mathcal{P}(U),$$

and we denote by

$$\mathsf{Soft}(U, S) := \{\mathcal{F} \mid \mathcal{F} : S \to \mathcal{P}(U)\}$$

the collection of all such soft sets on $(U, S)$.

Let $\Pi$ be a nonempty set of *meta-parameters.* A *MetaSoft Set on* $(U, S)$ *with meta-parameter set* $\Pi$ is a soft set over the universe $\mathsf{Soft}(U, S)$ with parameter set $\Pi$, that is, a pair

$$(\mathcal{G}, \Pi) \quad \text{where}$$
$$\mathcal{G} : \Pi \longrightarrow \mathcal{P}\big(\mathsf{Soft}(U, S)\big).$$

For each $\pi \in \Pi$, the value $\mathcal{G}(\pi) \subseteq \mathsf{Soft}(U, S)$ is interpreted as the *family of base soft sets* that satisfy the meta-criterion encoded by $\pi$.

**Example 2.29.3** (Real-life example of a MetaSoft Set: selecting suitable recommendation profiles)**.** Let $U$ be a set of customers of an online grocery service:

$$U = \{c_1, c_2, c_3, c_4\}.$$

Let $S$ be a set of product-category parameters:

$$S = \{s_{\mathrm{V}}, s_{\mathrm{G}}, s_{\mathrm{L}}\},$$

where $s_{\mathrm{V}}$ = "prefers vegan items", $s_{\mathrm{G}}$ = "prefers gluten-free items", $s_{\mathrm{L}}$ = "prefers low-sugar items".

A (base) soft set on $(U, S)$ is a map $\mathcal{F} : S \to \mathcal{P}(U)$. Consider three candidate recommendation profiles (three base soft sets) $\mathcal{F}_1, \mathcal{F}_2, \mathcal{F}_3 \in \mathsf{Soft}(U, S)$ defined by:

$$\mathcal{F}_1(s_{\mathrm{V}}) = \{c_1, c_3\}, \quad \mathcal{F}_1(s_{\mathrm{G}}) = \{c_2, c_3\}, \quad \mathcal{F}_1(s_{\mathrm{L}}) = \{c_1, c_2\},$$

$$\mathcal{F}_2(s_{\mathrm{V}}) = \{c_1, c_2, c_3\}, \quad \mathcal{F}_2(s_{\mathrm{G}}) = \{c_3\}, \quad \mathcal{F}_2(s_{\mathrm{L}}) = \{c_1\},$$

$$\mathcal{F}_3(s_{\mathrm{V}}) = \{c_4\}, \quad \mathcal{F}_3(s_{\mathrm{G}}) = \{c_2, c_4\}, \quad \mathcal{F}_3(s_{\mathrm{L}}) = \{c_2, c_3, c_4\}.$$

Thus $\{\mathcal{F}_1, \mathcal{F}_2, \mathcal{F}_3\} \subseteq \mathsf{Soft}(U, S)$ is a small universe of possible "recommendation rules" (soft-set profiles).

Now let $\Pi$ be a set of *meta-parameters* describing constraints on such profiles:

$$\Pi = \{\pi_{\mathrm{Bal}}, \pi_{\mathrm{Inc}}\},$$

where

$$\pi_{\mathrm{Bal}} = \text{“balanced coverage”}, \qquad \pi_{\mathrm{Inc}} = \text{“inclusive vegan”}.$$

Define $\mathcal{G} : \Pi \to \mathcal{P}(\mathsf{Soft}(U, S))$ by

$$\mathcal{G}(\pi_{\mathrm{Bal}}) = \Big\{ \mathcal{F} \in \mathsf{Soft}(U, S) \;\Big|\; |\mathcal{F}(s_{\mathrm{V}})| \geq 2, \; |\mathcal{F}(s_{\mathrm{G}})| \geq 2, \; |\mathcal{F}(s_{\mathrm{L}})| \geq 2 \Big\},$$

$$\mathcal{G}(\pi_{\mathrm{Inc}}) = \Big\{ \mathcal{F} \in \mathsf{Soft}(U, S) \;\Big|\; \mathcal{F}(s_{\mathrm{V}}) \supseteq \{c_1, c_3\} \Big\}.$$

Then $(\mathcal{G}, \Pi)$ is a MetaSoft Set: for each meta-parameter $\pi \in \Pi$, $\mathcal{G}(\pi)$ is a family of *base soft sets* (recommendation profiles) satisfying the meta-criterion $\pi$.

For the concrete profiles above, one checks that

$$\mathcal{F}_1 \in \mathcal{G}(\pi_{\mathrm{Bal}}), \qquad \mathcal{F}_2 \in \mathcal{G}(\pi_{\mathrm{Inc}}), \qquad \mathcal{F}_3 \notin \mathcal{G}(\pi_{\mathrm{Inc}}),$$

illustrating how a MetaSoft Set can be used to select or filter candidate soft-set profiles according to higher-level design requirements.

## 2.30 Double-framed Soft Set

A double-framed soft set assigns each parameter positive and negative approximation subsets, often constrained by an operation on parameters [136–139]. Moreover, further extensions have been studied, including $N$-framed soft sets [140–142], double-framed HyperSoft sets [143, 144], and Double-Framed SuperHyperSoft Set [145, 146].

**Definition 2.30.1** (Double-Framed Soft Set)**.** Let $U$ be a universal set and let $A$ be a (nonempty) set of parameters. Assume that $A$ is equipped with a binary operation

$$* : \; A \times A \longrightarrow A.$$

A *double-framed soft set* over $U$ (with parameter set $A$) is a triple

$$\langle (\alpha, \beta); \, A \rangle,$$

where

$$\alpha : \; A \to \mathcal{P}(U) \qquad \text{and} \qquad \beta : \; A \to \mathcal{P}(U)$$

are mappings. For each $x \in A$, the set $\alpha(x)$ is interpreted as the *positive frame* and $\beta(x)$ as the *negative frame*. Moreover, the following compatibility conditions are required: for all $x, y \in A$,

$$\alpha(x * y) \; \supseteq \; \alpha(x) \cap \alpha(y), \qquad \beta(x * y) \; \subseteq \; \beta(x) \cup \beta(y).$$

**Remark 2.30.2.** If one does not intend to use an algebraic operation on the parameter set, then a double-framed soft set may be taken simply as a pair of maps $\alpha, \beta : A \to \mathcal{P}(U)$ (i.e., the triple $\langle(\alpha, \beta); A\rangle$) without imposing the above $*$-compatibility axioms.

**Example 2.30.3** (Real-life example of a double-framed soft set: loan pre-screening with positive and negative evidence)**.** Let $U$ be a set of loan applicants:

$$U = \{u_1, u_2, u_3, u_4, u_5, u_6\}.$$

Let $A$ be a set of screening parameters:

$$A = \{a_{\text{Inc}}, a_{\text{Cred}}, a_{\text{IncCred}}\}.$$

Interpret $a_{\text{Inc}}$ = "high income", $a_{\text{Cred}}$ = "good credit history", and $a_{\text{IncCred}}$ = "high income and good credit".

Define a binary operation $* : A \times A \to A$ (parameter combination) by

$$a_{\text{Inc}} * a_{\text{Cred}} = a_{\text{IncCred}}, \qquad a_{\text{Cred}} * a_{\text{Inc}} = a_{\text{IncCred}}, \qquad x * x = x \quad (x \in A).$$

Define two set-valued maps $\alpha, \beta : A \to \mathcal{P}(U)$ as follows. For each parameter $x \in A$:

- $\alpha(x)$ collects applicants with *positive evidence* supporting $x$,
- $\beta(x)$ collects applicants with *negative evidence* against $x$.

Assume the bank's initial data yield:

$$\alpha(a_{\text{Inc}}) = \{u_1, u_2, u_4\}, \qquad \beta(a_{\text{Inc}}) = \{u_3, u_5\},$$
$$\alpha(a_{\text{Cred}}) = \{u_1, u_3, u_6\}, \qquad \beta(a_{\text{Cred}}) = \{u_2, u_4\},$$

and for the combined parameter take

$$\alpha(a_{\text{IncCred}}) = \{u_1\}, \qquad \beta(a_{\text{IncCred}}) = \{u_2, u_3, u_4, u_5\}.$$

Then the compatibility conditions hold:

$$\alpha(a_{\text{IncCred}}) \supseteq \alpha(a_{\text{Inc}}) \cap \alpha(a_{\text{Cred}}) = \{u_1, u_2, u_4\} \cap \{u_1, u_3, u_6\} = \{u_1\},$$

and

$$\beta(a_{\text{IncCred}}) \subseteq \beta(a_{\text{Inc}}) \cup \beta(a_{\text{Cred}}) = \{u_3, u_5\} \cup \{u_2, u_4\} = \{u_2, u_3, u_4, u_5\}.$$

Hence $\langle(\alpha, \beta); A\rangle$ is a double-framed soft set over $U$.

Interpretation: $\alpha$ lists candidates supported by a criterion, $\beta$ lists candidates contradicted by it, and the operation $*$ combines criteria so that positive support becomes at least as strict (intersection), while negative evidence is at most as broad (union).

## 2.31 Bijective Soft Set

A bijective soft set partitions the universe into disjoint parameter blocks, covering all elements, assigning each uniquely to one parameter [147–149]. As an extension, concepts such as bijective HyperSoft sets [150–152] have also been studied.

**Definition 2.31.1** (Bijective soft set)**.** [147–149] Let $U$ be a nonempty universe and let $E$ be a set of parameters. A *soft set* over $U$ is a pair $(F, B)$, where $B \subseteq E$ is a nonempty parameter set and $F : B \to \mathcal{P}(U)$.

The soft set $(F, B)$ is called a *bijective soft set* if the family of subsets $\{F(e)\}_{e\in B}$ forms a partition of $U$, i.e.,

(B1) **Covering:** $\bigcup_{e\in B} F(e) = U$;

(B2) **Pairwise disjointness:** for all $e_1, e_2 \in B$ with $e_1 \neq e_2$, one has $F(e_1) \cap F(e_2) = \varnothing$;

(B3) **Nontrivial blocks (optional but standard for literal bijectivity):** $F(e) \neq \varnothing$ for all $e \in B$.

Equivalently, for every $u \in U$ there exists a *unique* parameter $e \in B$ such that $u \in F(e)$.

If we denote the image family by

$$Y := \{\, F(e) \mid e \in B \,\} \;\subseteq\; \mathcal{P}(U),$$

then (under (B3)) the map $F : B \to Y$ is a bijection.

**Remark 2.31.2.** Many papers state bijective soft sets using only (B1)–(B2). In that case, literal bijectivity holds after discarding any parameters with empty images: set $B^* := \{e \in B \mid F(e) \neq \varnothing\}$ and restrict $F$ to $B^*$. Then $\{F(e)\}_{e\in B^*}$ is a partition of $U$ and $F : B^* \to \{F(e) \mid e \in B^*\}$ is bijective.

**Example 2.31.3** (Real-life example of a bijective soft set: unique department assignment)**.** Let $U$ be the set of employees in a company:

$$U = \{u_1, u_2, u_3, u_4, u_5, u_6, u_7\}.$$

Let $E$ be a set of parameters and choose a nonempty subset

$$B = \{e_{\mathrm{HR}}, e_{\mathrm{ENG}}, e_{\mathrm{FIN}}, e_{\mathrm{MKT}}\} \subseteq E,$$

where each parameter denotes a department:

$$e_{\mathrm{HR}} = \text{Human Resources}, \quad e_{\mathrm{ENG}} = \text{Engineering}, \quad e_{\mathrm{FIN}} = \text{Finance}, \quad e_{\mathrm{MKT}} = \text{Marketing}.$$

Define a mapping $F : B \to \mathcal{P}(U)$ by the employees assigned to each department:

$$F(e_{\mathrm{HR}}) = \{u_1, u_6\}, \qquad F(e_{\mathrm{ENG}}) = \{u_2, u_3, u_7\},$$

$$F(e_{\mathrm{FIN}}) = \{u_4\}, \qquad F(e_{\mathrm{MKT}}) = \{u_5\}.$$

Then $\bigcup_{e \in B} F(e) = U$ (every employee belongs to some department), and the sets

$$F(e_{\mathrm{HR}}), F(e_{\mathrm{ENG}}), F(e_{\mathrm{FIN}}), F(e_{\mathrm{MKT}})$$

are pairwise disjoint (no employee belongs to two departments simultaneously). Moreover, each $F(e)$ is nonempty.

Hence $\{F(e)\}_{e \in B}$ is a partition of $U$, so $(F, B)$ is a bijective soft set. Equivalently, each employee $u \in U$ has a unique department-parameter $e \in B$ such that $u \in F(e)$.

## 2.32 Ranked Soft Set

Ranked soft sets map each parameter to an ordered partition of the universe, expressing graded satisfaction levels for uncertain decision-making. The definition of the Ranked Soft Set is described as follows [153].

**Definition 2.32.1** (Ranked Soft Set)**.** [153] Let $U$ be a nonempty finite universe and let $E$ be a set of parameters. A *ranked partition* of $U$ is an ordered collection

$$V = (V^0, V^1, \ldots, V^k)$$

of subsets of $U$ satisfying:

1. $V^0 \cup V^1 \cup \cdots \cup V^k = U$,

2. For indices $i, j$ with $0 \leq i < j \leq k$, the elements in $V^j$ are regarded as satisfying the corresponding attribute with a higher degree than those in $V^i$.

A *ranked soft set* over $U$ is a pair $(R, E)$ where

$$R : E \to \mathcal{R}(U)$$

is a mapping from the set of parameters $E$ to the family $\mathcal{R}(U)$ of all ranked partitions of $U$. That is, for each $t \in E$, $R(t)$ is a ranked partition of $U$ representing the graded evaluation of the property $t$ on the elements of $U$.

**Example 2.32.2** (Real-life example of a ranked soft set: hotel recommendation by cleanliness)**.** Let $U$ be a finite set of hotels in a city:

$$U = \{h_1, h_2, h_3, h_4, h_5\}.$$

Let $E$ be a set of evaluation parameters and consider the parameter

$$t = \text{“cleanliness”} \in E.$$

Assume we use four ordered satisfaction levels (from worst to best), so $k = 3$ and we form a ranked partition

$$R(t) = (V^0, V^1, V^2, V^3),$$

where each $V^j \subseteq U$ collects hotels assessed at rank $j$, and higher $j$ means cleaner.

For instance, based on recent inspection reports and user reviews, suppose we obtain:

$$V^0 = \{h_4\} \quad \text{(poor cleanliness)}, \qquad V^1 = \{h_2\} \quad \text{(fair cleanliness)},$$

$$V^2 = \{h_3, h_5\} \quad \text{(good cleanliness)}, \qquad V^3 = \{h_1\} \quad \text{(excellent cleanliness)}.$$

Then $V^0 \cup V^1 \cup V^2 \cup V^3 = U$, and for $0 \le i < j \le 3$, the hotels in $V^j$ are regarded as satisfying the parameter $t$ more strongly than those in $V^i$.

Define a ranked soft set $R : E \to \mathcal{R}(U)$ by specifying such a ranked partition for each parameter. In particular, the value $R(t)$ above encodes a graded evaluation of *cleanliness* on $U$, which can be used to recommend hotels by prioritizing higher-ranked blocks (e.g., selecting from $V^3$ first, then $V^2$, etc.).

## 2.33 Refined Soft Set

Refined soft sets index multiple soft sets by secondary parameters, representing several evaluators' mappings from common attributes to subsets simultaneously. Related notions include *refined neutrosophic sets* [154–156]. The definition of the Refined Soft Set is described as follows [157–159].

**Definition 2.33.1** (Refined Soft Set)**.** [157] Let $U$ be a nonempty universe, and let $E$ and $F$ be two sets of parameters with

$$E \cap F = \emptyset.$$

Let $A$ be a nonempty subset of $E$. For each parameter $b \in F$, let

$$f_b : A \to \mathcal{P}(U)$$

be a soft set over $U$; that is, for each $a \in A$, we have $f_b(a) \subseteq U$. Then the collection

$$\{\, (f_b, A) \mid b \in F \,\}$$

is called a *refined soft set* over $U$ with respect to the parameter set $A$ and the indexing set $F$. Equivalently, a refined soft set may be viewed as a mapping

$$f : F \to \{\text{soft sets over } U \text{ with parameter set } A\},$$

defined by $f(b) = f_b$ for all $b \in F$.

**Example 2.33.2** (Real-life example of a refined soft set: multi-expert product screening)**.** Let $U$ be a set of job applicants for a software engineer position:

$$U = \{u_1, u_2, u_3, u_4, u_5, u_6\}.$$

Let $E$ be a parameter set of evaluation criteria and take a nonempty subset

$$A = \{\alpha_1, \alpha_2, \alpha_3\} \subseteq E,$$

where

$$\alpha_1 = \text{“strong algorithms”}, \qquad \alpha_2 = \text{“cloud experience”}, \qquad \alpha_3 = \text{“good communication”}.$$

Let $F$ be an indexing set of *evaluators* (secondary parameters), disjoint from $E$:

$$F = \{\beta_1, \beta_2, \beta_3\}, \qquad E \cap F = \varnothing,$$

where

$$\beta_1 = \text{HR}, \qquad \beta_2 = \text{Engineering manager}, \qquad \beta_3 = \text{Senior engineer}.$$

For each evaluator $\beta \in F$, define a soft set

$$f_\beta : A \longrightarrow \mathcal{P}(U),$$

where $f_\beta(\alpha)$ is the subset of applicants judged by evaluator $\beta$ to satisfy criterion $\alpha$. For instance, suppose the assessments are:

$$f_{\beta_1}(\alpha_1) = \{u_1, u_2, u_4\}, \quad f_{\beta_1}(\alpha_2) = \{u_2, u_3, u_5\}, \quad f_{\beta_1}(\alpha_3) = \{u_1, u_3, u_6\},$$

$$f_{\beta_2}(\alpha_1) = \{u_1, u_4, u_5\}, \quad f_{\beta_2}(\alpha_2) = \{u_2, u_5, u_6\}, \quad f_{\beta_2}(\alpha_3) = \{u_1, u_2, u_6\},$$

$$f_{\beta_3}(\alpha_1) = \{u_2, u_4, u_6\}, \quad f_{\beta_3}(\alpha_2) = \{u_1, u_3, u_6\}, \quad f_{\beta_3}(\alpha_3) = \{u_1, u_4, u_5\}.$$

Then the collection

$$\{(f_\beta, A) \mid \beta \in F\}$$

is a *refined soft set* over $U$: it records, for the same primary criteria set $A$, multiple soft sets indexed by evaluators in $F$. Equivalently, it is the mapping

$$f : F \longrightarrow \{\text{soft sets over } U \text{ with parameter set } A\}, \qquad f(\beta) = f_\beta.$$

In practice, such a refined soft set supports consensus or aggregation rules (e.g., selecting applicants satisfying $\alpha_1$ according to at least two evaluators).

## 2.34 MultiSoft Set

A MultiSoft set maps multiple parameters to subsets of the universe, representing simultaneous parameterized approximations for decision-making tasks [160–162]. Related notions also include *multi-fuzzy sets* [163, 164] and *multi-neutrosophic sets* [165, 166].

**Definition 2.34.1** (Multisoft set)**.** [160–162] Let $U$ be a nonempty universe of discourse and let $E$ be a (nonempty) set of parameters. Let $A \subseteq E$ be a (nonempty) subset of parameters, and write $\mathcal{P}(U)$ for the power set of $U$. A pair $(F, A)$ is called a *multisoft set* over $U$ if

$$F : A \longrightarrow \mathcal{P}(U).$$

For each parameter $a \in A$, the subset $F(a) \subseteq U$ is the *$a$-approximation* (or $a$-value set) of the multisoft set $(F, A)$.

**Example 2.34.2** (Real-life example of a multisoft set: smartphone selection by multiple criteria)**.** Let $U$ be a finite set of smartphone models:

$$U = \{s_1, s_2, s_3, s_4, s_5, s_6\}.$$

Let $E$ be a set of decision parameters and take the nonempty subset

$$A = \{a_{\mathrm{Cam}}, a_{\mathrm{Batt}}, a_{\mathrm{Price}}\} \subseteq E,$$

where $a_{\mathrm{Cam}}$ = "good camera", $a_{\mathrm{Batt}}$ = "long battery life", and $a_{\mathrm{Price}}$ = "affordable price".

Define a mapping $F : A \to \mathcal{P}(U)$ by listing the models that satisfy each criterion according to reviews and specifications:

$$F(a_{\mathrm{Cam}}) = \{s_1, s_3, s_5\}, \qquad F(a_{\mathrm{Batt}}) = \{s_2, s_3, s_6\}, \qquad F(a_{\mathrm{Price}}) = \{s_1, s_2, s_4\}.$$

Then $(F, A)$ is a multisoft set over $U$.

Interpretation: the family $\{F(a)\}_{a\in A}$ provides multiple parameterized approximations of $U$, supporting multi-criteria selection such as choosing phones in $F(a_{\mathrm{Cam}}) \cap F(a_{\mathrm{Batt}})$ (good camera *and* long battery life).

## 2.35 GraphicSoft Set

A GraphicSoft Set further generalizes this framework by mapping each subgraph of an attribute graph to a subset of the universe, thereby embedding inter-attribute relationships into the soft-set model [167].

**Definition 2.35.1** (GraphicSoft Set)**.** [167] Let $U$ be a universe of discourse, and let $G = (V, E)$ be a graph representing a set of attributes and their relationships. A *GraphicSoft Set* is defined as a mapping

$$F : \mathcal{P}(G) \to \mathcal{P}(U),$$

which assigns to each subgraph $H \in \mathcal{P}(G)$ a subset $F(H) \subseteq U$. Intuitively, $F(H)$ represents the set of objects in $U$ that possess the combined attributes described by the subgraph $H$.

**Example 2.35.2** (Real-life example of a GraphicSoft Set: diet-oriented product search)**.** Let $U$ be a small catalog of packaged foods:

$$U = \{p_1, p_2, p_3, p_4, p_5, p_6\},$$

where $p_1$ = tofu bowl, $p_2$ = chicken salad, $p_3$ = gluten-free quinoa crackers, $p_4$ = vegan protein bar, $p_5$ = Greek yogurt, $p_6$ = oat milk.

Let $G = (V, E)$ be an *attribute graph* whose vertices represent dietary attributes:

$$V = \{v_{\mathrm{V}}, v_{\mathrm{G}}, v_{\mathrm{L}}, v_{\mathrm{P}}\},$$

where $v_{\mathrm{V}}$ = Vegan, $v_{\mathrm{G}}$ = Gluten-free, $v_{\mathrm{L}}$ = Low-sugar, $v_{\mathrm{P}}$ = High-protein. Assume the edges encode *compatibility/certification* requirements between attributes:

$$E = \big\{\{v_{\mathrm{V}}, v_{\mathrm{G}}\}, \{v_{\mathrm{V}}, v_{\mathrm{L}}\}, \{v_{\mathrm{L}}, v_{\mathrm{P}}\}\big\}.$$

For each vertex $v \in V$, define the set of products having attribute $v$:

$$S(v) \subseteq U,$$

and for each edge $\{v, w\} \in E$, define the set of products satisfying the *joint requirement* (e.g., certified compatible) for $(v, w)$:

$$C(v, w) \subseteq U.$$

For concreteness, take

$$S(v_{\mathrm{V}}) = \{p_1, p_4, p_6\}, \quad S(v_{\mathrm{G}}) = \{p_1, p_3, p_4\}, \quad S(v_{\mathrm{L}}) = \{p_3, p_4, p_5\}, \quad S(v_{\mathrm{P}}) = \{p_1, p_2, p_4, p_5\},$$

and

$$C(v_{\mathrm{V}}, v_{\mathrm{G}}) = \{p_1, p_4\}, \quad C(v_{\mathrm{V}}, v_{\mathrm{L}}) = \{p_4\}, \quad C(v_{\mathrm{L}}, v_{\mathrm{P}}) = \{p_4, p_5\}.$$

Define a mapping $F : \mathcal{P}(G) \to \mathcal{P}(U)$ as follows. For any subgraph $H = (V(H), E(H))$ of $G$, set

$$F(H) := \Big( \bigcap_{v \in V(H)} S(v) \Big) \cap \Big( \bigcap_{\{v,w\} \in E(H)} C(v, w) \Big),$$

with the convention that the intersection over an empty family equals $U$.

Then $F(H)$ returns the products satisfying the attributes in $V(H)$ together with the relational constraints in $E(H)$. For instance:

1. If $H_1$ has $V(H_1) = \{v_{\mathrm{V}}, v_{\mathrm{G}}\}$ and $E(H_1) = \{\{v_{\mathrm{V}}, v_{\mathrm{G}}\}\}$, then

$$F(H_1) = S(v_{\mathrm{V}}) \cap S(v_{\mathrm{G}}) \cap C(v_{\mathrm{V}}, v_{\mathrm{G}}) = \{p_1, p_4\}.$$

2. If $H_2$ has $V(H_2) = \{v_{\mathrm{V}}, v_{\mathrm{L}}, v_{\mathrm{P}}\}$ and $E(H_2) = \{\{v_{\mathrm{V}}, v_{\mathrm{L}}\}, \{v_{\mathrm{L}}, v_{\mathrm{P}}\}\}$, then

$$F(H_2) = S(v_{\mathrm{V}}) \cap S(v_{\mathrm{L}}) \cap S(v_{\mathrm{P}}) \cap C(v_{\mathrm{V}}, v_{\mathrm{L}}) \cap C(v_{\mathrm{L}}, v_{\mathrm{P}}) = \{p_4\}.$$

Thus $(F, G)$ is a GraphicSoft Set modeling diet-oriented retrieval where edges encode compatibility/certification relations among attributes.

For reference, a comparison between Soft Sets and GraphicSoft Sets is provided in Table 2.6.

A related concept is that one can define a *DAG-soft set* as follows. A directed acyclic graph (DAG) is a directed graph containing no directed cycles, enabling topological ordering of vertices for computation [168–171].

**Definition 2.35.3** (DAGSoft set (acyclic-hierarchical parameter soft set))**.** Let $U$ be a nonempty universe and let $D = (A, \to)$ be a finite directed acyclic graph (DAG) of parameters. A *DAGSoft set* over $U$ is a soft set $(F, A)$, $F : A \to \mathcal{P}(U)$, satisfying the hierarchical coherence rule

$$a \to b \implies F(b) \subseteq F(a) \qquad (a, b \in A).$$

Table 2.6: Concise comparison between Soft Sets and GraphicSoft Sets over a universe $U$.

| **Aspect** | **Soft Set** | **GraphicSoft Set** |
|---|---|---|
| Universe | Fixed universe of discourse $U$. | Same: fixed universe $U$. |
| Parameter carrier | A (usually finite) parameter set $S \subseteq A$ (attributes). | An attribute *graph* $G = (V, E)$ encoding attributes ($V$) and their relations ($E$). |
| Indexing domain of the map | Parameters $e \in S$. | Subgraphs $H \in \mathrm{Sub}(G)$ (or equivalently $\mathcal{P}(G)$). |
| Basic data (definition) | A map $F : S \to \mathcal{P}(U)$; the soft set is $(F, S)$. | A map $F : \mathrm{Sub}(G) \to \mathcal{P}(U)$; the structure is $(F, G)$. |
| What the "parameter" means | A single attribute/criterion $e$ (treated independently unless combined externally). | A *pattern of interacting attributes*: a subgraph $H$ (vertices + edges) represents a chosen combination together with their relationships. |
| How attribute-relations are represented | Not represented intrinsically (no edges/constraints among parameters in the basic model). | Represented intrinsically via $E(H)$: edges record dependency, compatibility, adjacency, etc. |
| Admissible combinations | Typically handled by taking finite families of parameters (e.g., decision rules using multiple $e$'s), but this is *external* to $F$. | Built-in: any subgraph $H$ is an admissible "combined parameter"; $F(H)$ directly models objects satisfying the combined/related attributes encoded by $H$. |
| Granularity / expressiveness | "Unary" parameterization: $F$ assigns sets to individual parameters. | "Structured" parameterization: $F$ assigns sets to structured parameter-objects (subgraphs), capturing multi-attribute interactions explicitly. |
| Model size (typical) | Depends on $\|S\|$ evaluations of $F(e)$. | Potentially much larger: requires values for many subgraphs (in worst case exponential in $\|V\| + \|E\|$). |
| Natural reduction / embedding | Base model. | Contains soft-set-like information by restricting to subgraphs representing single attributes (e.g., isolated-vertex subgraphs). Conversely, a soft set can be viewed as a degenerate case where only "atomic" subgraphs are used. |
| Typical use-cases | Decision making with attribute-wise approximations; uncertainty via parameterized subsets. | Decision/knowledge modeling where *relations among attributes matter* (dependencies, synergies, conflicts), and where one wants selections indexed by attribute-interaction patterns. |

**Example 2.35.4** (Real-life example of a DAGSoft set: IT helpdesk ticket taxonomy with hierarchical parameters)**.** Let $U$ be a finite set of IT helpdesk tickets:

$$U = \{t_1, t_2, t_3, t_4, t_5, t_6, t_7\}.$$

Consider a parameter DAG $D = (A, \to)$ describing an issue taxonomy:

$$A = \{a_{\mathrm{Acc}}, a_{\mathrm{Conn}}, a_{\mathrm{WiFi}}, a_{\mathrm{VPN}}, a_{\mathrm{Auth}}, a_{\mathrm{SSO}}\},$$

where the intended meanings are

$$a_{\mathrm{Acc}} = \text{"access problem"}, \quad a_{\mathrm{Conn}} = \text{"connectivity problem"},$$

$$a_{\mathrm{WiFi}} = \text{"Wi-Fi problem"}, \quad a_{\mathrm{VPN}} = \text{"VPN problem"},$$

$$a_{\mathrm{Auth}} = \text{"authentication problem"}, \quad a_{\mathrm{SSO}} = \text{"SSO problem"}.$$

Let the directed edges encode refinement (child is more specific):

$$a_{\mathrm{Acc}} \to a_{\mathrm{Conn}}, \qquad a_{\mathrm{Acc}} \to a_{\mathrm{Auth}}, \qquad a_{\mathrm{Conn}} \to a_{\mathrm{WiFi}},$$

$$a_{\text{Conn}} \to a_{\text{VPN}}, \qquad a_{\text{Auth}} \to a_{\text{SSO}}.$$

This digraph is acyclic (a DAG).

Define a soft mapping $F : A \to \mathcal{P}(U)$ by

$$\begin{aligned} F(a_{\text{Acc}}) &= \{t_1, t_2, t_3, t_4, t_5, t_6, t_7\}, \\ F(a_{\text{Conn}}) &= \{t_1, t_2, t_4, t_6\}, \\ F(a_{\text{WiFi}}) &= \{t_1, t_2\}, \\ F(a_{\text{VPN}}) &= \{t_4, t_6\}, \\ F(a_{\text{Auth}}) &= \{t_3, t_5, t_7\}, \\ F(a_{\text{SSO}}) &= \{t_5, t_7\}. \end{aligned}$$

Then the hierarchical coherence rule in Definition 2.35.3 holds, because each directed edge $a \to b$ satisfies $F(b) \subseteq F(a)$; for instance,

$$F(a_{\text{WiFi}}) \subseteq F(a_{\text{Conn}}) \subseteq F(a_{\text{Acc}}), \qquad F(a_{\text{SSO}}) \subseteq F(a_{\text{Auth}}) \subseteq F(a_{\text{Acc}}).$$

Hence $(F, A)$ is a DAGSoft set over $U$, representing a parameterized classification of tickets in which more specific issue-types select subsets of the tickets selected by their broader parent types.

## 2.36 CycleSoft Set

A CycleSoft Set extends Soft Sets by organizing parameters in a cycle graph, mapping cycle subgraphs to subsets of a universal set for structured decision-making [172].

**Definition 2.36.1** (CycleSoft Set)**.** [172] Let $U$ be a universal set and let $C = (A, E_C)$ be a cycle graph, where $A$ is a set of parameters arranged in a cycle and

$$E_C = \{(a_i, a_{i+1}) \mid a_i, a_{i+1} \in A\} \cup \{(a_n, a_1)\}$$

describes the cyclic adjacency among the parameters. Define the power set of $C$ as

$$\mathcal{P}(C) = \{H \mid H \text{ is a subgraph of } C\}.$$

A *CycleSoft Set* is a mapping

$$F : \mathcal{P}(C) \to \mathcal{P}(U),$$

where for each subgraph $H \in \mathcal{P}(C)$, $F(H) \subseteq U$ represents the set of objects associated with the combination of parameters corresponding to $H$. A common aggregation is to define, for each $H$,

$$F(H)(x) = \bigcap_{a \in V(H)} f(a)(x),$$

with $f(a) : U \to [0, 1]$ (or characteristic functions in the crisp case).

**Example 2.36.2** (Real-life example of a CycleSoft Set: selecting restaurants by cyclic service attributes)**.** Let $U$ be a set of restaurants:

$$U = \{r_1, r_2, r_3, r_4, r_5, r_6\}.$$

Consider four service-related parameters arranged in a cycle:

$$A = \{a_1, a_2, a_3, a_4\},$$

where

$$a_1 = \text{“good food”}, \quad a_2 = \text{“good service”}, \quad a_3 = \text{“clean”}, \quad a_4 = \text{“good value”}.$$

Let $C = (A, E_C)$ be the cycle graph with edges

$$E_C = \{(a_1, a_2), (a_2, a_3), (a_3, a_4), (a_4, a_1)\}.$$

Let $\mathcal{P}(C)$ denote the family of all subgraphs of $C$.

First define a (crisp) soft mapping $f : A \to \mathcal{P}(U)$ by

$$f(a_1) = \{r_1, r_2, r_5\}, \quad f(a_2) = \{r_1, r_3, r_5, r_6\}, \quad f(a_3) = \{r_2, r_3, r_4, r_5\}, \quad f(a_4) = \{r_1, r_2, r_4\}.$$

For a subgraph $H \in \mathcal{P}(C)$, define the CycleSoft mapping $F : \mathcal{P}(C) \to \mathcal{P}(U)$ by

$$F(H) := \bigcap_{a \in V(H)} f(a),$$

with the convention $F(\varnothing) = U$.

For example, let $H_1$ be the path subgraph with vertices $\{a_1, a_2, a_3\}$ and edges $\{(a_1, a_2), (a_2, a_3)\}$. Then

$$F(H_1) = f(a_1) \cap f(a_2) \cap f(a_3) = \{r_1, r_2, r_5\} \cap \{r_1, r_3, r_5, r_6\} \cap \{r_2, r_3, r_4, r_5\} = \{r_5\}.$$

Thus $r_5$ is the (unique) restaurant satisfying the combined cyclic attribute pattern “good food $\to$ good service $\to$ clean”.

Hence $F$ is a CycleSoft Set on $U$ indexed by subgraphs of the cycle $C$.

## 2.37 ClusterSoft Set

A ClusterSoft Set groups multiple Soft Sets, capturing relationships among clustered attributes and mapping them to subsets of a universal set for decision modeling [172].

**Definition 2.37.1** (ClusterSoft Set)**.** [172] Let $\{F_i\}_{i \in I}$ be a finite family of soft sets over a universe $U$, where each soft set $F_i$ is a mapping

$$F_i : A_i \to \mathcal{P}(U)$$

for some set of attributes $A_i$. Suppose the index set $I$ is partitioned into clusters $\{C_j\}_{j \in J}$ with each $C_j \subseteq I$ and $C_j \cap C_k = \emptyset$ for $j \neq k$. A *ClusterSoft Set* is defined as a mapping

$$G : \{C_j : j \in J\} \to \mathcal{P}(U)$$

given by

$$G(C_j) = \bigcup_{i \in C_j} F_i^*(A_i),$$

where $F_i^*(A_i)$ denotes the set of objects in $U$ associated with soft set $F_i$ (possibly after appropriate aggregation or normalization). The union is taken in the usual set-theoretic sense.

**Example 2.37.2** (Real-life example of a ClusterSoft Set: grouping customer segments from multiple marketing soft sets)**.** Let $U$ be a set of customers of an online shop:

$$U = \{u_1, u_2, u_3, u_4, u_5, u_6, u_7, u_8\}.$$

Assume the marketing team maintains a finite family of soft sets $\{F_i\}_{i\in I}$, each capturing customers associated with certain attributes in a specific campaign. Let

$$I = \{1, 2, 3, 4\}.$$

Define, for each $i \in I$, a parameter set $A_i$ and a soft mapping $F_i : A_i \to \mathcal{P}(U)$.

**Campaign 1 (sports campaign).** Let $A_1 = \{a_{\text{Run}}, a_{\text{Gym}}\}$ and suppose

$$F_1(a_{\text{Run}}) = \{u_1, u_3, u_6\}, \qquad F_1(a_{\text{Gym}}) = \{u_2, u_3, u_7\}.$$

**Campaign 2 (outdoor campaign).** Let $A_2 = \{a_{\text{Hike}}\}$ and suppose

$$F_2(a_{\text{Hike}}) = \{u_1, u_4, u_6, u_8\}.$$

**Campaign 3 (family campaign).** Let $A_3 = \{a_{\text{Kids}}\}$ and suppose

$$F_3(a_{\text{Kids}}) = \{u_2, u_5, u_7\}.$$

**Campaign 4 (premium campaign).** Let $A_4 = \{a_{\text{Premium}}\}$ and suppose

$$F_4(a_{\text{Premium}}) = \{u_3, u_4, u_8\}.$$

For each soft set $F_i$, define the associated customer set

$$F_i^*(A_i) \;:=\; \bigcup_{a\in A_i} F_i(a) \;\subseteq\; U.$$

Then

$$F_1^*(A_1) = \{u_1, u_2, u_3, u_6, u_7\}, \quad F_2^*(A_2) = \{u_1, u_4, u_6, u_8\},$$
$$F_3^*(A_3) = \{u_2, u_5, u_7\}, \quad F_4^*(A_4) = \{u_3, u_4, u_8\}.$$

Now partition the index set $I$ into clusters (segments):

$$C_{\text{Active}} = \{1, 2\}, \qquad C_{\text{Lifestyle}} = \{3, 4\}, \qquad \{C_{\text{Active}}, C_{\text{Lifestyle}}\} \text{ is a partition of } I.$$

Define the ClusterSoft mapping $G : \{C_{\text{Active}}, C_{\text{Lifestyle}}\} \to \mathcal{P}(U)$ by

$$G(C) = \bigcup_{i\in C} F_i^*(A_i).$$

Hence

$$G(C_{\text{Active}}) = F_1^*(A_1) \cup F_2^*(A_2) = \{u_1, u_2, u_3, u_4, u_6, u_7, u_8\},$$
$$G(C_{\text{Lifestyle}}) = F_3^*(A_3) \cup F_4^*(A_4) = \{u_2, u_3, u_4, u_5, u_7, u_8\}.$$

Thus $G$ is a ClusterSoft Set: each cluster aggregates the customers associated with the soft sets belonging to that cluster, producing unified target segments for marketing actions.

## 2.38 Soft Expert Set

A *soft expert set* incorporates expert participation and their opinions: parameter–expert–opinion triples are mapped to subsets of $U$, equivalently yielding a mapping from $E \times X \times O$ into $\mathcal{P}(U)$ [173–176]. As extensions, *fuzzy soft expert sets* [177–179], *intuitionistic fuzzy soft expert sets* [180–182], *neutrosophic soft expert sets* [183–185], Soft MultiExpert Sets [186, 187], and *HyperSoft expert sets* [25, 152, 188, 189] are also known.

**Definition 2.38.1** (Soft Expert Set)**.** [173–176] Let $U$ be a universe, $E$ a set of parameters, $X$ a set of experts, and $O = \{0, 1\}$ a set of opinions. Put $Z := E \times X \times O$ and let $A \subseteq Z$ be nonempty. A *soft expert set* on $U$ is a pair $(G, A)$ where

$$G \colon A \longrightarrow \mathcal{P}(U)$$

assigns to each triple $\alpha = (e, x, o) \in A$ a subset $G(\alpha) \subseteq U$. Thus $G(\alpha)$ collects the elements of $U$ supported by expert $x$, with opinion $o$, under parameter $e$.

**Example 2.38.2** (Real-life example of a soft expert set: smartphone selection by multiple experts)**.** Let $U$ be a set of smartphone models:

$$U = \{s_1, s_2, s_3, s_4, s_5\}.$$

Let $E$ be a set of evaluation parameters:

$$E = \{e_{\text{Cam}}, e_{\text{Batt}}, e_{\text{Price}}\},$$

where $e_{\text{Cam}}$ = "good camera", $e_{\text{Batt}}$ = "long battery life", and $e_{\text{Price}}$ = "affordable price".

Let $X$ be a set of experts:

$$X = \{x_1, x_2, x_3\},$$

where $x_1$ is a reviewer, $x_2$ is an engineer, and $x_3$ is a budget-conscious user. Let $O = \{0, 1\}$ be the set of opinions, where 1 means *approve* and 0 means *disapprove.* Set $Z = E \times X \times O$ and choose $A = Z$ (i.e., all expert–parameter–opinion triples are allowed).

Define a mapping $G : A \to \mathcal{P}(U)$ by listing, for each triple $(e, x, o)$, the phones receiving opinion $o$ from expert $x$ under parameter $e$. For example, suppose:

$$G(e_{\text{Cam}}, x_1, 1) = \{s_1, s_3\}, \qquad G(e_{\text{Cam}}, x_1, 0) = \{s_2, s_4, s_5\},$$

$$G(e_{\text{Batt}}, x_2, 1) = \{s_2, s_3, s_5\}, \qquad G(e_{\text{Batt}}, x_2, 0) = \{s_1, s_4\},$$

$$G(e_{\text{Price}}, x_3, 1) = \{s_2, s_4\}, \qquad G(e_{\text{Price}}, x_3, 0) = \{s_1, s_3, s_5\},$$

and define $G(e, x, o)$ similarly for all remaining triples in $A$.

Then $(G, A)$ is a soft expert set on $U$: each triple $(e, x, o)$ determines the subset of smartphones supported (if $o = 1$) or rejected (if $o = 0$) by expert $x$ with respect to parameter $e$. This structure supports aggregation rules such as selecting phones approved by a majority of experts under key parameters.

## 2.39 Soft Rough Set

A soft rough set approximates a subset using a soft set: lower includes certainly contained elements, upper includes possibly related elements [16, 190]. Related notions such as *fuzzy soft rough sets* [191–193] and *neutrosophic soft rough sets* [190, 194, 195] are also known.

**Definition 2.39.1** (Soft Rough Set)**.** [16, 190] Let $S = (F, A)$ be a soft set over a universe $U$, and let $P = (U, S)$ be the corresponding *soft approximation space.* For $X \subseteq U$, the *soft $P$-lower* and *soft $P$-upper* approximations of $X$ are defined by

$$\underline{\mathrm{apr}}_P(X) = \{ u \in U : \exists a \in A \text{ such that } u \in F(a) \subseteq X \},$$
$$\overline{\mathrm{apr}}_P(X) = \{ u \in U : \exists a \in A \text{ such that } u \in F(a) \text{ and } F(a) \cap X \neq \emptyset \}.$$

The pair $(\underline{\mathrm{apr}}_P(X), \overline{\mathrm{apr}}_P(X))$ is called the *soft rough set* of $X$ with respect to $P$.

**Example 2.39.2** (Real-life example of a soft rough set: identifying "reliable suppliers" from parameterized checklists)**.** Let $U$ be a set of candidate suppliers:

$$U = \{s_1, s_2, s_3, s_4, s_5\}.$$

Let $A$ be a set of audit parameters (checklists):

$$A = \{a_{\mathrm{ISO}}, a_{\mathrm{OnTime}}, a_{\mathrm{LowDefect}}\}.$$

Define a soft set $S = (F, A)$ over $U$ by

$$F(a_{\mathrm{ISO}}) = \{s_1, s_2, s_4\} \quad \text{(ISO-certified suppliers)},$$
$$F(a_{\mathrm{OnTime}}) = \{s_1, s_3, s_4\} \quad \text{(historically on-time suppliers)},$$
$$F(a_{\mathrm{LowDefect}}) = \{s_2, s_4, s_5\} \quad \text{(low defect-rate suppliers)}.$$

Let $P = (U, S)$ be the induced soft approximation space.

Suppose the procurement team proposes a target set

$$X = \{s_1, s_4\} \subseteq U$$

of *preferred suppliers* (e.g., shortlisted by a manager).

**Lower approximation.** An element $u \in U$ belongs to $\underline{\mathrm{apr}}_P(X)$ if there exists a parameter $a \in A$ such that $u \in F(a) \subseteq X$. Here,

$$F(a_{\mathrm{ISO}}) = \{s_1, s_2, s_4\} \nsubseteq X, \quad F(a_{\mathrm{OnTime}}) = \{s_1, s_3, s_4\} \nsubseteq X, \quad F(a_{\mathrm{LowDefect}}) = \{s_2, s_4, s_5\} \nsubseteq X.$$

Thus no $F(a)$ is contained in $X$, and hence

$$\underline{\mathrm{apr}}_P(X) = \varnothing.$$

**Upper approximation.** An element $u \in U$ belongs to $\overline{\mathrm{apr}}_P(X)$ if there exists $a \in A$ such that $u \in F(a)$ and $F(a) \cap X \neq \varnothing$. Since

$$F(a_{\mathrm{ISO}}) \cap X = \{s_1, s_4\} \neq \varnothing, \quad F(a_{\mathrm{OnTime}}) \cap X = \{s_1, s_4\} \neq \varnothing, \quad F(a_{\mathrm{LowDefect}}) \cap X = \{s_4\} \neq \varnothing,$$

every supplier appearing in at least one of these $F(a)$ sets is possibly compatible with $X$. Therefore,

$$\overline{\mathrm{apr}}_P(X) = F(a_{\mathrm{ISO}}) \cup F(a_{\mathrm{OnTime}}) \cup F(a_{\mathrm{LowDefect}}) = \{s_1, s_2, s_3, s_4, s_5\} = U.$$

Hence the soft rough set of $X$ with respect to $P$ is

$$\big(\underline{\mathrm{apr}}_P(X), \overline{\mathrm{apr}}_P(X)\big) = (\varnothing, U).$$

Interpretation: based on these coarse checklists, no supplier is *certainly* in the preferred set, while every supplier is *possibly* related to it through at least one parameter block.

## 2.40 Weighted Soft Set

A weighted soft set maps each parameter to a universe subset and assigns a weight reflecting that parameter's relative significance [196–198].

**Definition 2.40.1** (Weighted Soft Set)**.** [198] Let $U$ be a finite universe of discourse and let $E$ be a finite set of parameters (attributes). A finite *weighted soft set* over $U$ is defined as a triple

$$(\mu, A, Q),$$

where:

- $A \subseteq E$ is a finite set of selected parameters,
- $\mu : A \to \mathcal{P}(U)$ is a mapping that assigns to each parameter $a \in A$ a subset $\mu(a) \subseteq U$, and
- $Q : A \to [0, 1]$ is a finite weight function that assigns to each parameter $a \in A$ a weight $Q(a)$, representing its relative importance.

Thus, the finite weighted soft set can be represented as

$$(\mu, A, Q) = \{(a, \mu(a), Q(a)) \mid a \in A\}.$$

**Example 2.40.2** (Real-life example of a weighted soft set: selecting apartments with weighted criteria)**.** Let $U$ be a finite set of apartments:

$$U = \{h_1, h_2, h_3, h_4, h_5\}.$$

Let $E$ be a finite set of decision parameters and choose

$$A = \{a_{\text{Rent}}, a_{\text{Comm}}, a_{\text{Safe}}\} \subseteq E,$$

where $a_{\text{Rent}}$ = "affordable rent", $a_{\text{Comm}}$ = "short commute", and $a_{\text{Safe}}$ = "safe neighborhood".

Define the soft mapping $\mu : A \to \mathcal{P}(U)$ by listing apartments that satisfy each criterion:

$$\mu(a_{\text{Rent}}) = \{h_1, h_3, h_5\}, \qquad \mu(a_{\text{Comm}}) = \{h_2, h_3, h_4\}, \qquad \mu(a_{\text{Safe}}) = \{h_1, h_2, h_4\}.$$

Assign a weight function $Q : A \to [0, 1]$ reflecting the user's priorities:

$$Q(a_{\text{Rent}}) = 0.50, \qquad Q(a_{\text{Comm}}) = 0.30, \qquad Q(a_{\text{Safe}}) = 0.20.$$

Then $(\mu, A, Q)$ is a weighted soft set over $U$.

Interpretation: affordability is the most important criterion (weight 0.50), followed by commute time and safety. A simple weighted choice score for an apartment $h \in U$ can be formed by summing the weights of parameters that accept $h$, e.g.,

$$\text{Score}(h) := \sum_{\substack{a \in A \\ h \in \mu(a)}} Q(a),$$

so that higher scores indicate better overall fit to the weighted criteria.

## 2.41 DSm-soft set

Dezert–Smarandache Theory is an information-fusion framework designed to handle uncertain, imprecise, conflicting, and potentially paradoxical evidence through generalized belief assignments [199–203]. Before defining a DSm-Number, we first recall the hyper-power set used in Dezert–Smarandache Theory [204, 205].

**Definition 2.41.1** (Hyper-Power Set)**.** [206, 207] Let $\Theta = \{\theta_1, \ldots, \theta_n\}$ be a finite frame of formal hypotheses, which need not be mutually exclusive. Under the free DSm model, its *hyper-power set* $D^\Theta$ is the finite distributive lattice generated by the symbols $\theta_1, \ldots, \theta_n$ under the formal operations $\cup$ and $\cap$, subject to the distributive lattice identities, with bottom $\varnothing$ and top $\theta_1 \cup \cdots \cup \theta_n$. Its elements are equivalence classes of formal expressions, not necessarily ordinary subsets of a set of disjoint atoms. In particular, $\theta_i \cap \theta_j$ is not forced to be $\varnothing$ merely because $i \neq j$. Model-specific integrity constraints may produce quotients of the free DSm lattice by identifying additional expressions.

Inspired by the relaxed completeness condition used for D Numbers, we introduce the following DSm-based numerical representation.

**Definition 2.41.2** (DSm-Number)**.** Let $D^\Theta$ be the hyper-power set associated with a finite frame $\Theta$. A *DSm-Number* is a mapping

$$N_{\mathrm{DSm}} : D^\Theta \longrightarrow [0, 1]$$

such that

$$N_{\mathrm{DSm}}(\varnothing) = 0, \qquad \sum_{A \in D^\Theta} N_{\mathrm{DSm}}(A) \leq 1.$$

The *completeness degree* of $N_{\mathrm{DSm}}$ is defined by

$$Q_{\mathrm{DSm}} = \sum_{A \in D^\Theta} N_{\mathrm{DSm}}(A).$$

The DSm-Number is said to be *complete* when

$$Q_{\mathrm{DSm}} = 1,$$

and *incomplete* when

$$Q_{\mathrm{DSm}} < 1.$$

If $Q_{\mathrm{DSm}} = 1$, then $N_{\mathrm{DSm}}$ takes the form of a generalized basic belief assignment defined on $D^\Theta$.

**Remark 2.41.3.** A DSm-Number replaces the ordinary power set used in the formulation of a D Number with the hyper-power set $D^\Theta$. Consequently, belief masses may be assigned explicitly not only to unions but also to intersections of hypotheses that are not required to be mutually exclusive.

**Definition 2.41.4** (DSm-Soft Set)**.** Let $\Theta$ be a finite nonempty frame and let $D^\Theta$ be its hyper-power set. Define

$$\mathsf{DSmNum}(\Theta) = \left\{ N : D^\Theta \to [0,1] \;\middle|\; N(\varnothing) = 0,\ \sum_{B \in D^\Theta} N(B) \le 1 \right\}.$$

Let $E$ be a nonempty parameter set and let $\varnothing \neq A \subseteq E$. A *DSm-Soft Set* over $\Theta$ is a pair $(F, A)$ such that

$$F : A \longrightarrow \mathsf{DSmNum}(\Theta).$$

Thus, for each parameter $e \in A$, $F(e) = N_e$ is a DSm-Number assigning masses to propositions in the hyper-power set $D^\Theta$.

**Example 2.41.5** (DSm-Soft Set)**.** Let

$$\Theta = \{\theta_1, \theta_2\}$$

be a finite frame whose hypotheses are not necessarily mutually exclusive.

Under the free DSm model, the hyper-power set is

$$D^\Theta = \{\varnothing, \theta_1, \theta_2, \theta_1 \cap \theta_2, \theta_1 \cup \theta_2\}.$$

Consider the parameter set

$$E = A = \{e_1, e_2\}.$$

Define

$$F : A \longrightarrow \mathsf{DSmNum}(\Theta)$$

by

$$F(e_1) = N_1, \qquad F(e_2) = N_2,$$

where

| $B \in D^\Theta$ | $N_1(B)$ | $N_2(B)$ |
|---|---|---|
| $\varnothing$ | 0 | 0 |
| $\theta_1$ | 0.4 | 0.1 |
| $\theta_2$ | 0 | 0.5 |
| $\theta_1 \cap \theta_2$ | 0.3 | 0.2 |
| $\theta_1 \cup \theta_2$ | 0.1 | 0.2 |

Then

$$N_1(\varnothing) = N_2(\varnothing) = 0.$$

Furthermore,

$$Q_{\mathrm{DSm}}(N_1) = 0.4 + 0.3 + 0.1 = 0.8 < 1,$$

and

$$Q_{\mathrm{DSm}}(N_2) = 0.1 + 0.5 + 0.2 + 0.2 = 1.$$

Thus, $N_1$ is an incomplete DSm-Number, whereas $N_2$ is a complete DSm-Number.

Since

$$N_1, N_2 \in \mathsf{DSmNum}(\Theta),$$

the pair

$$(F, A)$$

is a DSm-Soft Set over $\Theta$.

## 2.42 Geometric soft sets

Geometric soft sets map parameters to point-incidence subsets, capturing hyperplane or distance relations, enabling geometric realization and network analysis efficiently [208].

**Definition 2.42.1** (Geometric Soft Set)**.** Let $U \subseteq \mathbb{R}^d$ be a finite set of points in general position, and let $A \subseteq U$ be nonempty. For $r \geq 1$, write

$$\binom{A}{r} = \{B \subseteq A : |B| = r\}.$$

Let

$$E = \{e_0, e_1, \ldots, e_{|A|-1}\}.$$

A *Geometric Soft Set* on $A$ is a mapping

$$F_A : E \longrightarrow \mathcal{P}(\mathcal{P}(A))$$

such that $F_A(e_r) \subseteq \binom{A}{r+1}$, where membership

$$B \in F_A(e_r)$$

represents the incidence of the points of $B$ with an $r$-dimensional affine subspace. Moreover, the incidence structure is downward closed:

$$B \in F_A(e_r), \quad C \subseteq B, \quad |C| = s+1 \quad \Longrightarrow \quad C \in F_A(e_s).$$

**Example 2.42.2** (Geometric Soft Set)**.** Let

$$U = A = \{p_1, p_2, p_3\} \subseteq \mathbb{R}^2,$$

where

$$p_1 = (0,0), \qquad p_2 = (1,0), \qquad p_3 = (0,1).$$

The three points are noncollinear and therefore are in general position in $\mathbb{R}^2$.

Let

$$E = \{e_0, e_1, e_2\}.$$

Define

$$F_A : E \longrightarrow \mathcal{P}(\mathcal{P}(A))$$

by

$$F_A(e_0) = \big\{\{p_1\}, \{p_2\}, \{p_3\}\big\},$$

$$F_A(e_1) = \big\{\{p_1, p_2\}, \{p_1, p_3\}, \{p_2, p_3\}\big\},$$

and

$$F_A(e_2) = \big\{\{p_1, p_2, p_3\}\big\}.$$

Clearly,

$$F_A(e_r) \subseteq \binom{A}{r+1} \qquad (r = 0, 1, 2).$$

Each singleton represents a point, each pair determines an affine line, and the three noncollinear points determine a two-dimensional affine space.

Moreover, the incidence structure is downward closed.

For example,

$$\{p_1, p_2, p_3\} \in F_A(e_2)$$

implies

$$\{p_1, p_2\} \in F_A(e_1)$$

and

$$\{p_1\} \in F_A(e_0).$$

The same property holds for every nonempty subset of each selected incidence.

Therefore, $F_A$ is a Geometric Soft Set.

## 2.43 Multiparameterized soft sets

A multiparameterized soft set maps each multi-parameter tuple to a subset of the universe, capturing combined attribute approximations effectively simultaneously [209–211].

**Definition 2.43.1** (Multiparameterized Soft Set)**.** Let $U$ be a nonempty universe and let $\{E_i\}_{i\in I}$ be a family of pairwise disjoint parameter sets. Put

$$E = \bigcup_{i\in I} E_i,$$

and let $\varnothing \neq A \subseteq \mathcal{P}(E)$. A *Multiparameterized Soft Set* over $U$ is a pair $(F, A)$, where

$$F : A \longrightarrow \mathcal{P}(U).$$

Thus, each multiparameter $\alpha \in A$, possibly containing parameters from several $E_i$, is associated with a subset $F(\alpha) \subseteq U$.

**Example 2.43.2** (Multiparameterized Soft Set)**.** Let

$$U = \{u_1, u_2, u_3\}$$

be a universe of three laptop models.

Consider two disjoint parameter sets

$$E_1 = \{p_L, p_H\}, \qquad E_2 = \{b_L, b_S\},$$

where $p_L, p_H$ represent low and high price categories, respectively, and $b_L, b_S$ represent long and short battery-life categories, respectively.

Let

$$E = E_1 \cup E_2.$$

Define the multiparameter family

$$A = \{\{p_L, b_L\}, \{p_L, b_S\}, \{p_H, b_L\}\} \subseteq \mathcal{P}(E).$$

Consider the attribute assignments

$$H(p_L) = \{u_1, u_2\}, \qquad H(p_H) = \{u_3\},$$

and

$$H(b_L) = \{u_2, u_3\}, \qquad H(b_S) = \{u_1\}.$$

Define

$$F : A \longrightarrow \mathcal{P}(U)$$

by

$$F(\alpha) = \bigcap_{e \in \alpha} H(e).$$

Consequently,

$$F(\{p_L, b_L\}) = \{u_1, u_2\} \cap \{u_2, u_3\} = \{u_2\},$$

$$F(\{p_L, b_S\}) = \{u_1, u_2\} \cap \{u_1\} = \{u_1\},$$

and

$$F(\{p_H, b_L\}) = \{u_3\} \cap \{u_2, u_3\} = \{u_3\}.$$

Thus, every multiparameter $\alpha \in A$ is associated with a uniquely determined subset of $U$.

Therefore, $(F, A)$ is a Multiparameterized Soft Set.

## 2.44 Concave Soft sets

Concave soft sets are order-preserving: if $x \leq y$ then $F(x) \subseteq F(y)$, equivalently $F$ coincides with its closure $[[F]]$ for the given order [212–214].

**Definition 2.44.1** (Concave Soft Set)**.** Let $U$ be a nonempty universe and let $(S, \leq)$ be an ordered semigroup. A soft set $(F, S)$, with

$$F : S \longrightarrow \mathcal{P}(U),$$

is called a *Concave Soft Set* if

$$F(x) = \bigcap_{x \leq y} F(y) \qquad \text{for every } x \in S.$$

Equivalently,

$$x \leq y \quad \Longrightarrow \quad F(x) \subseteq F(y) \qquad \text{for all } x, y \in S.$$

**Example 2.44.2** (Concave Soft Set)**.** Let

$$U = \{u_1, u_2, u_3\}$$

and consider the ordered semigroup

$$(S, +, \leq) = (\mathbb{N}_0, +, \leq),$$

where $\leq$ is the usual order.

Define

$$F : S \longrightarrow \mathcal{P}(U)$$

by

$$F(n) = \{u_1, \dots, u_{\min\{n+1,3\}}\}.$$

In particular,

$$F(0) = \{u_1\},$$

$$F(1) = \{u_1, u_2\},$$

and

$$F(n) = U \qquad (n \geq 2).$$

For any $m, n \in \mathbb{N}_0$ with $n \leq m$, we have

$$\min\{n + 1, 3\} \leq \min\{m + 1, 3\}.$$

Hence,

$$F(n) \subseteq F(m).$$

Furthermore, since $n \leq n$, the family

$$\{F(m) : n \leq m\}$$

contains $F(n)$.

By monotonicity,

$$F(n) \subseteq F(m) \qquad \text{for every } m \geq n.$$

Therefore,

$$\bigcap_{n \leq m} F(m) = F(n).$$

Thus, $F$ satisfies

$$F(n) = \bigcap_{n \leq m} F(m) \qquad (n \in S).$$

Consequently, $(F, S)$ is a Concave Soft Set.

## 2.45 Incomplete Soft Set

An Incomplete Soft Set allows missing or unknown parameter-object information, extending soft sets when some evaluations are unavailable or unspecified [215–217].

**Definition 2.45.1** (Incomplete Soft Set)**.** Let $U$ be a nonempty universe, $E$ a set of parameters, and $\varnothing \neq A \subseteq E$. An *Incomplete Soft Set* over $U$ is a pair $(F, A)$, where

$$F : A \longrightarrow \{0, 1, *\}^U.$$

For $e \in A$ and $u \in U$,

$$F(e)(u) = \begin{cases} 1, & \text{if } u \text{ belongs to the approximation determined by } e, \\ 0, & \text{if } u \text{ does not belong to it}, \\ *, & \text{if the membership information is unknown.} \end{cases}$$

**Example 2.45.2** (Incomplete Soft Set)**.** Let

$$U = \{u_1, u_2, u_3\}$$

be a universe of three alternatives, and let

$$E = A = \{e_1, e_2\}$$

be a parameter set.

Define

$$F : A \longrightarrow \{0, 1, *\}^U$$

by the following table:

| | $u_1$ | $u_2$ | $u_3$ |
|---|---|---|---|
| $F(e_1)$ | 1 | 0 | $*$ |
| $F(e_2)$ | $*$ | 1 | 0 |

Thus,

$$F(e_1)(u_1) = 1, \qquad F(e_1)(u_2) = 0, \qquad F(e_1)(u_3) = *,$$

and

$$F(e_2)(u_1) = *, \qquad F(e_2)(u_2) = 1, \qquad F(e_2)(u_3) = 0.$$

The value 1 indicates confirmed membership, 0 indicates confirmed nonmembership, and $*$ indicates unknown membership information.

For example,

$$F(e_1)(u_3) = *$$

means that the membership of $u_3$ under parameter $e_1$ has not been determined.

Since every parameter is assigned a function from $U$ to $\{0, 1, *\}$, the mapping $F$ is well-defined.

Therefore, $(F, A)$ is an Incomplete Soft Set.

## 2.46 Dual Probabilistic Soft Set

A Dual Probabilistic Soft Set assigns each object a probability distribution over parameters, reversing the usual parameter-to-object probabilistic soft mapping [218, 219].

**Definition 2.46.1** (Dual Probabilistic Soft Set). Let $U$ be a nonempty finite universe, $E$ a finite set of parameters, and $\varnothing \neq A \subseteq E$. Define

$$D(A) = \left\{ p : A \to [0,1] \;\middle|\; \sum_{e \in A} p(e) = 1 \right\}.$$

A *Dual Probabilistic Soft Set* over $U$ with parameter set $A$ is a pair $(F, A)$, where

$$F : U \longrightarrow D(A).$$

Equivalently, for every $u \in U$,

$$F(u) : A \to [0,1], \qquad \sum_{e \in A} F(u)(e) = 1.$$

Thus, each object $u$ is associated with a probability distribution over the parameters in $A$.

**Example 2.46.2** (Dual Probabilistic Soft Set)**.** Let

$$U = \{u_1, u_2, u_3\}$$

be a universe of three candidate transportation routes.

Consider the parameter set

$$E = A = \{e_1, e_2, e_3\},$$

where $e_1, e_2, e_3$ represent low cost, short travel time, and high safety, respectively.

Define

$$F : U \longrightarrow D(A)$$

by

| | $e_1$ | $e_2$ | $e_3$ |
|---|---|---|---|
| $F(u_1)$ | 0.5 | 0.3 | 0.2 |
| $F(u_2)$ | 0.2 | 0.5 | 0.3 |
| $F(u_3)$ | 0.1 | 0.2 | 0.7 |

For each $u_i \in U$, the mapping

$$F(u_i) : A \longrightarrow [0, 1]$$

defines a probability distribution over the parameter set $A$.

Indeed,

$$\sum_{e \in A} F(u_1)(e) = 0.5 + 0.3 + 0.2 = 1,$$

$$\sum_{e \in A} F(u_2)(e) = 0.2 + 0.5 + 0.3 = 1,$$

and

$$\sum_{e \in A} F(u_3)(e) = 0.1 + 0.2 + 0.7 = 1.$$

Moreover, every assigned value belongs to $[0, 1]$.

Therefore,

$$F(u_i) \in D(A) \qquad (i = 1, 2, 3).$$

Consequently, $(F, A)$ is a Dual Probabilistic Soft Set.

Unlike a conventional parameter-to-object probabilistic mapping, this construction assigns each object a probability distribution over the parameters.

## 2.47 Rough Soft Set

A Rough Soft Set applies rough approximations to soft-set information, describing objects through lower and upper approximations induced by parameters [220–223].

**Definition 2.47.1** (Rough Soft Set)**.** Let $(U, R)$ be a Pawlak approximation space, where $R$ is an equivalence relation on $U$, and let $(F, A)$ be a soft set over $U$,

$$F : A \to \mathcal{P}(U).$$

For $X \subseteq U$, define

$$\underline{R}(X) = \{x \in U : [x]_R \subseteq X\}, \qquad \overline{R}(X) = \{x \in U : [x]_R \cap X \neq \varnothing\}.$$

The lower and upper rough approximations of $(F, A)$ are $(F_*, A)$ and $(F^*, A)$, respectively, where

$$F_*(e) = \underline{R}(F(e)), \qquad F^*(e) = \overline{R}(F(e)), \qquad e \in A.$$

If $(F_*, A) \neq (F^*, A)$, then $(F, A)$ is called a *Rough Soft Set*; otherwise, it is *definable*.

**Example 2.47.2** (Rough Soft Set)**.** Let

$$U = \{1, 2, 3, 4\}$$

and let $R$ be the equivalence relation whose equivalence classes are

$$[1]_R = [2]_R = \{1, 2\},$$

and

$$[3]_R = [4]_R = \{3, 4\}.$$

Consider the parameter set

$$A = \{\alpha, \beta\}$$

and define the soft set

$$F : A \longrightarrow \mathcal{P}(U)$$

by

$$F(\alpha) = \{1, 2, 3\}, \qquad F(\beta) = \{3, 4\}.$$

For parameter $\alpha$, the lower approximation is

$$\begin{aligned} F_*(\alpha) &= \underline{R}(F(\alpha)) \\ &= \{x \in U : [x]_R \subseteq \{1, 2, 3\}\} \\ &= \{1, 2\}. \end{aligned}$$

The upper approximation is

$$\begin{aligned} F^*(\alpha) &= \overline{R}(F(\alpha)) \\ &= \{x \in U : [x]_R \cap \{1, 2, 3\} \neq \varnothing\} \\ &= U. \end{aligned}$$

For parameter $\beta$, we obtain

$$F_*(\beta) = \underline{R}(\{3,4\}) = \{3,4\},$$

and

$$F^*(\beta) = \overline{R}(\{3,4\}) = \{3,4\}.$$

Thus, the lower and upper soft approximations are given by

| $e$ | $F_*(e)$ | $F^*(e)$ |
|---|---|---|
| $\alpha$ | $\{1,2\}$ | $U$ |
| $\beta$ | $\{3,4\}$ | $\{3,4\}$ |

Since

$$F_*(\alpha) \neq F^*(\alpha),$$

we have

$$(F_*, A) \neq (F^*, A).$$

Therefore, $(F, A)$ is a Rough Soft Set.

In particular, the approximation associated with $\beta$ is exact, whereas the approximation associated with $\alpha$ is not exact.

A comparison between Rough Soft Sets and Soft Rough Sets is presented in Table 2.7.

Table 2.7: Comparison of Rough Soft Sets and Soft Rough Sets.

| **Aspect** | **Rough Soft Set** | **Soft Rough Set** |
|---|---|---|
| Basic idea | Applies Pawlak rough approximations to each parameter-induced subset of a soft set. | Uses soft-set parameter images to approximate a target subset of the universe. |
| Underlying structure | A Pawlak approximation space $(U, R)$ and a soft set $(F, A)$. | A soft approximation space $P = (U, (F, A))$. |
| Approximation basis | Equivalence classes induced by $R$. | Parameter-induced subsets $F(a)$. |
| Approximation target | Each parameter image $F(a)$. | An arbitrary target subset $X \subseteq U$. |
| Lower approximation | $F_*(a) = \underline{R}(F(a))$. | $\mathrm{apr}_P(X) = \bigcup_{F(a)\subseteq X} F(a)$. |
| Upper approximation | $F^*(a) = \overline{R}(F(a))$. | $\overline{\mathrm{apr}}_P(X) = \bigcup_{F(a)\cap X\neq\varnothing} F(a)$. |
| Output | Two soft sets $(F_*, A)$ and $(F^*, A)$. | Two subsets $\mathrm{apr}_P(X)$ and $\overline{\mathrm{apr}}_P(X)$. |
| Main distinction | Rough approximations are applied parameterwise to soft-set images. | Soft-set images themselves determine the approximations of a target set. |

## 2.48 Partition Soft Set

A Partition Soft Set is a soft set whose parameter images form a disjoint cover of the universe, creating partitions [224–226].

**Definition 2.48.1** (Partition Soft Set)**.** Let $U$ be a nonempty universe, $E$ a set of parameters, and $\varnothing \neq A \subseteq E$. A soft set $(F, A)$, with

$$F : A \longrightarrow \mathcal{P}(U),$$

is called a *Partition Soft Set* if the family

$$\{F(a) : a \in A\}$$

forms a partition of $U$; that is,

$$\bigcup_{a\in A} F(a) = U$$

and, for all distinct $a, b \in A$,

$$F(a) \cap F(b) = \varnothing,$$

with $F(a) \neq \varnothing$ for every $a \in A$.

**Example 2.48.2** (Partition Soft Set)**.** Let

$$U = \{u_1, u_2, u_3, u_4, u_5\}$$

be a finite universe.

Consider the parameter set

$$E = A = \{e_1, e_2, e_3\}.$$

Define

$$F : A \longrightarrow \mathcal{P}(U)$$

by

$$F(e_1) = \{u_1, u_2\},$$

$$F(e_2) = \{u_3\},$$

and

$$F(e_3) = \{u_4, u_5\}.$$

Clearly,

$$F(e_i) \neq \varnothing \qquad (i = 1, 2, 3).$$

Furthermore,

$$F(e_1) \cap F(e_2) = \varnothing,$$

$$F(e_1) \cap F(e_3) = \varnothing,$$

and

$$F(e_2) \cap F(e_3) = \varnothing.$$

Thus, the parameter images are pairwise disjoint.

Moreover,

$$\begin{aligned}\bigcup_{e \in A} F(e) &= F(e_1) \cup F(e_2) \cup F(e_3) \\ &= \{u_1, u_2\} \cup \{u_3\} \cup \{u_4, u_5\} \\ &= U.\end{aligned}$$

Therefore,

$$\{F(e) : e \in A\}$$

forms a partition of $U$.

Consequently, $(F, A)$ is a Partition Soft Set.

## 2.49 G-Soft Set

A G-Number expresses an uncertain assessment through separate importance and necessity components, supporting decisions that distinguish significance from practical requirements [227,228].

**Definition 2.49.1** (G-Number)**.** [228] Let $\mathcal{L}$ be a set of linguistic evaluation values. A *G-Number* is an ordered pair

$$G = (I, N) \in \mathcal{L} \times \mathcal{L},$$

where $I$ represents the importance of an uncertain quantity and $N$ represents its necessity. Both components may be represented by linguistic terms or their corresponding numerical evaluations.

Let $\mathcal{X}$ be a nonempty universe of objects, $E$ a set of parameters, and $\varnothing \neq A \subseteq E$.

For a nonempty set $\mathcal{V}$ of admissible number-valued evaluations, a $\mathcal{V}$-valued soft set is a mapping

$$F : A \longrightarrow \mathcal{V}^{\mathcal{X}},$$

where $F(e)(x) \in \mathcal{V}$ for every $e \in A$ and $x \in \mathcal{X}$.

It induces an ordinary soft set

$$\Gamma_F : A \longrightarrow \mathcal{P}(\mathcal{X} \times \mathcal{V})$$

defined by

$$\Gamma_F(e) = \{(x, F(e)(x)) : x \in \mathcal{X}\}.$$

**Definition 2.49.2** (G-Soft Set)**.** Let $\mathcal{L}$ be a nonempty set of linguistic evaluation values. A *G-Soft Set* is a pair $(F_G, A)$, where

$$F_G : A \longrightarrow (\mathcal{L} \times \mathcal{L})^{\mathcal{X}}$$

assigns a G-Number to each object under each parameter.

For every $e \in A$ and $x \in \mathcal{X}$,

$$F_G(e)(x) = (I_e(x), N_e(x)),$$

where $I_e(x), N_e(x) \in \mathcal{L}$ represent importance and necessity, respectively.

**Theorem 2.49.3** (Well-definedness of G-Soft Sets)**.** *Every assignment of importance and necessity values from $\mathcal{L}$ determines a well-defined G-Soft Set.*

*Proof.* For each $e \in A$ and $x \in \mathcal{X}$, both $I_e(x)$ and $N_e(x)$ belong to $\mathcal{L}$. Hence,

$$(I_e(x), N_e(x)) \in \mathcal{L} \times \mathcal{L}.$$

Therefore, $F_G$ is a well-defined mapping into $(\mathcal{L} \times \mathcal{L})^{\mathcal{X}}$. Its graph representation $\Gamma_{F_G}$ is consequently an ordinary soft set. □

## 2.50 R-Soft Set

An R-Number models uncertain quantities using a risk-adjusted type-2 fuzzy representation incorporating risk perception, acceptable risk, and benefit-cost distinctions in decision-making [229–234].

**Definition 2.50.1** (R-Number)**.** [234] Let $\widetilde{B}$ be a fuzzy number with lower and upper bounds $l_B$ and $u_B$, respectively. An *R-Number* is a risk-adjusted type-2 fuzzy number represented by

$$R_s(\widetilde{B}) = \big(R_{1s}(\widetilde{B}), \widetilde{B}, R_{3s}(\widetilde{B})\big), \qquad s \in \{b, c\},$$

where $b$ and $c$ denote beneficial and cost-type evaluations, respectively.

Let $\widetilde{r}^{\pm}$, $\widetilde{AR}^{\pm}$, and $\widetilde{RP}^{\pm}$ denote fuzzy risk, acceptable risk, and risk perception, respectively. Define

$$d_{\pm} = \min\left\{ \frac{\widetilde{r}^{\pm}}{1 \ominus \widetilde{RP}^{\pm}}, \tau \right\} \otimes (1 \ominus \widetilde{AR}^{\pm}),$$

$$e_{\pm} = \frac{\widetilde{r}^{\pm}}{1 \ominus \widetilde{RP}^{\pm}} \otimes (1 \ominus \widetilde{AR}^{\pm}).$$

The risk-adjusted bounds are

$$R_{1b}(\widetilde{B}) = \max\{\widetilde{B} \otimes (1 \ominus d_-), l_B\},$$
$$R_{3b}(\widetilde{B}) = \min\{\widetilde{B} \otimes (1 \oplus e_+), u_B\},$$
$$R_{1c}(\widetilde{B}) = \max\{\widetilde{B} \otimes (1 \ominus d_+), l_B\},$$
$$R_{3c}(\widetilde{B}) = \min\{\widetilde{B} \otimes (1 \oplus e_-), u_B\}.$$

Here $\tau$ is a constant approaching 1, and the fuzzy arithmetic and clipping operations follow the original R-number construction, subject to its admissibility conditions.

**Definition 2.50.2** (R-Soft Set)**.** Let $\mathcal{R}_b$ and $\mathcal{R}_c$ denote the classes of admissible beneficial and cost-type R-Numbers, respectively, constructed using the preceding R-Number definition.

Set

$$\mathcal{R} = \mathcal{R}_b \cup \mathcal{R}_c.$$

An *R-Soft Set* is a pair $(F_R, A)$, where

$$F_R : A \longrightarrow \mathcal{R}^{\mathcal{X}}$$

assigns an admissible R-Number to each object under each parameter.

For $e \in A$ and $x \in \mathcal{X}$,

$$F_R(e)(x) = R_{s_{e,x}}(\widetilde{B}_{e,x}), \qquad s_{e,x} \in \{b, c\},$$

where the corresponding risk-adjustment data satisfy the admissibility conditions of the R-Number construction.

**Theorem 2.50.3** (Well-definedness of R-Soft Sets)**.** *Suppose that, for every $(e, x) \in A \times \mathcal{X}$, all fuzzy arithmetic operations in the R-Number construction are uniquely defined, their outputs are admissible fuzzy numbers, and the resulting R-Number belongs to $\mathcal{R}_{s_{e,x}}$.*

*Then $(F_R, A)$ is a well-defined R-Soft Set.*

*Proof.* For every $(e, x) \in A \times \mathcal{X}$, the admissibility assumptions ensure that the risk-adjusted quantities

$$R_{1s_{e,x}}(\widetilde{B}_{e,x}), \qquad R_{3s_{e,x}}(\widetilde{B}_{e,x})$$

exist uniquely and form an admissible R-Number together with $\widetilde{B}_{e,x}$.

Consequently,

$$F_R(e)(x) \in \mathcal{R}_{s_{e,x}} \subseteq \mathcal{R}.$$

Thus $F_R$ is a well-defined mapping, and $\Gamma_{F_R}$ is an ordinary soft set. □

## 2.51 H-Soft Set

An H-Number represents uncertain numerical assessments together with maximum and minimum utility values, supporting utility-based comparisons and decision-making under uncertainty [235, 236].

**Definition 2.51.1** (H-Number)**.** [236] Let $n = (a, b, c, d)$ be a trapezoidal numerical representation satisfying

$$a \leq b \leq c \leq d.$$

An *H-Number*, also called a *Utility-Number*, is an ordered triple

$$H = (n, M(n), N(n)),$$

where $M(n)$ and $N(n)$ denote the maximum and minimum utility values, respectively, with $N(n) \leq M(n)$.

An H-Number is called an *Immediate H-Number* if $N(n) = 0$, and an *Extended H-Number* if $N(n) \neq 0$.

**Definition 2.51.2** (H-Soft Set)**.** Let

$$\mathcal{Q} = \{(a, b, c, d) \in \mathbb{R}^4 : a \leq b \leq c \leq d\},$$

and define

$$\mathcal{H} = \left\{(n, M, N) \in \mathcal{Q} \times \mathbb{R}^2 : N \leq M\right\}.$$

An *H-Soft Set* is a pair $(F_H, A)$, where

$$F_H : A \longrightarrow \mathcal{H}^{\mathcal{X}}$$

assigns an H-Number to each object under each parameter.

For every $e \in A$ and $x \in \mathcal{X}$,

$$F_H(e)(x) = (n_e(x), M_e(x), N_e(x)),$$

where $n_e(x) \in \mathcal{Q}$ and

$$N_e(x) \leq M_e(x).$$

**Theorem 2.51.3** (Well-definedness of H-Soft Sets)**.** *Every assignment of admissible trapezoidal numerical values and ordered minimum and maximum utility values determines a well-defined H-Soft Set.*

*Proof.* For each $e \in A$ and $x \in \mathcal{X}$, the trapezoidal condition implies $n_e(x) \in \mathcal{Q}$. Moreover,

$$N_e(x) \leq M_e(x).$$

Therefore,

$$F_H(e)(x) \in \mathcal{H}.$$

Hence $F_H$ is a well-defined mapping into $\mathcal{H}^{\mathcal{X}}$, and $\Gamma_{F_H}$ is an ordinary soft set. □

## 2.52 U-Soft Set

A U-Number represents uncertain quantities using fuzzy restrictions and probabilistic usuality, extending Z-Number modeling to capture statements involving typical occurrences [237, 238].

**Definition 2.52.1** (U-Number)**.** Let $X$ be a random variable, and let $A$ be a fuzzy number on its domain. A *U-Number* is an ordered pair

$$U = (A, B_{\text{usually}}),$$

where $A$ represents a fuzzy restriction on $X$ and $B_{\text{usually}}$ is a fuzzy restriction on the probability measure of $A$.

For a discrete random variable with $P(X = x_i) = p_i$, the usuality grade is

$$\mu_{B_{\text{usually}}}\left(\sum_i p_i \mu_A(x_i)\right).$$

Thus, a U-Number is a special Z-Number whose reliability component represents usuality.

**Definition 2.52.2** (U-Soft Set)**.** Let $\Omega \subseteq \mathbb{R}$ be a nonempty domain, and let $\mathcal{F}(\Omega)$ denote the class of fuzzy numbers on $\Omega$.

Let $\mathcal{B}_{\text{usually}}$ be a nonempty class of fuzzy restrictions on $[0, 1]$ representing usuality.

Define

$$\mathcal{U} = \mathcal{F}(\Omega) \times \mathcal{B}_{\text{usually}}.$$

A *U-Soft Set* is a pair $(F_U, A)$, where

$$F_U : A \longrightarrow \mathcal{U}^{\mathcal{X}}$$

assigns a U-Number to each object under each parameter.

For every $e \in A$ and $x \in \mathcal{X}$,

$$F_U(e)(x) = (A_e(x), B_e(x)),$$

where

$$A_e(x) \in \mathcal{F}(\Omega), \qquad B_e(x) \in \mathcal{B}_{\text{usually}}.$$

**Theorem 2.52.3** (Well-definedness of U-Soft Sets)**.** *Every assignment of admissible fuzzy restrictions and usuality restrictions determines a well-defined U-Soft Set.*

*Furthermore, its probabilistic usuality grade is well-defined whenever the corresponding probability distribution is specified.*

*Proof.* For every $e \in A$ and $x \in \mathcal{X}$,

$$(A_e(x), B_e(x)) \in \mathcal{U}.$$

Thus $F_U$ is a well-defined mapping, and $\Gamma_{F_U}$ is an ordinary soft set.

Suppose that a discrete probability distribution $(p_i)$ is given, with

$$p_i \geq 0, \qquad \sum_i p_i = 1.$$

Since

$$0 \leq \mu_{A_e(x)}(\omega_i) \leq 1,$$

we obtain

$$0 \leq \sum_i p_i \mu_{A_e(x)}(\omega_i) \leq 1.$$

Therefore, the usuality grade

$$\mu_{B_e(x)}\left(\sum_i p_i \mu_{A_e(x)}(\omega_i)\right)$$

exists and belongs to $[0, 1]$. □

## 2.53 T-Soft Set

A T-Number extends triangular fuzzy numbers by incorporating trust and distrust assessments, adjusting uncertainty ranges for more reliable decision-making (cf. [239]).

**Definition 2.53.1** (T-Number)**.** Let $\widetilde{A} = (a, b, c)$ be a triangular fuzzy number, where $a \leq b \leq c$. A *T-Number* is an ordered triple

$$T = (\widetilde{A}, \widetilde{L}, \widetilde{U}),$$

where

$$\widetilde{L} = (e_l, f_l, g_l), \qquad \widetilde{U} = (e_u, f_u, g_u)$$

are triangular fuzzy numbers on $[-1, 1]$ representing trust or distrust in the left and right branches of $\widetilde{A}$, respectively.

Define

$$\alpha_l = \frac{e_l + 4f_l + g_l}{6}, \qquad \alpha_u = \frac{e_u + 4f_u + g_u}{6}.$$

The corresponding trust-adjusted triangular fuzzy number is

$$\mathrm{RATFN}(T) = (a + \alpha_l(b - a), b, c - \alpha_u(c - b)),$$

where $\alpha_l, \alpha_u \in [-1, 1]$. Positive values represent trust, whereas negative values represent distrust.

**Definition 2.53.2** (T-Soft Set)**.** For an interval $J \subseteq \mathbb{R}$, define

$$\mathrm{Tri}(J) = \{(a,b,c) \in J^3 : a \le b \le c\}.$$

Let

$$\mathcal{T} = \mathrm{Tri}(\mathbb{R}) \times \mathrm{Tri}([-1,1])^2.$$

A *T-Soft Set* is a pair $(F_T, A)$, where

$$F_T : A \longrightarrow \mathcal{T}^{\mathcal{X}}$$

assigns a T-Number to each object under each parameter.

For $e \in A$ and $x \in \mathcal{X}$,

$$F_T(e)(x) = (\widetilde{A}_{e,x}, \widetilde{L}_{e,x}, \widetilde{U}_{e,x}),$$

where

$$\widetilde{A}_{e,x} \in \mathrm{Tri}(\mathbb{R}),$$

and

$$\widetilde{L}_{e,x}, \widetilde{U}_{e,x} \in \mathrm{Tri}([-1,1]).$$

**Theorem 2.53.3** (Well-definedness of T-Soft Sets)**.** *Every assignment of admissible T-Numbers determines a well-defined T-Soft Set.*

*Moreover, the trust-adjusted representation of each assigned T-Number is a valid triangular fuzzy number.*

*Proof.* For every $e \in A$ and $x \in \mathcal{X}$, the component conditions imply

$$F_T(e)(x) \in \mathcal{T}.$$

Hence $F_T$ is well-defined, and $\Gamma_{F_T}$ is an ordinary soft set.

Now let

$$\widetilde{A}_{e,x} = (a,b,c), \qquad a \le b \le c.$$

Since the trust coefficients are weighted averages of values in $[-1,1]$,

$$-1 \le \alpha_l, \alpha_u \le 1.$$

Consequently,

$$a + \alpha_l(b-a) \le b$$

and

$$b \le c - \alpha_u(c-b).$$

Thus,

$$a + \alpha_l(b-a) \le b \le c - \alpha_u(c-b).$$

Therefore,

$$\mathrm{RATFN}(F_T(e)(x)) \in \mathrm{Tri}(\mathbb{R}),$$

which proves the claim. □

# Chapter 3

# Uncertain Soft Theory

In this chapter, we examine uncertainty-aware soft set models, including fuzzy soft sets and neutrosophic soft sets.

## 3.1 Fuzzy Soft Set

A fuzzy soft set assigns each parameter a fuzzy subset of the universe, giving graded membership degrees for objects under parameters [240–242]. Related notions include *fuzzy HyperSoft sets* [142, 243, 244] and *neutrosophic HyperSoft sets* [245–248].

**Definition 3.1.1** (Fuzzy Soft Set)**.** [240–242] Let $U$ be a nonempty universe and $E$ a set of parameters. Write $\mathcal{F}(U) := \{\mu : U \to [0,1]\}$ for the family of all fuzzy subsets of $U$. For a fixed subset $A \subseteq E$, a *fuzzy soft set* over $U$ (with respect to $A$) is a pair

$$(\Gamma_A, A), \qquad \Gamma_A : A \longrightarrow \mathcal{F}(U), \quad x \longmapsto \mu_x(\cdot),$$

so that it can be represented as the collection

$$\Gamma_A = \{\, (x, \mu_x) \mid x \in A,\ \mu_x : U \to [0,1] \,\}.$$

For $u \in U$ and $x \in A$, the value $\mu_x(u)$ is the degree to which $u$ approximately satisfies the parameter $x$. Equivalently, one may specify a map $\widetilde{\Gamma} : E \to \mathcal{F}(U)$ with $\widetilde{\Gamma}(x) = \mathbf{0}$ (the zero membership function) for all $x \notin A$.

## 3.2 Intuitionistic Fuzzy Soft Set

An intuitionistic fuzzy soft set assigns each parameter an intuitionistic fuzzy set, specifying membership and non-membership degrees with hesitation [249–252]. Intuitionistic Fuzzy Sets have been extensively studied in various publications in recent years [253–259]. Related concepts, such as Intuitionistic Fuzzy HyperSoft Sets [260–263], are also known.

**Definition 3.2.1** (Intuitionistic fuzzy set (Atanassov)). [5,264] Let $U$ be a nonempty universe. An *intuitionistic fuzzy set* (IFS) $A$ on $U$ is specified by a pair of functions

$$\mu_A : U \to [0,1], \qquad \nu_A : U \to [0,1],$$

called the *membership* and *non-membership* functions, respectively, such that

$$0 \;\le\; \mu_A(u) + \nu_A(u) \;\le\; 1 \qquad \text{for all } u \in U.$$

The *hesitation (indeterminacy) degree* is then

$$\pi_A(u) := 1 - \mu_A(u) - \nu_A(u) \in [0,1].$$

We write $\mathsf{IFS}(U)$ for the class of all intuitionistic fuzzy sets on $U$.

**Definition 3.2.2** (Intuitionistic fuzzy soft set). [249, 250] Let $U$ be a nonempty universe and let $E$ be a nonempty set of parameters. An *intuitionistic fuzzy soft set* (IFSS) over $U$ (with parameter set $E$) is a pair

$$(\widetilde{F}, A), \qquad A \subseteq E,$$

where

$$\widetilde{F} : A \longrightarrow \mathsf{IFS}(U)$$

is a mapping. Equivalently, an IFSS can be represented as a family of ordered pairs

$$(\widetilde{F}, A) \;=\; \big\{\, (e, \widetilde{F}(e)) \;\big|\; e \in A,\; \widetilde{F}(e) \in \mathsf{IFS}(U) \,\big\}.$$

If one prefers to keep the full parameter set $E$ as the domain, one may extend $\widetilde{F}$ to a mapping $\widetilde{F} : E \to \mathsf{IFS}(U)$ by setting, for each $e \in E \setminus A$, the *null IFS* $\widetilde{F}(e) = \widetilde{\varnothing}$ defined by

$$\mu_{\widetilde{\varnothing}}(u) = 0, \qquad \nu_{\widetilde{\varnothing}}(u) = 1 \quad (\forall u \in U),$$

which matches the common convention in the literature.

## 3.3 Neutrosophic Soft Set

A neutrosophic set assigns each element independent truth, indeterminacy, and falsity degrees, generalizing fuzzy and intuitionistic fuzzy sets [8, 9]. Neutrosophic sets have been extensively studied in numerous publications in recent years [265–274]. The Neutrosophic Soft Set is a concept that combines the principles of Neutrosophic Sets and Soft Sets [154, 275–281]. Related concepts, such as Neutrosophic HyperSoft Sets [28, 282–284], are also known. The definition is provided below.

**Definition 3.3.1** (Neutrosophic Soft Set [285, 286]). Let $U$ be a universe and $E$ a set of parameters. A *Neutrosophic Soft Set (NSS)* over $U$ is defined as a pair $\big(F, A\big)$, where $A \subseteq E$ and

$$F \;:\; A \;\longrightarrow\; P(U),$$

with $P(U)$ being the collection of *Neutrosophic Sets* on $U$. Hence for each parameter $e \in A$,

$$F(e) = \big(T_{F(e)}, I_{F(e)}, F_{F(e)}\big)$$

is a Neutrosophic Set on $U$, satisfying

$$0 \;\le\; T_{F(e)}(x) \;+\; I_{F(e)}(x) \;+\; F_{F(e)}(x) \;\le\; 3, \quad \forall\, x \in U.$$

## 3.4 Plithogenic Soft Set

A plithogenic set models multi-attribute membership using contradiction degrees among attribute values, generalizing fuzzy, intuitionistic, and neutrosophic sets [14, 287–289]. Plithogenic Sets have been extensively studied in various publications in recent years [290–292]. A plithogenic soft set maps attribute-value tuples to plithogenic evaluations and corresponding subsets, incorporating contradiction profiles relative to dominant values [293, 294].

**Definition 3.4.1** (Plithogenic soft set)**.** [293, 294] Let $U$ be a universe of discourse and let $z \in \{C, F, IF, N\}$. We write $\mathcal{P}_z(U)$ for the *z-power set* of $U$, defined by

$$\mathcal{P}_C(U) := \mathcal{P}(U), \qquad \mathcal{P}_F(U) := \{\, \mu : U \to [0,1] \,\},$$

$$\mathcal{P}_{IF}(U) := \Big\{ \, (\mu, \nu) : U \to [0,1]^2 \;\Big|\; 0 \le \mu(u) + \nu(u) \le 1 \ (\forall u \in U) \Big\},$$

and let $\mathcal{P}_N(U)$ denote the family of (single-valued) neutrosophic sets on $U$ (i.e., triples $(T, I, F) : U \to [0,1]^3$ satisfying the usual neutrosophic constraints, according to the chosen convention).

Let $a_1, a_2, \ldots, a_n$ $(n \ge 1)$ be distinct attributes with corresponding (pairwise disjoint) value sets $V_1, V_2, \ldots, V_n$ such that $V_i \cap V_j = \varnothing$ for $i \ne j$. Set

$$\Upsilon \;:=\; V_1 \times V_2 \times \cdots \times V_n.$$

Fix a *dominant value vector* $D = (D_1, \ldots, D_n) \in \Upsilon$ and, for each $i \in \{1, \ldots, n\}$, a *contradiction degree function*

$$c_i : V_i \times V_i \longrightarrow [0,1] \qquad \text{satisfying} \qquad c_i(v, v) = 0, \;\; c_i(v, w) = c_i(w, v).$$

Define

$$[0,1]^D \;:=\; [0,1]^n$$

and, for $v = (v_1, \ldots, v_n) \in \Upsilon$, define the contradiction vector relative to $D$ by

$$\mathbf{c}_D(v) \;:=\; \big(c_1(D_1, v_1),\ c_2(D_2, v_2),\ \ldots,\ c_n(D_n, v_n)\big) \in [0,1]^D.$$

A *z-plithogenic soft set* (briefly, *plithogenic soft set*) over $U$ is a pair $(F_P^z, \Upsilon)$ where

$$F_P^z :\ \Upsilon \longrightarrow [0,1]^D \times \mathcal{P}_z(U).$$

Equivalently, for each $v \in \Upsilon$ we can write

$$F_P^z(v) = \big(\mathbf{c}_D(v),\ S_v\big), \qquad \text{with } S_v \in \mathcal{P}_z(U),$$

so that each attribute-value tuple $v$ is assigned (i) its contradiction profile relative to the dominant tuple $D$, and (ii) a $z$-valued subset $S_v$ of the universe $U$.

## 3.5 Uncertain Soft Set

An Uncertain Set is any set-theoretic model assigning graded, possibly multi-component membership degrees to elements, generalizing fuzzy, intuitionistic, neutrosophic, plithogenic and related uncertainty frameworks unified [295–297].

**Definition 3.5.1** (Uncertain Set)**.** [295] Let $\mathbb{U}$ be the collection of all Uncertain Models. Fix

$$U \in \mathbb{U}, \quad \mathrm{Dom}(U) \subseteq [0,1]^r$$

for some integer $r \geq 1$, and let $X$ be a nonempty base set (universe of discourse).

An *Uncertain Set of type* $U$ *on* $X$ is a pair

$$\mathsf{A}^U \;=\; (X, \mu),$$

where

$$\mu \colon X \longrightarrow \mathrm{Dom}(U)$$

assigns to each element $x \in X$ a $U$–membership degree

$$\mu(x) \in \mathrm{Dom}(U).$$

Equivalently, once the base set $X$ and the Uncertain Model $U$ are fixed, we may identify the Uncertain Set with its membership function and simply write

$$\mathsf{A}^U \;:\; X \longrightarrow \mathrm{Dom}(U), \quad x \longmapsto \mu(x),$$

and view the collection of all Uncertain Sets of type $U$ on $X$ as the function space

$$\big(\mathrm{Dom}(U)\big)^X \;=\; \{\, \mu \mid \mu : X \to \mathrm{Dom}(U) \,\}.$$

In this sense, an Uncertain Set is a $U$–labeling of the base set $X$ by membership–degree tuples taken from $\mathrm{Dom}(U)$.

**Remark 3.5.2** (Recovery of classical fuzzy–type sets)**.** Let $X$ be a nonempty set and let $\mathsf{A}^U = (X, \mu)$ be an Uncertain Set of type $U$.

1. (*Fuzzy Set*) Take $U =$ Fuzzy with

$$\mathrm{Dom}(U) = [0,1] = [0,1]^1.$$

   Then an Uncertain Set of type $U$ is exactly a fuzzy set in the sense of Zadeh, since

$$\mu : X \to [0,1]$$

   is the usual fuzzy membership function.

2. (*Intuitionistic Fuzzy Set*) Take $U =$ Intuitionistic Fuzzy with

$$\mathrm{Dom}(U) = \{(\mu, \nu) \in [0,1]^2 \mid \mu + \nu \leq 1\} \subseteq [0,1]^2.$$

Then $\mathsf{A}^U = (X, \mu)$ coincides with an intuitionistic fuzzy set, because for each $x \in X$,

$$\mu(x) = (\mu_A(x), \nu_A(x)) \in [0,1]^2$$

satisfies $\mu_A(x) + \nu_A(x) \leq 1$.

3. (*Neutrosophic Set*) Take $U =$ Neutrosophic with

$$\mathrm{Dom}(U) = \{(T, I, F) \in [0,1]^3 \mid 0 \leq T + I + F \leq 3\} \subseteq [0,1]^3.$$

Then $\mathsf{A}^U = (X, \mu)$ is exactly a single–valued neutrosophic set, since

$$\mu(x) = (T_A(x), I_A(x), F_A(x)) \in [0,1]^3$$

with $0 \leq T_A(x) + I_A(x) + F_A(x) \leq 3$ for all $x \in X$.

4. (*Plithogenic Set*) For a Plithogenic Model $U =$ Plithogenic with degree–domain

$$\mathrm{Dom}(U) = \Big\{ (v, \mathrm{pdf}(x, v), \mathrm{pCF}(v_1, v_2)) \;\Big|\; v \in P_v,\ \mathrm{pdf}(x, v) \in [0,1]^s,\ \mathrm{pCF}(v_1, v_2) \in [0,1]^t \Big\} \subseteq [0,1]^{s+t+\ell},$$

an Uncertain Set of type $U$ on $X$ reproduces a Plithogenic Set on $X$, since each

$$\mu(x) \in \mathrm{Dom}(U)$$

encodes the Plithogenic degrees associated with $x \in X$.

Thus, by choosing different Uncertain Models $U \in \mathbb{U}$ and their corresponding domains $\mathrm{Dom}(U) \subseteq [0,1]^r$, the general notion of an Uncertain Set in Definition 3.5.1 unifies fuzzy sets, intuitionistic fuzzy sets, neutrosophic sets, plithogenic sets, and many other existing uncertainty–set frameworks.

As indicated above, the U-Set framework admits many model-dependent specializations. For reference, Table 3.1 lists representative uncertainty models according to the ambient dimension $k$ of their numerical degree-domains

$$\mathrm{Dom}(M) \subseteq \mathbb{R}^k.$$

The classification is dimensional only and does not imply equivalence among models having the same value of $k$ (cf. [298–300]).

**Definition 3.5.3** (Uncertainty domain and $\mathbb{D}$-uncertain sets)**.** Let $U$ be a nonempty universe. Fix an integer $r \geq 1$ and a nonempty set (degree-domain)

$$\mathbb{D} \subseteq [0,1]^r.$$

A $\mathbb{D}$-*uncertain set* on $U$ is a mapping

$$\mu : U \longrightarrow \mathbb{D}.$$

We denote the class of all $\mathbb{D}$-uncertain sets on $U$ by

$$\mathsf{Unc}_{\mathbb{D}}(U) := \mathbb{D}^U = \{\mu \mid \mu : U \to \mathbb{D}\}.$$

Table 3.1: Representative uncertainty-set models organized by the dimension $k$ of their numerical degree-domains $\mathrm{Dom}(M) \subseteq \mathbb{R}^k$ [298].

| $k$ | Restriction | Representative uncertain model(s) |
|---|---|---|
| 1 | | Fuzzy Set [4, 301]; N-Fuzzy Set [302–304]; Signed Fuzzy Set [305, 306]; Shadowed Set [307–309] |
| 2 | | Intuitionistic Fuzzy Set [5, 310]; Interval-Valued Fuzzy Set [311, 312]; Pythagorean Fuzzy Set [313, 314]; Fermatean Fuzzy Set [315, 316]; Hyperbolic Fuzzy Set [317–321]; Compact Fuzzy Set [322]; $q$-Rung Orthopair Fuzzy Set [323, 324]; Pseudo Fuzzy Set [325]; Vague Set [326, 327]; Bipolar Fuzzy Set (two-component representation) [328–330]; Variable Fuzzy Set [331–333]; Paraconsistent Fuzzy Set [334, 335]; Bifuzzy Set [336, 337] |
| 3 | | Single-Valued Neutrosophic Set [9, 18, 338]; Pythagorean Neutrosophic Set [339–341]; Fermatean Neutrosophic Set [342–344]; Indeterminacy Fuzzy Set [345, 346]; Inconsistent Intuitionistic Fuzzy Set [347, 348]; Ternary Fuzzy Set [349, 350]; Neutral Set [351–353]; Picture Fuzzy Set [7, 354]; Cubic Set [355, 356]; Kleene three-valued logic [357, 358]; Spherical Fuzzy Set [359–361]; T-Spherical Fuzzy Set [362, 363]; Tripolar Fuzzy Set (three-component formalisms) [364–367]; $q$-Rung Picture Fuzzy Set [368, 369]; Neutrosophic Vague Set [370, 371] |
| 4 | | Quadripartitioned Neutrosophic Set [10, 372]; Interval-Valued Intuitionistic Fuzzy Set [373, 374]; Interval-Valued Vague Set [375–377]; Interval-Valued Pythagorean Fuzzy Set [378–381]; Linear Diophantine Fuzzy Set [382, 383]; Decomposed Fuzzy Set [384]; Double-Valued Neutrosophic Set [385–387]; Support-Neutrosophic Set [388]; Local Neutrosophic Set [389]; Dual Hesitant Fuzzy Set [390, 391]; Ambiguous Set [392–394]; TINF-Neutrosophic Set; Quadri-Polar Fuzzy Set [395]; Belnap four-valued logic [396, 397]; Turiyam Neutrosophic Set [398–402] |
| 5 | | Pentapartitioned Neutrosophic Set [403–405]; Triple-Valued Neutrosophic Set [406–408] |
| 6 | | Hexapartitioned Neutrosophic Set [409, 410]; Quadruple-Valued Neutrosophic Set [407, 411]; Interval-Valued Neutrosophic Set [412–416]; Interval-Valued Picture Fuzzy Set [417, 418]; Bipolar Neutrosophic Set [419]; Interval-Valued T-Spherical Fuzzy Set [420, 421] |
| 7 | | Heptapartitioned Neutrosophic Set [422, 423]; Quintuple-Valued Neutrosophic Set [407, 424, 425] |
| 8 | | Octapartitioned Neutrosophic Set [409, 410] |
| 9 | | Nonapartitioned Neutrosophic Set [409, 410]; Neutrosophic Cubic Set [426–429]; Tripolar Neutrosophic Set [430, 431]; Cubic Picture Fuzzy Set [432–434] |
| $n$ | $n \geq 1$ | Multi-Valued Fuzzy Set [435]; MultiFuzzy Set [160, 436, 437]; $n$-valued Łukasiewicz logic [438]; Multipolar Fuzzy Set [72, 439]; $n$-Refined Fuzzy Set [156, 440] |
| $2n$ | $n \geq 1$ | $n$-Refined Intuitionistic Fuzzy Set [156]; Multi-Intuitionistic Fuzzy Set [160] |
| $3n$ | $n \geq 1$ | $n$-Refined Neutrosophic Set [156, 441–445]; Multipolar Neutrosophic Set [73]; Multi-Neutrosophic Set [160, 446] |

**Reading guide.** The table classifies representative models according to the ambient dimension of their numerical degree-domains. This dimensional classification does not imply that models with the same value of $k$ are mathematically or semantically equivalent. In particular, their admissibility conditions, coordinate interpretations, algebraic operations, and additional structural data may differ substantially.

(a) A common viewpoint in the literature is that neutrosophic sets form a broad umbrella encompassing several earlier multi-component fuzzy models and their generalizations; see [447].
(b) Ambiguous sets are often described as subclasses of certain four-component neutrosophic families; see [10, 372, 394].
(c) Turiyam neutrosophic sets are reported as subclasses of quadripartitioned neutrosophic sets; see [448].

**Definition 3.5.4** (Uncertain soft set)**.** Let $U$ be a nonempty universe and let $E$ be a nonempty set of parameters. Fix a nonempty uncertainty domain $\mathbb{D} \subseteq [0,1]^r$ as in Definition 3.5.3. Let $A \subseteq E$ be nonempty.

A $\mathbb{D}$-*uncertain soft set* (briefly, an *uncertain soft set*) over $U$ with parameter set $A$ is a pair $(F, A)$ where

$$F : A \longrightarrow \mathsf{Unc}_{\mathbb{D}}(U).$$

Equivalently, for each parameter $e \in A$, the value $F(e)$ is a $\mathbb{D}$-uncertain set on $U$, i.e., a function

$$F(e) : U \longrightarrow \mathbb{D}, \qquad u \longmapsto F(e)(u) \in \mathbb{D}.$$

**Remark 3.5.5** (Two standard choices of $\mathbb{D}$)**.** (i) **Fuzzy domain:** $\mathbb{D}_{\mathrm{Fuz}} := [0,1] \subseteq [0,1]^1$.

(ii) **Single-valued neutrosophic domain:**

$$\mathbb{D}_{\mathrm{Neu}} := \{(t,i,f) \in [0,1]^3 \mid 0 \le t+i+f \le 3\} \subseteq [0,1]^3.$$

(If one adopts a different conventional constraint for neutrosophic triples, replace $\mathbb{D}_{\mathrm{Neu}}$ accordingly; the theorem below remains valid.)

**Theorem 3.5.6** (Uncertain soft sets generalize fuzzy and neutrosophic soft sets)**.** *Let $U$ be a universe, $E$ a parameter set, and $A \subseteq E$ nonempty.*

(1) ***(Fuzzy soft sets as a special case)*** *If $\mathbb{D} = \mathbb{D}_{\mathrm{Fuz}} = [0,1]$, then $\mathbb{D}$-uncertain soft sets $(F,A)$ are exactly fuzzy soft sets over $U$ with parameter set $A$.*

(2) ***(Neutrosophic soft sets as a special case)*** *If $\mathbb{D} = \mathbb{D}_{\mathrm{Neu}}$, then $\mathbb{D}$-uncertain soft sets $(F,A)$ are exactly (single-valued) neutrosophic soft sets over $U$ with parameter set $A$.*

*Proof.* **(1)** Assume $\mathbb{D} = [0,1]$. Then

$$\mathsf{Unc}_{\mathbb{D}}(U) = \mathbb{D}^U = [0,1]^U,$$

the set of all membership functions on $U$. Hence an uncertain soft set is a map

$$F : A \to [0,1]^U.$$

For each $e \in A$, put $\mu_e := F(e) \in [0,1]^U$. Thus $(F,A)$ assigns to each parameter $e$ a fuzzy subset $\mu_e$ of $U$, which is precisely the definition of a fuzzy soft set. Conversely, any fuzzy soft set $(\Gamma, A)$ with $\Gamma(e) = \mu_e : U \to [0,1]$ is exactly a map $\Gamma : A \to [0,1]^U = \mathsf{Unc}_{\mathbb{D}}(U)$, hence a $\mathbb{D}$-uncertain soft set. Therefore the two notions coincide when $\mathbb{D} = [0,1]$.

**(2)** Assume $\mathbb{D} = \mathbb{D}_{\mathrm{Neu}} \subseteq [0,1]^3$. Then $\mathsf{Unc}_{\mathbb{D}}(U) = \mathbb{D}^U$ is the class of all functions

$$U \to \mathbb{D}_{\mathrm{Neu}}, \quad u \mapsto (T(u), I(u), F(u)),$$

i.e., the class of (single-valued) neutrosophic sets on $U$ under the chosen constraint. An uncertain soft set is a map

$$F : A \to \mathbb{D}_{\mathrm{Neu}}^U,$$

so each $e \in A$ is assigned a neutrosophic set $F(e)$ on $U$. This is exactly the definition of a (single-valued) neutrosophic soft set. Conversely, any neutrosophic soft set $(G,A)$ is by definition a map $G : A \to \mathbb{D}_{\mathrm{Neu}}^U = \mathsf{Unc}_{\mathbb{D}}(U)$, hence a $\mathbb{D}$-uncertain soft set. Thus the two notions coincide when $\mathbb{D} = \mathbb{D}_{\mathrm{Neu}}$. □

**Remark 3.5.7** (Optional extension: parameter-dependent uncertainty types)**.** If one wishes to allow different uncertainty domains per parameter, one may fix a family $\{\mathbb{D}_e \subseteq [0,1]^{r_e}\}_{e \in A}$ and define an "untyped" uncertain soft set as a map $F(e) : U \to \mathbb{D}_e$. The specialization arguments in Theorem 3.5.6 then apply componentwise.

An *Uncertain Soft Set* may be obtained by parameterizing an uncertainty-set model. Let $U$ be a nonempty universe, let $E$ be a parameter set, and let $\varnothing \neq A \subseteq E$. For an uncertainty model $M$, define

$$\mathfrak{S}_M(U) = \{\mu : U \to \mathrm{Dom}(M) \mid \mu \text{ satisfies the model-specific admissibility conditions of } M\},$$

where

$$\mathrm{Dom}(M) \subseteq \mathbb{R}^k$$

is the numerical degree-domain associated with $M$. An $M$-based *Uncertain Soft Set* is then a pair $(F, A)$ with

$$F : A \longrightarrow \mathfrak{S}_M(U).$$

Thus, for every parameter $e \in A$, the value $F(e)$ is an uncertain subset of $U$ described according to the model $M$.

As indicated above, the Uncertain Soft Set framework admits many model-dependent specializations. For reference, Table 3.2 lists representative induced Uncertain Soft Set models according to the ambient dimension $k$ of their numerical degree-domains

$$\mathrm{Dom}(M) \subseteq \mathbb{R}^k.$$

The classification is dimensional only and does not imply equivalence among models having the same value of $k$.

## 3.6 Z-Soft Set

A Z-number combines two fuzzy components: one describes a restriction on uncertain values, while the other represents its associated reliability [17]. A Z-soft set maps each parameter and object to a Z-number, combining fuzzy assessment with fuzzy reliability information.

**Definition 3.6.1** (Fuzzy number)**.** [513,514] A *fuzzy number* on $[0, 1]$ is a fuzzy set $\widetilde{A} : [0, 1] \to [0, 1]$ such that:

1. $\widetilde{A}$ is *normal*: $\sup_{x\in[0,1]} \widetilde{A}(x) = 1$;

2. $\widetilde{A}$ is *convex*: for every $\alpha \in (0, 1]$, the $\alpha$-cut

$$[\widetilde{A}]_\alpha := \{x \in [0, 1] \mid \widetilde{A}(x) \geq \alpha\}$$

is a (possibly degenerate) closed interval;

3. $\widetilde{A}$ is upper semicontinuous and has compact support in $[0, 1]$.

We write $\mathsf{FN}([0, 1])$ for the family of all fuzzy numbers on $[0, 1]$.

Table 3.2: Representative Uncertain Soft Set models organized by the dimension $k$ of their numerical degree-domains $\mathrm{Dom}(M) \subseteq \mathbb{R}^k$.

| $k$ | Restriction | Representative Uncertain Soft Set model(s) |
|---|---|---|
| 1 | | Fuzzy Soft Set; N-Fuzzy Soft Set; Signed Fuzzy Soft Set; Shadowed Soft Set [449, 450] |
| 2 | | Intuitionistic Fuzzy Soft Set [249, 250]; Interval-Valued Fuzzy Soft Set [451–453]; Pythagorean Fuzzy Soft Set [454–456]; Fermatean Fuzzy Soft Set [457–459]; Hyperbolic Fuzzy Soft Set; $q$-Rung Orthopair Fuzzy Soft Set; Pseudo Fuzzy Soft Set; Vague Soft Set [460–462]; Bipolar Fuzzy Soft Set [463–465] |
| 3 | | Single-Valued Neutrosophic Soft Set [285, 286]; Pythagorean Neutrosophic Soft Set [466–468]; Fermatean Neutrosophic Soft Set [469, 470]; Ternary Fuzzy Soft Set; Picture Fuzzy Soft Set [471–474]; Cubic Soft Set [99, 475]; Spherical Fuzzy Soft Set [476–478]; T-Spherical Fuzzy Soft Set [479–481]; Tripolar Fuzzy Soft Set; $q$-Rung Picture Fuzzy Soft Set; Neutrosophic Vague Soft Set |
| 4 | | Quadripartitioned Neutrosophic Soft Set [482, 483]; Interval-Valued Intuitionistic Fuzzy Soft Set [484–486]; Interval-Valued Vague Soft Set; Interval-Valued Pythagorean Fuzzy Soft Set [487–489]; Linear Diophantine Fuzzy Soft Set [490–492]; Double-Valued Neutrosophic Soft Set; Dual Hesitant Fuzzy Soft Set; Ambiguous Soft Set; TINF-Neutrosophic Soft Set; Turiyam Neutrosophic Soft Set |
| 5 | | Pentapartitioned Neutrosophic Soft Set [493–496]; Triple-Valued Neutrosophic Soft Set |
| 6 | | Hexapartitioned Neutrosophic Soft Set; Quadruple-Valued Neutrosophic Soft Set; Interval-Valued Neutrosophic Soft Set [497–499]; Interval-Valued Picture Fuzzy Soft Set [500–502]; Bipolar Neutrosophic Soft Set [503–505]; Interval-Valued T-Spherical Fuzzy Soft Set |
| 7 | | Heptapartitioned Neutrosophic Soft Set; Quintuple-Valued Neutrosophic Soft Set |
| 8 | | Octapartitioned Neutrosophic Soft Set |
| 9 | | Nonapartitioned Neutrosophic Soft Set; Neutrosophic Cubic Soft Set [506–508]; Tripolar Neutrosophic Soft Set; Cubic Picture Fuzzy Soft Set [509] |
| $n$ | $n \geq 1$ | Multi-Valued Fuzzy Soft Set; MultiFuzzy Soft Set; $n$-Valued Łukasiewicz Soft Set; Multipolar Fuzzy Soft Set; $n$-Refined Fuzzy Soft Set |
| $2n$ | $n \geq 1$ | $n$-Refined Intuitionistic Fuzzy Soft Set; Multi-Intuitionistic Fuzzy Soft Set |
| $3n$ | $n \geq 1$ | $n$-Refined Neutrosophic Soft Set [275, 276, 510, 511]; Multipolar Neutrosophic Soft Set [512]; Multi-Neutrosophic Soft Set |

**Reading guide.** Each entry should be understood as the soft-set counterpart induced by the corresponding uncertainty model $M$: for each parameter $e \in A$, the value $F(e)$ is an $M$-type uncertain subset of $U$. The table classifies these models only according to the ambient dimension of their numerical degree-domains. Hence, models with the same value of $k$ need not be mathematically or semantically equivalent. Their admissibility conditions, coordinate interpretations, algebraic operations, and additional structural data may differ substantially.

The terminology in the table is model-induced: when a dedicated Soft Set version has not been introduced independently in the literature, the corresponding name denotes the natural parameterized construction

$$F : A \to \mathfrak{S}_M(U)$$

rather than asserting an independently established theory under exactly that terminology.

**Definition 3.6.2** (Z-number)**.** [17] A *Z-number* is an ordered pair

$$Z = (\widetilde{A}, \widetilde{B}),$$

where $\widetilde{A} \in \mathsf{FN}([0,1])$ represents a fuzzy restriction (e.g., a fuzzy assessment of a degree), and $\widetilde{B} \in \mathsf{FN}([0,1])$ represents the reliability (certainty) of $\widetilde{A}$. Let

$$\mathcal{Z} := \mathsf{FN}([0,1]) \times \mathsf{FN}([0,1])$$

denote the set of all such Z-numbers.

**Definition 3.6.3** (Z-valued fuzzy set)**.** Let $U$ be a nonempty universe. A *Z-valued fuzzy set* on $U$ is a mapping

$$\mu : U \longrightarrow \mathcal{Z}.$$

We denote the family of all Z-valued fuzzy sets on $U$ by

$$\mathsf{ZFS}(U) := \mathcal{Z}^U.$$

**Definition 3.6.4** (Z-soft set)**.** Let $U$ be a nonempty universe and let $E$ be a nonempty set of parameters. Let $A \subseteq E$ be nonempty. A *Z-soft set* over $U$ with parameter set $A$ is a pair $(F, A)$ where

$$F : A \longrightarrow \mathsf{ZFS}(U).$$

Equivalently, $(F, A)$ may be identified with a single mapping

$$\mu : A \times U \longrightarrow \mathcal{Z}, \qquad \mu(e, u) = \big(\widetilde{A}_{e,u}, \widetilde{B}_{e,u}\big),$$

where $\widetilde{A}_{e,u}$ encodes the (fuzzy) assessment of $u$ under parameter $e$, and $\widetilde{B}_{e,u}$ encodes the reliability of that assessment.

## 3.7 ZD-Soft Set

A ZD-Soft Set assigns a ZD-Number to each parameter, representing uncertain value assessments and their reliability through parameter-dependent D-Numbers.

**Definition 3.7.1** (ZD-Number)**.** [515] Let $\Theta$ and $\Lambda$ be finite nonempty sets of value labels and reliability labels, respectively. Let

$$g : \mathcal{P}(\Lambda) \setminus \{\varnothing\} \longrightarrow [0, 1]$$

be a reliability semantics map.

A *ZD-Number* is an ordered pair

$$Z = (D_X, D_R),$$

where

$$D_X : \mathcal{P}(\Theta) \longrightarrow [0, 1], \qquad D_R : \mathcal{P}(\Lambda) \longrightarrow [0, 1]$$

are D-Numbers satisfying

$$D_X(\varnothing) = D_R(\varnothing) = 0,$$

and

$$\sum_{A \subseteq \Theta} D_X(A) \leq 1, \qquad \sum_{B \subseteq \Lambda} D_R(B) \leq 1.$$

Here, $D_X$ represents the uncertain value assessment, while $D_R$ represents its reliability.

The induced reliability degree is defined by

$$\rho(Z) = \sum_{\substack{B \subseteq \Lambda \\ B \neq \varnothing}} D_R(B) g(B).$$

The reliability-discounted value component is

$$D_X^Z(A) = \rho(Z) D_X(A), \qquad A \subseteq \Theta.$$

**Definition 3.7.2** (ZD-Soft Set)**.** Let $\Theta$ and $\Lambda$ be finite nonempty sets of value labels and reliability labels, respectively. Define

$$\mathcal{D}(\Omega) = \left\{ D : \mathcal{P}(\Omega) \to [0,1] \;\middle|\; D(\varnothing) = 0,\ \sum_{B \subseteq \Omega} D(B) \le 1 \right\}$$

for $\Omega \in \{\Theta, \Lambda\}$.

Let

$$g : \mathcal{P}(\Lambda) \setminus \{\varnothing\} \longrightarrow [0,1]$$

be a fixed reliability semantics map.

Define the family of ZD-Numbers by

$$\mathsf{ZD}(\Theta, \Lambda; g) = \mathcal{D}(\Theta) \times \mathcal{D}(\Lambda).$$

Let $E$ be a nonempty parameter set and $\varnothing \neq A \subseteq E$.

A *ZD-Soft Set* over $(\Theta, \Lambda, g)$ is a pair $(F, A)$, where

$$F : A \longrightarrow \mathsf{ZD}(\Theta, \Lambda; g).$$

For every $e \in A$,

$$F(e) = (D_e^X, D_e^R),$$

where

$$D_e^X \in \mathcal{D}(\Theta), \qquad D_e^R \in \mathcal{D}(\Lambda).$$

The induced reliability degree is

$$\rho_e = \sum_{\substack{B \subseteq \Lambda \\ B \neq \varnothing}} D_e^R(B) g(B).$$

The reliability-discounted value component is

$$\widetilde{D}_e^X(S) = \rho_e D_e^X(S), \qquad S \subseteq \Theta.$$

**Theorem 3.7.3** (Well-Definedness of ZD-Soft Sets)**.** *Let $\Theta, \Lambda, E, A,$ and $g$ be as above. Suppose that, for every $e \in A$,*

$$D_e^X : \mathcal{P}(\Theta) \to [0,1], \qquad D_e^R : \mathcal{P}(\Lambda) \to [0,1]$$

*satisfy*

$$D_e^X(\varnothing) = D_e^R(\varnothing) = 0$$

*and*

$$\sum_{S \subseteq \Theta} D_e^X(S) \le 1, \qquad \sum_{B \subseteq \Lambda} D_e^R(B) \le 1.$$

*Then $F(e) = (D_e^X, D_e^R)$ defines a well-defined ZD-Soft Set.*

*Moreover, for every $e \in A$,*

$$0 \le \rho_e \le 1, \qquad \widetilde{D}_e^X \in \mathcal{D}(\Theta).$$

*Proof.* For each $e \in A$, the assumptions imply

$$D_e^X \in \mathcal{D}(\Theta), \qquad D_e^R \in \mathcal{D}(\Lambda).$$

Hence,

$$F(e) \in \mathsf{ZD}(\Theta, \Lambda; g),$$

so $F$ is well-defined.

Since $0 \le g(B) \le 1$, we obtain

$$0 \le \rho_e \le \sum_{B \subseteq \Lambda} D_e^R(B) \le 1.$$

Furthermore,

$$\widetilde{D}_e^X(\varnothing) = 0$$

and

$$\sum_{S \subseteq \Theta} \widetilde{D}_e^X(S) = \rho_e \sum_{S \subseteq \Theta} D_e^X(S) \le 1.$$

Thus,

$$\widetilde{D}_e^X \in \mathcal{D}(\Theta).$$

Therefore, the ZD-Soft Set and its reliability-discounted components are well-defined. □

## 3.8 ZDSm-Soft Set

A ZDSm-Number combines DSm-based value assessments and reliability information through two generalized belief assignments on hyper-power sets.

**Definition 3.8.1** (ZDSm-Number)**.** Let $\Theta$ and $\Lambda$ be finite nonempty frames, and let $D^\Theta$ and $D^\Lambda$ denote their hyper-power sets.

For $\Omega \in \{\Theta, \Lambda\}$, define

$$\mathcal{M}_{\mathrm{DSm}}(\Omega) = \left\{ N : D^\Omega \to [0,1] \;\middle|\; N(\varnothing) = 0,\ \sum_{B \in D^\Omega} N(B) \le 1 \right\}.$$

Let

$$g : D^\Lambda \setminus \{\varnothing\} \longrightarrow [0,1]$$

be a reliability semantics map.

A *ZDSm-Number* is an ordered pair

$$Z_{\mathrm{DSm}} = (N_X, N_R),$$

where

$$N_X \in \mathcal{M}_{\mathrm{DSm}}(\Theta), \qquad N_R \in \mathcal{M}_{\mathrm{DSm}}(\Lambda).$$

Here, $N_X$ represents the uncertain value assessment, while $N_R$ represents its reliability.

The induced reliability degree is

$$\rho(Z_{\mathrm{DSm}}) = \sum_{\substack{B \in D^{\Lambda} \\ B \neq \varnothing}} N_R(B) g(B).$$

The reliability-discounted value component is

$$N_X^Z(C) = \rho(Z_{\mathrm{DSm}}) N_X(C), \qquad C \in D^{\Theta}.$$

**Example 3.8.2** (ZDSm-Number)**.** Let

$$\Theta = \{\theta_1, \theta_2\}, \qquad \Lambda = \{\lambda_1, \lambda_2\}$$

be two finite frames whose elements are not necessarily mutually exclusive.

Under the free DSm model, their hyper-power sets are

$$D^{\Theta} = \{\varnothing, \theta_1, \theta_2, \theta_1 \cap \theta_2, \theta_1 \cup \theta_2\},$$

and

$$D^{\Lambda} = \{\varnothing, \lambda_1, \lambda_2, \lambda_1 \cap \lambda_2, \lambda_1 \cup \lambda_2\}.$$

Define the value component

$$N_X : D^{\Theta} \to [0,1]$$

by

$$\begin{aligned} N_X(\varnothing) &= 0, \\ N_X(\theta_1) &= 0.3, \\ N_X(\theta_2) &= 0.2, \\ N_X(\theta_1 \cap \theta_2) &= 0.2, \\ N_X(\theta_1 \cup \theta_2) &= 0.1. \end{aligned}$$

Similarly, define the reliability component

$$N_R : D^{\Lambda} \to [0,1]$$

by

$$\begin{aligned} N_R(\varnothing) &= 0, \\ N_R(\lambda_1) &= 0.4, \\ N_R(\lambda_2) &= 0.2, \\ N_R(\lambda_1 \cap \lambda_2) &= 0.1, \\ N_R(\lambda_1 \cup \lambda_2) &= 0.2. \end{aligned}$$

Clearly,

$$\sum_{C \in D^{\Theta}} N_X(C) = 0.8 \le 1,$$

and

$$\sum_{B \in D^{\Lambda}} N_R(B) = 0.9 \le 1.$$

Thus,

$$N_X \in \mathcal{M}_{\mathrm{DSm}}(\Theta), \qquad N_R \in \mathcal{M}_{\mathrm{DSm}}(\Lambda).$$

Define the reliability semantics map

$$g : D^{\Lambda} \setminus \{\varnothing\} \to [0,1]$$

by

$$\begin{aligned} g(\lambda_1) &= 0.9, \\ g(\lambda_2) &= 0.5, \\ g(\lambda_1 \cap \lambda_2) &= 0.7, \\ g(\lambda_1 \cup \lambda_2) &= 0.8. \end{aligned}$$

The induced reliability degree is

$$\begin{aligned} \rho(Z_{\mathrm{DSm}}) &= \sum_{\substack{B \in D^{\Lambda} \\ B \neq \varnothing}} N_R(B) g(B) \\ &= 0.4(0.9) + 0.2(0.5) \\ &\quad + 0.1(0.7) + 0.2(0.8) \\ &= 0.69. \end{aligned}$$

Therefore, the reliability-discounted value component is

$$N_X^Z(C) = 0.69 N_X(C).$$

More explicitly,

$$\begin{aligned} N_X^Z(\varnothing) &= 0, \\ N_X^Z(\theta_1) &= 0.207, \\ N_X^Z(\theta_2) &= 0.138, \\ N_X^Z(\theta_1 \cap \theta_2) &= 0.138, \\ N_X^Z(\theta_1 \cup \theta_2) &= 0.069. \end{aligned}$$

Furthermore,

$$\sum_{C \in D^{\Theta}} N_X^Z(C) = 0.69(0.8) = 0.552 \le 1.$$

Hence,

$$N_X^Z \in \mathcal{M}_{\mathrm{DSm}}(\Theta).$$

Consequently,

$$Z_{\mathrm{DSm}} = (N_X, N_R)$$

is a well-defined ZDSm-Number.

This example illustrates how incomplete and potentially overlapping value assessments can be adjusted using a separate DSm-based reliability component.

A ZDSm-Soft Set assigns parameter-dependent ZDSm-Numbers, integrating uncertain value assessments and reliability information within a soft-set framework.

**Definition 3.8.3** (ZDSm-Soft Set)**.** Let $\Theta$, $\Lambda$, and $g$ be as above. Define

$$\mathsf{ZDSm}(\Theta, \Lambda; g) = \mathcal{M}_{\mathrm{DSm}}(\Theta) \times \mathcal{M}_{\mathrm{DSm}}(\Lambda).$$

Let $E$ be a nonempty parameter set and $\varnothing \neq A \subseteq E$.

A *ZDSm-Soft Set* is a pair $(F, A)$, where

$$F : A \longrightarrow \mathsf{ZDSm}(\Theta, \Lambda; g).$$

For each $e \in A$,

$$F(e) = (N_{X,e}, N_{R,e})$$

is a ZDSm-Number.

Its induced reliability degree is

$$\rho_e = \sum_{\substack{B \in D^\Lambda \\ B \neq \varnothing}} N_{R,e}(B) g(B),$$

and its reliability-discounted value component is

$$\widetilde{N}_{X,e}(C) = \rho_e N_{X,e}(C), \qquad C \in D^\Theta.$$

**Theorem 3.8.4** (Well-Definedness)**.** *Let $(F, A)$ be a ZDSm-Soft Set. Then, for every $e \in A$,*

$$0 \leq \rho_e \leq 1$$

*and*

$$\widetilde{N}_{X,e} \in \mathcal{M}_{\mathrm{DSm}}(\Theta).$$

*Proof.* Since $0 \le g(B) \le 1$ and $N_{R,e} \in \mathcal{M}_{\mathrm{DSm}}(\Lambda)$,

$$0 \le \rho_e \le \sum_{B \in D^{\Lambda}} N_{R,e}(B) \le 1.$$

Furthermore,

$$\widetilde{N}_{X,e}(\varnothing) = 0$$

and

$$\sum_{C \in D^{\Theta}} \widetilde{N}_{X,e}(C) = \rho_e \sum_{C \in D^{\Theta}} N_{X,e}(C) \le 1.$$

Thus,

$$\widetilde{N}_{X,e} \in \mathcal{M}_{\mathrm{DSm}}(\Theta),$$

which completes the proof. □

## 3.9 Functorial Soft Set

Functorial Sets provide an additional categorical layer: they organize such assignments across objects and morphisms so that structure transport is systematic and compositional [295, 298]. A functorial soft set treats parameters as objects in a category, assigning subsets functorially so morphisms induce consistent transformations naturally.

**Definition 3.9.1** (Functorial Set)**.** [295] Let $\mathcal{C}$ be a category and let

$$F : \mathcal{C} \longrightarrow \mathbf{Set}$$

be a covariant functor. The pair $(\mathcal{C}, F)$ is called a *Functorial Set.* For each object $X \in \mathrm{Ob}(\mathcal{C})$, the set $F(X)$ is interpreted as the collection of $F$-structures attached to $X$. Every morphism $f : X \to Y$ induces a structure-preserving map

$$F(f) : F(X) \longrightarrow F(Y),$$

such that $F(\mathrm{id}_X) = \mathrm{id}_{F(X)}$ and

$$F(g \circ f) = F(g) \circ F(f)$$

for all composable morphisms $f, g$ in $\mathcal{C}$.

**Definition 3.9.2** (Covariant power-set functor)**.** Let **Set** be the category of sets and functions. The *covariant power-set functor*

$$\mathcal{P} : \mathbf{Set} \longrightarrow \mathbf{Set}$$

is defined on objects by $\mathcal{P}(X) = \{A \mid A \subseteq X\}$ and on morphisms $f : X \to Y$ by the *direct image* map

$$\mathcal{P}(f) : \mathcal{P}(X) \longrightarrow \mathcal{P}(Y), \qquad \mathcal{P}(f)(A) := f[A] = \{f(x) \mid x \in A\}.$$

Then $\mathcal{P}(\mathrm{id}_X) = \mathrm{id}_{\mathcal{P}(X)}$ and $\mathcal{P}(g \circ f) = \mathcal{P}(g) \circ \mathcal{P}(f)$.

**Definition 3.9.3** (Functorial soft set)**.** Let $\mathcal{C}$ be a category. Let

$$\mathcal{U} : \mathcal{C} \longrightarrow \mathbf{Set} \quad \text{and} \quad \mathcal{E} : \mathcal{C} \longrightarrow \mathbf{Set}$$

be covariant functors, interpreted as a *universe functor* and a *parameter functor*, respectively. A *functorial soft set* on $(\mathcal{C}, \mathcal{U}, \mathcal{E})$ is a natural transformation

$$\mathcal{F} : \mathcal{E} \Longrightarrow \mathcal{P} \circ \mathcal{U}.$$

Equivalently, it is the data of maps

$$\mathcal{F}_X : \mathcal{E}(X) \longrightarrow \mathcal{P}(\mathcal{U}(X)) \qquad (X \in \mathrm{Ob}(\mathcal{C}))$$

such that for every morphism $f : X \to Y$ in $\mathcal{C}$ the following naturality condition holds:

$$\mathcal{P}(\mathcal{U}(f)) \circ \mathcal{F}_X \;=\; \mathcal{F}_Y \circ \mathcal{E}(f).$$

In elementwise form, for all $e \in \mathcal{E}(X)$,

$$\mathcal{U}(f)[\mathcal{F}_X(e)] \;=\; \mathcal{F}_Y(\mathcal{E}(f)(e)) \;\subseteq\; \mathcal{U}(Y).$$

**Theorem 3.9.4** (Functorial soft sets generalize soft sets and functorial sets)**.**(1) ***(Soft sets as a special case)*** *Let $\mathcal{C}$ be the terminal category with one object $*$ and only $\mathrm{id}_*$. Let $\mathcal{U}, \mathcal{E} : \mathcal{C} \to \mathbf{Set}$ be constant functors with $\mathcal{U}(*) = U$ and $\mathcal{E}(*) = E$. Then every functorial soft set $\mathcal{F} : \mathcal{E} \Rightarrow \mathcal{P} \circ \mathcal{U}$ is uniquely the same as a classical soft set mapping*

$$F : \; E \longrightarrow \mathcal{P}(U).$$

(2) ***(Functorial sets embed into functorial soft sets)*** *Let $(\mathcal{C}, F)$ be a functorial set, i.e., $F : \mathcal{C} \to \mathbf{Set}$. Define $\mathcal{U} := F$ and $\mathcal{E} := F$, and define components*

$$\mathcal{F}_X : \; F(X) \longrightarrow \mathcal{P}(F(X)), \qquad \mathcal{F}_X(x) := \{x\}.$$

*Then $\mathcal{F} : \mathcal{E} \Rightarrow \mathcal{P} \circ \mathcal{U}$ is a natural transformation, hence a functorial soft set. Moreover, from this functorial soft set one recovers the original functorial set by taking either $\mathcal{U}$ or $\mathcal{E}$ (both equal $F$).*

*Proof.* **(1)** In the terminal category $\mathcal{C}$, a functor $\mathcal{U} : \mathcal{C} \to \mathbf{Set}$ is determined by a single set $U := \mathcal{U}(*)$, and likewise $\mathcal{E}$ is determined by a single set $E := \mathcal{E}(*)$. A natural transformation

$$\mathcal{F} : \mathcal{E} \Rightarrow \mathcal{P} \circ \mathcal{U}$$

is determined by its single component

$$\mathcal{F}_* : \; E \longrightarrow \mathcal{P}(U).$$

Since the only morphism is $\mathrm{id}_*$, the naturality condition is automatic. Thus, setting $F := \mathcal{F}_*$ yields precisely a soft set mapping $F : E \to \mathcal{P}(U)$, and conversely any such mapping defines a unique natural transformation. Hence the notions coincide.

**(2)** Let $f : X \to Y$ be a morphism in $\mathcal{C}$ and let $x \in F(X)$. Compute the left-hand side of naturality:

$$\mathcal{P}(\mathcal{U}(f))(\mathcal{F}_X(x)) = \mathcal{P}(F(f))(\{x\}) = F(f)[\{x\}] = \{F(f)(x)\}.$$

Compute the right-hand side:

$$\mathcal{F}_Y(\mathcal{E}(f)(x)) = \mathcal{F}_Y(F(f)(x)) = \{F(f)(x)\}.$$

Thus $\mathcal{P}(\mathcal{U}(f)) \circ \mathcal{F}_X = \mathcal{F}_Y \circ \mathcal{E}(f)$ holds for all $f$, so $\mathcal{F}$ is natural and hence defines a functorial soft set. Finally, since $\mathcal{U} = \mathcal{E} = F$ by construction, the original functorial set is recovered immediately from the functorial soft set. □

## 3.10 Uncertain-Parameterized Soft Set

An Uncertain-Parameterized Soft Set assigns uncertainty-valued importance to parameters and uncertainty-valued evaluations to objects under each parameter in one framework.

**Definition 3.10.1** (Uncertain-Parameterized Soft Set)**.** Let $U$ be a nonempty universe, $E$ a set of parameters, and $\varnothing \neq A \subseteq E$. Let $M_P$ be an uncertainty model for parameter importance and $M_U$ an uncertainty model for object evaluation. Define

$$\mathrm{Unc}_{M_U}(U) = \{\mu : U \to \mathrm{Dom}(M_U)\}.$$

An $(M_P, M_U)$-*Uncertain-Parameterized Soft Set* over $U$ is a pair

$$(\zeta, F),$$

where

$$\zeta : A \to \mathrm{Dom}(M_P)$$

assigns an uncertainty-valued importance to each parameter, and

$$F : A \to \mathrm{Unc}_{M_U}(U)$$

assigns an $M_U$-uncertain subset of $U$ to each parameter.

Equivalently, for $e \in A$ and $u \in U$,

$$F(e)(u) \in \mathrm{Dom}(M_U),$$

or, in flattened form,

$$\widetilde{\mu} : A \times U \to \mathrm{Dom}(M_U), \qquad \widetilde{\mu}(e, u) = F(e)(u).$$

Related concepts, such as Fuzzy-Parameterized Soft Sets [516,517], Intuitionistic Fuzzy-Parameterized Soft Sets [518, 519], Neutrosophic-Parameterized Soft Sets [520, 521], and Fuzzy-Parameterized HyperSoft Sets [522, 523], are also known.

## 3.11 Z-Parameterized Soft Set

A Z-Parameterized Soft Set associates each parameter with a subset of the universe and a Z-number encoding its importance and the reliability of that assessment.

**Definition 3.11.1** (Z-Parameterized Soft Set)**.** Let $U$ be a nonempty universe, $E$ a nonempty parameter set, and $\varnothing \neq A \subseteq E$.

Let $\mathcal{Z}$ denote the family of admissible Z-Numbers.

A *Z-Parameterized Soft Set* over $U$ is a triple

$$(F, Z, A),$$

where

$$F : A \longrightarrow \mathcal{P}(U)$$

is a soft approximation mapping and

$$Z : A \longrightarrow \mathcal{Z}$$

assigns a Z-Number to each parameter.

For every $e \in A$,

$$Z(e) = (V_e, R_e),$$

where $V_e$ represents the uncertain evaluation or importance of parameter $e$, and $R_e$ represents its reliability.

**Example 3.11.2** (Z-Parameterized Soft Set)**.** Let

$$U = \{u_1, u_2, u_3\}$$

be a universe of three laptop models, and let

$$E = A = \{e_1, e_2, e_3\},$$

where

$$e_1 = \text{Low price}, \qquad e_2 = \text{Long battery life}, \qquad e_3 = \text{Lightweight}.$$

Define the soft approximation mapping

$$F : A \longrightarrow \mathcal{P}(U)$$

by

$$F(e_1) = \{u_1, u_2\},$$
$$F(e_2) = \{u_2, u_3\},$$

and

$$F(e_3) = \{u_1, u_3\}.$$

Let $(a, b, c)$ denote a triangular fuzzy number on $[0, 1]$, where

$$0 \le a < b < c \le 1.$$

Define the Z-Number mapping

$$Z : A \longrightarrow \mathcal{Z}$$

by

$$Z(e_1) = \big((0.5, 0.7, 0.9), (0.65, 0.8, 0.95)\big),$$
$$Z(e_2) = \big((0.7, 0.85, 0.95), (0.6, 0.75, 0.9)\big),$$

and

$$Z(e_3) = \big((0.3, 0.5, 0.7), (0.7, 0.85, 0.95)\big).$$

For every $e_i \in A$, write

$$Z(e_i) = (V_{e_i}, R_{e_i}),$$

where $V_{e_i}$ represents the uncertain importance of parameter $e_i$, and $R_{e_i}$ represents the reliability of that importance assessment.

For example,

$$Z(e_2) = \big((0.7, 0.85, 0.95), (0.6, 0.75, 0.9)\big)$$

indicates that long battery life has a relatively high importance, with an associated fuzzy reliability assessment.

Since

$$F(e_i) \subseteq U$$

and

$$Z(e_i) \in \mathcal{Z} \qquad (i = 1, 2, 3),$$

both mappings

$$F : A \to \mathcal{P}(U), \qquad Z : A \to \mathcal{Z}$$

are well-defined.

Therefore,

$$(F, Z, A)$$

is a Z-Parameterized Soft Set over $U$.

This example demonstrates how parameterized object selection can be combined with uncertain parameter importance and reliability within a unified soft-set framework.

## 3.12 D-Parameterized Soft Set

A D-Parameterized Soft Set associates each parameter with a subset of the universe and a D-number representing its uncertain importance through incomplete or nonexclusive evidence.

**Definition 3.12.1** (D-Parameterized Soft Set)**.** Let $U$ be a nonempty universe, $E$ a nonempty parameter set, and $\varnothing \neq A \subseteq E$. Let $\Lambda$ be a finite nonempty set of parameter-evaluation labels.

Define

$$\mathcal{D}(\Lambda) = \left\{ d : \mathcal{P}(\Lambda) \to [0,1] \;\middle|\; d(\varnothing) = 0,\ \sum_{B \subseteq \Lambda} d(B) \leq 1 \right\}.$$

A *D-Parameterized Soft Set* over $U$ is a triple

$$(F, D, A),$$

where

$$F : A \longrightarrow \mathcal{P}(U)$$

is a soft approximation mapping and

$$D : A \longrightarrow \mathcal{D}(\Lambda)$$

assigns a D-Number to each parameter.

For every $e \in A$, the D-Number

$$D(e) = D_e$$

represents the uncertain importance or evaluation of parameter $e$, allowing incomplete and nonexclusive information.

**Theorem 3.12.2** (Well-Definedness of D-Parameterized Soft Sets). *Let $U$ and $E$ be nonempty sets, $\varnothing \neq A \subseteq E$, and $\Lambda$ a finite nonempty set.*

*Suppose that, for every $e \in A$, there exist a subset $F_e \subseteq U$ and a mapping*

$$D_e : \mathcal{P}(\Lambda) \to [0, 1]$$

*satisfying*

$$D_e(\varnothing) = 0, \qquad \sum_{B \subseteq \Lambda} D_e(B) \leq 1.$$

*Then the assignments*

$$F(e) = F_e, \qquad D(e) = D_e$$

*define a unique D-Parameterized Soft Set $(F, D, A)$ over $U$.*

*Moreover, the class of D-Parameterized Soft Sets is nonempty.*

*Proof.* Since $F_e \subseteq U$ for every $e \in A$, the mapping

$$F : A \to \mathcal{P}(U)$$

is well-defined.

Furthermore, each $D_e$ satisfies the defining conditions of a D-Number. Hence,

$$D_e \in \mathcal{D}(\Lambda)$$

for every $e \in A$, so

$$D : A \to \mathcal{D}(\Lambda)$$

is well-defined.

Both mappings are uniquely determined by their prescribed values.

Therefore, $(F, D, A)$ is a well-defined D-Parameterized Soft Set.

Finally, choosing

$$F(e) = \varnothing, \qquad D(e)(B) = 0$$

for all $e \in A$ and $B \subseteq \Lambda$ provides an explicit instance. Thus, the class is nonempty. □

## 3.13 Rough-Parameterized Soft Set

The Rough-Parameterized Soft Set is defined as follows. This concept is developed by applying the basic idea underlying the Fuzzy-Parameterized Soft Set [524–526] to a rough-set-based parameterization framework.

**Definition 3.13.1** (Rough-Parameterized Soft Set)**.** Let $U$ be a nonempty universe and let $E$ be a nonempty parameter universe. Let

$$F : E \longrightarrow \mathcal{P}(U)$$

be a Soft Set mapping.

Let

$$(E, R_E)$$

be a Pawlak approximation space on the parameter universe $E$, and let

$$P \subseteq E$$

be a target set of parameters. Define

$$\underline{R_E}(P) = \{e \in E : [e]_{R_E} \subseteq P\},$$

and

$$\overline{R_E}(P) = \{e \in E : [e]_{R_E} \cap P \neq \varnothing\}.$$

The corresponding lower and upper parameterized Soft Sets are

$$\underline{\mathcal{S}}_{R_E}(P, F) = \left(F|_{\underline{R_E}(P)}, \underline{R_E}(P)\right)$$

and

$$\overline{\mathcal{S}}_{R_E}(P, F) = \left(F|_{\overline{R_E}(P)}, \overline{R_E}(P)\right),$$

respectively.

The pair

$$\mathcal{RS}_{R_E}(P, F) = \left(\underline{\mathcal{S}}_{R_E}(P, F), \overline{\mathcal{S}}_{R_E}(P, F)\right)$$

is called a *Rough-Parameterized Soft Set* over $U$.

The parameters in

$$\underline{R_E}(P)$$

are interpreted as certainly admissible parameters, whereas those in

$$\overline{R_E}(P)$$

are interpreted as possibly admissible parameters.

**Remark 3.13.2.** A Rough-Parameterized Soft Set should be distinguished from both a Rough Soft Set and a Soft Rough Set. In a Rough Soft Set [527, 528], rough approximations are applied to the subsets $F(e) \subseteq U$ associated with individual parameters. In a Soft Rough Set [529, 530], a Soft Set is used as an approximation structure for a target subset of $U$.

By contrast, in a Rough-Parameterized Soft Set, the rough approximation is applied directly to a subset $P \subseteq E$ of the parameter universe. Thus, roughness occurs at the *parameter level*, while the mapping

$$F : E \to \mathcal{P}(U)$$

retains the ordinary Soft Set structure.

# Chapter 4

# Applications of Soft Set

In this chapter, we describe several studies on extended concepts developed using soft set theory.

## 4.1 Soft Graph

A soft graph assigns each parameter a subgraph of a fixed graph, modeling uncertain, parameterized relationships among vertices and edges [531,532]. As extensions, concepts such as HyperSoft Graphs [533–535], Fuzzy Soft Graphs [536, 537], Neutrosophic Soft Graphs [538, 539], Soft HyperGraphs [540, 541], Soft SuperHyperGraphs [542], and Soft Directed Graphs [543, 544] have also been studied.

**Definition 4.1.1** (Soft Graph)**.** [531,532] Let $G = (V, E)$ be a simple undirected graph and $C$ a nonempty set of parameters. A *soft graph* over $G$ with parameter set $C$ is a quadruple

$$\big(G,\, C,\, A,\, B\big),$$

where

$$A : C \longrightarrow \mathcal{P}(V), \qquad B : C \longrightarrow \mathcal{P}(E),$$

and for each $c \in C$,

$$B(c) \;\subseteq\; \big\{\{u, v\} \in E : u \in A(c),\ v \in A(c)\big\}.$$

The pair $\big(A(c), B(c)\big)$ is called the *soft subgraph* at parameter $c$.

**Example 4.1.2** (Example of a soft graph: a friendship network under different interaction contexts)**.** Let $G = (V, E)$ be a simple undirected graph modeling friendships among six people:

$$V = \{v_1, v_2, v_3, v_4, v_5, v_6\},$$

$$E = \big\{\{v_1, v_2\}, \{v_1, v_3\}, \{v_2, v_3\}, \{v_2, v_4\}, \{v_3, v_5\}, \{v_4, v_5\}, \{v_5, v_6\}\big\}.$$

Let $C$ be a set of parameters describing interaction contexts:

$$C = \{c_{\text{Work}}, c_{\text{Sport}}, c_{\text{Online}}\}.$$

Define $A : C \to \mathcal{P}(V)$ (active people) and $B : C \to \mathcal{P}(E)$ (active friendships) by:

$$A(c_{\text{Work}}) = \{v_1, v_2, v_3, v_4\}, \qquad B(c_{\text{Work}}) = \{\{v_1, v_2\}, \{v_1, v_3\}, \{v_2, v_3\}, \{v_2, v_4\}\},$$

$$A(c_{\text{Sport}}) = \{v_2, v_3, v_5, v_6\}, \qquad B(c_{\text{Sport}}) = \{\{v_2, v_3\}, \{v_3, v_5\}, \{v_5, v_6\}\},$$

$$A(c_{\text{Online}}) = \{v_1, v_3, v_5\}, \qquad B(c_{\text{Online}}) = \{\{v_1, v_3\}, \{v_3, v_5\}\}.$$

For each $c \in C$, every edge in $B(c)$ has both endpoints in $A(c)$, hence

$$B(c) \subseteq \big\{\{u, v\} \in E : \ u, v \in A(c)\big\}.$$

Therefore, $\big(G, C, A, B\big)$ is a soft graph over $G$.

Interpretation: the underlying friendship network is $G$, while each parameter $c$ extracts a context-dependent subgraph (e.g., work interactions, sports interactions, online interactions).

## 4.2 Soft HyperGraph

A Soft HyperGraph associates each parameter with vertex and hyperedge subsets, ensuring selected hyperedges lie entirely within their assigned vertices [540, 541].

**Definition 4.2.1** (Soft HyperGraph)**.** Let $H = (V, \mathcal{E})$ be a hypergraph, where

$$\mathcal{E} \subseteq \mathcal{P}(V) \setminus \{\varnothing\},$$

and let $A$ be a nonempty parameter set.

A *Soft HyperGraph* over $H$ is a quadruple

$$\mathcal{H} = (H, F, K, A),$$

where

$$F : A \longrightarrow \mathcal{P}(V), \qquad K : A \longrightarrow \mathcal{P}(\mathcal{E}),$$

such that, for every $a \in A$,

$$K(a) \subseteq \{e \in \mathcal{E} : e \subseteq F(a)\}.$$

Thus, each parameter $a \in A$ determines a subhypergraph

$$H_a = (F(a), K(a))$$

of $H$, called the *soft subhypergraph* associated with $a$.

## 4.3 Soft Topological Space

A soft topological space is a parameterized family of soft open sets closed under soft unions and finite intersections, containing the null and absolute soft sets [412, 545]. Related notions include hypersoft topological spaces [546–548], fuzzy soft topological spaces [549,550], intuitionistic fuzzy soft topological spaces [551, 552], and neutrosophic soft topological spaces [496, 545, 553, 554].

**Definition 4.3.1** (Soft topology and soft topological space)**.** [412, 545] Let $X$ be a nonempty universe and let $A$ be a nonempty set of parameters. A *soft set* over $X$ (with parameter set $A$) is a pair $(F, A)$ where $F : A \to \mathcal{P}(X)$. Denote by $\mathsf{SS}(X, A)$ the collection of all such soft sets.

Define the $A$-*null* and $A$-*absolute* soft sets by

$$0_A = (F_0, A), \quad F_0(a) = \varnothing \ (a \in A), \qquad 1_A = (F_1, A), \quad F_1(a) = X \ (a \in A).$$

For a family $\{(F_i, A)\}_{i\in I} \subseteq \mathsf{SS}(X, A)$, define the soft union $\bigsqcup_{i\in I}(F_i, A) = (F, A)$ by

$$F(a) = \bigcup_{i\in I} F_i(a) \qquad (a \in A),$$

and for $(F, A), (G, A) \in \mathsf{SS}(X, A)$ define the soft intersection $(F, A) \sqcap (G, A) = (H, A)$ by

$$H(a) = F(a) \cap G(a) \qquad (a \in A).$$

A subfamily $\tau \subseteq \mathsf{SS}(X, A)$ is called a *soft topology* on $X$ (with parameter set $A$) if

(ST1) $0_A$, $1_A \in \tau$;

(ST2) if $(F, A), (G, A) \in \tau$, then $(F, A) \sqcap (G, A) \in \tau$;

(ST3) if $(F_i, A) \in \tau$ for all $i \in I$, then $\bigsqcup_{i\in I}(F_i, A) \in \tau$.

In this case, the triple $(X, \tau, A)$ is called a *soft topological space*. Elements of $\tau$ are called *soft open sets*; a soft set $(F, A)$ is *soft closed* if its soft complement $(F, A)^c$ belongs to $\tau$.

**Example 4.3.2** (Example of a soft topological space: neighborhood-based accessibility under different criteria)**.** Let

$$X = \{x_1, x_2, x_3\}$$

be a set of locations (e.g., three service points), and let

$$A = \{a_{\text{Walk}}, a_{\text{Drive}}\}$$

be a set of parameters, where $a_{\text{Walk}}$ means "reachable on foot" and $a_{\text{Drive}}$ means "reachable by car".

Define the following soft sets over $X$ (all with parameter set $A$):

$$0_A(a_{\text{Walk}}) = 0_A(a_{\text{Drive}}) = \varnothing, \qquad 1_A(a_{\text{Walk}}) = 1_A(a_{\text{Drive}}) = X,$$

and a nontrivial soft set $(F, A)$ by

$$F(a_{\text{Walk}}) = \{x_1, x_2\}, \qquad F(a_{\text{Drive}}) = \{x_1, x_2, x_3\}.$$

Now define

$$\tau = \{\, 0_A,\ 1_A,\ (F, A)\,\} \subseteq \mathsf{SS}(X, A).$$

We verify that $\tau$ is a soft topology on $X$:

(ST1) $0_A, 1_A \in \tau$ by definition.

(ST2) Finite soft intersections remain in $\tau$:

$$(F, A) \sqcap 1_A = (F, A), \qquad (F, A) \sqcap 0_A = 0_A, \qquad (F, A) \sqcap (F, A) = (F, A).$$

(ST3) Arbitrary soft unions of members of $\tau$ remain in $\tau$:

$$0_A \bigsqcup (F, A) = (F, A), \qquad (F, A) \bigsqcup 1_A = 1_A, \qquad 0_A \bigsqcup 1_A = 1_A,$$

and any union of copies of $(F, A)$ is still $(F, A)$.

Hence $(X, \tau, A)$ is a soft topological space.

Interpretation: under the walking criterion, the "open" accessible region is $\{x_1, x_2\}$, while under driving it is $X$; the soft topology $\tau$ contains the null region, the whole region, and this parameter-dependent accessibility region.

## 4.4 Soft Algebra

A soft algebra is a family of soft sets closed under soft complement and finite soft unions, containing the null soft set [555–558].

**Definition 4.4.1** (Soft algebra)**.** [555, 556] Let $X$ be a nonempty universe and let $Q$ be a nonempty set of parameters. A *soft set* over $X$ (with parameter set $Q$) is a pair $(F, Q)$ where

$$F : Q \longrightarrow \mathcal{P}(X).$$

Denote by $\mathsf{S}_Q(X)$ the family of all such soft sets.

Define the *null soft set* $\Phi_Q = (F_0, Q)$ and the *absolute soft set* $X_Q = (F_1, Q)$ by

$$F_0(q) = \varnothing, \qquad F_1(q) = X \qquad (q \in Q).$$

For $(F, Q) \in \mathsf{S}_Q(X)$ define the *soft complement* $(F, Q)^c = (F^c, Q)$ by

$$F^c(q) = X \setminus F(q) \qquad (q \in Q).$$

For $(F, Q), (G, Q) \in \mathsf{S}_Q(X)$ define the *soft union* and *soft intersection* by

$$(F, Q) \,\widetilde{\cup}\, (G, Q) = (H, Q),\ \ H(q) = F(q) \cup G(q), \qquad (F, Q) \,\widetilde{\cap}\, (G, Q) = (K, Q),\ \ K(q) = F(q) \cap G(q).$$

A subfamily $\Sigma \subseteq \mathsf{S}_Q(X)$ is called a *soft algebra* on $X$ (parameterized by $Q$) if:

(SA1) $\Phi_Q \in \Sigma$;

(SA2) if $(F, Q) \in \Sigma$, then $(F, Q)^c \in \Sigma$;

(SA3) if $(F_i, Q) \in \Sigma$ for $i = 1, 2, \ldots, k$, then

$$\widetilde{\bigcup}_{i=1}^{k} (F_i, Q) \in \Sigma,$$

equivalently, $\Sigma$ is closed under finite soft unions.

**Example 4.4.2** (Example of a soft algebra: "fixed-or-empty" soft sets)**.** Let

$$X = \{1, 2, 3, 4\} \quad \text{and} \quad Q = \{q_1, q_2\}$$

be a universe and a parameter set. Fix two subsets of $X$:

$$A = \{1, 2\}, \qquad B = \{3, 4\}.$$

Consider the following subfamily $\Sigma \subseteq \mathsf{S}_Q(X)$ consisting of all soft sets $(F, Q)$ such that for each parameter $q \in Q$ one has

$$F(q) \in \{\varnothing,\ A,\ B,\ X\}.$$

Equivalently,

$$\Sigma = \Big\{ (F, Q) \in \mathsf{S}_Q(X) \ \Big|\ F(q_1), F(q_2) \in \{\varnothing, A, B, X\} \Big\}.$$

Then $\Sigma$ is a soft algebra on $X$ (parameterized by $Q$):

(SA1) The null soft set $\Phi_Q$ belongs to $\Sigma$ since $\Phi_Q(q_1) = \Phi_Q(q_2) = \varnothing$.

(SA2) If $(F, Q) \in \Sigma$, then for each $q \in Q$ we have $F(q) \in \{\varnothing, A, B, X\}$, and hence

$$F^c(q) = X \setminus F(q) \in \{X, B, A, \varnothing\} \subseteq \{\varnothing, A, B, X\}.$$

Thus $(F, Q)^c \in \Sigma$.

(SA3) If $(F_1, Q), \ldots, (F_k, Q) \in \Sigma$, then for each $q \in Q$ every $F_i(q)$ is one of $\varnothing, A, B, X$, and therefore

$$\bigcup_{i=1}^{k} F_i(q) \in \{\varnothing, A, B, X\}.$$

Hence $\widetilde{\bigcup}_{i=1}^{k} (F_i, Q) \in \Sigma$.

Interpretation: $\Sigma$ models a simplified decision system where, for each parameter, only four outcomes are allowed (select nothing, select group $A$, select group $B$, or select all $X$), and the family remains stable under combining rules (union) and negating rules (complement).

## 4.5 Soft Lattice

A soft lattice is a lattice structure on soft sets (often modulo soft equality), using soft union and intersection operations [559–561]. Related notions include fuzzy soft lattices [562, 563], intuitionistic fuzzy soft lattices [564–566], and neutrosophic soft lattices [559].

**Definition 4.5.1** (Soft Lattice on a Fixed Parameter Set)**.** Let $U$ be a nonempty universe and let $E$ be a nonempty fixed parameter set. Consider the family

$$\mathsf{S}_E(U) = \{F : E \to \mathcal{P}(U)\}.$$

For $F, G \in \mathsf{S}_E(U)$, define the pointwise operations

$$(F \vee G)(e) = F(e) \cup G(e), \qquad (F \wedge G)(e) = F(e) \cap G(e) \quad (e \in E).$$

Write $F \leq G$ when $F(e) \subseteq G(e)$ for every $e \in E$. A *fixed-parameter Soft Lattice* is the ordered structure

$$\big(\mathsf{S}_E(U), \vee, \wedge, \leq\big).$$

The parameter set is fixed so that these operations do not require choosing representatives from equivalence classes.

**Theorem 4.5.2** (Complete Distributive Soft Lattice)**.** *The structure in Definition 4.5.1 is a complete distributive lattice, with bottom and top given by*

$$\bot(e) = \varnothing, \qquad \top(e) = U \quad (e \in E).$$

*Proof.* For any family $(F_j)_{j \in J} \subseteq \mathsf{S}_E(U)$, define

$$\left(\bigvee_{j \in J} F_j\right)(e) = \bigcup_{j \in J} F_j(e), \qquad \left(\bigwedge_{j \in J} F_j\right)(e) = \bigcap_{j \in J} F_j(e).$$

Each value is a subset of $U$, so the maps belong to $\mathsf{S}_E(U)$. They are respectively the least upper bound and greatest lower bound under pointwise inclusion. The usual identities for union and intersection of subsets hold at each $e$, and hence distributivity and absorption hold for the corresponding soft-set operations. The empty join and meet yield $\bot$ and $\top$. $\square$

**Remark 4.5.3** (Limitation of quotient constructions)**.** Extended union and restricted intersection on soft sets with varying parameter domains need not be compatible with relations that identify soft sets merely by mutual coverage of their value families. Consequently, their induced operations on such quotient classes cannot be assumed to be well-defined without proving congruence. The fixed-domain construction avoids this difficulty.

## 4.6 Soft Vector

A soft vector is a parameter-indexed mapping into a vector space, representing context-dependent vectors and enabling membership in soft sets [567].

**Definition 4.6.1** (Soft vector (soft element of a vector space))**.** [567] Let $V$ be a vector space over a field $\mathbb{F}$, and let $A$ be a nonempty set of parameters. A *soft vector* over $V$ (with respect to $A$) is a mapping

$$\tilde{v} : \ A \longrightarrow V.$$

If $(F, A)$ is a soft set over $V$, i.e. $F : A \to \mathcal{P}(V)$, then a soft vector $\tilde{v}$ is said to *belong* to $(F, A)$ (denoted $\tilde{v} \in (F, A)$) if

$$\tilde{v}(a) \in F(a) \qquad \text{for all } a \in A.$$

**Remark 4.6.2.** A soft vector $\tilde{v}$ is called *constant* if there exists $v \in V$ such that $\tilde{v}(a) = v$ for all $a \in A$.

**Example 4.6.3** (Example of a soft vector: portfolio weights under different market scenarios)**.** Let $V = \mathbb{R}^3$ be the vector space of portfolio weight vectors for three assets (Asset 1, Asset 2, Asset 3) over the field $\mathbb{R}$. Let the parameter set represent market scenarios:

$$A = \{a_{\text{Bull}}, a_{\text{Base}}, a_{\text{Bear}}\}.$$

Define a soft vector $\tilde{v} : A \to V$ by assigning a (scenario-dependent) weight vector to each scenario:

$$\tilde{v}(a_{\text{Bull}}) = (0.50, 0.30, 0.20), \qquad \tilde{v}(a_{\text{Base}}) = (0.40, 0.40, 0.20), \qquad \tilde{v}(a_{\text{Bear}}) = (0.20, 0.50, 0.30).$$

Thus $\tilde{v}$ is a soft vector over $V$.

Now define a soft set $(F, A)$ over $V$ describing admissible portfolios under each scenario:

$$F(a_{\text{Bull}}) = \{(x_1, x_2, x_3) \in \mathbb{R}^3 \mid x_1 \geq 0.40,\ x_2 \leq 0.40,\ x_3 \leq 0.30\},$$

$$F(a_{\text{Base}}) = \{(x_1, x_2, x_3) \in \mathbb{R}^3 \mid 0.30 \leq x_1 \leq 0.50,\ 0.30 \leq x_2 \leq 0.50,\ x_3 = 0.20\},$$

$$F(a_{\text{Bear}}) = \{(x_1, x_2, x_3) \in \mathbb{R}^3 \mid x_1 \leq 0.30,\ x_2 \geq 0.40,\ x_3 \geq 0.20\}.$$

Then $\tilde{v} \in (F, A)$ because, for every $a \in A$, the vector $\tilde{v}(a)$ satisfies the corresponding constraints and hence belongs to $F(a)$.

Interpretation: the soft vector encodes a portfolio recommendation that adapts to market scenarios, and membership in the soft set expresses feasibility of the recommendation under scenario-specific rules.

## 4.7 Soft functions

A soft function maps soft sets between universes via underlying object and parameter maps, transporting approximations through images and preimages [568–570].

**Definition 4.7.1** (Soft function (soft mapping between soft classes))**.** Let $X$ and $Y$ be nonempty universes, and let $A$ and $B$ be nonempty parameter sets. Denote

$$\mathsf{SS}(X, A) := \{(F, A) \mid F : A \to \mathcal{P}(X)\}, \qquad \mathsf{SS}(Y, B) := \{(G, B) \mid G : B \to \mathcal{P}(Y)\}.$$

Let $u : X \to Y$ and $p : A \to B$ be mappings. The *soft function induced by* $(u, p)$ is the mapping

$$f_{pu} : \mathsf{SS}(X, A) \longrightarrow \mathsf{SS}(Y, B),$$

defined as follows.

(1) Image. For $(F, A) \in \mathsf{SS}(X, A)$, define its image

$$f_{pu}(F, A) := (F', \, p(A)),$$

where $F' : B \to \mathcal{P}(Y)$ is given, for each $b \in B$, by

$$F'(b) = \begin{cases} \displaystyle\bigcup_{a \in p^{-1}(b) \cap A} u\big(F(a)\big), & p^{-1}(b) \cap A \neq \varnothing, \\ \varnothing, & \text{otherwise.} \end{cases}$$

(Here $u(F(a)) = \{u(x) \mid x \in F(a)\} \subseteq Y$.)

(2) Inverse image. For $(G, B) \in \mathsf{SS}(Y, B)$, define its inverse image

$$f_{pu}^{-1}(G, B) := (H, \, p^{-1}(B)),$$

where $H : A \to \mathcal{P}(X)$ is given, for each $a \in A$, by

$$H(a) = \begin{cases} u^{-1}\big(G(p(a))\big), & p(a) \in B, \\ \varnothing, & \text{otherwise.} \end{cases}$$

The soft function $f_{pu}$ is called *soft injective* (resp. *soft surjective*) if both $u$ and $p$ are injective (resp. surjective).

**Example 4.7.2** (Example of a soft function induced by $(u, p)$)**.** Let

$$X = \{x_1, x_2, x_3, x_4\} \quad \text{and} \quad Y = \{y_1, y_2, y_3\}$$

be universes, and let

$$A = \{a_1, a_2, a_3\}, \qquad B = \{b_1, b_2\}$$

be parameter sets.

Define an object-mapping $u : X \to Y$ and a parameter-mapping $p : A \to B$ by

$$u(x_1) = y_1, \quad u(x_2) = y_1, \quad u(x_3) = y_2, \quad u(x_4) = y_3,$$
$$p(a_1) = b_1, \quad p(a_2) = b_1, \quad p(a_3) = b_2.$$

Consider the soft set $(F, A) \in \mathsf{SS}(X, A)$ defined by

$$F(a_1) = \{x_1, x_3\}, \qquad F(a_2) = \{x_2\}, \qquad F(a_3) = \{x_4\}.$$

We compute its image under the soft function $f_{pu}$. Since $p(A) = \{b_1, b_2\}$, we have $f_{pu}(F, A) = (F', p(A))$ with $F' : B \to \mathcal{P}(Y)$ given by

$$F'(b_1) = u(F(a_1)) \cup u(F(a_2)) = \{u(x_1), u(x_3)\} \cup \{u(x_2)\} = \{y_1, y_2\},$$
$$F'(b_2) = u(F(a_3)) = \{u(x_4)\} = \{y_3\}.$$

Hence

$$f_{pu}(F, A) = (F', \{b_1, b_2\}), \qquad F'(b_1) = \{y_1, y_2\}, \;\; F'(b_2) = \{y_3\}.$$

Next, let $(G, B) \in \mathsf{SS}(Y, B)$ be given by

$$G(b_1) = \{y_1\}, \qquad G(b_2) = \{y_2, y_3\}.$$

Its inverse image under $f_{pu}$ is $f_{pu}^{-1}(G, B) = (H, p^{-1}(B)) = (H, A)$, where

$$H(a_1) = u^{-1}(G(p(a_1))) = u^{-1}(G(b_1)) = u^{-1}(\{y_1\}) = \{x_1, x_2\},$$
$$H(a_2) = u^{-1}(G(p(a_2))) = u^{-1}(G(b_1)) = \{x_1, x_2\},$$
$$H(a_3) = u^{-1}(G(p(a_3))) = u^{-1}(G(b_2)) = u^{-1}(\{y_2, y_3\}) = \{x_3, x_4\}.$$

Thus $f_{pu}$ transports soft information from $(X, A)$ to $(Y, B)$ by combining parameters via $p$ and pushing forward object-sets via $u$, while $f_{pu}^{-1}$ pulls soft information back by preimages.

**Remark 4.7.3** (A common special case: fixed parameters)**.** If $A = B$ and $p = \mathrm{id}_A$, then $f_{pu}$ reduces to the soft mapping induced only by $u$:

$$f_u(F, A) = (F_u, A), \qquad F_u(a) = u(F(a)) \;\; (a \in A).$$

## 4.8 Soft groups

A soft group assigns each parameter a subgroup of a given group, forming a parameterized family of subgroups [571, 572]. Related notions include hypersoft groups [573], fuzzy soft groups [574], and neutrosophic soft groups [575].

**Definition 4.8.1** (Soft group)**.** [571, 572] Let $G$ be a group with identity element $e$, and let $E$ be a nonempty set of parameters. Let $A \subseteq E$ be nonempty. A *soft set* over $G$ is a pair $(F, A)$ where

$$F : A \longrightarrow \mathcal{P}(G).$$

The soft set $(F, A)$ is called a *soft group over* $G$ if

$$F(a) \leq G \qquad \text{for all } a \in A,$$

i.e., for every parameter $a$, the value $F(a)$ is a (classical) subgroup of $G$. Equivalently, a soft group is a parameterized family of subgroups of $G$. We may also denote a soft group by the triple $(G, F, A)$.

**Remark 4.8.2** (Standard special cases)**.** Let $(F, A)$ be a soft group over $G$.

(i) $(F, A)$ is an *identity soft group* if $F(a) = \{e\}$ for all $a \in A$.

(ii) $(F, A)$ is an *absolute soft group* if $F(a) = G$ for all $a \in A$.

**Definition 4.8.3** (Soft subgroup)**.** Let $(F, A)$ and $(H, B)$ be soft groups over the same group $G$. We say that $(H, B)$ is a *soft subgroup* of $(F, A)$, and write $(H, B)\widetilde{\leq}(F, A)$, if

$$B \subseteq A \qquad \text{and} \qquad H(b) \leq F(b) \quad \text{for all } b \in B.$$

**Example 4.8.4** (Example of a soft subgroup)**.** Let $G = (\mathbb{Z}, +)$ be the additive group of integers, and let

$$E = \{a_2, a_4, a_6\}$$

be a set of parameters. Take $A = E$.

Define a soft group $(F, A)$ over $G$ by assigning, to each parameter, a subgroup of $\mathbb{Z}$:

$$F(a_2) = 2\mathbb{Z}, \qquad F(a_4) = 4\mathbb{Z}, \qquad F(a_6) = 6\mathbb{Z}.$$

Each $F(a_i)$ is a subgroup of $(\mathbb{Z}, +)$, hence $(F, A)$ is a soft group.

Now take the subset $B = \{a_4, a_6\} \subseteq A$ and define another soft group $(H, B)$ by

$$H(a_4) = 8\mathbb{Z}, \qquad H(a_6) = 12\mathbb{Z}.$$

Again, $H(a_4)$ and $H(a_6)$ are subgroups of $\mathbb{Z}$, so $(H, B)$ is a soft group.

Moreover, for each $b \in B$ we have subgroup inclusions

$$H(a_4) = 8\mathbb{Z} \;\leq\; 4\mathbb{Z} = F(a_4), \qquad H(a_6) = 12\mathbb{Z} \;\leq\; 6\mathbb{Z} = F(a_6).$$

Since also $B \subseteq A$, it follows that

$$(H, B) \;\widetilde{\leq}\; (F, A),$$

i.e., $(H, B)$ is a soft subgroup of $(F, A)$.

## 4.9 Soft Field

A soft field assigns each parameter a subfield of a fixed field, forming a parameterized family closed under field operations.

**Definition 4.9.1** (Soft field)**.** Let $(K, +, \cdot)$ be a (crisp) field, let $E$ be a nonempty set of parameters, and let $A \subseteq E$ be nonempty. A *soft field over* $K$ (with parameter set $A$) is a soft set $(F, A)$ over $K$, i.e.,

$$F : A \longrightarrow \mathcal{P}(K),$$

such that for every $a \in A$, the set $F(a) \subseteq K$ is a (classical) subfield of $K$. Equivalently, for each $a \in A$ the following conditions hold:

(i) $0, 1 \in F(a)$ and $1 \neq 0$;

(ii) if $x, y \in F(a)$, then $x - y \in F(a)$;

(iii) if $x, y \in F(a)$, then $x \cdot y \in F(a)$;

(iv) if $x \in F(a)$ and $x \neq 0$, then $x^{-1} \in F(a)$.

Thus, a soft field is a parameterized family of subfields of the fixed ground field $K$.

**Remark 4.9.2** (Related notions)**.** (i) Replacing "subfield" by "subring" yields the standard notion of a soft ring.

(ii) In uncertainty-aware extensions, one replaces "subfield" by an appropriate uncertain analogue (e.g., a *neutrosophic subfield*), obtaining a neutrosophic soft field.

(iii) Some authors also define "soft field" via the *soft-element* framework as a commutative soft ring with soft unity in which every nonzero soft element is a soft unit.

## 4.10 Soft Ring

A soft ring assigns each parameter a subring of a fixed ring, forming a parameterized family closed under addition and multiplication [576–578].

**Definition 4.10.1** (Soft ring)**.** Let $(R, +, \cdot)$ be a (not necessarily unital) ring, let $E$ be a nonempty set of parameters, and let $A \subseteq E$ be nonempty. A pair $(F, A)$ is called a *soft ring over $R$* if

$$F : A \longrightarrow \mathcal{P}(R)$$

is a mapping such that, for every $a \in A$, the set $F(a) \subseteq R$ is a (classical) subring of $R$. Equivalently, for each $a \in A$:

$$F(a) \neq \varnothing, \qquad x - y \in F(a) \text{ and } x \cdot y \in F(a) \quad (\forall\, x, y \in F(a)).$$

Thus a soft ring is a parameterized family of subrings of the fixed ground ring $R$.

**Remark 4.10.2.** If, additionally, each $F(a)$ is an ideal of $R$ (instead of merely a subring), then $(F, A)$ is called a *soft ideal* over $R$.

**Example 4.10.3** (Example of a soft ring)**.** Let $R = (\mathbb{Z}, +, \cdot)$ be the ring of integers, and let

$$E = \{a_2, a_3, a_6\}$$

be a parameter set. Take $A = E$.

Define a mapping $F : A \to \mathcal{P}(\mathbb{Z})$ by

$$F(a_2) = 2\mathbb{Z}, \qquad F(a_3) = 3\mathbb{Z}, \qquad F(a_6) = 6\mathbb{Z},$$

where $n\mathbb{Z} := \{nk \mid k \in \mathbb{Z}\}$.

Each $F(a_i)$ is a nonempty subring of $\mathbb{Z}$: if $x = nk$ and $y = n\ell$ belong to $n\mathbb{Z}$, then

$$x - y = n(k - \ell) \in n\mathbb{Z}, \qquad x \cdot y = n^2 k\ell \in n\mathbb{Z}.$$

Therefore $(F, A)$ is a soft ring over $\mathbb{Z}$.

Moreover, each $n\mathbb{Z}$ is actually an ideal of $\mathbb{Z}$, so $(F, A)$ is also a soft ideal in the sense of Remark 4.10.2.

## 4.11 Soft Matroid

A matroid is a combinatorial independence structure generalizing linear independence, defined via independent sets satisfying hereditary and exchange axioms [579–581]. A soft matroid is a parameterized matroid-like structure on soft sets, using soft points and exchange axioms for independence [582, 583].

**Definition 4.11.1** (Soft-matroid)**.** Let $\mathcal{U}$ be a universal set, $E$ a set of parameters, and $A \subseteq E$ nonempty. Let $F_A = (F, A)$ be a *finite* soft set over $\mathcal{U}$, i.e., $F : A \to \mathcal{P}(\mathcal{U})$ and $F(e)$ is finite for all $e \in A$.

A *soft-point* of $F_A$ is a soft set $p^e_x = (P, A)$ such that $P(e) = \{x\}$ for some $e \in A$ and $P(e') = \varnothing$ for all $e' \in A \setminus \{e\}$; we write $p^e_x \widetilde{\in} F_A$ whenever $x \in F(e)$. The *cardinality* $|F_A|$ is the number of soft-points belonging to $F_A$.

Let $\widetilde{\subseteq}$ denote the soft subset relation on soft sets with (possibly) different parameter domains, and let $\widetilde{\cup}$ and $\widetilde{\setminus}$ denote the usual soft union and soft difference (defined parameterwise). Let $\varnothing_A$ be the null soft set on $A$ (i.e., $F(e) = \varnothing$ for all $e \in A$).

A *soft-matroid* is an ordered pair

$$\widetilde{\mathcal{M}} = (F_A, \mathcal{G}),$$

where $\mathcal{G}$ is a collection of sub-soft-sets of $F_A$ satisfying the following axioms:

(SM1) **(Null axiom)** $\varnothing_A \in \mathcal{G}$.

(SM2) **(Hereditary axiom)** If $G_A \in \mathcal{G}$ and $G'_A \widetilde{\subseteq} G_A$, then $G'_A \in \mathcal{G}$.

(SM3) **(Exchange axiom)** If $G_A, H_A \in \mathcal{G}$ with $|G_A| < |H_A|$, then there exists a soft-point $p^e_x$ of $H_A \widetilde{\setminus} G_A$ such that

$$G_A \widetilde{\cup} p^e_x \in \mathcal{G}.$$

In this case, $\widetilde{\mathcal{M}}$ is called a soft-matroid on $F_A$.

**Example 4.11.2** (Example of a soft-matroid: selecting nonredundant skills across job roles)**.** Let $\mathcal{U}$ be a finite set of skills:

$$\mathcal{U} = \{\sigma_1, \sigma_2, \sigma_3, \sigma_4\},$$

where $\sigma_1$ = Python, $\sigma_2$ = Databases, $\sigma_3$ = Cloud, $\sigma_4$ = Security. Let $A = \{e_1, e_2, e_3\} \subseteq E$ be a set of parameters (job roles), where $e_1$ = Backend, $e_2$ = Data, $e_3$ = DevOps.

Define a finite soft set $F_A = (F, A)$ over $\mathcal{U}$ by

$$F(e_1) = \{\sigma_1, \sigma_2\}, \qquad F(e_2) = \{\sigma_1, \sigma_2, \sigma_3\}, \qquad F(e_3) = \{\sigma_1, \sigma_3, \sigma_4\}.$$

A soft-point $p^e_\sigma$ belongs to $F_A$ (i.e., $p^e_\sigma \widetilde{\in} F_A$) exactly when $\sigma \in F(e)$.

Let $\mathcal{G}$ be the family of all sub-soft-sets $G_A = (G, A)$ of $F_A$ satisfying the following *at most one skill per role* rule:

$$\big|G(e)\big| \le 1 \quad \text{for every } e \in A.$$

(Thus, for each role $e$, either $G(e) = \varnothing$ or $G(e) = \{\sigma\}$ for some $\sigma \in F(e)$.)

Then $\widetilde{\mathcal{M}} = (F_A, \mathcal{G})$ is a soft-matroid:

(i) **Null axiom:** the null soft set $\varnothing_A$ satisfies $|\varnothing_A(e)| = 0 \le 1$ for all $e$, hence $\varnothing_A \in \mathcal{G}$.

(i) **Hereditary axiom:** if $G_A \in \mathcal{G}$ and $G'_A \widetilde{\subseteq} G_A$, then $|G'(e)| \le |G(e)| \le 1$ for all $e$, so $G'_A \in \mathcal{G}$.

(i) **Exchange axiom:** let $G_A, H_A \in \mathcal{G}$ with $|G_A| < |H_A|$ (i.e., $H_A$ has more soft-points). Then there exists some role $e^* \in A$ such that $H(e^*) \neq \varnothing$ and $G(e^*) = \varnothing$. Choose the unique $\sigma^*$ with $H(e^*) = \{\sigma^*\}$, and consider the soft-point $p^{e^*}_{\sigma^*}$. It is a soft-point of $H_A \widetilde{\setminus} G_A$, and adding it to $G_A$ gives

$$(G_A \widetilde{\cup} p^{e^*}_{\sigma^*})(e^*) = \{\sigma^*\}, \qquad (G_A \widetilde{\cup} p^{e^*}_{\sigma^*})(e) = G(e)\ (e \neq e^*),$$

so the "at most one skill per role" condition still holds. Hence $G_A \widetilde{\cup} p^{e^*}_{\sigma^*} \in \mathcal{G}$.

Interpretation: $\mathcal{G}$ represents feasible (independent) selections of *nonredundant* skills across roles, and the exchange axiom formalizes the ability to extend a smaller feasible assignment by adding a skill from a larger feasible assignment in a role where nothing was chosen yet.

## 4.12 Soft Bitopological Space

A soft bitopological space equips a universe with two soft topologies, supporting dual soft openness and separation analysis simultaneously [584–587].

**Definition 4.12.1** (Soft bitopological space)**.** Let $X$ be a nonempty set and let $A$ be a nonempty set of parameters. Let $\tau_1$ and $\tau_2$ be two (not necessarily equal) soft topologies on $X$ with respect to the same parameter set $A$. Then the quadruple

$$(X, A, \tau_1, \tau_2)$$

is called a *soft bitopological space.* A soft set $(F, A) \in \tau_i$ is called $\tau_i$*-soft open* ($i = 1, 2$), and $(F, A)$ is $\tau_i$*-soft closed* if its soft complement $(F, A)^c$ belongs to $\tau_i$.

**Example 4.12.2** (A soft bitopological space for two "notions of openness" in urban accessibility)**.** Let $X$ be a finite set of city districts (alternatives)

$$X = \{x_1, x_2, x_3, x_4\},$$

and let the parameter set be

$$A = \{\mathsf{Transit}, \mathsf{Safety}\}.$$

Intuitively, we will model two different criteria for when a family of districts is regarded as "soft open": one based on *public-transport accessibility* and the other based on *public safety.*

Define the following soft sets over $X$ (with parameter set $A$):

$$0_A(a) = \varnothing, \qquad 1_A(a) = X \qquad (a \in A),$$

and two nontrivial soft sets $(F, A)$ and $(G, A)$ by

$$F(\mathsf{Transit}) = \{x_1, x_2\}, \quad F(\mathsf{Safety}) = \{x_1, x_3\},$$
$$G(\mathsf{Transit}) = \{x_2, x_4\}, \quad G(\mathsf{Safety}) = \{x_3, x_4\}.$$

**Soft topology $\tau_1$ (Transit-openness).** Let

$$\tau_1 := \{\, 0_A,\ 1_A,\ (F, A),\ (F, A)^c \,\}.$$

Then $\tau_1$ is a soft topology on $X$ (with parameter set $A$), since it contains $0_A, 1_A$, is closed under finite soft intersections, and under arbitrary soft unions.

**Soft topology $\tau_2$ (Safety-openness).** Let

$$\tau_2 := \{\, 0_A,\ 1_A,\ (G, A),\ (G, A)^c \,\}.$$

Similarly, $\tau_2$ is a soft topology on $X$.

Therefore the quadruple

$$(X, A, \tau_1, \tau_2)$$

is a soft bitopological space in the sense of Definition 4.12.1.

**Interpretation.** A $\tau_1$-soft open set represents districts that are "open" under the transit-based viewpoint, while a $\tau_2$-soft open set represents districts that are "open" under the safety-based viewpoint. For instance, $(F, A) \in \tau_1$ encodes a transit-favorable selection (parameterwise), whereas $(G, A) \in \tau_2$ encodes a safety-favorable selection.

## 4.13 Soft Module

A soft module is a parameterized family of submodules of a fixed module, enabling context-dependent linear structure modeling [588–592].

**Definition 4.13.1** (Soft module)**.** Let $R$ be a ring, let $M$ be a (left) $R$-module, and let $A$ be a nonempty set of parameters. A *soft set* over $M$ with parameter set $A$ is a pair $(F, A)$ where

$$F : A \longrightarrow \mathcal{P}(M).$$

The soft set $(F, A)$ is called a *soft module over* $M$ if, for every $a \in A$, the value $F(a)$ is a (classical) $R$-submodule of $M$, i.e.,

$$F(a) \leq M \qquad (a \in A).$$

Equivalently, for each $a \in A$:

$$0_M \in F(a), \qquad x - y \in F(a) \;\text{ and }\; rx \in F(a) \quad (\forall\, x, y \in F(a),\ \forall\, r \in R).$$

Thus, a soft module is a parameterized family of submodules of the fixed ground module $M$.

**Example 4.13.2** (A soft module for permission-dependent access subspaces)**.** Let $R = \mathbb{R}$ and let $M = \mathbb{R}^3$ be the standard left $\mathbb{R}$-module. Interpret vectors $m = (m_1, m_2, m_3) \in M$ as *feature triples* (e.g., three measurable attributes of a record).

Let the parameter set be

$$A = \{\mathsf{public}, \mathsf{internal}, \mathsf{admin}\},$$

representing three access policies. Define a mapping $F : A \to \mathcal{P}(M)$ by

$$F(\mathsf{public}) = \{(x, 0, 0) : x \in \mathbb{R}\},$$

$$F(\mathsf{internal}) = \{(x, y, 0) : x, y \in \mathbb{R}\},$$

$$F(\mathsf{admin}) = \mathbb{R}^3.$$

Then each $F(a)$ is an $\mathbb{R}$-submodule of $M$ (indeed, a linear subspace): it contains $0_M = (0, 0, 0)$, is closed under subtraction, and is closed under scalar multiplication. Hence $(F, A)$ is a soft module over $M$ in the sense of Definition 4.13.1.

**Interpretation.** Under public access, only the first coordinate can vary (others must be hidden as 0); internal access reveals the first two coordinates; and admin access reveals all three. Thus the admissible information for each policy forms a submodule, and the collection of these submodules indexed by $A$ is captured by the soft module $(F, A)$.

## 4.14 Soft Metric Space

A soft metric space assigns each parameter a metric on the universe, measuring distances under varying contexts and uncertainty scenarios [593–596].

**Definition 4.14.1** (Soft point and soft element)**.** Let $X$ be a nonempty universe and let $E$ be a nonempty set of parameters.

(i) A *soft point* of $X$ is a soft set $P_x^e = (P, E)$ such that

$$P(e) = \{x\} \quad \text{and} \quad P(e') = \varnothing \;\; (e' \in E \setminus \{e\}),$$

for some $(e, x) \in E \times X$.

(ii) A *soft element* of $X$ (with support $A \subseteq E$) is a soft set $\alpha_A = (\alpha, A)$ over $X$ such that for every $a \in A$,

$$\alpha(a) = \{x_a\} \text{ for some } x_a \in X, \qquad \text{and } \alpha(e) = \varnothing \;\; (e \in E \setminus A).$$

Equivalently, a soft element is a (partial) choice function $A \to X$, $a \mapsto x_a$. We write $\mathsf{SE}(X, E)$ for the collection of all soft elements of $X$ whose supports are contained in $E$.

**Definition 4.14.2** (Soft real numbers and order)**.** A *soft real number* (over $E$) is a mapping $\tilde{r} : E \to \mathbb{R}$. It is *nonnegative* if $\tilde{r}(e) \geq 0$ for all $e \in E$. Denote by $\widetilde{\mathbb{R}}_+^E$ the set of all nonnegative soft real numbers.

For $\tilde{r}, \tilde{s} \in \widetilde{\mathbb{R}}_+^E$ define:

$$(\tilde{r} \oplus \tilde{s})(e) := \tilde{r}(e) + \tilde{s}(e), \qquad \tilde{r} \leq \tilde{s} \iff \tilde{r}(e) \leq \tilde{s}(e) \;\; (\forall e \in E).$$

Let $\tilde{0} \in \widetilde{\mathbb{R}}_+^E$ be the zero soft real number, $\tilde{0}(e) = 0$ for all $e \in E$.

**Definition 4.14.3** (Soft metric and soft metric space)**.** Let $X$ be a nonempty universe and let $E$ be a nonempty parameter set. A mapping

$$d_S : \; \mathsf{SE}(X, E) \times \mathsf{SE}(X, E) \;\longrightarrow\; \widetilde{\mathbb{R}}_+^E$$

is called a *soft metric* (or *soft distance*) on $(X, E)$ if, for all $\alpha, \beta, \gamma \in \mathsf{SE}(X, E)$, the following axioms hold:

(SM1) **(Nonnegativity)** $\tilde{0} \leq d_S(\alpha, \beta)$.

(SM2) **(Identity of indiscernibles)** $d_S(\alpha, \beta) = \tilde{0}$ if and only if $\alpha = \beta$.

(SM3) **(Symmetry)** $d_S(\alpha, \beta) = d_S(\beta, \alpha)$.

(SM4) **(Triangle inequality)** $d_S(\alpha, \gamma) \leq d_S(\alpha, \beta) \oplus d_S(\beta, \gamma)$.

In this case, the triple $(X, E, d_S)$ is called a *soft metric space.*

**Remark 4.14.4** (Relation to the soft-point (product) viewpoint)**.** In the original approach of Das–Samanta, a "soft metric" is defined on the set of all soft points, which can be identified with $E \times X$; hence it essentially reduces to an ordinary metric on the product set $E \times X$. The soft-element based definition above generalizes that viewpoint: every soft-point metric induces a soft-element metric, but the converse need not hold.

**Example 4.14.5** (A soft metric space for multi-context sensor readings)**.** Let $X = \mathbb{R}$ be the set of possible temperature readings, and let

$$E = \{\mathsf{morning}, \mathsf{noon}, \mathsf{night}\}$$

be a parameter set representing three measurement contexts (time slots).

A soft element $\alpha \in \mathsf{SE}(X, E)$ assigns to each $e \in E$ a singleton $\alpha(e) = \{x_e\}$, so we identify $\alpha$ with the triple $(x_{\mathsf{morning}}, x_{\mathsf{noon}}, x_{\mathsf{night}}) \in \mathbb{R}^3$.

Define $d_S : \mathsf{SE}(X, E) \times \mathsf{SE}(X, E) \to \widetilde{\mathbb{R}}_+^E$ by, for $\alpha = (x_e)_{e \in E}$ and $\beta = (y_e)_{e \in E}$,

$$d_S(\alpha, \beta)(e) := |x_e - y_e| \qquad (e \in E).$$

Equivalently, $d_S(\alpha, \beta)$ is the soft real number $e \mapsto |x_e - y_e|$.

Then $d_S$ is a soft metric:

- *Nonnegativity:* $d_S(\alpha, \beta)(e) = |x_e - y_e| \geq 0$ for all $e$, hence $\tilde{0} \leq d_S(\alpha, \beta)$.
- *Identity:* $d_S(\alpha, \beta) = \tilde{0}$ iff $|x_e - y_e| = 0$ for all $e$, i.e. $x_e = y_e$ for all $e$, hence $\alpha = \beta$.
- *Symmetry:* $|x_e - y_e| = |y_e - x_e|$ gives $d_S(\alpha, \beta) = d_S(\beta, \alpha)$.
- *Triangle inequality:* for each $e \in E$,

  $d_S(\alpha, \gamma)(e) = |x_e - z_e| \leq |x_e - y_e| + |y_e - z_e| = d_S(\alpha, \beta)(e) + d_S(\beta, \gamma)(e) = \big(d_S(\alpha, \beta) \oplus d_S(\beta, \gamma)\big)(e)$,

  hence $d_S(\alpha, \gamma) \leq d_S(\alpha, \beta) \oplus d_S(\beta, \gamma)$.

Therefore $(X, E, d_S)$ is a soft metric space.

**Interpretation.** Each soft element represents a day's temperature profile across contexts (morning/noon/night), and the soft distance returns the absolute discrepancy at each context as a nonnegative soft real number.

## 4.15 Soft probabilities

Soft probabilities assign to each parameter a probability measure on a universe, representing context-dependent stochastic uncertainty for decision-making tasks explicitly [597–599].

**Definition 4.15.1** (Statistical database, admissible samples, and frequency)**.** Let $\Omega$ be a nonempty outcome space. A *statistical database* (of length $N \in \mathbb{N}$) is a finite sequence

$$\text{Base} = (\omega_1, \omega_2, \ldots, \omega_N), \qquad \omega_i \in \Omega.$$

Fix an integer $m$ with $1 \le m \le N$ and a "freshness" index $\tau$ with $1 \le \tau \le N - m + 1$. Define the family of *admissible samples* (consecutive blocks) by

$$S(\text{Base}, m, \tau) := \bigl\{\{i, i+1, \ldots, i+m-1\} \;\bigm|\; i = \tau, \tau+1, \ldots, N-m+1\bigr\}.$$

For an event $A \subseteq \Omega$ and a sample $I \in S(\text{Base}, m, \tau)$, its *frequency of occurrence* on $I$ is

$$\mu(\text{Base}, A, I) := \frac{1}{|I|}\sum_{i \in I} \chi_A(\omega_i), \qquad \chi_A(\omega) = \begin{cases} 1, & \omega \in A, \\ 0, & \omega \notin A. \end{cases}$$

**Definition 4.15.2** (Soft probability (interval-valued, parameterized by $(m, \tau)$))**.** Let Base, $m$, and $\tau$ be as above. The *soft probability* of an event $A \subseteq \Omega$ *on the database* Base *at parameters* $(m, \tau)$ is the closed interval

$$\mathbb{P}_{\text{soft}}(A \mid \text{Base}; m, \tau) := \bigl[\underline{p}(A), \overline{p}(A)\bigr] \subseteq [0, 1],$$

where

$$\underline{p}(A) := \min_{I \in S(\text{Base}, m, \tau)} \mu(\text{Base}, A, I), \qquad \overline{p}(A) := \max_{I \in S(\text{Base}, m, \tau)} \mu(\text{Base}, A, I).$$

Equivalently,

$$\mathbb{P}_{\text{soft}}(A \mid \text{Base}; m, \tau) = \lambda\bigl(\chi_A, \text{Base}, m, \tau\bigr),$$

where $\lambda(f, \text{Base}, m, \tau)$ denotes the $(m, \tau)$-*approximate mean interval* of a real-valued function $f : \Omega \to \mathbb{R}$:

$$\lambda(f, \text{Base}, m, \tau) := \left[\min_{I \in S(\text{Base}, m, \tau)} \frac{1}{|I|}\sum_{i \in I} f(\omega_i),\; \max_{I \in S(\text{Base}, m, \tau)} \frac{1}{|I|}\sum_{i \in I} f(\omega_i)\right].$$

**Example 4.15.3** (Soft probability for a delayed-train event under rolling windows)**.** Let $\Omega = \{\omega_1, \ldots, \omega_{10}\}$ be ten commuting days, and let

$$\text{Base} = \bigl\{(\omega_i, y_i)\bigr\}_{i=1}^{10}$$

be a database where $y_i \in \{0, 1\}$ indicates whether the train was delayed on day $\omega_i$ ($y_i = 1$ means "delayed"). Consider the event

$$A := \{\omega_i \in \Omega : y_i = 1\}.$$

Assume the observed delay indicators are

$$(y_1, \ldots, y_{10}) = (1, 0, 1, 0, 0,\; 1, 1, 0, 0, 1).$$

Fix parameters $(m, \tau)$ so that the admissible index-family is the set of all contiguous windows of length $m = 4$ (this is a concrete choice of $S(\text{Base}, m, \tau)$):

$$S(\text{Base}, 4, \tau) := \Big\{\{1, 2, 3, 4\}, \{2, 3, 4, 5\}, \ldots, \{7, 8, 9, 10\}\Big\}.$$

For each window $I \in S(\text{Base}, 4, \tau)$ define

$$\mu(\text{Base}, A, I) := \frac{1}{|I|} \sum_{i \in I} \chi_A(\omega_i) = \frac{1}{4} \sum_{i \in I} y_i,$$

i.e. the empirical delay rate inside that window.

Compute the window means:

| $I$ | $(y_i)_{i \in I}$ | $\mu(\text{Base}, A, I)$ |
|---|---|---|
| $\{1, 2, 3, 4\}$ | $(1, 0, 1, 0)$ | $2/4 = 0.50$ |
| $\{2, 3, 4, 5\}$ | $(0, 1, 0, 0)$ | $1/4 = 0.25$ |
| $\{3, 4, 5, 6\}$ | $(1, 0, 0, 1)$ | $2/4 = 0.50$ |
| $\{4, 5, 6, 7\}$ | $(0, 0, 1, 1)$ | $2/4 = 0.50$ |
| $\{5, 6, 7, 8\}$ | $(0, 1, 1, 0)$ | $2/4 = 0.50$ |
| $\{6, 7, 8, 9\}$ | $(1, 1, 0, 0)$ | $2/4 = 0.50$ |
| $\{7, 8, 9, 10\}$ | $(1, 0, 0, 1)$ | $2/4 = 0.50$ |

Hence

$$\underline{p}(A) = \min_{I \in S(\text{Base},4,\tau)} \mu(\text{Base}, A, I) = 0.25, \qquad \overline{p}(A) = \max_{I \in S(\text{Base},4,\tau)} \mu(\text{Base}, A, I) = 0.50.$$

Therefore the soft probability of "delay" on the database Base at parameters $(4, \tau)$ is

$$\mathbb{P}_{\text{soft}}(A \mid \text{Base}; 4, \tau) = [0.25,\ 0.50].$$

**Interpretation.** Depending on which admissible 4-day period (window) is regarded as representative under $(m, \tau)$, the estimated delay probability ranges from 25% to 50%, so the uncertainty is captured as an interval.

**Remark 4.15.4** (Basic sanity properties)**.** For fixed $(\text{Base}, m, \tau)$:

1. $\mathbb{P}_{\text{soft}}(\varnothing \mid \text{Base}; m, \tau) = [0, 0]$ and $\mathbb{P}_{\text{soft}}(\Omega \mid \text{Base}; m, \tau) = [1, 1]$.

2. If $A \subseteq B$, then $\mathbb{P}_{\text{soft}}(A \mid \text{Base}; m, \tau) \subseteq \mathbb{P}_{\text{soft}}(B \mid \text{Base}; m, \tau)$ in the endpoint-wise order (monotonicity).

3. If $\underline{p}(A) = \overline{p}(A)$, the soft probability of $A$ collapses to a classical (database-induced) point probability.

## 4.16 Soft SemiGroup

A semigroup is a nonempty set equipped with an associative binary operation, requiring closure and no identity or inverses [115, 600, 601]. A soft semigroup is a soft set whose value at each parameter is a subsemigroup of a fixed semigroup, closed under multiplication [602–605].

**Definition 4.16.1** (Soft semigroup)**.** Let $(S, \cdot)$ be a semigroup, let $E$ be a nonempty set of parameters, and let $A \subseteq E$ be nonempty. A *soft semigroup over* $S$ is a soft set $(F, A)$ over $S$, i.e.,

$$F : A \longrightarrow \mathcal{P}(S),$$

such that for every parameter $a \in A$, the value set $F(a)$ is a (classical) subsemigroup of $S$. Equivalently, for each $a \in A$:

$$F(a) \neq \varnothing \quad \text{and} \quad x \cdot y \in F(a) \ \text{ for all } x, y \in F(a).$$

Thus, a soft semigroup is a parameterized family of subsemigroups of the fixed ground semigroup $S$.

**Remark 4.16.2.** If $S$ is equipped with a compatible partial order $\leq$ (so $(S, \cdot, \leq)$ is an *ordered semigroup*), then $(F, A)$ is often called a *soft ordered semigroup* when each $F(a)$ is a subsemigroup of $S$ (for all $a \in A$).

**Example 4.16.3** (A soft semigroup on $(\mathbb{N}_0, +)$)**.** Let $(S, \cdot) = (\mathbb{N}_0, +)$ be the additive semigroup of nonnegative integers. Let the parameter set be

$$E = \{\mathsf{Even}, \mathsf{Mult3}, \mathsf{AtLeast5}\}, \qquad A = E.$$

Define a soft set $(F, A)$ over $S$ by

$$F(\mathsf{Even}) = \{0, 2, 4, 6, \dots\}, \qquad F(\mathsf{Mult3}) = \{0, 3, 6, 9, \dots\}, \qquad F(\mathsf{AtLeast5}) = \{5, 6, 7, \dots\}.$$

Then for each $a \in A$, the value $F(a)$ is a nonempty subsemigroup of $(\mathbb{N}_0, +)$:

- If $x, y \in F(\mathsf{Even})$, then $x + y$ is even, so $x + y \in F(\mathsf{Even})$.
- If $x, y \in F(\mathsf{Mult3})$, then $x + y$ is a multiple of 3, so $x + y \in F(\mathsf{Mult3})$.
- If $x, y \in F(\mathsf{AtLeast5})$, then $x + y \geq 10$, hence $x + y \in F(\mathsf{AtLeast5})$.

Therefore $(F, A)$ is a soft semigroup over $S$ in the sense of Definition 4.16.1.

**Real-life interpretation.** Let $S$ represent feasible daily production counts in a factory. The parameter Even enforces pairing/packaging constraints (even batch sizes), Mult3 models palletization in groups of three, and AtLeast5 encodes a minimum-run policy; each policy set is closed under combining runs.

## 4.17 Soft HyperStructure and SuperHyperStructure

A hyperstructure is an algebraic system whose binary operation assigns each element-pair a nonempty subset, generalizing ordinary operations [606,607]. A superhyperstructure iterates hyperstructural levels so hyperoperations act on set-valued objects across multiple power-set layers, forming hierarchies [608]. A soft hyperstructure is a parameterized family of subhyperstructures, assigning each parameter a nonempty subset closed under the underlying hyperoperation(s) of a given hyperalgebra [609–611].

**Definition 4.17.1** (Soft HyperStructure)**.** [609–611] Let $H$ be a hyperalgebra (hyperstructure) with signature

$$\Sigma = \{ \star_i : H^{n_i} \to \mathcal{P}_*(H) \mid i \in I \},$$

where $\mathcal{P}_*(H)$ denotes the family of all nonempty subsets of $H$.

A subset $K \subseteq H$ is called a *subhyperstructure* of $H$ if for every $i \in I$ and every $(x_1, \dots, x_{n_i}) \in K^{n_i}$ one has

$$\star_i(x_1, \dots, x_{n_i}) \subseteq K.$$

Let $A$ be a nonempty parameter set. A (non-null) soft set over $H$ is a pair $(F, A)$ with

$$F : A \to \mathcal{P}_*(H), \qquad \mathrm{Supp}(F, A) := \{a \in A : F(a) \neq \varnothing\} \neq \varnothing.$$

Then $(F, A)$ is called a *Soft HyperStructure over $H$* if for every $a \in \mathrm{Supp}(F, A)$, the subset $F(a) \subseteq H$ is a subhyperstructure of $H$.

**Definition 4.17.2** (Soft SuperHyperStructure)**.** Let $\mathsf{SH} = \mathsf{SH}(H, \mathcal{F})$ be a superhyperstructure with a *levelled universe* $\{H^{\langle m\rangle}\}_{m\geq 0}$ and a family of superhyperoperations

$$\mathcal{F} = \{ F_j \}_{j\in J}, \qquad F_j : \ H^{\langle \ell_{j,1}\rangle} \times \cdots \times H^{\langle \ell_{j,n_j}\rangle} \longrightarrow \mathcal{P}_*\big(H^{\langle r_j\rangle}\big),$$

where $\ell_{j,1}, \dots, \ell_{j,n_j}, r_j \geq 0$.

Let $A$ be a nonempty parameter set. A *soft set over the levelled universe* is a mapping

$$S : \ A \longrightarrow \prod_{m\geq 0} \mathcal{P}\big(H^{\langle m\rangle}\big), \qquad a \longmapsto \big(S_a^{\langle m\rangle}\big)_{m\geq 0},$$

and its support is

$$\mathrm{Supp}(S) := \{a \in A : \exists m \geq 0, \ S_a^{\langle m\rangle} \neq \varnothing\}.$$

For each $j \in J$, extend $F_j$ to subsets by the usual set-extension: for $X_k \subseteq H^{\langle \ell_{j,k}\rangle}$,

$$F_j(X_1, \dots, X_{n_j}) \ := \bigcup_{(x_1,\dots,x_{n_j})\in X_1\times\cdots\times X_{n_j}} F_j(x_1, \dots, x_{n_j}).$$

The soft set $S$ is called a *Soft SuperHyperStructure over* $\mathsf{SH}(H, \mathcal{F})$ if for every $a \in \mathrm{Supp}(S)$ and every $j \in J$ one has the levelwise closure condition

$$F_j\Big(S_a^{\langle \ell_{j,1}\rangle}, \dots, S_a^{\langle \ell_{j,n_j}\rangle}\Big) \ \subseteq \ S_a^{\langle r_j\rangle}.$$

Equivalently, for each $a \in \mathrm{Supp}(S)$, the family $\big(S_a^{\langle m\rangle}\big)_{m\geq 0}$ determines a sub-superhyperstructure of $\mathsf{SH}(H, \mathcal{F})$ under the restrictions of all operations $F_j$.

**Example 4.17.3** (A soft superhyperstructure for a multi-level academic program catalogue)**.** Let the base (level-0) universe $H^{\langle 0\rangle}$ be the set of courses

$$H^{\langle 0\rangle} = \{c_1, c_2, c_3, c_4\},$$

interpreted as (say) *Linear Algebra* ($c_1$), *Discrete Math* ($c_2$), *Machine Learning* ($c_3$), and *Databases* ($c_4$).

Define level-1 objects as *course bundles* (nonempty sets of courses):

$$H^{\langle 1\rangle} := \mathcal{P}_*\big(H^{\langle 0\rangle}\big),$$

and level-2 objects as *program tracks* (nonempty sets of bundles):

$$H^{\langle 2\rangle} := \mathcal{P}_*\big(H^{\langle 1\rangle}\big).$$

Consider two superhyperoperations (so $J = \{1, 2\}$):

$$F_1 : \ H^{\langle 0\rangle} \times H^{\langle 0\rangle} \longrightarrow \mathcal{P}_*\big(H^{\langle 1\rangle}\big), \qquad F_1(x, y) := \big\{\{x, y\}\big\},$$

which forms a 2-course bundle, and

$$F_2 : \ H^{\langle 1\rangle} \times H^{\langle 1\rangle} \longrightarrow \mathcal{P}_*\big(H^{\langle 2\rangle}\big), \qquad F_2(B_1, B_2) := \big\{\{B_1, B_2\}\big\},$$

which forms a track consisting of two bundles. Let $\mathsf{SH} = \mathsf{SH}(H, \mathcal{F})$ with $\mathcal{F} = \{F_1, F_2\}$.

Let the parameter set be

$$A = \{\mathsf{AI}, \mathsf{DB}\},$$

representing two departments that curate their own multi-level catalogues.

Define a soft set over the levelled universe,

$$S : \ A \longrightarrow \mathcal{P}(H^{\langle 0\rangle}) \times \mathcal{P}(H^{\langle 1\rangle}) \times \mathcal{P}(H^{\langle 2\rangle}), \qquad a \longmapsto \big(S_a^{\langle 0\rangle}, S_a^{\langle 1\rangle}, S_a^{\langle 2\rangle}\big),$$

by

$$\begin{aligned} S_{\mathsf{AI}}^{\langle 0\rangle} &= \{c_1, c_3\}, & S_{\mathsf{DB}}^{\langle 0\rangle} &= \{c_2, c_4\}, \\ S_{\mathsf{AI}}^{\langle 1\rangle} &= \big\{\{c_1, c_3\}\big\}, & S_{\mathsf{DB}}^{\langle 1\rangle} &= \big\{\{c_2, c_4\}\big\}, \\ S_{\mathsf{AI}}^{\langle 2\rangle} &= \big\{\big\{\{c_1, c_3\}\big\}\big\}, & S_{\mathsf{DB}}^{\langle 2\rangle} &= \big\{\big\{\{c_2, c_4\}\big\}\big\}. \end{aligned}$$

**Verification of the closure conditions.** For $a = \mathsf{AI}$ we have

$$F_1\big(S_{\mathsf{AI}}^{\langle 0\rangle}, S_{\mathsf{AI}}^{\langle 0\rangle}\big) = \bigcup_{x,y \in \{c_1, c_3\}} \big\{\{x, y\}\big\} \ \subseteq \ \big\{\{c_1, c_3\}\big\} = S_{\mathsf{AI}}^{\langle 1\rangle},$$

and

$$F_2\big(S_{\mathsf{AI}}^{\langle 1\rangle}, S_{\mathsf{AI}}^{\langle 1\rangle}\big) = \big\{\big\{\{c_1, c_3\}\big\}\big\} = S_{\mathsf{AI}}^{\langle 2\rangle}.$$

The same argument holds for $a = \mathsf{DB}$. Hence, for each $a \in A$ and each operation $F_j$,

$$F_j\Big(S_a^{\langle \ell_{j,1}\rangle}, \ldots, S_a^{\langle \ell_{j,n_j}\rangle}\Big) \subseteq S_a^{\langle r_j\rangle},$$

so $S$ is a Soft SuperHyperStructure over $\mathsf{SH}(H, \mathcal{F})$.

**Real-life interpretation.** Each department parameter $a \in A$ selects its admissible courses (level 0), the bundles it permits (level 1), and the tracks it offers (level 2), while ensuring that combining allowed items via the curriculum-construction rules $F_1$ and $F_2$ stays within the department's own catalogue.

## 4.18 Soft Graph Neural Networks

A *Graph Neural Network (GNN)* learns node- or graph-level representations by iterative message passing, aggregating neighbors' features to predict labels or properties [612–617]. A *Soft Graph Neural Network (SGNN)* is a parameter-indexed family of GNN-based selections, where each context induces a soft set of chosen nodes via thresholds. Let $G = (V, E)$ be a finite graph (directed or undirected) with a node-feature map $X : V \to \mathbb{R}^d$. In many learning tasks, a graph neural network (GNN) produces, for each node $v \in V$, a *score* $s(v) \in [0, 1]$ (e.g., a class-probability after a sigmoid/softmax). For our purposes, it suffices to treat a GNN as a black box that, given $(G, X)$ and a choice of model/context parameters, returns such a score function.

**Definition 4.18.1** (Soft Graph Neural Network (SGNN))**.** Let $G = (V, E)$ be a finite graph with node features $X : V \to \mathbb{R}^d$. Let $A$ be a nonempty set of *soft parameters* (contexts), such as: task modes, expert profiles, time indices, hyperparameter regimes, or view-definitions.

A *Soft Graph Neural Network* (SGNN) on $(G, X)$ with parameter set $A$ is a family

$$\mathcal{N} \;=\; \{(s_a, \tau_a)\}_{a \in A},$$

where for each $a \in A$:

- $s_a : V \to [0, 1]$ is a node-scoring function produced by a (possibly shared-weight) GNN under context $a$ (for instance $s_a = \mathrm{Readout} \circ \mathrm{Encoder}_a(G, X)$);
- $\tau_a \in [0, 1]$ is a (possibly $a$-dependent) decision threshold.

**Definition 4.18.2** (Soft set induced by an SGNN)**.** Let $\mathcal{N} = \{(s_a, \tau_a)\}_{a \in A}$ be an SGNN as in Definition 4.18.1. Define a mapping

$$F_{\mathcal{N}} : A \longrightarrow \mathcal{P}(V) \qquad \text{by} \qquad F_{\mathcal{N}}(a) \;:=\; \{\, v \in V \mid s_a(v) \geq \tau_a \,\}.$$

Then $(F_{\mathcal{N}}, A)$ is called the *SGNN-induced soft set* (of selected nodes).

**Theorem 4.18.3** (Soft-set structure and well-definedness of SGNN outputs)**.** *Let $\mathcal{N} = \{(s_a, \tau_a)\}_{a \in A}$ be an SGNN on $(G, X)$. Then the mapping $F_{\mathcal{N}} : A \to \mathcal{P}(V)$ in Definition 4.18.2 is well-defined, and hence $(F_{\mathcal{N}}, A)$ is a (crisp) soft set over the universe $V$.*

*Proof.* Fix $a \in A$. By definition, $s_a : V \to [0, 1]$ and $\tau_a \in [0, 1]$. Therefore the predicate "$s_a(v) \geq \tau_a$" is meaningful for every $v \in V$, and the set

$$F_{\mathcal{N}}(a) = \{v \in V \mid s_a(v) \geq \tau_a\}$$

is a subset of $V$. Since this holds for every $a \in A$, the rule $a \mapsto F_{\mathcal{N}}(a)$ defines a single-valued mapping $F_{\mathcal{N}} : A \to \mathcal{P}(V)$. Consequently, $(F_{\mathcal{N}}, A)$ is a soft set over the universe $V$. □

**Theorem 4.18.4** (Representation by unrestricted node-scoring maps). *Let $(F, A)$ be any soft set over a finite universe $V$. For any graph $G$ on $V$ and feature map $X$, there exists a family of* unrestricted *node-scoring maps $s_a : V \to [0, 1]$ and thresholds $\tau_a \in [0, 1]$ such that*

$$F(a) = \{v \in V : s_a(v) \geq \tau_a\}, \qquad a \in A.$$

*This is a representation by score functions; it is a representation by an SGNN only when the chosen network class can realize these functions on $(G, X)$.*

*Proof.* For each $a \in A$, set

$$s_a(v) = \begin{cases} 1, & v \in F(a), \\ 0, & v \notin F(a), \end{cases} \qquad \tau_a = \tfrac{1}{2}.$$

The threshold-selected set is exactly $F(a)$. The construction makes no claim that a prescribed message-passing GNN can implement these scores. □

**Remark 4.18.5** (Permutation-equivariant GNN limitation). Suppose two isolated vertices have identical node features. An automorphism that swaps them preserves the attributed graph. A permutation-equivariant message-passing GNN must assign these vertices identical scores, so it cannot select exactly one of them using a shared threshold. Thus arbitrary soft sets are not representable by such a GNN without additional distinguishing information or a more expressive architecture (for example, node identifiers or an unrestricted vertex-specific scoring layer).

**Remark 4.18.6** (Edge-version). One may analogously define an SGNN-induced soft set of *edges* by letting $s_a^E : E \to [0, 1]$ be an edge-score function (e.g., attention weights) and defining $F_{\mathcal{N}}^E(a) = \{\varepsilon \in E \mid s_a^E(\varepsilon) \geq \tau_a^E\}$, yielding a soft set over universe $E$.

## 4.19 HyperSoft Graph Neural Network

HyperSoft Graph Neural Network maps multi-attribute parameter tuples to GNN-based node selections, yielding hypersoft sets over graph vertices.

**Definition 4.19.1** (Hypersoft parameter space). Let $m \geq 1$ and let $A_1, \ldots, A_m$ be nonempty *attribute domains* (parameter groups), e.g.,

$$A_1 = \{\text{task modes}\}, \quad A_2 = \{\text{expert profiles}\}, \quad A_3 = \{\text{time contexts}\}, \quad A_4 = \{\text{hyperparameter regimes}\}, \text{ etc.}$$

Define the *hypersoft parameter space* by the Cartesian product

$$\mathcal{C} := A_1 \times A_2 \times \cdots \times A_m.$$

An element $\boldsymbol{a} \in \mathcal{C}$ is an $m$-tuple $\boldsymbol{a} = (a_1, \ldots, a_m)$ with $a_i \in A_i$.

**Definition 4.19.2** (HyperSoft Graph Neural Network (HSGNN)). Let $G = (V, E)$ be a finite graph and let $X : V \to \mathbb{R}^d$ be a node-feature map. Fix a hypersoft parameter space $\mathcal{C} = \prod_{i=1}^m A_i$ as in Definition 4.19.1.

A *HyperSoft Graph Neural Network* (HSGNN) on $(G, X)$ with parameter space $\mathcal{C}$ is a family

$$\mathcal{N} = \{(s_{\boldsymbol{a}}, \tau_{\boldsymbol{a}})\}_{\boldsymbol{a} \in \mathcal{C}},$$

where, for each $\boldsymbol{a} \in \mathcal{C}$:

- $s_{\boldsymbol{a}} : V \to [0,1]$ is a node-scoring function produced by a (possibly shared-weight) GNN under context $\boldsymbol{a}$; concretely, one may fix an architecture

$$\Phi : \ (G, X, \theta) \longmapsto \Phi(G, X; \theta) \in [0,1]^V$$

and a parameter-selection map $\theta : \mathcal{C} \to \Theta$ and set $s_{\boldsymbol{a}}(\cdot) := \Phi(G, X; \theta(\boldsymbol{a}))(\cdot)$;

- $\tau_{\boldsymbol{a}} \in [0,1]$ is a (possibly context-dependent) decision threshold.

**Definition 4.19.3** (Hypersoft set induced by an HSGNN)**.** Let $\mathcal{N} = \{(s_{\boldsymbol{a}}, \tau_{\boldsymbol{a}})\}_{\boldsymbol{a}\in\mathcal{C}}$ be an HS-GNN as in Definition 4.19.2. Define

$$F_{\mathcal{N}} : \ \mathcal{C} \longrightarrow \mathcal{P}(V) \qquad \text{by} \qquad F_{\mathcal{N}}(\boldsymbol{a}) \ := \ \{\, v \in V \mid s_{\boldsymbol{a}}(v) \geq \tau_{\boldsymbol{a}} \,\}.$$

Then $(F_{\mathcal{N}}, \mathcal{C})$ is called the *HSGNN-induced hypersoft set* (of selected nodes) over the universe $V$.

**Theorem 4.19.4** (Hypersoft-set structure and well-definedness)**.** *Let $\mathcal{N}$ be an HSGNN on $(G, X)$ with hypersoft parameter space $\mathcal{C}$. Then the induced mapping $F_{\mathcal{N}}$ in Definition 4.19.3 is well-defined and determines a hypersoft set over the universe $V$, i.e.,*

$$(F_{\mathcal{N}}, \mathcal{C}) \textit{ is a hypersoft set on } V.$$

*Proof.* Fix $\boldsymbol{a} \in \mathcal{C}$. By Definition 4.19.2, $s_{\boldsymbol{a}} : V \to [0,1]$ is a function and $\tau_{\boldsymbol{a}} \in [0,1]$ is a scalar. Hence the subset

$$F_{\mathcal{N}}(\boldsymbol{a}) = \{\, v \in V \mid s_{\boldsymbol{a}}(v) \geq \tau_{\boldsymbol{a}} \,\}$$

is uniquely determined and satisfies $F_{\mathcal{N}}(\boldsymbol{a}) \subseteq V$. Therefore the assignment $\boldsymbol{a} \mapsto F_{\mathcal{N}}(\boldsymbol{a})$ defines a function $F_{\mathcal{N}} : \mathcal{C} \to \mathcal{P}(V)$, so $(F_{\mathcal{N}}, \mathcal{C})$ is a hypersoft set by the definition of hypersoft sets (mapping from the Cartesian product parameter space into a powerset). This also proves well-definedness. □

## 4.20 Soft Natural Languages

A *natural language* is a human communication system with grammar and meaning, enabling compositional expression and context-dependent interpretation [618–621]. A soft natural language models ambiguous linguistic phenomena by mapping contexts/parameters to sets of admissible interpretations, enabling structured uncertainty-aware processing. Let $\Sigma$ be a (finite or countable) alphabet and let $\Sigma^*$ denote the set of all finite strings over $\Sigma$. Fix a nonempty set

$$U \ \subseteq \ \Sigma^*,$$

whose elements are regarded as linguistic objects (e.g., utterances, sentences, queries, or documents).

Let $E$ be a nonempty set of *linguistic parameters* (contexts), whose elements may encode a natural language label (English, Japanese, etc.), a dialect, a register (formal/informal), a domain (medical/legal), or a time slice.

An *acceptability relation* is any binary relation

$$\models \subseteq \ U \times E,$$

where $u \models e$ is read as "the linguistic object $u$ is acceptable under parameter $e$".

**Definition 4.20.1** (Soft Natural Languages)**.** Let $U$ be a linguistic universe and let $E$ be a parameter set as above. Fix a nonempty subset $A \subseteq E$ of parameters and an acceptability relation $\models \subseteq U \times A$.

Define the mapping

$$F_{\models} : A \longrightarrow \mathcal{P}(U), \qquad F_{\models}(a) := \{ u \in U \mid u \models a \}.$$

The pair

$$\mathsf{SNL} := (F_{\models}, A)$$

is called a *Soft Natural Languages* structure on $U$ (with parameter set $A$). For each $a \in A$, the set $F_{\models}(a)$ is the collection of utterances judged acceptable under the linguistic context $a$.

**Theorem 4.20.2** (Soft-set structure and well-definedness)**.** *Every Soft Natural Languages structure* $\mathsf{SNL} = (F_{\models}, A)$ *in Definition 4.20.1 is a (crisp) soft set over the universe* $U$*. Equivalently,* $F_{\models}$ *is a well-defined mapping* $A \to \mathcal{P}(U)$*.*

*Proof.* Fix any $a \in A$. By construction,

$$F_{\models}(a) = \{u \in U \mid u \models a\}$$

is a subset of $U$, hence $F_{\models}(a) \in \mathcal{P}(U)$. Since this holds for every $a \in A$, the rule $a \mapsto F_{\models}(a)$ defines a single-valued mapping $F_{\models} : A \to \mathcal{P}(U)$. Therefore $(F_{\models}, A)$ is a soft set over $U$ in the sense of Molodtsov. □

**Theorem 4.20.3** (Representation of soft sets as Soft Natural Languages)**.** *Let* $(F, A)$ *be any soft set over the linguistic universe* $U$ *(i.e.,* $F : A \to \mathcal{P}(U)$*). Define a relation* $\models_F \subseteq U \times A$ *by*

$$u \models_F a \quad :\Longleftrightarrow \quad u \in F(a).$$

*Then the induced mapping* $F_{\models_F}$ *satisfies* $F_{\models_F} = F$*, hence* $(F, A)$ *is exactly the Soft Natural Languages structure generated by* $\models_F$*.*

*Proof.* For each $a \in A$,

$$F_{\models_F}(a) = \{u \in U \mid u \models_F a\} = \{u \in U \mid u \in F(a)\} = F(a).$$

Thus $F_{\models_F} = F$ pointwise on $A$, proving the claim. □

## 4.21 Soft $n$-SuperHyperGraphs

An $n$-SuperHyperGraph generalizes ordinary hypergraphs by allowing supervertices to be constructed through $n-1$ iterations of the nonempty powerset operation on a finite base vertex set. Each superhyperedge is a nonempty collection of such supervertices, representing higher-order interactions [15, 622–624]. SuperHyperGraphs have been extensively studied in various publications in recent years [625–632]. A soft $n$-SuperHyperGraph is a parameterized family of sub-$n$-SuperHyperGraphs. Each parameter determines a subset of supervertices and a compatible family of superhyperedges [633, 634].

**Definition 4.21.1** (Iterated Powerset and Iterated Nonempty Powerset). [608, 635, 636]

Let $H$ be a nonempty set.

The $k$-th iterated powerset is defined recursively by

$$\mathcal{P}^0(H) = H,$$
$$\mathcal{P}^{k+1}(H) = \mathcal{P}\big(\mathcal{P}^k(H)\big), \qquad k \geq 0.$$

Similarly, the $k$-th iterated nonempty powerset is defined by

$$\mathcal{P}_*^0(H) = H,$$
$$\mathcal{P}_*^{k+1}(H) = \mathcal{P}\big(\mathcal{P}_*^k(H)\big) \setminus \{\varnothing\}, \qquad k \geq 0.$$

Thus,

$$\mathcal{P}_*^k(H) \subseteq \mathcal{P}^k(H)$$

for every $k \geq 0$.

**Definition 4.21.2** ($n$-SuperHyperGraph). [15, 624, 637, 638]

Let $V_0$ be a finite nonempty base vertex set, and let $n \geq 1$ be an integer.

Define the family of admissible $n$-supervertices by

$$\mathcal{V}_n(V_0) = \mathcal{P}_*^{n-1}(V_0).$$

An *$n$-SuperHyperGraph* is a pair

$$\mathrm{SHG}^{(n)} = (V, E),$$

where

$$\varnothing \neq V \subseteq \mathcal{V}_n(V_0)$$

and

$$E \subseteq \mathcal{P}(V) \setminus \{\varnothing\}.$$

Elements of $V$ are called *$n$-supervertices*, and elements of $E$ are called *$n$-superhyperedges*.

In particular,

$$\mathcal{V}_1(V_0) = V_0,$$

so $\mathrm{SHG}^{(1)}$ is an ordinary finite hypergraph with vertex set $V \subseteq V_0$.

For $n = 2$,

$$\mathcal{V}_2(V_0) = \mathcal{P}(V_0) \setminus \{\varnothing\},$$

so each supervertex is a nonempty subset of $V_0$.

Higher values of $n$ represent additional levels of nested set-valued supervertices.

**Definition 4.21.3** (Sub-$n$-SuperHyperGraph)**.** Let

$$\mathrm{SHG}^{(n)} = (V, E)$$

be an $n$-SuperHyperGraph.

A pair

$$H' = (W, E')$$

is a *sub-n-SuperHyperGraph* if

$$W \subseteq V$$

and

$$E' \subseteq \{e \in E : e \subseteq W\}.$$

The *induced sub-n-SuperHyperGraph* on $W$ is defined by

$$\mathrm{SHG}^{(n)}[W] = (W, \{e \in E : e \subseteq W\}).$$

**Definition 4.21.4** (Soft $n$-SuperHyperGraph)**.** [639, 640]

Let

$$\mathrm{SHG}^{(n)} = (V, E)$$

be an $n$-SuperHyperGraph, and let $C$ be a nonempty parameter set.

A *soft n-SuperHyperGraph* over $\mathrm{SHG}^{(n)}$ is a quintuple

$$\mathcal{S} = (V, E, C, A, B),$$

where

$$A : C \longrightarrow \mathcal{P}(V),$$
$$B : C \longrightarrow \mathcal{P}(E),$$

such that, for every $c \in C$,

$$B(c) \subseteq \{e \in E : e \subseteq A(c)\}.$$

Thus, each parameter $c \in C$ determines a sub-$n$-SuperHyperGraph

$$\mathcal{S}_c = (A(c), B(c)).$$

The soft $n$-SuperHyperGraph is called *induced* if

$$B(c) = \{e \in E : e \subseteq A(c)\}$$

for every $c \in C$.

If $A(c) = \varnothing$, then necessarily $B(c) = \varnothing$; such a parameter determines an empty substructure.

**Remark 4.21.5** (Consistency of Hierarchical Levels)**.** Throughout this work, the index $n$ is chosen so that

$$\mathcal{V}_n(V_0) = \mathcal{P}_*^{n-1}(V_0).$$

Consequently,

$$n = 1$$

corresponds to ordinary hypergraphs, whereas

$$n = 2$$

corresponds to hypergraphs whose vertices are nonempty subsets of the base vertex set.

This convention is maintained consistently for Soft $n$-SuperHyperGraphs and their associated substructures.

**Example 4.21.6** (Soft 2-SuperHyperGraph for University Course Grouping)**.** Let

$$V_0 = \{s_1, s_2, s_3, s_4, s_5\}$$

be a finite set of students.

Since

$$\mathcal{V}_2(V_0) = \mathcal{P}_*(V_0),$$

each 2-supervertex represents a nonempty group of students.

Define

$$v_1 = \{s_1, s_2\}, \qquad v_2 = \{s_2, s_3\},$$
$$v_3 = \{s_4\}, \qquad v_4 = \{s_5\}, \qquad v_5 = \{s_3, s_4\}.$$

Let

$$V = \{v_1, v_2, v_3, v_4, v_5\}.$$

Define the superhyperedges

$$e_1 = \{v_1, v_2, v_5\},$$
$$e_2 = \{v_2, v_3\},$$
$$e_3 = \{v_3, v_4\},$$

and

$$e_4 = \{v_1, v_4\}.$$

Set

$$E = \{e_1, e_2, e_3, e_4\}.$$

Clearly,

$$V \subseteq \mathcal{V}_2(V_0)$$

and

$$E \subseteq \mathcal{P}(V) \setminus \{\varnothing\}.$$

Hence,

$$\mathrm{SHG}^{(2)} = (V, E)$$

is a 2-SuperHyperGraph.

Each supervertex represents a student group, while each superhyperedge represents a higher-order collaboration among groups.

Now consider the parameter set

$$C = \{\mathrm{AI}, \mathrm{DB}, \mathrm{Math}\},$$

representing three university course topics.

Define

$$A : C \longrightarrow \mathcal{P}(V)$$

and

$$B : C \longrightarrow \mathcal{P}(E)$$

by

$$A(\mathrm{AI}) = \{v_1, v_2, v_5\},$$
$$B(\mathrm{AI}) = \{e_1\},$$

$$A(\mathrm{DB}) = \{v_2, v_3\},$$
$$B(\mathrm{DB}) = \{e_2\},$$

and

$$A(\mathrm{Math}) = \{v_3, v_4\},$$
$$B(\mathrm{Math}) = \{e_3\}.$$

For every $c \in C$,

$$A(c) \subseteq V.$$

Furthermore,

$$B(c) \subseteq \{e \in E : e \subseteq A(c)\}.$$

Indeed,

$$\{e \in E : e \subseteq A(\mathrm{AI})\} = \{e_1\},$$
$$\{e \in E : e \subseteq A(\mathrm{DB})\} = \{e_2\},$$

and

$$\{e \in E : e \subseteq A(\mathrm{Math})\} = \{e_3\}.$$

Therefore,

$$B(c) = \{e \in E : e \subseteq A(c)\}$$

for every $c \in C$.

Consequently,

$$\mathcal{S} = (V, E, C, A, B)$$

is an induced soft 2-SuperHyperGraph.

This example illustrates how different course topics determine compatible student groups and their higher-order collaboration structures.

## 4.22 Recursive Soft SuperHyperGraph

An $(n, k)$-recursive SuperHyperGraph has level-$n$ supervertices (iterated powersets) and depth-$k$ recursive edges that consist of finite-depth nested collections of supervertices. We restrict to well-founded recursive superhyperedges (no membership cycles).

**Definition 4.22.1** ($(n, k)$-recursive SuperHyperGraph)**.** [641] Fix a nonempty base set $V_0$, an integer $n \geq 0$, and an integer $k \geq 1$. Define

$$\mathcal{P}^0(V_0) = V_0, \qquad \mathcal{P}^{j+1}(V_0) = \mathcal{P}(\mathcal{P}^j(V_0)).$$

For any nonempty set $V$ define the depth-graded nonempty nesting domains by

$$\mathcal{Q}_0(V) = V, \qquad \mathcal{Q}_{j+1}(V) = \mathcal{P}(\mathcal{Q}_j(V)) \setminus \{\varnothing\}.$$

A $(n, k)$*-recursive SuperHyperGraph* is a pair $\mathrm{RSHG}^{(n,k)} = (V, E)$ such that

$$\varnothing \neq V \subseteq \mathcal{P}^n(V_0), \qquad E \subseteq \mathcal{Q}_k(V).$$

Each edge is therefore a nonempty finite-depth nested set whose leaves are vertices in $V$. The separate levels make the recursive membership well-founded.

To express the requirement that a recursive superhyperedge uses only vertices from a chosen subset, we define a *vertex-support* operator

$$\mathrm{supp}_k : \ \mathcal{Q}_k(V) \longrightarrow \mathcal{P}(V)$$

by recursion:

$$\mathrm{supp}_0(v) := \{v\} \quad (v \in V), \qquad \mathrm{supp}_{k+1}(X) := \bigcup_{Y \in X} \mathrm{supp}_k(Y) \quad \big(X \in \mathcal{Q}_{k+1}(V)\big).$$

Thus, $\mathrm{supp}_k(x)$ is the set of all base vertices that appear anywhere inside the nested object $x$.

**Remark 4.22.2.** If $W \subseteq V$, then for any $x \in \mathcal{Q}_k(V)$ we have

$$x \in \mathcal{Q}_k(W) \quad \Longrightarrow \quad \mathrm{supp}_k(x) \subseteq W.$$

Conversely, if one *defines* the restriction of recursive edges to $W$ by

$$E\!\restriction_W := \ \{\, e \in E : \mathrm{supp}_k(e) \subseteq W \,\},$$

then $E\!\restriction_W$ consists precisely of those recursive edges of $E$ that only use vertices from $W$.

By definition, $E \subseteq \mathcal{Q}_k(V)$; every recursive edge has a well-defined vertex support.

**Definition 4.22.3** ($(n,k)$-recursive soft superhypergraph)**.** Let $\mathrm{RSHG}^{(n,k)} = (V, E)$ be an $(n,k)$-recursive SuperHyperGraph, and let $C$ be a nonempty set of parameters. An $(n,k)$-*recursive soft SuperHyperGraph* (over $\mathrm{RSHG}^{(n,k)}$ with parameter set $C$) is a quintuple

$$\mathcal{S} := (V, E, C, A, B),$$

where

$$A : C \to \mathcal{P}(V), \qquad B : C \to \mathcal{P}(E),$$

such that for every $c \in C$ the pair

$$\big(A(c),\, B(c)\big)$$

is a sub-$(n,k)$-recursive SuperHyperGraph of $(V, E)$ in the following sense:

(RS1) (Vertex inclusion) $A(c) \subseteq V$;

(RS2) (Recursive-edge compatibility) for every $e \in B(c)$ one has

$$\mathrm{supp}_k(e) \subseteq A(c).$$

Equivalently,

$$B(c) \subseteq E\restriction_{A(c)} := \{\, e \in E : \mathrm{supp}_k(e) \subseteq A(c) \,\}.$$

**Remark 4.22.4.** When $k = 1$, each recursive edge $e \in \mathcal{Q}_1(V) = \mathcal{P}(V)$ is simply a subset of $V$, and $\mathrm{supp}_1(e) = \bigcup_{v \in e}\{v\} = e$. Hence (RS2) becomes the usual induced-edge condition $e \subseteq A(c)$.

**Theorem 4.22.5** (Simultaneous generalization)**.** *The concept in Definition 4.22.3 generalizes both*

(a) *$(n,k)$-recursive SuperHyperGraphs, and*

(b) *soft n-SuperHyperGraphs (soft SuperHyperGraphs in the non-recursive sense).*

*Proof.* (a) Let $\mathrm{RSHG}^{(n,k)} = (V, E)$ be any $(n,k)$-recursive SuperHyperGraph. Take the singleton parameter set $C := \{*\}$ and define

$$A(*) := V, \qquad B(*) := E.$$

Then (RS1) holds trivially. For (RS2), every $e \in B(*) = E$ satisfies $\mathrm{supp}_k(e) \subseteq V = A(*)$ by definition of $\mathrm{supp}_k$. Hence $(V, E, C, A, B)$ is an $(n,k)$-recursive soft SuperHyperGraph. Moreover, forgetting the (trivial) parameterization recovers exactly $(V, E)$.

(b) Let $(V, E, C, A, B)$ be an $(n,1)$-recursive soft SuperHyperGraph. Since $k = 1$, we have $E \subseteq \mathcal{Q}_1(V) \setminus \{\emptyset\} = \mathcal{P}(V) \setminus \{\varnothing\}$, so each edge $e \in E$ is a nonempty subset of $V$. By Remark 4.22.4, condition (RS2) becomes

$$e \in B(c) \implies e \subseteq A(c),$$

equivalently,

$$B(c) \subseteq \{\, e \in E : \ e \subseteq A(c) \,\}.$$

Thus $(V, E, C, A, B)$ is precisely a soft $n$-SuperHyperGraph in the standard (non-recursive) sense.

Conversely, any soft $n$-SuperHyperGraph $(V, E, C, A, B)$ (with $E \subseteq \mathcal{P}(V) \setminus \{\varnothing\}$ and $B(c) \subseteq \{e \in E : e \subseteq A(c)\}$) is an $(n, 1)$-recursive soft SuperHyperGraph because $\mathrm{supp}_1(e) = e$ for all $e \subseteq V$. □

## 4.23 Hierarchical Soft SuperHyperGraph

A hierarchical superhypergraph permits vertices from several powerset levels and allows edges to join *mixed-level* vertices, while enforcing a downward-closure coherence principle [296]. A Hierarchical Soft SuperHyperGraph assigns parameter-dependent vertex and hyperedge subsets across multiple levels, preserving downward closure and valid edge incidences.

**Definition 4.23.1** (Tagged nonempty powerset tower and hierarchical universe)**.** Let $V_0$ be a finite nonempty base set and $r \in \mathbb{N}_0$. Define disjoint tagged levels recursively by

$$\mathcal{T}_0(V_0) = \{(0, v) : v \in V_0\},$$

$$\mathcal{T}_{j+1}(V_0) = \{(j+1, X) : \varnothing \neq X \subseteq \mathcal{T}_j(V_0)\}.$$

The hierarchical universe is the disjoint union

$$\mathcal{U}_r(V_0) = \bigcup_{j=0}^{r} \mathcal{T}_j(V_0).$$

For a tagged object $x = (j, X) \in \mathcal{T}_j(V_0)$, its level is $\ell(x) = j$. For $j \geq 1$ define its immediate-child family $\mathrm{Ch}(x) = X$. The canonical tagging maps from the nonempty powerset tower are defined recursively by

$$\iota_0(v) = (0, v), \qquad \iota_{j+1}(Y) = (j+1, \{\iota_j(y) : y \in Y\}).$$

Each $\iota_j : \mathcal{P}_*^j(V_0) \to \mathcal{T}_j(V_0)$ is a bijection. Tagging distinguishes vertices at different levels even when their untagged set-theoretic representations coincide.

**Definition 4.23.2** (Downward closure)**.** Let $W \subseteq \mathcal{U}_r(V_0)$, and put

$$D_0(W) = W, \qquad D_{t+1}(W) = D_t(W) \cup \bigcup_{\substack{x \in D_t(W) \\ \ell(x) \geq 1}} \mathrm{Ch}(x).$$

Define $\mathrm{dcl}(W) = D_r(W)$. This is the smallest subset of $\mathcal{U}_r(V_0)$ containing $W$ and closed under the operation Ch. In particular, a downward closure starting at level $j$ adds objects only at levels at most $j$, never additional objects of level $j$.

**Definition 4.23.3** (Hierarchical SuperHyperGraph of height $r$)**.** Let $V_0$ be a finite nonempty base set and fix $r \in \mathbb{N}_0$. Let $\mathcal{U}_r(V_0)$ and $\ell(\cdot)$ be as in Definition 4.23.1. A *hierarchical SuperHyperGraph of height* $r$ on $V_0$ is a pair

$$\mathbb{H}^{\langle r \rangle} = (V, E)$$

satisfying:

(H1) *(Hierarchical vertex set).* $V$ is a finite nonempty set with $V \subseteq \mathcal{U}_r(V_0)$.

(H2) *(Cross-level hyperedges).* $E \subseteq \mathcal{P}(V) \setminus \{\emptyset\}$.

(H3) *(Coherence / downward closure).* If $x \in V$ and $\ell(x) \geq 1$, then $\mathrm{Ch}(x) \subseteq V$.

For each $k \in \{0, \ldots, r\}$, define the $k$-th layer

$$V_k \ := \ \{\, x \in V : \ \ell(x) = k \,\}, \qquad \text{so that} \qquad V = \dot{\bigcup}_{k=0}^{r} V_k.$$

**Definition 4.23.4** (Hierarchical Soft SuperHyperGraph). Let $\mathbb{H}^{\langle r \rangle} = (V, E)$ be a hierarchical SuperHyperGraph of height $r$ on a base set $V_0$ in the sense of Definition 4.23.3. Let $C$ be a nonempty set of parameters.

A *hierarchical soft SuperHyperGraph* (of height $r$) over $\mathbb{H}^{\langle r \rangle}$ is a quintuple

$$\mathbb{S} \ = \ \big(V, E, C, A, B\big),$$

where

$$A : C \longrightarrow \mathcal{P}(V), \qquad B : C \longrightarrow \mathcal{P}(E),$$

such that for every $c \in C$:

(HS1) *(Downward-closed slice vertices).* If $x \in A(c)$ and $\ell(x) \geq 1$, then $\mathrm{Ch}(x) \subseteq A(c)$.

(HS2) *(Induced slice edges).* $B(c) \subseteq \{\, e \in E : \ e \subseteq A(c) \,\}$.

For each $c \in C$, the pair

$$\mathbb{S}[c] \ := \ \big(A(c),\, B(c)\big)$$

is called the *parameter slice* (a sub-hierarchical-superhypergraph) at $c$.

**Proposition 4.23.5** (Each slice is a hierarchical SuperHyperGraph). *Let $\mathbb{S} = (V, E, C, A, B)$ be a hierarchical soft SuperHyperGraph as in Definition 4.23.4. Then for every $c \in C$, the slice $\mathbb{S}[c] = (A(c), B(c))$ is a hierarchical SuperHyperGraph (possibly empty-vertex if one allows it; if one requires nonempty vertex sets, assume $A(c) \neq \varnothing$).*

*Proof.* Fix $c \in C$. Since $A(c) \subseteq V \subseteq \mathcal{U}_r(V_0)$, the vertex condition holds. Also $B(c) \subseteq E \subseteq \mathcal{P}(V) \setminus \{\emptyset\}$ and by (HS2) each $e \in B(c)$ satisfies $e \subseteq A(c)$, hence $B(c) \subseteq \mathcal{P}(A(c)) \setminus \{\emptyset\}$. Finally, (HS1) is exactly the coherence (downward-closure) axiom for the slice. $\square$

**Theorem 4.23.6** (Hierarchical SuperHyperGraphs are special cases). *Let $\mathbb{H}^{\langle r \rangle} = (V, E)$ be a hierarchical SuperHyperGraph. Define a singleton parameter set $C := \{c_0\}$ and maps*

$$A(c_0) := V, \qquad B(c_0) := E.$$

*Then $\mathbb{S} = (V, E, C, A, B)$ is a hierarchical soft SuperHyperGraph, and its unique slice $\mathbb{S}[c_0]$ coincides with $\mathbb{H}^{\langle r \rangle}$.*

*Proof.* Since $A(c_0) = V$, the slice-vertex downward-closure (HS1) holds because $V$ satisfies (H3). Also $B(c_0) = E \subseteq \{e \in E : e \subseteq V\}$, so (HS2) holds. Thus $\mathbb{S}$ is hierarchical soft, and $\mathbb{S}[c_0] = (V, E) = \mathbb{H}^{\langle r\rangle}$. □

**Theorem 4.23.7** (Faithful tagged embedding of soft $n$-SuperHyperGraphs)**.** *Let $n \geq 1$, let $V_0$ be a finite nonempty base set, and let $(V_n, E, C, A_n, B)$ be a soft $n$-SuperHyperGraph, so that $V_n \subseteq \mathcal{P}_*^{n-1}(V_0)$. Put $r = n - 1$ and define the tagged top-level set*

$$\widehat{V}_n = \iota_r(V_n) \subseteq \mathcal{T}_r(V_0),$$

*with tagged edge family*

$$\widehat{E} = \{\widehat{e} : e \in E\}, \quad \widehat{e} = \{\iota_r(v) : v \in e\}.$$

*For each $c \in C$, put*

$$\widehat{A}(c) = \mathrm{dcl}(\iota_r(A_n(c))), \qquad \widehat{B}(c) = \{\widehat{e} : e \in B(c)\}.$$

*Then*

$$\widehat{\mathbb{S}} = (\mathrm{dcl}(\widehat{V}_n), \widehat{E}, C, \widehat{A}, \widehat{B})$$

*is a hierarchical soft SuperHyperGraph of height $r$. Moreover, the original vertex and edge selections are recovered exactly by*

$$A_n(c) = \iota_r^{-1}(\widehat{A}(c) \cap \widehat{V}_n), \qquad B(c) = \{e \in E : \widehat{e} \in \widehat{B}(c)\}.$$

*Proof.* The tagging map $\iota_r$ is injective, and hence $\widehat{V}_n$ and $\widehat{E}$ are well-defined. Downward closure satisfies the hierarchical vertex axiom, while each tagged edge is a nonempty subset of $\widehat{V}_n \subseteq \mathrm{dcl}(\widehat{V}_n)$. For each $c$, the set $\widehat{A}(c)$ is downward closed. If $\widehat{e} \in \widehat{B}(c)$, the corresponding original edge $e$ satisfies $e \subseteq A_n(c)$, so $\widehat{e} \subseteq \widehat{A}(c)$. Thus the required vertex and edge compatibility axioms hold. Finally, tagging keeps all levels disjoint, and downward closure introduces only elements of strictly lower level. Therefore

$$\widehat{A}(c) \cap \widehat{V}_n = \iota_r(A_n(c)).$$

Applying $\iota_r^{-1}$ recovers $A_n(c)$; injectivity also recovers each original edge and $B(c)$ uniquely. □

## 4.24 Soft Tree-width

Tree-width measures how closely a graph resembles a tree by minimizing the largest bag size minus one over all tree decompositions [642–644]. Related concepts such as HyperTree-width [645, 646], Fuzzy Tree-width [647], and SuperHyperTree-width [648] are also known. Soft Tree-width measures parameter-dependent graph complexity by taking the maximum classical tree-width among the graphs induced by the considered soft parameters. Related concepts such as Fuzzy Tree-width and Neutrosophic Tree-width are also known.

**Definition 4.24.1** (Soft Tree-Width)**.** Let

$$G = (V, E)$$

be a finite simple graph, let $A$ be a nonempty finite parameter set, and let

$$F_V : A \to \mathcal{P}(V), \qquad F_E : A \to \mathcal{P}(E)$$

satisfy

$$F_E(a) \subseteq \{\{u,v\} \in E : u,v \in F_V(a)\} \qquad \text{for every } a \in A.$$

Assume that $F_V(a) \neq \varnothing$ for all $a \in A$. For each $a \in A$, define the parameter-induced graph

$$G_a = \big(F_V(a), F_E(a)\big).$$

The *Soft Tree-Width profile* is the mapping

$$\mathrm{stw}_{\mathcal{S}} : A \to \mathbb{N}_0, \qquad \mathrm{stw}_{\mathcal{S}}(a) = \mathrm{tw}(G_a),$$

where $\mathrm{tw}(G_a)$ denotes the ordinary tree-width of $G_a$.

The *Soft Tree-Width* of the soft graph

$$\mathcal{S} = (G, A, F_V, F_E)$$

is defined by

$$\mathrm{stw}(\mathcal{S}) = \max_{a \in A} \mathrm{tw}(G_a).$$

Thus, the Soft Tree-Width measures the largest tree-width occurring among the graph structures induced by the soft parameters.

**Remark 4.24.2.** If $A = \{a\}$ and $G_a = G$, then

$$\mathrm{stw}(\mathcal{S}) = \mathrm{tw}(G),$$

so the classical tree-width is recovered as a special case.

**Example 4.24.3** (Soft Tree-width)**.** Let

$$V = \{1,2,3,4\}, \qquad E = \{\{1,2\},\{2,3\},\{1,3\},\{3,4\}\},$$

and let $G = (V, E)$.

Consider the parameter set

$$A = \{\alpha, \beta\}.$$

Define

$$F_V(\alpha) = \{1,2,3\}, \qquad F_E(\alpha) = \{\{1,2\},\{2,3\},\{1,3\}\},$$

and

$$F_V(\beta) = \{3,4\}, \qquad F_E(\beta) = \{\{3,4\}\}.$$

The parameter-induced graphs are

$$G_\alpha \cong K_3, \qquad G_\beta \cong K_2.$$

Consequently,

$$\mathrm{tw}(G_\alpha) = 2, \qquad \mathrm{tw}(G_\beta) = 1.$$

Therefore,

$$\mathrm{stw}_{\mathcal{S}}(\alpha) = 2, \qquad \mathrm{stw}_{\mathcal{S}}(\beta) = 1,$$

and

$$\mathrm{stw}(\mathcal{S}) = \max\{2,1\} = 2.$$

Thus, the Soft Tree-width is determined by the parameter-induced graph having the largest classical tree-width.

## 4.25 Soft Filter, Soft Ideal, Soft Grill, Soft Primal

A Soft Filter is an upward-closed family of soft sets, excluding the null soft set and closed under finite intersections [649–651]. A Soft Ideal is a downward-closed family of soft sets, excluding the absolute soft set and closed under finite unions [652, 653]. A Soft Grill is an upward-closed family excluding the null set, where a union belongs only if one component belongs [654–656]. A Soft Primal is a downward-closed family excluding the absolute soft set, where intersection membership forces membership of one factor.

Let $X$ be a nonempty universe and $E$ a nonempty parameter set. Write

$$SS_E(X) = \{(F, E) : F : E \to \mathcal{P}(X)\}.$$

Let $\Phi$ and $\widetilde{X}$ denote the null and absolute soft sets, respectively. For $(F, E), (G, E) \in SS_E(X)$, write

$$(F, E)\widetilde{\subseteq}(G, E) \iff F(e) \subseteq G(e) \text{ for all } e \in E,$$

and let $\widetilde{\cup}$ and $\widetilde{\cap}$ denote the pointwise soft union and intersection.

**Definition 4.25.1** (Soft Filter)**.** A family $\mathcal{F} \subseteq SS_E(X)$ is called a *Soft Filter* on $X$ if

$$\Phi \notin \mathcal{F},$$

$$(F_1, E), (F_2, E) \in \mathcal{F} \Longrightarrow (F_1, E)\widetilde{\cap}(F_2, E) \in \mathcal{F},$$

and

$$(F_1, E) \in \mathcal{F}, \quad (F_1, E)\widetilde{\subseteq}(F_2, E) \Longrightarrow (F_2, E) \in \mathcal{F}.$$

**Example 4.25.2** (Soft Filter)**.** Let

$$X = \{u, v\}, \qquad E = \{p, q\}.$$

Define

$$\mathcal{F} = \{(H, E) \in SS_E(X) : u \in H(p)\}.$$

Then $\mathcal{F}$ is a Soft Filter.

Indeed, the null soft set does not belong to $\mathcal{F}$, since

$$u \notin \varnothing.$$

If $(H_1, E), (H_2, E) \in \mathcal{F}$, then

$$u \in H_1(p) \cap H_2(p),$$

and hence

$$(H_1, E)\widetilde{\cap}(H_2, E) \in \mathcal{F}.$$

Furthermore, if

$$(H_1, E)\widetilde{\subseteq}(H_2, E)$$

and $(H_1, E) \in \mathcal{F}$, then

$$u \in H_2(p).$$

Therefore,

$$(H_2, E) \in \mathcal{F}.$$

Thus, all Soft Filter conditions are satisfied.

**Definition 4.25.3** (Soft Ideal)**.** A family $\mathcal{I} \subseteq SS_E(X)$ is called a *Soft Ideal* on $X$ if

$$\widetilde{X} \notin \mathcal{I},$$

$$(F_1, E), (F_2, E) \in \mathcal{I} \Longrightarrow (F_1, E)\widetilde{\cup}(F_2, E) \in \mathcal{I},$$

and

$$(F_1, E) \in \mathcal{I}, \quad (F_2, E)\widetilde{\subseteq}(F_1, E) \Longrightarrow (F_2, E) \in \mathcal{I}.$$

**Example 4.25.4** (Soft Ideal)**.** Let

$$X = \{u, v\}, \qquad E = \{p, q\}.$$

Define

$$\mathcal{I} = \{(H, E) \in SS_E(X) : u \notin H(p)\}.$$

The absolute soft set does not belong to $\mathcal{I}$, because

$$u \in X.$$

If $(H_1, E), (H_2, E) \in \mathcal{I}$, then

$$u \notin H_1(p) \cup H_2(p).$$

Consequently,

$$(H_1, E)\widetilde{\cup}(H_2, E) \in \mathcal{I}.$$

Moreover, if

$$(H_2, E)\widetilde{\subseteq}(H_1, E)$$

and $(H_1, E) \in \mathcal{I}$, then

$$u \notin H_2(p).$$

Therefore, $\mathcal{I}$ is a Soft Ideal.

**Definition 4.25.5** (Soft Grill)**.** A family $\mathcal{G} \subseteq SS_E(X)$ is called a *Soft Grill* on $X$ if

$$\Phi \notin \mathcal{G},$$

$$(G_1, E) \in \mathcal{G}, \quad (G_1, E)\widetilde{\subseteq}(G_2, E) \Longrightarrow (G_2, E) \in \mathcal{G},$$

and

$$(G_1, E)\widetilde{\cup}(G_2, E) \in \mathcal{G} \Longrightarrow (G_1, E) \in \mathcal{G} \text{ or } (G_2, E) \in \mathcal{G}.$$

**Example 4.25.6** (Soft Grill)**.** Let

$$X = \{u, v\}, \qquad E = \{p, q\},$$

and define

$$\mathcal{G} = \{(H, E) \in SS_E(X) : u \in H(p)\}.$$

Clearly,

$$\Phi \notin \mathcal{G}.$$

The family is upward closed because

$$u \in H_1(p) \subseteq H_2(p)$$

implies

$$(H_2, E) \in \mathcal{G}.$$

Furthermore, suppose that

$$(H_1, E)\widetilde{\cup}(H_2, E) \in \mathcal{G}.$$

Then

$$u \in H_1(p) \cup H_2(p).$$

Hence,

$$u \in H_1(p) \quad \text{or} \quad u \in H_2(p).$$

Therefore,

$$(H_1, E) \in \mathcal{G} \quad \text{or} \quad (H_2, E) \in \mathcal{G}.$$

Thus, $\mathcal{G}$ is a Soft Grill.

**Definition 4.25.7** (Soft Primal)**.** A family $\mathcal{P} \subseteq SS_E(X)$ is called a *Soft Primal* on $X$ if

$$\widetilde{X} \notin \mathcal{P},$$

$$(F_1, E) \in \mathcal{P}, \quad (F_2, E)\widetilde{\subseteq}(F_1, E) \Longrightarrow (F_2, E) \in \mathcal{P},$$

and

$$(F_1, E)\widetilde{\cap}(F_2, E) \in \mathcal{P} \Longrightarrow (F_1, E) \in \mathcal{P} \text{ or } (F_2, E) \in \mathcal{P}.$$

**Example 4.25.8** (Soft Primal)**.** Let

$$X = \{u, v\}, \qquad E = \{p, q\}.$$

Define

$$\mathcal{P} = \{(H, E) \in SS_E(X) : u \notin H(p)\}.$$

Clearly,

$$\widetilde{X} \notin \mathcal{P}.$$

If

$$(H_2, E)\widetilde{\subseteq}(H_1, E)$$

and $(H_1, E) \in \mathcal{P}$, then

$$u \notin H_2(p),$$

so

$$(H_2, E) \in \mathcal{P}.$$

Now suppose that

$$(H_1, E)\widetilde{\cap}(H_2, E) \in \mathcal{P}.$$

Then

$$u \notin H_1(p) \cap H_2(p).$$

Therefore,

$$u \notin H_1(p) \quad \text{or} \quad u \notin H_2(p).$$

Consequently,

$$(H_1, E) \in \mathcal{P} \quad \text{or} \quad (H_2, E) \in \mathcal{P}.$$

Hence, $\mathcal{P}$ is a Soft Primal.

## 4.26 Convex Soft Geometry

A Convex Soft Geometry is an intersection-closed family of soft sets satisfying an accessibility condition through adding suitable soft points [657].

**Definition 4.26.1** (Convex Soft Geometry)**.** Let $X$ be a nonempty finite set, let $E$ be a nonempty parameter set, and let $SS_E(X)$ denote the family of all soft sets over $X$ with parameter set $E$. Let $\Phi$ and $\widetilde{X}$ denote the null and absolute soft sets, respectively.

For $x \in X$, define the constant soft point

$$\widetilde{x}(e) = \{x\}, \qquad e \in E.$$

A family

$$\mathcal{C} \subseteq SS_E(X)$$

is called a *Convex Soft Geometry* if:

1. $\Phi \in \mathcal{C}$;

2. if $F, G \in \mathcal{C}$, then

$$F\widetilde{\cap}G \in \mathcal{C};$$

3. if $F \in \mathcal{C}$ and $F \neq \widetilde{X}$, then there exists $x \in X$ such that

$$\widetilde{x} \,\widetilde{\not\subseteq} F \qquad \text{and} \qquad F \widetilde{\cup} \widetilde{x} \in \mathcal{C}.$$

The members of $\mathcal{C}$ are called *soft convex sets*.

**Example 4.26.2** (Convex Soft Geometry)**.** Let

$$X = \{1, 2, 3\}, \qquad E = \{p, q\}.$$

For every $B \subseteq X$, define the constant soft set $(F_B, E)$ by

$$F_B(e) = B, \qquad e \in E.$$

Consider the family

$$\mathcal{C} = \{(F_B, E) : B \subseteq X\}.$$

Clearly,

$$\Phi = (F_{\varnothing}, E) \in \mathcal{C}.$$

For any $B, C \subseteq X$,

$$(F_B, E) \widetilde{\cap} (F_C, E) = (F_{B \cap C}, E) \in \mathcal{C}.$$

Moreover, if $B \neq X$, choose

$$x \in X \setminus B.$$

Then

$$\widetilde{x} \,\widetilde{\not\subseteq} (F_B, E),$$

and

$$(F_B, E) \widetilde{\cup} \widetilde{x} = (F_{B \cup \{x\}}, E) \in \mathcal{C}.$$

Thus, $\mathcal{C}$ satisfies the intersection-closure and accessibility conditions and is a Convex Soft Geometry.

## 4.27 Soft Semiring

A Soft Semiring is a parameterized family of subsemirings, assigning each parameter a subsemiring of one fixed underlying semiring structure [658–660]. Related concepts such as fuzzy semirings [661, 662] and neutrosophic semirings [274, 663] are also known.

**Definition 4.27.1** (Soft Semiring)**.** Let $S = (S, +, \cdot)$ be a semiring and let $A$ be a nonempty parameter set. A soft set

$$F : A \longrightarrow \mathcal{P}(S)$$

is called a *Soft Semiring* over $S$ if, for every $a \in A$ with $F(a) \neq \varnothing$,

$$F(a) \leq S,$$

where $F(a) \leq S$ means that $F(a)$ is a subsemiring of $S$.

**Example 4.27.2** (Soft Semiring)**.** Let

$$S = \mathbb{N}_0[x]$$

be the semiring of polynomials with nonnegative integer coefficients.

Consider

$$A = \{\alpha, \beta\},$$

and define

$$F(\alpha) = \mathbb{N}_0[x^2], \qquad F(\beta) = \mathbb{N}_0[x].$$

Both sets contain 0 and 1 and are closed under ordinary polynomial addition and multiplication.

Hence,

$$F(\alpha) \leq S, \qquad F(\beta) \leq S.$$

Therefore, the soft set

$$F : A \longrightarrow \mathcal{P}(S)$$

is a Soft Semiring.

For instance,

$$1 + x^2 \in F(\alpha), \qquad x \notin F(\alpha),$$

whereas

$$x \in F(\beta).$$

## 4.28 Soft Relation

A Soft Relation assigns each parameter a binary relation on a universe, providing parameter-dependent relational information between ordered object pairs.

**Definition 4.28.1** (Soft Relation)**.** Let $X$ be a nonempty set and let $A$ be a nonempty parameter set. A *Soft Relation* on $X$ is a soft set

$$\sigma : A \longrightarrow \mathcal{P}(X \times X).$$

Thus, for each $a \in A$,

$$\sigma(a) \subseteq X \times X$$

is an ordinary binary relation on $X$.

**Example 4.28.2** (Soft Relation)**.** Let

$$X = \{1, 2, 3\}, \qquad A = \{\alpha, \beta\}.$$

Define

$$\sigma(\alpha) = \{(1, 1), (2, 2), (3, 3)\},$$

and

$$\sigma(\beta) = \{(1, 2), (1, 3), (2, 3)\}.$$

Then

$$\sigma : A \longrightarrow \mathcal{P}(X \times X)$$

is a Soft Relation.

The relation $\sigma(\alpha)$ is the identity relation on $X$, whereas $\sigma(\beta)$ is the strict order relation induced by the usual ordering on $X$.

Thus, different parameters describe different relational structures on the same universe.

## 4.29 Soft $\sigma$-Algebra

A Soft $\sigma$-Algebra is a family of soft sets containing the null set and closed under complements and countable unions [664–666].

**Definition 4.29.1** (Soft $\sigma$-Algebra)**.** Let $X$ be a nonempty universe, $E$ a nonempty parameter set, and $SS_E(X)$ the family of all soft sets over $X$ with parameter set $E$. A family

$$\Sigma \subseteq SS_E(X)$$

is called a *Soft $\sigma$-Algebra* if:

1. the null soft set satisfies
$$\Phi \in \Sigma;$$

2. if $F \in \Sigma$, then
$$F^c \in \Sigma;$$

3. for every sequence $F_1, F_2, \ldots \in \Sigma$,
$$\widetilde{\bigcup}_{n=1}^{\infty} F_n \in \Sigma.$$

**Example 4.29.2** (Soft $\sigma$-Algebra)**.** Let
$$X = \{u, v\}, \qquad E = \{p, q\}.$$

For each $B \subseteq X$, define
$$F_B(e) = B, \qquad e \in E.$$

Consider
$$\Sigma = \{(F_B, E) : B \subseteq X\}.$$

Since
$$\Phi = (F_\varnothing, E) \in \Sigma,$$
the null soft set belongs to $\Sigma$.

For every $B \subseteq X$,
$$(F_B, E)^c = (F_{X \setminus B}, E) \in \Sigma.$$

Furthermore, for any sequence $\{B_n\}_{n \geq 1}$ of subsets of $X$,
$$\widetilde{\bigcup}_{n=1}^{\infty} (F_{B_n}, E) = \big(F_{\bigcup_{n=1}^{\infty} B_n}, E\big) \in \Sigma.$$

Therefore, $\Sigma$ is a Soft $\sigma$-Algebra.

## 4.30 Soft Compact Space

A Soft Compact Space is a soft topological space in which every soft open cover admits a finite soft subcover [667–669]. Related concepts such as Fuzzy Compact Spaces [670,671] and Neutrosophic Compact Spaces [672, 673] are also known.

**Definition 4.30.1** (Soft Compact Space)**.** Let $(X, \tau, E)$ be a soft topological space and let $\widetilde{X}$ denote the absolute soft set. A family

$$\{F_i : i \in I\} \subseteq \tau$$

is a *soft open cover* of $X$ if

$$\widetilde{X} \widetilde{\subseteq} \widetilde{\bigcup}_{i \in I} F_i.$$

The space $(X, \tau, E)$ is called *Soft Compact* if every soft open cover has a finite subcover; that is, there exist $i_1, \ldots, i_m \in I$ such that

$$\widetilde{X} \widetilde{\subseteq} F_{i_1} \widetilde{\cup} \cdots \widetilde{\cup} F_{i_m}.$$

**Example 4.30.2** (Soft Compact Space)**.** Let

$$X = \{u, v\}, \qquad E = \{p, q\},$$

and consider the discrete soft topology

$$\tau = SS_E(X).$$

Then

$$(X, \tau, E)$$

is a Soft Compact Space.

Indeed, let

$$\{F_i : i \in I\} \subseteq \tau$$

be an arbitrary soft open cover of $\widetilde{X}$.

For each

$$(e, x) \in E \times X,$$

there exists an index $i(e, x) \in I$ such that

$$x \in F_{i(e,x)}(e).$$

Since

$$|E \times X| = 4,$$

at most four selected soft open sets are sufficient to cover every parameter-element pair.

Hence, the cover admits a finite soft subcover.

Therefore,

$$(X, \tau, E)$$

is Soft Compact.

## 4.31 Soft Separation Axioms

Soft Separation Axioms distinguish points using soft open sets, extending classical $T_0$, $T_1$, and $T_2$ separation properties to soft topology [667, 674–676].

**Definition 4.31.1** (Soft Separation Axioms)**.** Let $(X, \tau, E)$ be a soft topological space. For $x \in X$, let $\widetilde{x}$ denote the constant soft point defined by

$$\widetilde{x}(e) = \{x\}, \qquad e \in E.$$

The space is called:

1. *Soft* $T_0$ if, for every distinct $x, y \in X$, there exists $F \in \tau$ containing one of $\widetilde{x}, \widetilde{y}$ but not the other;

2. *Soft* $T_1$ if, for every distinct $x, y \in X$, there exist $F, G \in \tau$ such that

$$\widetilde{x}\widetilde{\in}F, \quad \widetilde{y}\,\widetilde{\notin}F, \qquad \widetilde{y}\widetilde{\in}G, \quad \widetilde{x}\,\widetilde{\notin}G;$$

3. *Soft* $T_2$ if, for every distinct $x, y \in X$, there exist $F, G \in \tau$ such that

$$\widetilde{x}\widetilde{\in}F, \qquad \widetilde{y}\widetilde{\in}G, \qquad F\widetilde{\cap}G = \Phi.$$

**Example 4.31.2** (Soft $T_0$ Space)**.** Let

$$X = \{0, 1\}, \qquad E = \{p\}.$$

For every $B \subseteq X$, let

$$F_B(p) = B.$$

Define

$$\tau = \{(F_\varnothing, E), (F_{\{1\}}, E), (F_X, E)\}.$$

The distinct points 0 and 1 are distinguished by the soft open set $(F_{\{1\}}, E)$, since

$$1 \in F_{\{1\}}(p), \qquad 0 \notin F_{\{1\}}(p).$$

Thus, the space satisfies the Soft $T_0$ separation axiom.

However, no soft open set contains 0 without containing 1.

Consequently, this space is Soft $T_0$ but not Soft $T_1$.

**Example 4.31.3** (Soft $T_1$ Space)**.** Let

$$X = \mathbb{N}, \qquad E = \{p\}.$$

For every $U \subseteq X$, define

$$F_U(p) = U.$$

Let

$$\tau = \{(F_U, E) : U = \varnothing \text{ or } X \setminus U \text{ is finite}\}.$$

This is the soft topology induced by the cofinite topology on $\mathbb{N}$.

For any distinct $x, y \in X$, the soft open sets associated with

$$U = X \setminus \{y\}, \qquad V = X \setminus \{x\}$$

satisfy

$$x \in U, \quad y \notin U,$$

and

$$y \in V, \quad x \notin V.$$

Hence, the space is Soft $T_1$.

However, any two nonempty open subsets of an infinite cofinite space intersect.

Therefore, the space is not Soft $T_2$.

**Example 4.31.4** (Soft $T_2$ Space)**.** Let

$$X = \{u, v\}, \qquad E = \{p\},$$

and consider the discrete soft topology

$$\tau = SS_E(X).$$

Define the soft open sets $F$ and $G$ by

$$F(p) = \{u\}, \qquad G(p) = \{v\}.$$

Then

$$\widetilde{u}\widetilde{\in}F, \qquad \widetilde{v}\widetilde{\in}G,$$

and

$$F\widetilde{\cap}G = \Phi.$$

Therefore, the space satisfies the Soft $T_2$ separation axiom.

In particular, it also satisfies the Soft $T_1$ and Soft $T_0$ separation axioms.

## 4.32 Soft Topological Group

A Soft Topological Group combines a group with a soft topology such that multiplication and inversion are soft continuous mappings [677–680].

**Definition 4.32.1** (Soft Topological Group)**.** Let $G$ be a group, let $A$ be a nonempty parameter set, and let $(F, A)$ be a non-null soft set over $G$,

$$F : A \longrightarrow \mathcal{P}(G).$$

Let $\tau$ be a topology on $G$. The triple

$$(F, A, \tau)$$

is called a *Soft Topological Group* if, for every $a \in A$, $F(a)$ is a subgroup of $G$ and $F(a)$, equipped with the subspace topology induced by $\tau$, is a topological group.

**Example 4.32.2** (Soft Topological Group)**.** Let

$$G = (\mathbb{R}, +)$$

be the additive group of real numbers equipped with the usual topology $\tau$.

Consider

$$A = \{\alpha, \beta\},$$

and define

$$F(\alpha) = \mathbb{Z}, \qquad F(\beta) = \mathbb{R}.$$

Both

$$(\mathbb{Z}, +) \quad \text{and} \quad (\mathbb{R}, +)$$

are subgroups of $G$.

The subspace topology on $\mathbb{Z}$ is discrete, and hence $(\mathbb{Z}, +)$ is a topological group.

Moreover, $(\mathbb{R}, +)$ is a topological group under the usual topology.

Thus, each parameter determines a topological subgroup of $G$.

Consequently,

$$(F, A, \tau)$$

is a Soft Topological Group.

## 4.33 Soft BCK/BCI-Algebra

A BCK/BCI-Algebra is an abstract algebraic structure with a binary operation and constant zero, satisfying axioms derived from implication systems [681–683]. A Soft BCK/BCI-Algebra assigns each parameter a subalgebra of a fixed BCK or BCI algebra, forming a parameterized algebraic family [684–687]. Related concepts such as Fuzzy BCK/BCI-Algebras [688, 689] and Neutrosophic BCK/BCI-Algebras [690, 691] are also known.

**Definition 4.33.1** (Soft BCK/BCI-Algebra)**.** Let

$$(X, *, 0)$$

be a BCK-algebra or BCI-algebra, and let $A$ be a nonempty parameter set. A soft set

$$F : A \longrightarrow \mathcal{P}(X)$$

is called a *Soft BCK/BCI-Algebra* over $X$ if, for every $a \in A$ with $F(a) \neq \varnothing$,

$$(F(a), *, 0)$$

is a BCK-subalgebra or BCI-subalgebra of $X$, respectively.

**Example 4.33.2** (Soft BCI-Algebra)**.** Let

$$X = \mathbb{Z}$$

and define

$$x * y = x - y, \qquad x, y \in \mathbb{Z}.$$

Then

$$(X, *, 0)$$

is a BCI-algebra.

Consider

$$A = \{\alpha, \beta\},$$

and define

$$F(\alpha) = 2\mathbb{Z}, \qquad F(\beta) = \mathbb{Z}.$$

Both parameter-induced sets contain 0 and are closed under subtraction.

Indeed,

$$2m * 2n = 2(m - n) \in 2\mathbb{Z}$$

for all $m, n \in \mathbb{Z}$.

Thus,

$$F(\alpha) \quad \text{and} \quad F(\beta)$$

are BCI-subalgebras of $X$.

Consequently, $F$ is a Soft BCI-Algebra.

## 4.34 Soft Normed Linear Space

A Normed Linear Space is a vector space equipped with a norm measuring vector magnitude and inducing a natural metric structure [692, 693]. A Soft Normed Linear Space assigns parameter-dependent norms to soft vectors, measuring vector magnitude consistently across parameter contexts and conditions [113, 694–696].

**Definition 4.34.1** (Soft Normed Linear Space)**.** Let $V$ be a vector space over $\mathbb{K}$, where $\mathbb{K} = \mathbb{R}$ or $\mathbb{C}$, and let $A$ be a nonempty parameter set. Define the soft vector space

$$\widetilde{V} = V^A = \{\widetilde{x} : A \to V\}.$$

Suppose that, for each $a \in A$, $\|\cdot\|_a$ is a norm on $V$. Define

$$\|\widetilde{x}\|_s : A \to [0, \infty), \qquad \|\widetilde{x}\|_s(a) = \|\widetilde{x}(a)\|_a.$$

Then

$$(\widetilde{V}, \|\cdot\|_s)$$

is called a *Soft Normed Linear Space.*

**Example 4.34.2** (Soft Normed Linear Space)**.** Let

$$V = \mathbb{R}^2, \qquad A = \{\alpha, \beta\}.$$

Define two norms on $V$ by

$$\|(x, y)\|_\alpha = |x| + |y|,$$

and

$$\|(x, y)\|_\beta = \sqrt{x^2 + y^2}.$$

Let

$$\widetilde{V} = V^A.$$

For $\widetilde{x} \in \widetilde{V}$, define

$$\|\widetilde{x}\|_s(a) = \|\widetilde{x}(a)\|_a, \qquad a \in A.$$

For example, suppose

$$\widetilde{x}(\alpha) = (1, -1), \qquad \widetilde{x}(\beta) = (3, 4).$$

Then

$$\|\widetilde{x}\|_s(\alpha) = 2, \qquad \|\widetilde{x}\|_s(\beta) = 5.$$

Since both parameter-dependent functions are ordinary norms,

$$(\widetilde{V}, \|\cdot\|_s)$$

is a Soft Normed Linear Space.

## 4.35 Soft Banach Space

A Soft Banach Space is a complete soft normed linear space where every soft Cauchy sequence converges to a vector [696].

**Definition 4.35.1** (Soft Banach Space)**.** Let

$$(\widetilde{V}, \|\cdot\|_s)$$

be a Soft Normed Linear Space induced by the family $\{\|\cdot\|_a\}_{a\in A}$. It is called a *Soft Banach Space* if

$$(V, \|\cdot\|_a)$$

is a Banach space for every $a \in A$. Equivalently, the space is complete with respect to parameterwise soft Cauchy convergence.

**Example 4.35.2** (Soft Banach Space)**.** Let

$$V = \mathbb{R}^2, \qquad A = \{\alpha, \beta\}.$$

Equip $V$ with the norms

$$\|(x, y)\|_\alpha = |x| + |y|$$

and

$$\|(x, y)\|_\beta = \sqrt{x^2 + y^2}.$$

Both normed spaces

$$(V, \|\cdot\|_\alpha) \quad \text{and} \quad (V, \|\cdot\|_\beta)$$

are complete because every finite-dimensional normed vector space is a Banach space.

Consequently, every soft Cauchy sequence converges parameterwise to an element of

$$\widetilde{V} = V^A.$$

Therefore,

$$(\widetilde{V}, \|\cdot\|_s)$$

is a Soft Banach Space.

## 4.36 Soft Inner Product Space

A Soft Inner Product Space assigns parameter-dependent inner products to soft vectors, inducing parameter-wise norms, orthogonality, angles, and geometric structure [697–700]. Related concepts, such as Fuzzy Inner Product Spaces [701, 702], are also known.

**Definition 4.36.1** (Soft Inner Product Space)**.** Let $V$ be a vector space over $\mathbb{K}$, let $A$ be a nonempty parameter set, and let

$$\widetilde{V} = V^A.$$

For every $a \in A$, let

$$\langle \cdot, \cdot \rangle_a$$

be an inner product on $V$. Define the *soft inner product* by

$$\langle \widetilde{x}, \widetilde{y} \rangle_s(a) = \langle \widetilde{x}(a), \widetilde{y}(a) \rangle_a, \qquad a \in A.$$

Then

$$(\widetilde{V}, \langle \cdot, \cdot \rangle_s)$$

is called a *Soft Inner Product Space.*

**Example 4.36.2** (Soft Inner Product Space)**.** Let

$$V = \mathbb{R}^2, \qquad A = \{\alpha, \beta\}.$$

For

$$x = (x_1, x_2), \qquad y = (y_1, y_2),$$

define

$$\langle x, y \rangle_\alpha = x_1 y_1 + x_2 y_2,$$

and

$$\langle x, y \rangle_\beta = 2x_1 y_1 + 3x_2 y_2.$$

Both mappings are inner products on $\mathbb{R}^2$.

For

$$\widetilde{x}, \widetilde{y} \in V^A,$$

define

$$\langle \widetilde{x}, \widetilde{y} \rangle_s(a) = \langle \widetilde{x}(a), \widetilde{y}(a) \rangle_a.$$

For example, let

$$\widetilde{x}(\alpha) = (1, 0), \qquad \widetilde{y}(\alpha) = (0, 1),$$

and

$$\widetilde{x}(\beta) = (1, 2), \qquad \widetilde{y}(\beta) = (3, 4).$$

Then

$$\langle \widetilde{x}, \widetilde{y} \rangle_s(\alpha) = 0,$$

whereas

$$\langle \widetilde{x}, \widetilde{y} \rangle_s(\beta) = 2(1)(3) + 3(2)(4) = 30.$$

Thus,

$$(\widetilde{V}, \langle \cdot, \cdot \rangle_s)$$

is a Soft Inner Product Space.

## 4.37 Soft Hilbert Space

A Hilbert space is a complete inner product space, providing a rigorous framework for orthogonality, projection, convergence, and infinite-dimensional analysis. A Soft Hilbert Space is a complete soft inner product space combining parameterized inner products with completeness under the induced norm [703–706]. Related concepts, such as Fuzzy Hilbert Spaces [707, 708] and Neutrosophic Hilbert Spaces [709], are also known.

**Definition 4.37.1** (Soft Hilbert Space)**.** A Soft Inner Product Space

$$(\widetilde{V}, \langle \cdot, \cdot \rangle_s)$$

is called a *Soft Hilbert Space* if, for every $a \in A$,

$$(V, \langle \cdot, \cdot \rangle_a)$$

is a Hilbert space. Equivalently, it is complete with respect to the induced soft norm

$$\|\widetilde{x}\|_s(a) = \sqrt{\langle \widetilde{x}(a), \widetilde{x}(a) \rangle_a}.$$

**Example 4.37.2** (Soft Hilbert Space)**.** Let

$$V = \mathbb{R}^2, \qquad A = \{\alpha, \beta\}.$$

Define

$$\langle x, y \rangle_\alpha = x_1 y_1 + x_2 y_2,$$

and

$$\langle x, y \rangle_\beta = 2x_1 y_1 + 3x_2 y_2.$$

The corresponding induced norms are

$$\|x\|_\alpha = \sqrt{x_1^2 + x_2^2},$$

and

$$\|x\|_\beta = \sqrt{2x_1^2 + 3x_2^2}.$$

Both inner product spaces are finite-dimensional and therefore complete.

Hence,

$$(V, \langle \cdot, \cdot \rangle_\alpha)$$

and

$$(V, \langle \cdot, \cdot \rangle_\beta)$$

are Hilbert spaces.

Consequently, the parameterwise soft inner product space

$$(\widetilde{V}, \langle \cdot, \cdot \rangle_s), \qquad \widetilde{V} = V^A,$$

is a Soft Hilbert Space.

## 4.38 Soft Continuous Mapping

A Soft Continuous Mapping preserves soft topology by requiring inverse images of soft open sets to remain open under parameters [710–712].

**Definition 4.38.1** (Soft Continuous Mapping)**.** Let

$$(X, \tau_X, A) \qquad \text{and} \qquad (Y, \tau_Y, A)$$

be soft topological spaces with the same parameter set $A$, and let

$$f : X \to Y.$$

For a soft set $(G, A)$ over $Y$, define

$$f^{-1}(G, A) = (H, A), \qquad H(a) = f^{-1}(G(a)).$$

The mapping $f$ is called *Soft Continuous* if

$$(G, A) \in \tau_Y \quad \Longrightarrow \quad f^{-1}(G, A) \in \tau_X.$$

**Example 4.38.2** (Soft Continuous Mapping)**.** Let

$$X = \{1, 2, 3\}, \qquad Y = \{u, v\}, \qquad A = \{p, q\}.$$

Equip both universes with their discrete soft topologies:

$$\tau_X = SS_A(X), \qquad \tau_Y = SS_A(Y).$$

Define

$$f : X \longrightarrow Y$$

by

$$f(1) = u, \qquad f(2) = u, \qquad f(3) = v.$$

Let $(G, A) \in \tau_Y$ be arbitrary.

Its inverse image is the soft set $(H, A)$, where

$$H(a) = f^{-1}(G(a)), \qquad a \in A.$$

Since

$$H(a) \subseteq X$$

for every $a \in A$, one has

$$(H, A) \in SS_A(X) = \tau_X.$$

Therefore,

$$f^{-1}(G, A) \in \tau_X$$

for every $(G, A) \in \tau_Y$.

Thus, $f$ is Soft Continuous.

For example, if

$$G(p) = \{u\}, \qquad G(q) = \{v\},$$

then

$$H(p) = \{1, 2\}, \qquad H(q) = \{3\}.$$

## 4.39 Soft Connected Space

A Soft Connected Space cannot be decomposed into two nonempty disjoint soft open sets whose union equals the absolute set [713, 714].

**Definition 4.39.1** (Soft Connected Space)**.** Let $(X, \tau, A)$ be a soft topological space, with null soft set $\Phi$ and absolute soft set $\widetilde{X}$. The space is called *Soft Connected* if there do not exist $F, G \in \tau$ such that

$$F \neq \Phi, \qquad G \neq \Phi,$$

$$F \widetilde{\cap} G = \Phi,$$

and

$$F \widetilde{\cup} G = \widetilde{X}.$$

Otherwise, the space is called *Soft Disconnected*.

**Example 4.39.2** (Soft Connected Space)**.** Let

$$X = \{u, v\}, \qquad A = \{\alpha, \beta\}.$$

Consider the indiscrete soft topology

$$\tau = \{\Phi, \widetilde{X}\}.$$

The only non-null soft open set is

$$\widetilde{X}.$$

Consequently, there do not exist two non-null soft open sets $F, G \in \tau$ satisfying

$$F \widetilde{\cap} G = \Phi$$

and

$$F \widetilde{\cup} G = \widetilde{X}.$$

Therefore,

$$(X, \tau, A)$$

is a Soft Connected Space.

## 4.40 Soft Measure Space

A Soft Measure Space equips a soft sigma-algebra with a countably additive measure, quantifying parameterized sets consistently under disjoint unions [517]. Related concepts, such as Fuzzy Measure Spaces [715–717] and Neutrosophic Measure Spaces [718–720], are also known.

**Definition 4.40.1** (Soft Measure Space)**.** Let $X$ be a nonempty universe, $A$ a nonempty parameter set, and let $\Sigma$ be a Soft $\sigma$-Algebra of soft sets over $X$. A mapping

$$\mu : \Sigma \to [0, \infty]$$

is called a *Soft Measure* if

$$\mu(\Phi) = 0$$

and, for every pairwise soft-disjoint sequence

$$F_1, F_2, \ldots \in \Sigma, \qquad F_i \widetilde{\cap} F_j = \Phi \quad (i \neq j),$$

one has

$$\mu\left(\widetilde{\bigcup}_{n=1}^{\infty} F_n\right) = \sum_{n=1}^{\infty} \mu(F_n).$$

The quadruple

$$(X, \Sigma, \mu, A)$$

is called a *Soft Measure Space.*

**Example 4.40.2** (Soft Measure Space)**.** Let

$$X = \{u, v\}, \qquad A = \{\alpha, \beta\},$$

and let

$$\Sigma = SS_A(X).$$

Define

$$\mu : \Sigma \longrightarrow [0, \infty]$$

by

$$\mu(F) = \sum_{a \in A} |F(a)|.$$

Clearly,

$$\mu(\Phi) = 0.$$

Suppose that

$$F_1, F_2, \ldots \in \Sigma$$

are pairwise soft-disjoint.

Then, for every $a \in A$, the sets

$$F_1(a), F_2(a), \ldots$$

are pairwise disjoint.

Therefore,

$$\mu\left(\widetilde{\bigcup}_{n=1}^{\infty} F_n\right) = \sum_{a\in A}\left|\bigcup_{n=1}^{\infty} F_n(a)\right| = \sum_{a\in A}\sum_{n=1}^{\infty}|F_n(a)| = \sum_{n=1}^{\infty}\sum_{a\in A}|F_n(a)| = \sum_{n=1}^{\infty}\mu(F_n).$$

Thus, $\mu$ is countably additive.

For instance, if

$$F(\alpha) = \{u, v\}, \qquad F(\beta) = \{v\},$$

then

$$\mu(F) = 2 + 1 = 3.$$

Consequently,

$$(X, \Sigma, \mu, A)$$

is a Soft Measure Space.

# Chapter 5

# Soft Decision-Making

In this chapter, we examine several soft decision-making methods.

## 5.1 Soft decision-making

Soft decision-making ranks finitely many alternatives by weighted counts of satisfied parameters in a soft set, and selects the argmax scorers. Related frameworks include *fuzzy decision-making* [721, 722] and *neutrosophic decision-making* [723, 724].

**Definition 5.1.1** (Soft decision-making model induced by a soft set)**.** Let $U$ be a nonempty *finite* set of alternatives (objects) and let $E$ be a nonempty set of parameters (criteria). Fix a nonempty *finite* subset $A \subseteq E$ and a soft set $(F, A)$ over $U$, i.e.,

$$F : A \longrightarrow \mathcal{P}(U).$$

Let $w : A \to [0, \infty)$ be a weight (importance) function. The finiteness of $A$ ensures that the score below is a finite sum; alternatively, for infinite $A$ one must assume that the nonnegative weight family is summable.

For each alternative $u \in U$ define the *(crisp) satisfaction indicator* under $a \in A$ by

$$\chi_F(u, a) \ := \ \begin{cases} 1, & u \in F(a), \\ 0, & u \notin F(a). \end{cases}$$

The *soft decision score* of $u$ (with respect to $(F, A)$ and $w$) is the real number

$$S_{(F,A),w}(u) \ := \ \sum_{a \in A} w(a)\, \chi_F(u, a).$$

The induced *soft decision correspondence* is

$$\mathsf{DM}(F, A; w) \ := \ \arg\max_{u \in U} S_{(F,A),w}(u) \ \subseteq \ U,$$

i.e., the set of all alternatives having maximum score. Equivalently, one may define the induced (pre)order $\succeq_{(F,A),w}$ on $U$ by

$$u \succeq_{(F,A),w} v \quad \Longleftrightarrow \quad S_{(F,A),w}(u) \geq S_{(F,A),w}(v),$$

and then $\mathsf{DM}(F, A; w)$ is the set of $\succeq_{(F,A),w}$-maximal elements.

**Theorem 5.1.2** (Soft-set structure and well-definedness)**.** *In the setting of Definition 5.1.1:*

(i) ***Soft-set structure.*** *The data used by the decision model is exactly a soft set* $(F, A)$ *over* $U$ *(together with a weight map* $w$*).*

(ii) ***Well-defined score.*** *The mapping* $S_{(F,A),w} : U \to \mathbb{R}$ *is well-defined (unique value for each* $u \in U$*).*

(iii) ***Existence of optimal decisions.*** *The decision correspondence* $\mathsf{DM}(F, A; w)$ *is well-defined and nonempty.*

(iv) ***Invariance under extension to the full parameter set.*** *Define the extension* $\widetilde{F} : E \to \mathcal{P}(U)$ *by*

$$\widetilde{F}(e) = \begin{cases} F(e), & e \in A, \\ \varnothing, & e \in E \setminus A. \end{cases}$$

*Then the decision outcome computed from* $(F, A)$ *equals the outcome computed from* $\widetilde{F}$ *restricted to* $A$*; in particular, adding "unused" parameters with empty images does not change* $\mathsf{DM}(F, A; w)$*.*

*Proof.* (i) By assumption, $A \subseteq E$ is nonempty and $F : A \to \mathcal{P}(U)$; hence $(F, A)$ is, by definition, a soft set over $U$. The additional map $w : A \to [0, \infty)$ only provides importance weights and does not alter the underlying soft-set structure.

(ii) Fix $u \in U$. For each $a \in A$, the statement $u \in F(a)$ is unambiguous, so $\chi_F(u, a) \in \{0, 1\}$ is uniquely determined. Since $A$ is finite and $w(a)\chi_F(u, a) \in [0, \infty)$ for each $a$, the finite sum $S_{(F,A),w}(u) = \sum_{a \in A} w(a)\chi_F(u, a)$ is a uniquely determined real number. Thus $S_{(F,A),w} : U \to \mathbb{R}$ is well-defined.

(iii) Because $U$ is finite and $S_{(F,A),w}(u) \in \mathbb{R}$ is well-defined for every $u \in U$ by (ii), the maximum value $\max_{u \in U} S_{(F,A),w}(u)$ exists. Therefore the argmax set $\mathsf{DM}(F, A; w) = \{u \in U : S_{(F,A),w}(u) = \max_{v \in U} S_{(F,A),w}(v)\}$ is a well-defined subset of $U$ and is nonempty.

(iv) For $a \in A$ we have $\widetilde{F}(a) = F(a)$ by definition of $\widetilde{F}$, hence $\chi_{\widetilde{F}}(u, a) = \chi_F(u, a)$ for all $u \in U$ and $a \in A$. Therefore, for every $u \in U$,

$$S_{(\widetilde{F}|_A, A),w}(u) = \sum_{a \in A} w(a)\chi_{\widetilde{F}}(u, a) = \sum_{a \in A} w(a)\chi_F(u, a) = S_{(F,A),w}(u).$$

Thus the score functions coincide on $U$, and their argmax sets coincide:

$$\mathsf{DM}(\widetilde{F}|_A, A; w) = \mathsf{DM}(F, A; w).$$

□

## 5.2 HyperSoft TOPSIS and SuperHyperSoft TOPSIS

TOPSIS ranks alternatives by distance to positive ideal and negative ideal points, using normalized weighted criteria values [37, 725–727]. HyperSoft TOPSIS ranks alternatives using TOPSIS on multi-attribute value tuples (hypersoft parameters) with weighted distances to ideal solutions. SuperHyperSoft TOPSIS extends HyperSoft TOPSIS by allowing each attribute to be a value-subset, enabling set-valued criteria tuples.

**Definition 5.2.1** (Hypersoft criterion domain). Let $U = \{u_1, \dots, u_n\}$ be a finite set of alternatives. Let $\mathcal{A}_1, \dots, \mathcal{A}_k$ be pairwise-disjoint attribute-value sets (domains), and define the *hypersoft parameter space*

$$\mathcal{C} := \mathcal{A}_1 \times \mathcal{A}_2 \times \cdots \times \mathcal{A}_k.$$

A finite set of *criteria-tuples* is a subset

$$\Lambda = \{\lambda_1, \dots, \lambda_m\} \subseteq \mathcal{C},$$

where each $\lambda_j = (a_{1j}, \dots, a_{kj})$ specifies a combined multi-attribute value tuple.

**Definition 5.2.2** (Numeric hypersoft evaluation). A *numeric hypersoft evaluation* is a map

$$x : U \times \Lambda \to \mathbb{R}_{\geq 0}, \qquad (u_i, \lambda_j) \mapsto x_{ij}.$$

Equivalently, $x$ is the *hypersoft decision matrix* $X = (x_{ij}) \in \mathbb{R}_{\geq 0}^{n \times m}$.

(*Crisp hypersoft set as a special case.*) If one is given a crisp hypersoft set $G : \Lambda \to \mathcal{P}(U)$, then it induces a numeric evaluation via

$$x_{ij} := \mathbf{1}_{\{u_i \in G(\lambda_j)\}} \in \{0, 1\}.$$

More generally, if evaluations are given in a graded/uncertain hypersoft form (fuzzy, neutrosophic, picture fuzzy, interval-valued, etc.), one first applies a *score* map Score : (grade object) $\to \mathbb{R}_{\geq 0}$ and sets $x_{ij} :=$ Score(grade of $u_i$ at $\lambda_j$).

**Definition 5.2.3** (HyperSoft TOPSIS). Assume the hypersoft decision matrix $X = (x_{ij}) \in \mathbb{R}_{\geq 0}^{n \times m}$ is given. Let $w = (w_1, \dots, w_m)$ be criterion weights with

$$w_j > 0, \qquad \sum_{j=1}^{m} w_j = 1.$$

Let $B \subseteq \{1, \dots, m\}$ be the index set of *benefit* criteria and $C \subseteq \{1, \dots, m\}$ be the index set of *cost* criteria, with $B \cup C = \{1, \dots, m\}$ and $B \cap C = \varnothing$.

**(1) Vector normalization.** For each $j \in \{1, \dots, m\}$ define

$$d_j := \sqrt{\sum_{i=1}^{n} x_{ij}^2}.$$

The normalized matrix $R = (r_{ij})$ is

$$r_{ij} := \begin{cases} \dfrac{x_{ij}}{d_j}, & d_j > 0, \\ 0, & d_j = 0. \end{cases}$$

**(2) Weighted normalized matrix.** Define $V = (v_{ij})$ by

$$v_{ij} := w_j \, r_{ij}.$$

**(3) Hypersoft positive/negative ideal solutions.** For each criterion $j$, define the ideal values

$$v_j^+ := \begin{cases} \max_{1 \le i \le n} v_{ij}, & j \in B, \\ \min_{1 \le i \le n} v_{ij}, & j \in C, \end{cases} \qquad v_j^- := \begin{cases} \min_{1 \le i \le n} v_{ij}, & j \in B, \\ \max_{1 \le i \le n} v_{ij}, & j \in C. \end{cases}$$

Let $A^+ := (v_1^+, \ldots, v_m^+)$ and $A^- := (v_1^-, \ldots, v_m^-)$.

**(4) Separation measures.** For each alternative $u_i$, define the (Euclidean) distances to the ideals:

$$S_i^+ := \sqrt{\sum_{j=1}^{m} (v_{ij} - v_j^+)^2}, \qquad S_i^- := \sqrt{\sum_{j=1}^{m} (v_{ij} - v_j^-)^2}.$$

**(5) Closeness coefficient and ranking.** Define the *hypersoft TOPSIS closeness coefficient*

$$\mathbb{C}_i := \begin{cases} \dfrac{S_i^-}{S_i^+ + S_i^-}, & S_i^+ + S_i^- > 0, \\ \dfrac{1}{2}, & S_i^+ + S_i^- = 0, \end{cases}$$

and rank alternatives by decreasing $\mathbb{C}_i$:

$$u_p \succcurlyeq u_q \iff \mathbb{C}_p \ge \mathbb{C}_q.$$

Any total order refining $\succcurlyeq$ (e.g., by tie-breaking rules) is called a *HyperSoft TOPSIS ranking.*

**Definition 5.2.4** (SuperHyperSoft parameter space and criteria family)**.** Let $U = \{u_1, \ldots, u_n\}$ be a finite set of alternatives. Let $A_1, \ldots, A_k$ be pairwise-disjoint attribute-value sets ($k \ge 1$). Define the *superhypersoft parameter space*

$$\mathcal{C}^{\mathrm{SH}} := \mathcal{P}(A_1) \times \mathcal{P}(A_2) \times \cdots \times \mathcal{P}(A_k),$$

whose elements $\gamma = (\alpha_1, \ldots, \alpha_k)$ are tuples of *value-subsets* $\alpha_t \subseteq A_t$. A finite set of *super-criteria* is a subset

$$\Gamma = \{\gamma_1, \ldots, \gamma_m\} \subseteq \mathcal{C}^{\mathrm{SH}}.$$

**Definition 5.2.5** (Numeric superhypersoft evaluation and decision matrix)**.** A *numeric superhypersoft evaluation* (on $\Gamma$) is a map

$$x^{\mathrm{SH}} : U \times \Gamma \to \mathbb{R}_{\geq 0}, \qquad (u_i, \gamma_j) \mapsto x_{ij}^{\mathrm{SH}}.$$

Equivalently, $x^{\mathrm{SH}}$ is represented by the *superhypersoft decision matrix* $X^{\mathrm{SH}} = (x_{ij}^{\mathrm{SH}}) \in \mathbb{R}_{\geq 0}^{n \times m}$.

(*Crisp/graded inputs.*) If the raw information is given as a (crisp) SuperHyperSoft set $F : \Gamma \to \mathcal{P}(U)$, define

$$x_{ij}^{\mathrm{SH}} := \mathbf{1}_{\{u_i \in F(\gamma_j)\}}.$$

If the raw information is graded (fuzzy, neutrosophic, interval-valued, etc.), assume a fixed score map Score to $\mathbb{R}_{\geq 0}$ and set $x_{ij}^{\mathrm{SH}} := \mathrm{Score}(\text{grade of } u_i \text{ under } \gamma_j)$.

**Definition 5.2.6** (SuperHyperSoft TOPSIS)**.** Assume the superhypersoft decision matrix $X^{\mathrm{SH}} = (x_{ij}^{\mathrm{SH}}) \in \mathbb{R}_{\geq 0}^{n \times m}$ is given. Let $w = (w_1, \ldots, w_m)$ be weights with $w_j > 0$ and $\sum_{j=1}^{m} w_j = 1$. Let $B$ (benefit) and $C$ (cost) be a partition of $\{1, \ldots, m\}$.

**(1) Vector normalization.** For each $j$ let

$$d_j^{\mathrm{SH}} := \sqrt{\sum_{i=1}^{n} (x_{ij}^{\mathrm{SH}})^2}, \qquad r_{ij}^{\mathrm{SH}} := \begin{cases} \dfrac{x_{ij}^{\mathrm{SH}}}{d_j^{\mathrm{SH}}}, & d_j^{\mathrm{SH}} > 0, \\ 0, & d_j^{\mathrm{SH}} = 0. \end{cases}$$

Let $R^{\mathrm{SH}} = (r_{ij}^{\mathrm{SH}})$.

**(2) Weighted normalized matrix.** Define $V^{\mathrm{SH}} = (v_{ij}^{\mathrm{SH}})$ by

$$v_{ij}^{\mathrm{SH}} := w_j \, r_{ij}^{\mathrm{SH}}.$$

**(3) Ideal solutions.** For each $j$ define

$$v_j^{\mathrm{SH},+} := \begin{cases} \max_{1 \leq i \leq n} v_{ij}^{\mathrm{SH}}, & j \in B, \\ \min_{1 \leq i \leq n} v_{ij}^{\mathrm{SH}}, & j \in C, \end{cases} \qquad v_j^{\mathrm{SH},-} := \begin{cases} \min_{1 \leq i \leq n} v_{ij}^{\mathrm{SH}}, & j \in B, \\ \max_{1 \leq i \leq n} v_{ij}^{\mathrm{SH}}, & j \in C. \end{cases}$$

Let $A^{\mathrm{SH},+} = (v_1^{\mathrm{SH},+}, \ldots, v_m^{\mathrm{SH},+})$ and $A^{\mathrm{SH},-} = (v_1^{\mathrm{SH},-}, \ldots, v_m^{\mathrm{SH},-})$.

**(4) Separation measures.** For each alternative $u_i$ set

$$S_i^{\mathrm{SH},+} := \sqrt{\sum_{j=1}^{m} \left(v_{ij}^{\mathrm{SH}} - v_j^{\mathrm{SH},+}\right)^2}, \qquad S_i^{\mathrm{SH},-} := \sqrt{\sum_{j=1}^{m} \left(v_{ij}^{\mathrm{SH}} - v_j^{\mathrm{SH},-}\right)^2}.$$

**(5) Closeness coefficient and ranking.** Define

$$\mathbb{C}_i^{\mathrm{SH}} := \begin{cases} \dfrac{S_i^{\mathrm{SH},-}}{S_i^{\mathrm{SH},+} + S_i^{\mathrm{SH},-}}, & S_i^{\mathrm{SH},+} + S_i^{\mathrm{SH},-} > 0, \\ \dfrac{1}{2}, & S_i^{\mathrm{SH},+} + S_i^{\mathrm{SH},-} = 0, \end{cases}$$

and rank alternatives by decreasing $\mathbb{C}_i^{\mathrm{SH}}$.

**Definition 5.2.7** (Singleton embedding of hypersoft parameters)**.** Let $\mathcal{C}^{\mathrm{H}} := A_1 \times \cdots \times A_k$ be the hypersoft parameter space. Define the map

$$\iota : \mathcal{C}^{\mathrm{H}} \to \mathcal{C}^{\mathrm{SH}}, \qquad \iota(a_1, \ldots, a_k) \ := \ (\{a_1\}, \ldots, \{a_k\}).$$

For $\Lambda = \{\lambda_1, \ldots, \lambda_m\} \subseteq \mathcal{C}^{\mathrm{H}}$, set

$$\Gamma \ := \ \iota(\Lambda) = \{\iota(\lambda_1), \ldots, \iota(\lambda_m)\} \subseteq \mathcal{C}^{\mathrm{SH}}.$$

**Theorem 5.2.8** (Singleton reduction of SuperHyperSoft TOPSIS to HyperSoft TOPSIS)**.** *Let $U$ be a set of alternatives and let $\Lambda = \{\lambda_1, \ldots, \lambda_m\} \subseteq A_1 \times \cdots \times A_k$ be a hypersoft criterion family. Let $x^{\mathrm{H}} : U \times \Lambda \to \mathbb{R}_{\geq 0}$ be a numeric hypersoft evaluation and define $\Gamma = \iota(\Lambda)$ as in Theorem 5.2.7. Define a superhypersoft evaluation $x^{\mathrm{SH}} : U \times \Gamma \to \mathbb{R}_{\geq 0}$ by*

$$x^{\mathrm{SH}}(u, \iota(\lambda)) \ := \ x^{\mathrm{H}}(u, \lambda), \qquad (u \in U, \ \lambda \in \Lambda).$$

*Fix the same weight vector $w$ and the same benefit/cost partition $(B, C)$ for both methods. Then the SuperHyperSoft TOPSIS closeness coefficients equal the HyperSoft TOPSIS closeness coefficients:*

$$\mathbb{C}_i^{\mathrm{SH}} \ = \ \mathbb{C}_i^{\mathrm{H}} \qquad \textit{for all } i \in \{1, \ldots, n\},$$

*and hence both methods produce the same ranking of alternatives (with the same tie convention). This establishes faithful singleton reduction, not a claim of strictly greater ranking expressivity.*

*Proof.* Index $\Lambda = \{\lambda_1, \ldots, \lambda_m\}$ and $\Gamma = \{\gamma_1, \ldots, \gamma_m\}$ with $\gamma_j = \iota(\lambda_j)$. By definition of $x^{\mathrm{SH}}$, the two decision matrices coincide entrywise:

$$x_{ij}^{\mathrm{SH}} = x^{\mathrm{SH}}(u_i, \gamma_j) = x^{\mathrm{SH}}(u_i, \iota(\lambda_j)) = x^{\mathrm{H}}(u_i, \lambda_j) = x_{ij}^{\mathrm{H}}.$$

Therefore, for each column $j$ the normalization denominators are equal:

$$d_j^{\mathrm{SH}} = \sqrt{\sum_{i=1}^{n} (x_{ij}^{\mathrm{SH}})^2} = \sqrt{\sum_{i=1}^{n} (x_{ij}^{\mathrm{H}})^2} = d_j^{\mathrm{H}}.$$

Hence $r_{ij}^{\mathrm{SH}} = r_{ij}^{\mathrm{H}}$ for all $i, j$, and thus also

$$v_{ij}^{\mathrm{SH}} = w_j r_{ij}^{\mathrm{SH}} = w_j r_{ij}^{\mathrm{H}} = v_{ij}^{\mathrm{H}}.$$

Since the weighted normalized matrices coincide, their componentwise maxima/minima over $i$ coincide as well; thus $A^{\mathrm{SH},+} = A^{\mathrm{H},+}$ and $A^{\mathrm{SH},-} = A^{\mathrm{H},-}$. Consequently, the separation measures are identical for each $i$:

$$S_i^{\mathrm{SH},\pm} = \sqrt{\sum_{j=1}^{m} (v_{ij}^{\mathrm{SH}} - v_j^{\mathrm{SH},\pm})^2} = \sqrt{\sum_{j=1}^{m} (v_{ij}^{\mathrm{H}} - v_j^{\mathrm{H},\pm})^2} = S_i^{\mathrm{H},\pm}.$$

Plugging into the closeness coefficient formula yields $\mathbb{C}_i^{\mathrm{SH}} = \mathbb{C}_i^{\mathrm{H}}$ for all $i$. Therefore the induced rankings coincide. □

## 5.3 Soft, HyperSoft, and SuperHyperSoft AHP

AHP is an MCDM method using pairwise comparisons to derive ratio-scale weights, aggregate priorities hierarchically, and rank alternatives [728–731]. Soft AHP applies pairwise comparison matrices to soft parameters, derives criteria weights and alternative priorities, then aggregates them into rankings. HyperSoft AHP extends Soft AHP by using multiple attribute value tuples as criteria, enabling richer parameterized pairwise evaluations for alternatives. SuperHyperSoft AHP generalizes HyperSoft AHP by allowing set valued attribute choices per criterion tuple, improving decision flexibility under uncertainty contexts.

**Definition 5.3.1** (Positive reciprocal (pairwise-comparison) matrix)**.** Let $n \in \mathbb{N}$. A matrix $A = (a_{ij}) \in \mathbb{R}^{n\times n}$ is called a *positive reciprocal matrix* if

$$a_{ij} > 0, \qquad a_{ii} = 1, \qquad a_{ij} = \frac{1}{a_{ji}} \quad (1 \le i, j \le n).$$

We denote by $\mathcal{R}_n$ the set of all positive reciprocal matrices of size $n$.

**Definition 5.3.2** (Priority vector operator)**.** Let $A \in \mathcal{R}_n$. Since $A$ is a positive matrix, by the Perron–Frobenius theorem $A$ has a largest eigenvalue $\lambda_{\max}(A) > 0$ with a strictly positive eigenvector. Define the *priority vector* $\pi(A) \in \mathbb{R}^n_{>0}$ as any Perron eigenvector normalized to sum 1:

$$A\,\pi(A) = \lambda_{\max}(A)\,\pi(A), \qquad \sum_{i=1}^{n} \pi_i(A) = 1.$$

(When the Perron eigenvalue is simple, $\pi(A)$ is unique.)

**Remark 5.3.3** (Consistency)**.** A matrix $A \in \mathcal{R}_n$ is (multiplicatively) *consistent* iff $a_{ij}a_{jk} = a_{ik}$ for all $i, j, k$. In that case there exists $w \in \mathbb{R}^n_{>0}$ such that $a_{ij} = w_i/w_j$, and then $\pi(A)$ recovers $w$ up to normalization.

**Definition 5.3.4** (Soft AHP decision instance)**.** Let $U = \{u_1, \ldots, u_n\}$ be a finite set of alternatives and let $E$ be a set of parameters (criteria). Fix a *soft parameter set* $S = \{e_1, \ldots, e_m\} \subseteq E$.

A *Soft AHP decision instance* is a tuple

$$\mathsf{SAHP} = (U, E, S,\ A^{(0)},\ \{A^{(e)}\}_{e\in S}),$$

where

(i) $A^{(0)} \in \mathcal{R}_m$ is the criteria pairwise-comparison matrix indexed by $S$;

(ii) for each $e \in S$, $A^{(e)} \in \mathcal{R}_n$ is the alternative pairwise-comparison matrix under criterion $e$.

The *criteria weight vector* is $w = \pi(A^{(0)}) \in \mathbb{R}^m_{>0}$, and the *local alternative weight* under $e_j$ is $p^{(j)} = \pi(A^{(e_j)}) \in \mathbb{R}^n_{>0}$.

The *global priority* (overall score) of alternatives is the vector

$$P := \sum_{j=1}^{m} w_j \, p^{(j)} \in \mathbb{R}^n_{>0}, \qquad \text{so} \quad P_i = \sum_{j=1}^{m} w_j \, p_i^{(j)}.$$

A *Soft AHP ranking* is any ordering of $U$ that is nonincreasing in $P_i$.

**Definition 5.3.5** (HyperSoft AHP decision instance)**.** Let $U = \{u_1, \dots, u_n\}$ be alternatives. Let $\mathcal{A}_1, \dots, \mathcal{A}_k$ be pairwise-disjoint attribute-value sets and define the hypersoft parameter space $\mathcal{C}^{\mathrm{H}} = \mathcal{A}_1 \times \cdots \times \mathcal{A}_k$. Fix a finite *criteria-tuples family*

$$\Lambda = \{\lambda_1, \dots, \lambda_m\} \subseteq \mathcal{C}^{\mathrm{H}}.$$

A *HyperSoft AHP decision instance* is a tuple

$$\mathsf{HAHP} = (U, \mathcal{A}_1, \dots, \mathcal{A}_k, \Lambda, \; A^{(0)}, \; \{A^{(\lambda)}\}_{\lambda \in \Lambda}),$$

where $A^{(0)} \in \mathcal{R}_m$ compares the criteria-tuples in $\Lambda$, and for each $\lambda \in \Lambda$, $A^{(\lambda)} \in \mathcal{R}_n$ compares alternatives under criterion-tuple $\lambda$.

Define $w = \pi(A^{(0)}) \in \mathbb{R}^m_{>0}$ and $p^{(j)} = \pi(A^{(\lambda_j)}) \in \mathbb{R}^n_{>0}$. The *global priority* is

$$P := \sum_{j=1}^{m} w_j \, p^{(j)} \in \mathbb{R}^n_{>0},$$

and alternatives are ranked by nonincreasing $P_i$.

**Theorem 5.3.6** (HyperSoft AHP generalizes Soft AHP)**.** *Every Soft AHP decision instance can be realized as a HyperSoft AHP decision instance, and under this realization both methods produce identical global priorities and rankings.*

*Proof.* Let $\mathsf{SAHP} = (U, E, S, A^{(0)}, \{A^{(e)}\}_{e \in S})$ with $S = \{e_1, \dots, e_m\}$. Set $k = 1$ and let $\mathcal{A}_1 := E$. Then $\mathcal{C}^{\mathrm{H}} = \mathcal{A}_1 = E$. Define $\Lambda := S \subseteq E$ and identify $\lambda_j \equiv e_j$.

Now define a HyperSoft AHP instance by taking the same criteria matrix $A^{(0)} \in \mathcal{R}_m$ and, for each $\lambda = e \in \Lambda$, set $A^{(\lambda)} := A^{(e)}$. Then by construction, the criteria weight vector $w = \pi(A^{(0)})$ is the same in both models, and each local vector satisfies

$$\pi(A^{(\lambda_j)}) = \pi(A^{(e_j)}).$$

Hence the global priority vectors coincide:

$$P^{\mathrm{H}} = \sum_{j=1}^{m} w_j \, \pi(A^{(\lambda_j)}) = \sum_{j=1}^{m} w_j \, \pi(A^{(e_j)}) = P^{\mathrm{S}}.$$

Therefore the induced rankings are identical. ∎

**Definition 5.3.7** (SuperHyperSoft AHP decision instance)**.** Let $U = \{u_1, \ldots, u_n\}$ be alternatives. Let $A_1, \ldots, A_k$ be pairwise-disjoint attribute-value sets and define the superhypersoft parameter space

$$\mathcal{C}^{\mathrm{SH}} = \mathcal{P}(A_1) \times \mathcal{P}(A_2) \times \cdots \times \mathcal{P}(A_k).$$

Fix a finite *super-criteria family*

$$\Gamma = \{\gamma_1, \ldots, \gamma_m\} \subseteq \mathcal{C}^{\mathrm{SH}}, \qquad \gamma_j = (\alpha_{1j}, \ldots, \alpha_{kj}),\ \alpha_{tj} \subseteq A_t.$$

A *SuperHyperSoft AHP decision instance* is

$$\mathsf{SHAHP} = (U, A_1, \ldots, A_k, \Gamma,\ A^{(0)},\ \{A^{(\gamma)}\}_{\gamma\in\Gamma}),$$

where $A^{(0)} \in \mathcal{R}_m$ compares the elements of $\Gamma$, and for each $\gamma \in \Gamma$, $A^{(\gamma)} \in \mathcal{R}_n$ compares alternatives under super-criterion $\gamma$.

Let $w = \pi(A^{(0)}) \in \mathbb{R}^m_{>0}$ and $p^{(j)} = \pi(A^{(\gamma_j)}) \in \mathbb{R}^n_{>0}$. The *global priority* is

$$P \ := \ \sum_{j=1}^{m} w_j\, p^{(j)} \in \mathbb{R}^n_{>0},$$

and alternatives are ranked by nonincreasing $P_i$.

**Definition 5.3.8** (Singleton embedding)**.** Let $\mathcal{C}^{\mathrm{H}} = A_1 \times \cdots \times A_k$ and $\mathcal{C}^{\mathrm{SH}} = \mathcal{P}(A_1) \times \cdots \times \mathcal{P}(A_k)$. Define

$$\iota : \mathcal{C}^{\mathrm{H}} \to \mathcal{C}^{\mathrm{SH}}, \qquad \iota(a_1, \ldots, a_k) := (\{a_1\}, \ldots, \{a_k\}).$$

For $\Lambda = \{\lambda_1, \ldots, \lambda_m\} \subseteq \mathcal{C}^{\mathrm{H}}$, set $\Gamma := \iota(\Lambda)$ and index $\Gamma = \{\gamma_1, \ldots, \gamma_m\}$ by $\gamma_j = \iota(\lambda_j)$.

**Theorem 5.3.9** (SuperHyperSoft AHP generalizes HyperSoft AHP)**.** *Every HyperSoft AHP decision instance can be realized as a SuperHyperSoft AHP decision instance via the singleton embedding $\iota$. Under this realization, both methods yield identical global priorities and rankings.*

*Proof.* Let $\mathsf{HAHP} = (U, A_1, \ldots, A_k, \Lambda, A^{(0)}, \{A^{(\lambda)}\}_{\lambda\in\Lambda})$ with $\Lambda = \{\lambda_1, \ldots, \lambda_m\}$. Define $\Gamma = \iota(\Lambda)$ and $\gamma_j = \iota(\lambda_j)$ as in Theorem 5.3.8. Construct a SuperHyperSoft AHP instance by taking the same criteria matrix $A^{(0)} \in \mathcal{R}_m$ and defining

$$A^{(\gamma_j)} \ := \ A^{(\lambda_j)} \qquad (j = 1, \ldots, m).$$

Then the criteria weight vector is the same $w = \pi(A^{(0)})$ in both models, and each local alternative priority vector satisfies

$$\pi(A^{(\gamma_j)}) = \pi(A^{(\lambda_j)}).$$

Hence the global priority vectors coincide:

$$P^{\mathrm{SH}} = \sum_{j=1}^{m} w_j\, \pi(A^{(\gamma_j)}) = \sum_{j=1}^{m} w_j\, \pi(A^{(\lambda_j)}) = P^{\mathrm{H}}.$$

Therefore the induced rankings are identical. □

## 5.4 Soft, HyperSoft, and SuperHyperSoft VIKOR

VIKOR is an MCDM method ranking alternatives by compromise using group utility and individual regret, controlled by parameter v [732–735]. Soft VIKOR ranks alternatives using soft parameters, computes best worst values, group utility S, individual regret R, compromise Q index. HyperSoft VIKOR replaces single parameters with attribute value tuples, then applies VIKOR normalization, S and R aggregation, Q ranking procedure. SuperHyperSoft VIKOR allows set valued attribute choices per criterion, embedding HyperSoft via singleton sets, preserving S R Q outputs exactly.

**Notation 5.4.1** (Alternatives, criteria-orientation, and weights)**.** *Let $U = \{u_1, \ldots, u_n\}$ be a finite set of alternatives and let $m \in \mathbb{N}$. A* criterion *(parameter) will always be equipped with an* orientation

$$\tau \in \{\mathrm{ben}, \mathrm{cost}\},$$

*meaning that larger values are preferred for* ben *and smaller values are preferred for* cost*.*

*A* weight vector *on a finite criterion-family $C = \{c_1, \ldots, c_m\}$ is $w = (w_1, \ldots, w_m) \in [0,1]^m$ with $\sum_{j=1}^{m} w_j = 1$.*

**Definition 5.4.2** (Ideal best/worst values)**.** Let $C = \{c_1, \ldots, c_m\}$ be a criterion-family with orientations $\tau_j \in \{\mathrm{ben}, \mathrm{cost}\}$. Let $f : U \times C \to \mathbb{R}$ be an evaluation function and write $f_{ij} := f(u_i, c_j)$.

For each $j \in \{1, \ldots, m\}$ define the *ideal best* value $f_j^*$ and *ideal worst* value $f_j^-$ by

$$(f_j^*, f_j^-) := \begin{cases} \big(\max_{1\le i\le n} f_{ij},\ \min_{1\le i\le n} f_{ij}\big), & \tau_j = \mathrm{ben}, \\ \big(\min_{1\le i\le n} f_{ij},\ \max_{1\le i\le n} f_{ij}\big), & \tau_j = \mathrm{cost}. \end{cases}$$

**Definition 5.4.3** (Normalized loss (distance from the ideal))**.** Under the hypotheses of Theorem 5.4.2, define the *normalized loss* $d_{ij} \in [0,1]$ by

$$d_{ij} := \begin{cases} 0, & f_j^* = f_j^-, \\ \dfrac{f_j^* - f_{ij}}{f_j^* - f_j^-}, & \tau_j = \mathrm{ben} \text{ and } f_j^* \ne f_j^-, \\ \dfrac{f_{ij} - f_j^*}{f_j^- - f_j^*}, & \tau_j = \mathrm{cost} \text{ and } f_j^* \ne f_j^-. \end{cases}$$

Equivalently, $d_{ij} = 0$ iff $u_i$ attains the ideal best on criterion $c_j$ (or the criterion is constant).

**Definition 5.4.4** (Group utility and individual regret)**.** Let $d_{ij}$ be as in Theorem 5.4.3 and let $w \in [0,1]^m$ with $\sum_j w_j = 1$. Define for each alternative $u_i$:

$$S_i := \sum_{j=1}^{m} w_j\, d_{ij} \qquad \text{and} \qquad R_i := \max_{1\le j\le m} (w_j\, d_{ij}).$$

Set

$$S^* := \min_{1\le i\le n} S_i, \quad S^- := \max_{1\le i\le n} S_i, \qquad R^* := \min_{1\le i\le n} R_i, \quad R^- := \max_{1\le i\le n} R_i.$$

**Definition 5.4.5** (Compromise index $Q$)**.** Let $v \in [0,1]$ be fixed (often $v = 1/2$). With $S_i, R_i, S^*, S^-, R^*, R^-$ as in Theorem 5.4.4, define

$$Q_i \;:=\; v\,\frac{S_i - S^*}{S^- - S^*} \;+\; (1-v)\,\frac{R_i - R^*}{R^- - R^*},$$

using the convention that $\frac{0}{0} := 0$ (so if $S^- = S^*$ then the first fraction is set to 0, and similarly for $R^- = R^*$).

**Definition 5.4.6** (Soft VIKOR decision instance and solution)**.** Let $E$ be a parameter set and let $S = \{e_1, \ldots, e_m\} \subseteq E$ be a finite soft-parameter set. A *Soft VIKOR decision instance* is a tuple

$$\mathsf{SVIKOR} = (U, E, S,\ \tau,\ w,\ f,\ v),$$

where

(i) $\tau : S \to \{\text{ben}, \text{cost}\}$ assigns an orientation $\tau_j := \tau(e_j)$ to each parameter;

(ii) $w = (w_1, \ldots, w_m) \in [0,1]^m$ with $\sum_{j=1}^{m} w_j = 1$ is the criterion-weight vector;

(iii) $f : U \times S \to \mathbb{R}$ is an evaluation function (decision matrix) with $f_{ij} := f(u_i, e_j)$;

(iv) $v \in [0,1]$ is the compromise coefficient.

Compute $f_j^*, f_j^-$ by Theorem 5.4.2, then $d_{ij}$ by Theorem 5.4.3, then $S_i, R_i$ by Theorem 5.4.4, and finally $Q_i$ by Theorem 5.4.5. A *Soft VIKOR ranking* is any ordering of $U$ that is nondecreasing in $Q_i$.

(Optionally) If $n \geq 2$, let $u_{(1)}$ and $u_{(2)}$ be the first and second alternatives under the $Q$-ordering. Define $DQ := \frac{1}{n-1}$. For $n = 1$, the unique alternative is selected without applying this advantage test. If (a) $Q(u_{(2)}) - Q(u_{(1)}) \geq DQ$ (acceptable advantage) and (b) $u_{(1)}$ is also best by $S$ or by $R$ (acceptable stability), then $u_{(1)}$ is called the *(unique) compromise solution.* Otherwise one may output a compromise set consisting of the top few $Q$-alternatives.

**Definition 5.4.7** (HyperSoft VIKOR decision instance)**.** Let $\mathcal{A}_1, \ldots, \mathcal{A}_k$ be (pairwise-disjoint) attribute-value sets and let

$$\mathcal{C}^{\mathrm{H}} := \mathcal{A}_1 \times \cdots \times \mathcal{A}_k$$

be the hypersoft parameter domain. Fix a finite criteria-tuples family $\Lambda = \{\lambda_1, \ldots, \lambda_m\} \subseteq \mathcal{C}^{\mathrm{H}}$.

A *HyperSoft VIKOR decision instance* is a tuple

$$\mathsf{HVIKOR} = (U, \mathcal{A}_1, \ldots, \mathcal{A}_k, \Lambda,\ \tau,\ w,\ f,\ v),$$

where $\tau : \Lambda \to \{\text{ben}, \text{cost}\}$, $w \in [0,1]^m$ with $\sum_j w_j = 1$, $f : U \times \Lambda \to \mathbb{R}$, and $v \in [0,1]$.

With indices $f_{ij} := f(u_i, \lambda_j)$ and $\tau_j := \tau(\lambda_j)$, define $f_j^*, f_j^-$, $d_{ij}$, $S_i, R_i$, and $Q_i$ exactly as in Theorems 5.4.2 to 5.4.5. The ranking is nondecreasing in $Q_i$.

**Theorem 5.4.8** (HyperSoft VIKOR generalizes Soft VIKOR). *Every Soft VIKOR decision instance can be realized as a HyperSoft VIKOR decision instance. Under this realization, all computed quantities* $(f_j^*, f_j^-, d_{ij}, S_i, R_i, Q_i)$ *coincide, hence the rankings (and compromise solutions/sets) coincide.*

*Proof.* Let $\mathsf{SVIKOR} = (U, E, S, \tau, w, f, v)$ with $S = \{e_1, \ldots, e_m\}$. Set $k = 1$ and define $\mathcal{A}_1 := E$, so $\mathcal{C}^{\mathrm{H}} = \mathcal{A}_1 = E$. Let $\Lambda := S \subseteq E$ and identify $\lambda_j \equiv e_j$.

Define the HyperSoft instance by keeping the same $v$ and the same weight vector $w$, setting $\tau(\lambda_j) := \tau(e_j)$, and defining $f(u, \lambda) := f(u, e)$ under the identification $\lambda \equiv e$.

Then for every $i, j$ we have identical entries $f_{ij}$ in both models, hence the best/worst values $f_j^*, f_j^-$ from Theorem 5.4.2 coincide, and therefore the normalized losses $d_{ij}$ from Theorem 5.4.3 coincide. With the same $w$, this forces $S_i$ and $R_i$ from Theorem 5.4.4 to coincide, and consequently $Q_i$ from Theorem 5.4.5 coincides. Thus the induced rankings and any compromise outputs coincide. □

**Definition 5.4.9** (SuperHyperSoft VIKOR decision instance). Let $A_1, \ldots, A_k$ be (pairwise-disjoint) attribute-value sets and define the superhypersoft domain

$$\mathcal{C}^{\mathrm{SH}} := \mathcal{P}(A_1) \times \mathcal{P}(A_2) \times \cdots \times \mathcal{P}(A_k).$$

Fix a finite family $\Gamma = \{\gamma_1, \ldots, \gamma_m\} \subseteq \mathcal{C}^{\mathrm{SH}}$ of *set-valued* criteria-tuples.

A *SuperHyperSoft VIKOR decision instance* is a tuple

$$\mathsf{SHVIKOR} = (U, A_1, \ldots, A_k, \Gamma,\ \tau,\ w,\ f,\ v),$$

where $\tau : \Gamma \to \{\mathrm{ben}, \mathrm{cost}\}$, $w \in [0,1]^m$ with $\sum_j w_j = 1$, $f : U \times \Gamma \to \mathbb{R}$, and $v \in [0,1]$.

Define $f_j^*, f_j^-$, $d_{ij}$, $S_i, R_i$, and $Q_i$ exactly as in Theorems 5.4.2 to 5.4.5 with $\gamma_j$ in place of $c_j$. Rank alternatives nondecreasingly by $Q_i$.

**Definition 5.4.10** (Singleton embedding). Let $\mathcal{C}^{\mathrm{H}} := A_1 \times \cdots \times A_k$ and $\mathcal{C}^{\mathrm{SH}} := \mathcal{P}(A_1) \times \cdots \times \mathcal{P}(A_k)$. Define

$$\iota : \mathcal{C}^{\mathrm{H}} \to \mathcal{C}^{\mathrm{SH}}, \qquad \iota(a_1, \ldots, a_k) := (\{a_1\}, \ldots, \{a_k\}).$$

For $\Lambda = \{\lambda_1, \ldots, \lambda_m\} \subseteq \mathcal{C}^{\mathrm{H}}$ set $\Gamma := \iota(\Lambda)$, indexed as $\gamma_j := \iota(\lambda_j)$.

**Theorem 5.4.11** (SuperHyperSoft VIKOR generalizes HyperSoft VIKOR). *Every HyperSoft VIKOR decision instance can be realized as a SuperHyperSoft VIKOR decision instance via the singleton embedding* $\iota$. *Under this realization, all computed quantities* $(f_j^*, f_j^-, d_{ij}, S_i, R_i, Q_i)$ *coincide, hence the rankings coincide.*

*Proof.* Let $\mathsf{HVIKOR} = (U, A_1, \ldots, A_k, \Lambda, \tau, w, f, v)$ with $\Lambda = \{\lambda_1, \ldots, \lambda_m\} \subseteq A_1 \times \cdots \times A_k$. Let $\Gamma = \iota(\Lambda)$ and $\gamma_j = \iota(\lambda_j)$ as in Theorem 5.4.10.

Define the SuperHyperSoft instance by keeping the same $v$ and $w$, setting

$$\tau(\gamma_j) := \tau(\lambda_j), \qquad f(u, \gamma_j) := f(u, \lambda_j) \quad (u \in U,\ j = 1, \ldots, m).$$

Then for every $i, j$, the entries $f_{ij}$ coincide under the identification $\gamma_j \leftrightarrow \lambda_j$. Therefore $f_j^*, f_j^-$ coincide (same extrema over the same numbers), hence $d_{ij}$ coincide. With identical weights $w$, the aggregates $S_i$ and $R_i$ coincide, and thus $Q_i$ coincides. Consequently the induced rankings coincide. □

Other decision-making methods, such as Soft ELECTRE [736, 737], Soft PROMETHEE [738, 739], and Soft WASPAS [740, 741], are also known.

# Chapter 6

# Conclusion

In this book, we provided a survey-style overview of Soft Set Theory and its major developments. In addition to the basic Soft Set framework, we examined a broad range of extensions, including HyperSoft Sets, SuperHyperSoft Sets, hierarchical and uncertainty-aware models, as well as related mathematical structures and decision-making methods.

These concepts illustrate the flexibility of soft parameterization for representing different forms of structured and uncertain information. At the same time, distinct models may use different parameter spaces, compatibility rules, inference semantics, and computational assumptions; such differences should not be suppressed by notation alone. In particular, a reduction obtained by singleton parameter embedding does not establish strict expressive superiority, while an arbitrary node-scoring representation does not imply that a standard permutation-equivariant GNN can realize every soft set.

Further research may examine algebraic and order-theoretic operations, precise conditions for structure-preserving mappings, computational complexity, approximation algorithms, and empirical validation of decision-making models. Higher-order graph constructions will also benefit from explicit level tagging and rigorously defined incidence and closure conditions. These directions may strengthen the formal foundations and applications of Soft Set Theory.

# Disclaimer

## Funding

This study did not receive any financial or external support from organizations or individuals.

## Acknowledgments

We extend our sincere gratitude to everyone who provided insights, inspiration, and assistance throughout this research. We particularly thank our readers for their interest and acknowledge the authors of the cited works for laying the foundation that made our study possible. We also appreciate the support from individuals and institutions that provided the resources and infrastructure needed to produce and share this book. Finally, we are grateful to all those who supported us in various ways during this project.

## Data Availability

This research is purely theoretical, involving no data collection or analysis. We encourage future researchers to pursue empirical investigations to further develop and validate the concepts introduced here.

## Ethical Approval

As this research is entirely theoretical in nature and does not involve human participants or animal subjects, no ethical approval is required.

## Use of Generative AI and AI-Assisted Tools

I use generative AI and AI-assisted tools for tasks such as English grammar checking, and I do not employ them in any way that violates ethical standards.

## Conflicts of Interest

The authors confirm that there are no conflicts of interest related to the research or its publication.

## Disclaimer

This work presents theoretical concepts that have not yet undergone practical testing or validation. Future researchers are encouraged to apply and assess these ideas in empirical contexts. While every effort has been made to ensure accuracy and appropriate referencing, unintentional errors or omissions may still exist. Readers are advised to verify referenced materials on their own. The views and conclusions expressed here are the authors' own and do not necessarily reflect those of their affiliated organizations.

# Appendix (List of Tables)

*

# Bibliography

[1] Pradip Kumar Maji, Ranjit Biswas, and A Ranjan Roy. Soft set theory. *Computers & mathematics with applications*, 45(4-5):555–562, 2003.

[2] Dmitriy Molodtsov. Soft set theory-first results. *Computers & mathematics with applications*, 37(4-5):19–31, 1999.

[3] Thomas Jech. *Set theory: The third millennium edition, revised and expanded.* Springer, 2003.

[4] Lotfi A Zadeh. Fuzzy sets. *Information and control*, 8(3):338–353, 1965.

[5] Krassimir T Atanassov. Circular intuitionistic fuzzy sets. *Journal of Intelligent & Fuzzy Systems*, 39(5):5981–5986, 2020.

[6] Vicenç Torra. Hesitant fuzzy sets. *International journal of intelligent systems*, 25(6):529–539, 2010.

[7] Bui Cong Cuong. Picture fuzzy sets. *Journal of Computer Science and Cybernetics*, 30:409, 2015.

[8] Said Broumi, Mohamed Talea, Assia Bakali, and Florentin Smarandache. Single valued neutrosophic graphs. *Journal of New theory*, 10:86–101, 2016.

[9] Haibin Wang, Florentin Smarandache, Yanqing Zhang, and Rajshekhar Sunderraman. *Single valued neutrosophic sets.* Infinite study, 2010.

[10] R Radha, A Stanis Arul Mary, and Florentin Smarandache. Quadripartitioned neutrosophic pythagorean soft set. *International Journal of Neutrosophic Science (IJNS) Volume 14, 2021*, page 11, 2021.

[11] Rama Mallick and Surapati Pramanik. *Pentapartitioned neutrosophic set and its properties*, volume 36. Infinite Study, 2020.

[12] Lin Wei. An integrated decision-making framework for blended teaching quality evaluation in college english courses based on the double-valued neutrosophic sets. *J. Intell. Fuzzy Syst.*, 45:3259–3266, 2023.

[13] Hu Zhao and Hong-Ying Zhang. On hesitant neutrosophic rough set over two universes and its application. *Artificial Intelligence Review*, 53:4387–4406, 2020.

[14] Florentin Smarandache. *Plithogenic set, an extension of crisp, fuzzy, intuitionistic fuzzy, and neutrosophic sets-revisited.* Infinite study, 2018.

[15] Florentin Smarandache. *Extension of HyperGraph to n-SuperHyperGraph and to Plithogenic n-SuperHyperGraph, and Extension of HyperAlgebra to n-ary (Classical-/Neutro-/Anti-) HyperAlgebra.* Infinite Study, 2020.

[16] Feng Feng, Xiaoyan Liu, Violeta Leoreanu-Fotea, and Young Bae Jun. Soft sets and soft rough sets. *Information Sciences*, 181(6):1125–1137, 2011.

[17] Lotfi A Zadeh. A note on z-numbers. *Information sciences*, 181(14):2923–2932, 2011.

[18] Florentin Smarandache. A unifying field in logics: Neutrosophic logic. In *Philosophy*, pages 1–141. American Research Press, 1999.

[19] Naeem Jan, Tahir Mahmood, Lemnaouar Zedam, and Zeeshan Ali. Multi-valued picture fuzzy soft sets and their applications in group decision-making problems. *Soft Computing*, 24:18857 – 18879, 2020.

[20] Eugenio Aguirre and Antonio González. Fuzzy behaviors for mobile robot navigation: design, coordination and fusion. *International Journal of Approximate Reasoning*, 25(3):255–289, 2000.

[21] Alberto Fernandez, Francisco Herrera, Oscar Cordon, Maria Jose del Jesus, and Francesco Marcelloni. Evolutionary fuzzy systems for explainable artificial intelligence: Why, when, what for, and where to? *IEEE Computational intelligence magazine*, 14(1):69–81, 2019.

[22] Yasmine M Ibrahim, Reem Essameldin, and Saad M Darwish. An adaptive hate speech detection approach using neutrosophic neural networks for social media forensics. *Computers, Materials & Continua*, 79(1), 2024.

[23] OM Khaled, AA Salama, Mostafa Herajy, MM El-Kirany, Huda E Khalid, Ahmed K Essa, and Ramiz Sabbagh. A novel approach for cyber-attack detection in iot networks with neutrosophic neural networks. *Neutrosophic Sets and Systems*, 86(1):48, 2025.

[24] Florentin Smarandache. *New types of soft sets "hypersoft set, indetermsoft set, indetermhypersoft set, and treesoft set": an improved version.* Infinite Study, 2023.

[25] Muhammad Ihsan, Atiqe Ur Rahman, and Muhammad Haris Saeed. Hypersoft expert set with application in decision making for recruitment process. In *Neutrosophic Sets and Systems*, 2021.

[26] Florentin Smarandache. Extension of soft set to hypersoft set, and then to plithogenic hypersoft set. *Neutrosophic sets and systems*, 22(1):168–170, 2018.

[27] Dongmei Li, Himanshu Dhumras, Bingda Zhang, Jian Zhou, Zhicheng Li, and Zhe Liu. Integrating q-rung picture fuzzy hypersoft sets with aczel–alsina operators for pharmaceutical supply chain management. *OPSEARCH*, pages 1–34, 2026.

[28] Edwin Sánchez-León, Eduardo Espinoza-Solís, Huber Echeverría, Antonio Blázquez-Zaballos, and Purificación Galindo-Villardón. Symmetric and asymmetric trends in sparse tucker decomposition based on plithogenic neutrosophic hypersoft sets: A bibliometric and conceptual review. *Symmetry (20738994)*, 18(9):1532, 2026.

[29] Haitham Qawaqneh, Abdallah Shihadeh, Wael Mahmoud Mohammad Salameh, Alaa M Abd El-latif, MI Elashiry, Muhammad Zeeshan, and Takaaki Fujita. Bonferroni-based aggregation operators and decision-making framework under complex fuzzy hypersoft sets for selection of smart healthcare systems. *Scientific Reports*, 2026.

[30] Aram Noori Qadir, Atiqe Ur Rahman, and Pishtiwan Othman Sabir. An integrated approach of complex fuzzy hypersoft set and possibility degree setting to manage uncertainties associated with multi-attribute decision-making. *Journal of Uncertain Systems*, page 2650019, 2026.

[31] Muhammad Abdullah, Khuram Ali Khan, Atiqe Ur Rahman, and Rostin Matendo Mabela. A novel multi-attribute decision-making strategy for industry 4.0 with uncertainties using rough approximate operators of fermatean fuzzy hypersoft sets. *International Journal of Computational Intelligence Systems*, 2026.

[32] Muhammad Sajid, Khuram Ali Khan, Atiqe Ur Rahman, and Rostin Matendo Mabela. Enhanced multi-attribute decision-making method to analyze renewable energy sources for sustainable transportation by synergizing the cubic set with picture fuzzy hypersoft sets. *Scientific Reports*, 2026.

[33] Orhan Dalkılıç. Virtual logic-based fuzzy parameterized fuzzy hypersoft sets: An algorithmic framework for improved decision-making. *El-Cezeri*, 13(2):233–244, 2026.

[34] Pairote Yiarayong. Advancing hepatitis diagnosis with hyperbolic fuzzy hypersoft sets in multiple criteria decision applications. *International Journal of Machine Learning and Cybernetics*, 17(4):172, 2026.

[35] Oswaldo Edison García Brito, Andrea Sofía Ribadeneira Vacacela, Carmen Hortensia Sánchez Burneo, and Mónica Cecilia Jimbo Galarza. English for specific purposes in the medical sciences to strengthen the professional profile of the higher education medicine student: a knowledge representation using superhypersoft sets. *Neutrosophic Sets and Systems*, 74(1):10, 2024.

[36] Mona Mohamed, Alaa Elmor, Florentin Smarandache, and Ahmed A Metwaly. An efficient superhypersoft framework for evaluating llms-based secure blockchain platforms. *Neutrosophic Sets and Systems*, 72:1–21, 2024.

[37] T Kiruthika, M Karpagadevi, S Krishnaprakash, and G Deepa. Superhypersoft sets using python and its applications in neutrosophic superhypersoft sets under topsis method. *Neutrosophic Sets and Systems*, 91:586–616, 2025.

[38] Florentin Smarandache. Foundation of the superhypersoft set and the fuzzy extension superhypersoft set: A new vision. *Neutrosophic Systems with Applications*, 11:48–51, 2023.

[39] Dua Sajid, Anns Uzair, Aiman Ishtiaq, Rashid Jan, Muhammad Talha Azhar, and Qamar Noureen. Nanotechnology prioritization through complex fermatean neutrosophic super-hypersoft structures. *Journal of Contemporary Decision Science*, pages 1–18, 2026.

[40] Takaaaki Fujita, Mithun Datta, and Kalyani Debnath. Weighted hypersoft and superhypersoft sets for high-dimensional decision models. *Journal of Decisions & Operations Research*, 11(1):1, 2026.

[41] G Hemalatha and A Francina Shalini. Applications of a few operations on pythagorean neutrosophic super hypersoft matrices. *SuperHyperTopologies and SuperHyperStructures with their Applications*, page 268, 2026.

[42] Takaaki Fujita and Florentin Smarandache. *Harnessing quantum superposition in soft set theory: Introducing quantum hypersoft and superhypersoft sets*. Infinite Study, 2025.

[43] Takaaki Fujita, Ajoy Kanti Das, Suman Das, Sankar Prasad Mondal, Volkan Duran, and Mithun Datta. DiGraphicSoft Sets and BiDiGraphicSoft Sets: Directed and Bidirected Extensions of GraphicSoft Modeling. *Neutrosophic Systems with Applications*, 26(7), 2026.

[44] Ali Alqazzaz and Karam M Sallam. Evaluation of sustainable waste valorization using treesoft set with neutrosophic sets. *Neutrosophic Sets and Systems*, 65(1):9, 2024.

[45] Edwin Collazos Paucar, Jeri G Ramón Ruffner de Vega, Efrén S Michue Salguedo, Agustina C Torres-Rodríguez, and Patricio A Santiago-Saturnino. Analysis using treesoft set of the strategic development plan for extreme poverty municipalities. *Neutrosophic Sets and Systems*, 69(1):3, 2024.

[46] G Dhanalakshmi, S Sandhiya, Florentin Smarandache, et al. Selection of the best process for desalination under a treesoft set environment using the multi-criteria decision-making method. *International Journal of Neutrosophic Science*, 23(3):140–40, 2024.

[47] Mona Gharib, Fatima Rajab, and Mona Mohamed. Harnessing tree soft set and soft computing techniques' capabilities in bioinformatics: Analysis, improvements, and applications. *Neutrosophic sets and systems*, 61:579–597, 2023.

[48] Takaaki Fujita. Polytree-soft sets and polyforest-soft sets: A directed acyclic framework for soft set modeling. *HyperSoft Set Methods in Engineering*, 4:11–23, 2025.

[49] Takaaki Fujita, Arif Mehmood, Ajoy Kanti Das, Suman Das, Volkan Duran, Arkan A Ghaib, and Talal Al-Hawary. Multitree-soft, pseudotree-soft set, hypertree-soft, andtree-to-tree-soft set. *Neutrosophic Computing and Machine Learning. ISSN 2574-1101*, 42:65–87, 2026.

[50] Florentin Smarandache. New types of soft sets: Hypersoft set, indetermsoft set, indetermhypersoft set, and treesoft set. *International Journal of Neutrosophic Science*, 2023.

[51] Hairong Luo. Forestsoft set approach for estimating innovation and entrepreneurship education in universities through a hierarchical and uncertainty-aware analytical framework. *Neutrosophic Sets and Systems*, 86(1):21, 2025.

[52] Takaaki Fujita, Ajoy Kanti Das, Arif Mehmood, Suman Das, and Volkan Duran. Decision analytics applications of the relationship between treesoft graphs and forestsoft graphs. *Applied Decision Analytics*, 2(1):73–92, 2026.

[53] Takaaki Fujita and Florentin Smarandache. *Quantum-TreeSoft Set and Quantum-ForestSoft Set.* Infinite Study, 2025.

[54] P Sathya, Nivetha Martin, and Florentine Smarandache. Plithogenic forest hypersoft sets in plithogenic contradiction based multi-criteria decision making. *Neutrosophic Sets and Systems*, 73:668–693, 2024.

[55] Viviana del Rocío Marfetan Marfetan, Lesly Gissela Tipanguano Chicaiza, Styven Andrés Pila Chicaiza, and Estephany Monserrath Ojeda Sanchez. Classification of cases of animal abuse in ecuador using indetermsoft and c4. 5 algorithms. *Neutrosophic Sets and Systems*, 92:121–133, 2025.

[56] Erick González Caballero, Ketty Marilú Moscoso-Paucarchuco, Noel Batista Hernandez, Lorenzo Jovanny Cevallos Torres, Maikel Leyva, and Victor Gustavo Gómez Rodríguez. Algorithms of designing decision trees from indeterm soft sets. In *Neutrosophic and Plithogenic Inventory Models for Applied Mathematics*, pages 561–586. IGI Global Scientific Publishing, 2025.

[57] Wei Wei and Pingting Peng. Weighted indetermsoft set for prioritized decision-making with indeterminacy and its application to green competitiveness evaluation in equipment manufacturing enterprises. *Neutrosophic Sets and Systems*, 85:1018–1026, 2025.

[58] Hai Yang and Cuijuan Lin. A recursive indetermtree soft set (rit-soft set) for dynamic and uncertain performance evaluation in college competitive sports. *Neutrosophic Sets and Systems*, 85:874–886, 2025.

[59] Tao Shen and Chunmei Mao. Sustainability impact of online consumption behavior from the perspective of digital empowerment: Indetermsoft set with application. *Neutrosophic Sets and Systems*, 82(1):24, 2025.

[60] Bhargavi Krishnamurthy and Sajjan G Shiva. Indetermsoft-set-based d* extra lite framework for resource provisioning in cloud computing. *Algorithms*, 17(11):479, 2024.

[61] Florentin Smarandache. *Introduction to SuperHyperAlgebra and Neutrosophic SuperHyperAlgebra.* Infinite Study, 2022.

[62] Yan Xu. A neutrosophic $\alpha$-discounting indetermhypersoft framework for evaluating agricultural product export trade quality under uncertainty. *Neutrosophic Sets and Systems*, 87:533–542, 2025.

[63] Takaaki Fujita. A comprehensive discussion on fuzzy hypersoft expert, superhypersoft, and indetermsoft graphs. *Neutrosophic Sets and Systems*, 77:241–263, 2025.

[64] Takaaki Fujita and Florentin Smarandache. Indetermsuperhypersoft sets, indetermforestsoft sets, and indetermtreesoft sets. *SuperHyperTopologies and SuperHyperStructures with their Applications*, page 23, 2025.

[65] Lingling Chen. A comprehensive indetermhypersoft set model for evaluating university literature education effectiveness: Integrating cultural context, argumentation skills, and dynamic progress. *Neutrosophic Sets and Systems*, 87:295–309, 2025.

[66] Takaaki Fujita and Florentin Smarandache. An introduction to advanced soft set variants: Superhypersoft sets, indetermsuperhypersoft sets, indetermtreesoft sets, bihypersoft sets, graphicsoft sets, and beyond. *Neutrosophic Sets and Systems*, 82:817–843, 2025.

[67] Takaaki Fujita and Florentin Smarandache. *Navigating Bipolar Indeterminacy: Bipolar IndetermSoft Sets and Bipolar IndetermHyperSoft Sets for Knowledge Representation.* Infinite Study, 2026.

[68] Takaaki Fujita, Raed Hatamleh, and Ahmed Salem Heilat. Contrasoft set and contrarough set with using upside-down logic. *Statistics, Optimization & Information Computing*, 2025.

[69] Vicenç Torra and Yasuo Narukawa. On hesitant fuzzy sets and decision. In *2009 IEEE international conference on fuzzy systems*, pages 1378–1382. IEEE, 2009.

[70] Yiwei Chen, Qiu Xie, Xiaoyu Ma, and Yuwei Li. Optimizing site selection for construction and demolition waste resource treatment plants using a hesitant neutrosophic set: a case study in xiamen, china. *Engineering Optimization*, pages 1–22, 2024.

[71] Muhammad Riaz, Muhammad Tahir Hamid, Deeba Afzal, Dragan Pamucar, and Yuming Chu. Multi-criteria decision making in robotic agri-farming with q-rung orthopair m-polar fuzzy sets. *PLoS ONE*, 16, 2021.

[72] Juanjuan Chen, Shenggang Li, Shengquan Ma, and Xueping Wang. m-polar fuzzy sets: an extension of bipolar fuzzy sets. *The scientific world journal*, 2014(1):416530, 2014.

[73] V Rajam and N Rajesh. Multipolar neutrosophic subalgebras/ideals of up-algebras. *International Journal of Neutrosophic Science (IJNS)*, 23(4), 2024.

[74] Muhammad Saqlain, Muhammad Riaz, Natasha Kiran, Poom Kumam, and Miin-Shen Yang. Water quality evaluation using generalized correlation coefficient for m-polar neutrosophic hypersoft sets. *Neutrosophic Sets and Systems, vol. 55/2023: An International Journal in Information Science and Engineering*, page 58, 2024.

[75] M Sivakumar, Rabıyathul Basarıya, Abdul Rajak, M Senthil, T Vetriselvi, G Raja, and R Rajavarman. Transforming arabic text analysis: Integrating applied linguistics with m-polar neutrosophic set mood change and depression on social media. *International Journal of Neutrosophic Science (IJNS)*, 25(2), 2025.

[76] Hind Y Saleh, Areen A Salih, Baravan A Asaad, and Ramadhan A Mohammed. Binary bipolar soft points and topology on binary bipolar soft sets with their symmetric properties. *Symmetry*, 16(1):23, 2023.

[77] Asghar Khan, Muhammad Izhar, and Mohammed M. Khalaf. Generalised multi-fuzzy bipolar soft sets and its application in decision making. *J. Intell. Fuzzy Syst.*, 37:2713–2725, 2019.

[78] Maha M Saeed, Sagvan Y Musa, Baravan A Asaad, and Zanyar A Ameen. Pythagorean fuzzy n-bipolar soft sets-based multi-criteria decision-making framework for sustainability evaluation and risk assessment in manufacturing industries. *Scientific Reports*, 15(1):29648, 2025.

[79] Sagvan Y. Musa and Baravan A. Asaad. Topological structures via bipolar hypersoft sets. *Journal of Mathematics*, 2022.

[80] Sagvan Y Musa and Baravan A Asaad. Mappings on bipolar hypersoft classes. *Neutrosophic Sets and Systems*, 53(1):36, 2023.

[81] Sagvan Y Musa and Baravan A Asaad. A progressive approach to multi-criteria group decision-making: N-bipolar hypersoft topology perspective. *Plos one*, 19(5):e0304016, 2024.

[82] T. Fujita and A. Mehmood. Extending classical uncertainty models via hyperpolar structures: Fuzzy, neutrosophic, and soft set perspectives. *Galoitica: J. Math. Struct. Appl.*, 12:24–39, 2025.

[83] Muhammad Saeed. An introduction to dynamic soft sets: A framework for modeling temporal uncertainty. *Available at SSRN 5820784*, 2025.

[84] Muhammad Saeed, Fatima Razaq, and Muhammad Hassan. Dynamic soft set topology: A novel topological framework incorporating evolving parameter structures, 2025.

[85] Muhammad Saeed, Fatima Razaq, Muhammad Hassan, and Dr Atiqe Ur Rahman. Dynamic soft graphs: A unified framework for modeling time-indexed uncertainty in evolving networks, 2025.

[86] Himanshukumar R Patel and Vipul A Shah. General type-2 fuzzy logic systems using shadowed sets: a new paradigm towards fault-tolerant control. In *2021 Australian & New Zealand Control Conference (ANZCC)*, pages 116–121. IEEE, 2021.

[87] Mohammad Hossein Azadi, Khaled Nawaser, Ali Vafaei-Zadeh, Seyed Najmodin Mousavi, Razieh Bagherzadeh Khodashahri, and Haniruzila Hanifah. Investigating antecedents of customer relationship management using interval type-2 fuzzy fmea approach. *International Journal of Business Innovation and Research*, 34(2):139–165, 2024.

[88] Marwan H Hassan, Saad M Darwish, and Saleh M Elkaffas. Type-2 neutrosophic set and their applications in medical databases deadlock resolution. *Computers, Materials & Continua*, 74(2), 2023.

[89] Muslem Al-Saidi, Áron Ballagi, Oday Ali Hassen, and Saad M Saad. Type-2 neutrosophic markov chain model for subject-independent sign language recognition: A new uncertainty–aware soft sensor paradigm. *Sensors (Basel, Switzerland)*, 24(23):7828, 2024.

[90] Soumen Kumar Das, F Yu Vincent, Sankar Kumar Roy, and Gerhard Wilhelm Weber. Location–allocation problem for green efficient two-stage vehicle-based logistics system: A type-2 neutrosophic multi-objective modeling approach. *Expert Systems with Applications*, 238:122174, 2024.

[91] Khizar Hayat, Muhammad Irfan Ali, Bing yuan Cao, and Xiaopeng Yang. A new type-2 soft set: Type-2 soft graphs and their applications. *Adv. Fuzzy Syst.*, 2017:6162753:1–6162753:17, 2017.

[92] Khizar Hayat, Bing-Yuan Cao, Muhammad Irfan Ali, Faruk Karaaslan, and Zejian Qin. Characterizations of certain types of type 2 soft graphs. *Discrete Dynamics in Nature and Society*, 2018(1):8535703, 2018.

[93] Musavarah Sarwar and Muhammad Akram. Certain hybrid rough models with type-2 soft information. *Journal of Multiple-Valued Logic & Soft Computing*, 40, 2023.

[94] Shumaila Manzoor, Saima Mustafa, Kanza Gulzar, Asim Gulzar, Sadia Nishat Kazmi, Syed Muhammad Abrar Akber, Rasool Bukhsh, Sheraz Aslam, and Syed Muhammad Mohsin. Multifuzztops: A fuzzy multi-criteria decision-making model using type-2 soft sets and topsis. *Symmetry*, 16(6):655, 2024.

[95] Guzide Senel. Soft topology generated by l-soft sets. *Journal of New Theory*, 24:88–100, 2018.

[96] Arif Mehmood Khattak, Nazia Hanif, Fawad Nadeem, Muhammad Zamir, Choonkil Park, Giorgio Nordo, and Shamoona Jabeen. *Soft b-separation axioms in neutrosophic soft topological structures*. Infinite Study, 2019.

[97] Saleem Abdullah, Imran Khan, and Muhammad Aslam. A new approach to soft set through applications of cubic set. *arXiv preprint arXiv:1210.6517*, 2012.

[98] Srinivasan Vijayabalaji and Kaliyaperumal Punniyamoorthy. Cubic inverse soft set. In *Soft Computing*, pages 87–94. CRC Press, 2023.

[99] G Muhiuddin and Abdullah M Al-roqi. Cubic soft sets with applications in bck/bci-algebras. *Annals of Fuzzy Mathematics and Informatics*, 8(2):291–304, 2014.

[100] Fatia Fatimah, Dedi Rosadi, RB Fajriya Hakim, and José Carlos R. Alcantud. Probabilistic soft sets and dual probabilistic soft sets in decision-making. *Neural Computing and Applications*, 31:397–407, 2019.

[101] Ping Zhu and Qiaoyan Wen. Probabilistic soft sets. In *2010 IEEE international conference on granular computing*, pages 635–638. IEEE, 2010.

[102] Bindu Nila and Jagannath Roy. Analysis of critical success factors of logistics 4.0 using d-number based pythagorean fuzzy dematel method. *Decision Making Advances*, 2(1):92–104, 2024.

[103] Yuzhen Li and Yabin Shao. Fuzzy cognitive maps based on d-number theory. *IEEE Access*, 10:72702–72716, 2022.

[104] Takaaki Fujita and Florentin Smarandache. *A Survey of Complex-Valued Mathematical Structures*. Neutrosophic Science International Association (NSIA) Publishing House, 2026.

[105] Nuttapong Wattanasiripong, Nuchanat Tiprachot, and Somsak Lekkoksung. On tripolar complex fuzzy sets and their application in ordered semigroups. *International Journal of Analysis and Applications*, 23:139–139, 2025.

[106] Songsong Dai. Linguistic complex fuzzy sets. *Axioms*, 12(4):328, 2023.

[107] Faisal Al-Sharqi, Ashraf Al-Quran, et al. Similarity measures on interval-complex neutrosophic soft sets with applications to decision making and medical diagnosis under uncertainty. *Neutrosophic Sets and Systems*, 51:495–515, 2022.

[108] Said Broumi, Mohamed Talea, Assia Bakali, and Florentin Smarandache. Complex neutrosophic graphs of type. *Collected Papers. Volume VI: On Neutrosophic Theory and Applications*, page 204, 2022.

[109] Naveed Yaqoob and Muhammad Akram. *Complex neutrosophic graphs*. Infinite Study, 2018.

[110] Tahir Mahmood and Ubaid ur Rehman. A novel approach towards bipolar complex fuzzy sets and their applications in generalized similarity measures. *International Journal of Intelligent Systems*, 37:535 – 567, 2021.

[111] Daniel Ramot, Menahem Friedman, Gideon Langholz, Ron Milo, and Abraham Kandel. On complex fuzzy sets. *10th IEEE International Conference on Fuzzy Systems. (Cat. No.01CH37297)*, 3:1160–1163 vol.2, 2001.

[112] Güzide Şenel. A new construction of spheres via soft real numbers and soft points. *Mathematics Letters*, 4(3):39–43, 2018.

[113] Sujoy Das and SK Samanta. Soft linear operators in soft normed linear spaces. *Ann. Fuzzy Math. Inform*, 6(2):295–314, 2013.

[114] Sujoy Das and SK Samanta. Soft real sets, soft real numbers and their properties. *J. fuzzy Math*, 20(3):551–576, 2012.

[115] Seok Zun Song, Hee Sik Kim, and Young Bae Jun. Ideal theory in semigroups based on intersectional soft sets. *The Scientific World Journal*, 2014(1):136424, 2014.

[116] Eun Hwan Roh and Young Bae Jun. Positive implicative ideals of bck-algebras based on intersectional soft sets. *Journal of Applied Mathematics*, 2013(1):853907, 2013.

[117] G Muhiuddin. Intersectional soft sets theory applied to generalized hypervector spaces. *Analele ştiinţifice ale Universităţii" Ovidius" Constanţa. Seria Matematică*, 28(3):171–191, 2020.

[118] Young Bae Jun, Chul Hwan Park, and Noura Omair Alshehri. Hypervector spaces based on intersectional soft sets. In *Abstract and Applied Analysis*. Wiley Online Library, 2014.

[119] Hüseyin Kamac and Subramanian Petchimuthu. Bipolar n-soft set theory with applications. *Soft Computing*, 24:16727 – 16743, 2020.

[120] Muhammad Akram, Arooj Adeel, and José Carlos Rodriguez Alcantud. Group decision-making methods based on hesitant n-soft sets. *Expert Syst. Appl.*, 115:95–105, 2019.

[121] Fatia Fatimah and José Carlos Rodriguez Alcantud. The multi-fuzzy n-soft set and its applications to decision-making. *Neural Computing and Applications*, 33:11437 – 11446, 2021.

[122] Fatia Fatimah, Fatia Fatimah, Dedi Rosadi, R. B. Fajriya Hakim, and José Carlos Rodriguez Alcantud. N-soft sets and their decision making algorithms. *Soft Computing*, 22:3829 – 3842, 2017.

[123] Sagvan Y Musa, Ramadhan A Mohammed, and Baravan A Asaad. N-hypersoft sets: An innovative extension of hypersoft sets and their applications. *Symmetry*, 15(9):1795, 2023.

[124] Sagvan Y Musa. N-bipolar hypersoft sets: Enhancing decision-making algorithms. *Plos one*, 19(1):e0296396, 2024.

[125] Orhan Dalkılıç. Unifying relationships in uncertain environments: examining relations in binary soft sets for expressing inter-object correspondence. *The Journal of Supercomputing*, 81(16):1–29, 2025.

[126] Ahu Açıkgöz and Nihal Tas. Binary soft set theory. *EUROPEAN JOURNAL OF PURE AND APPLIED MATHEMATICS*, 9(4):452–463, 2016.

[127] Muhammad Saqlain, Poom Kumam, and Wiyada Kumam. Multi-criteria decision-making method based on weighted and geometric aggregate operators of linguistic fuzzy-valued hypersoft set with application. *Journal of Fuzzy Extension and Applications*, 6(2):344–370, 2025.

[128] Muhammad Saqlain, Poom Kumam, and Wiyada Kumam. Linguistic hypersoft set with application to multi-criteria decision-making to enhance rural health services. *Neutrosophic Sets and Systems*, 61:28–52, 2023.

[129] Srinivasan Vijayabalaji and Adhimoolam Ramesh. Uncertain multiplicative linguistic soft sets and their application to group decision making. *Journal of Intelligent & Fuzzy Systems*, 35(3):3883–3893, 2018.

[130] Hongjun Guan, Shuang Guan, and Aiwu Zhao. Intuitionistic fuzzy linguistic soft sets and their application in multi-attribute decision-making. *Journal of Intelligent & Fuzzy Systems*, 31(6):2869–2879, 2016.

[131] Zhao Aiwu and Guan Hongjun. Fuzzy-valued linguistic soft set theory and multi-attribute decision-making application. *Chaos, Solitons & Fractals*, 89:2–7, 2016.

[132] Zhifu Tao, Huayou Chen, Ligang Zhou, and Jinpei Liu. 2-tuple linguistic soft set and its application to group decision making. *Soft computing*, 19:1201–1213, 2015.

[133] NLA Mohd Kamal, Lazim Abdullah, and Ilyani Abdullah. Multi-valued neutrosophic linguistic soft set and its application in multi-criteria decision-making. *Journal of Advanced Research in Dynamical and Control Systems*, 11:12, 2019.

[134] Takaaki Fujita. Metafuzzy, metaneutrosophic, metasoft, and metarough set. *Journal of Bangladesh Academy of Sciences*, 50(1):107–145, 2026.

[135] Takaaki Fujita. Metastructure, meta-hyperstructure, and meta-superhyper structure. *Journal of Computers and Applications*, 1(1):1–22, 2025.

[136] Asghar Khan, Muhammad Izhar, and Mohammed M Khalaf. Double-framed soft la-semigroups. *Journal of Intelligent & Fuzzy Systems*, 33(6):3339–3353, 2017.

[137] Yanbin Liu, Peina Liang, and Jingjie Ma. An empirical study on the quality of industry-linked education in vocational colleges: Double-framed treesoft set framework. *Neutrosophic Sets and Systems*, 85(1):32, 2025.

[138] Muhammad Saeed, Hafiz Inam ul Haq, and Mubashir Ali. Extension of double frame soft set to double frame hypersoft set (dfss to dfhss). *HyperSoft Set Methods in Engineering*, 2:18–27, 2024.

[139] Muhammad Izhar, Tariq Mahmood, Asghar Khan, Muhammad Farooq, and Kostaq Hila. Double-framed soft set theory applied to abel-grassmann's hypergroupoids. *New Mathematics and Natural Computation*, 18(03):819–841, 2022.

[140] Muhammad Saeed, Muhammad Rayees Ahmad, Muhammad Saqlain, and Muhammad Riaz. Rudiments of n-framed soft sets. *Punjab University Journal of Mathematics*, 52(5), 2020.

[141] Muhammad Rayees Ahmad, Usman Afzal, Nadir Omer, Ali Delham Algarni, Sara A Ghorashi, and Huda Eltayeb. A computational diagnostic model for infectious diseases via similarity measures on n-framed plithogenic hypersoft sets. *Alexandria Engineering Journal*, 127:1209–1219, 2025.

[142] Usman Afzal, Muhammad Rayees Ahmad, Nazek Alessa, Nauman Raza, Fathea MO Birkea, Salem Alkhalaf, and Nader Omer. Intelligent faculty evaluation and ranking system based on n-framed plithogenic fuzzy hypersoft set and extended nr-topsis. *Alexandria Engineering Journal*, 109:18–28, 2024.

[143] Ajoy Kanti Das, Florentin Smarandache, Rakhal Das, and Suman Das. A comprehensive study on decision-making algorithms in retail and project management using double framed hypersoft sets. *HyperSoft Set Methods in Engineering*, 2:62–71, 2024.

[144] Minyan Chen. Double framed hypersoft set for studies factors that influence and ways to improve vocational college instruction in innovation and entrepreneurship. *Neutrosophic Sets and Systems*, 85(1):5, 2025.

[145] Lingling Chen. Measuring teaching success in college foreign literature programs: An evaluation perspective using double framed superhypersoft set. *Neutrosophic Sets and Systems*, 85(1):7, 2025.

[146] Takaaki Fujita. Double-framed superhypersoft set and double-framed treesoft set. *Advancing Uncertain Combinatorics through Graphization, Hyperization, and Uncertainization: Fuzzy, Neutrosophic, Soft, Rough, and Beyond*, page 71, 2025.

[147] Ke Gong, Panpan Wang, and Zhi Xiao. Bijective soft set decision system based parameters reduction under fuzzy environments. *Applied Mathematical Modelling*, 37(6):4474–4485, 2013.

[148] Ke Gong, Zhi Xiao, and Xia Zhang. The bijective soft set with its operations. *Comput. Math. Appl.*, 60:2270–2278, 2010.

[149] Varun Kumar Tiwari, Prashant Kumar Jain, and Puneet Tandon. An integrated shannon entropy and topsis for product design concept evaluation based on bijective soft set. *Journal of Intelligent Manufacturing*, 30:1645 – 1658, 2017.

[150] Atiqe Ur Rahman, Muhammad Saeed, and Abida Hafeez. Theory of bijective hypersoft set with application in decision making. *Punjab University Journal of Mathematics*, 53(7), 2021.

[151] Takaaki Fujita. N-superhypersoft set and bijective superhypersoft set. *Advancing Uncertain Combinatorics through Graphization, Hyperization, and Uncertainization: Fuzzy, Neutrosophic, Soft, Rough, and Beyond*, page 138, 2025.

[152] Muhammad Ihsan, Muhammad Saeed, Atiqe Ur Rahman, and Florentin Smarandache. Multi-attribute decision support model based on bijective hypersoft expert set. *Punjab University Journal of Mathematics*, 54(1), 2022.

[153] Gustavo Santos-García and José Carlos R Alcantud. Ranked soft sets. *Expert Systems*, 40(6):e13231, 2023.

[154] Irfan Deli. Refined neutrosophic sets and refined neutrosophic soft sets: theory and applications. In *Handbook of research on generalized and hybrid set structures and applications for soft computing*, pages 321–343. IGI Global, 2016.

[155] Takaaki Fujita and Florentin Smarandache. *Some types of hyperneutrosophic set (6): Multineutrosophic set and refined neutrosophic set.* Infinite Study, 2025.

[156] Florentin Smarandache. n-valued refined neutrosophic logic and its applications to physics. *Infinite study*, 4:143–146, 2013.

[157] Anjan Mukherjee, Mithun Datta, and Abhijit Saha. Refined soft sets and its applications. *Journal of New Theory*, 14:10–25, 2016.

[158] Faruk Karaaslan. Correlation coefficients of single-valued neutrosophic refined soft sets and their applications in clustering analysis. *Neural Computing and Applications*, 28(9):2781–2793, 2017.

[159] R Anitha Cruz. Neutrosophic soft cubic refined sets. *Neutrosophic Sets & Systems*, 73, 2024.

[160] Fujita Takaaki and Arif Mehmood. Iterative multifuzzy set, iterative multineutrosophic set, iterative multisoft set, and multiplithogenic sets. *Neutrosophic Computing and Machine Learning*, 41:1–30, 2025.

[161] Shawkat Alkhazaleh, Abdul Razak Salleh, Nasruddin Hassan, and Abd Ghafur Ahmad. Multisoft sets. In *Proc. 2nd International Conference on Mathematical Sciences*, pages 910–917, 2010.

[162] Florentin Smarandache. *Practical applications of IndetermSoft Set and IndetermHyperSoft Set and introduction to TreeSoft Set as an extension of the MultiSoft Set.* Infinite Study, 2022.

[163] Sabu Sebastian and TV Ramakrishnan. Multi-fuzzy sets: An extension of fuzzy sets. *Fuzzy Information and Engineering*, 3:35–43, 2011.

[164] Mahalakshmi Pethaperumal, Vimala Jeyakumar, Jeevitha Kannan, and Ashma Banu. An algebraic analysis on exploring q-rung orthopair multi-fuzzy sets. *Journal of fuzzy extension and applications*, 4(3):235–245, 2023.

[165] Johanna Estefanía Street Imbaquingo Street, Karen Milagros Díaz Street Salambay, Jordy Alexis Vargas Yumbo, and Kevin Christopher Carrasco Azogue. Intercultural education from ancestral logic: Application of the ayni method multi-neutrosophic to strengthen mixed methodology in the waorani community. *Neutrosophic Sets and Systems*, 92:263–283, 2025.

[166] Ennio Jesús Mérida Córdova, Elizabeth Esther Vergel Parejo, and Raúl López Fernández. Scholarai scholarly article search strategies with the ayni method multi-neutrosophic for ethical information management in ai. *Neutrosophic Sets and Systems*, 92:380–397, 2025.

[167] Takaaki Fujita and Florentin Smarandache. An introduction to advanced soft set variants: Superhypersoft sets, indetermsuperhypersoft sets, indetermtreesoft sets, bihypersoft sets, graphicsoft sets, and beyond. *Neutrosophic Sets and Systems*, 82:817–843, 2025.

[168] Ari M Lipsky and Sander Greenland. Causal directed acyclic graphs. *JAMA*, 327(11):1083–1084, 2022.

[169] Takaaki Fujita. Directed acyclic superhypergraphs (dash): A general framework for hierarchical dependency modeling. *Neutrosophic Knowledge*, 6:72–86, 2025.

[170] Peter WG Tennant, Eleanor J Murray, Kellyn F Arnold, Laurie Berrie, Matthew P Fox, Sarah C Gadd, Wendy J Harrison, Claire Keeble, Lynsie R Ranker, Johannes Textor, et al. Use of directed acyclic graphs (dags) to identify confounders in applied health research: review and recommendations. *International journal of epidemiology*, 50(2):620–632, 2021.

[171] Weizhou Shen, Siyue Wu, Yunyi Yang, and Xiaojun Quan. Directed acyclic graph network for conversational emotion recognition. *arXiv preprint arXiv:2105.12907*, 2021.

[172] Takaaki Fujita and Florentin Smarandache. An introduction to advanced soft set variants: Superhypersoft sets, indetermsuperhypersoft sets, indetermtreesoft sets, bihypersoft sets, graphicsoft sets, and beyond. *Neutrosophic Sets and Systems*, 82:817–843, 2025.

[173] Mehmet Şahin, İrfan Deli, and Vakkas Uluçay. *Bipolar Neutrosophic Soft Expert Sets.* Infinite Study, 2016.

[174] Faisal Al-Sharqi, Abd Ghafur Ahmad, and Ashraf Al-Quran. Interval-valued neutrosophic soft expert set from real space to complex space. *CMES-Computer Modeling in Engineering & Sciences*, 132(1), 2022.

[175] Ashraf Al-Quran and Nasruddin Hassan. The complex neutrosophic soft expert set and its application in decision making. *Journal of Intelligent & Fuzzy Systems*, 34(1):569–582, 2018.

[176] Ashraf Al-Quran, Nasruddin Hassan, and Shawkat Alkhazaleh. Fuzzy parameterized complex neutrosophic soft expert set for decision under uncertainty. *Symmetry*, 11(3):382, 2019.

[177] Fathima Perveen PA, Sunil Jacob John, et al. On spherical fuzzy soft expert sets. In *AIP conference proceedings*. AIP Publishing, 2020.

[178] Anwar Bataihah. A decision making framework using parameterized interval valued fuzzy soft expert sets. *Statistics, Optimization & Information Computing*, 15(4):2829–2854, 2026.

[179] Yousef Al-Qudah and Nasruddin Hassan. Fuzzy parameterized complex multi-fuzzy soft expert sets. *THE 2018 UKM FST POSTGRADUATE COLLOQUIUM: Proceedings of the Universiti Kebangsaan Malaysia, Faculty of Science and Technology 2018 Postgraduate Colloquium*, 2019.

[180] Srinivasan Vijayabalaji, Shanmugam Kalaiselvan, Akin Osman Atagün, and Hüseyin Kamaci. A study on intuitionistic fuzzy soft expert groups. *Int. J. Open Problems Compt. Math*, 17(2), 2024.

[181] Said Broumi and Florentin Smarandache. Intuitionistic fuzzy soft expert sets and its application in decision making. *Journal of new theory*, (1):89–105, 2015.

[182] Ganeshsree Selvachandran and Abdul Razak Salleh. Possibility intuitionistic fuzzy soft expert set theory and its application in decision making. *International Journal of Mathematics and Mathematical Sciences*, 2015(1):314285, 2015.

[183] Mehmet Sahin, Shawkat Alkhazaleh, and Vakkas Ulucay. Neutrosophic soft expert sets. *Applied Mathematics-a Journal of Chinese Universities Series B*, 06:116–127, 2015.

[184] Faisal Al-Sharqi, Yousef Al-Qudah, and Naif Alotaibi. Decision-making techniques based on similarity measures of possibility neutrosophic soft expert sets. *Neutrosophic Sets and Systems, vol. 55/2023: An International Journal in Information Science and Engineering*, page 358, 2024.

[185] Sumyyah Al-Hijjawi, Abd Ghafur Ahmad, and Shawkat Alkhazaleh. Effective neutrosophic soft expert set and its application. *International Journal of Neutrosophic Science (IJNS)*, 23(1), 2024.

[186] Takaaki Fujita. Note for soft multiexpert graph and multisoft multiexpert graph. *Integra: Journal of Integrated Mathematics and Computer Science*, 3(1):6–15, 2026.

[187] Takaaki Fujita, Iqbal M. Batiha, Nidal Anakira, Shadi Al-Ahmad, Mohammad S. Hijazi, and Abed Al-Rahman M. Malkawi. Note for soft multiexpert set and multisoft multiexpert set. *Boletim da Sociedade Paranaense de Matemática*, 44:1–20, June 2026.

[188] Muhammad IHSAN, Muhammad SAEED, and Muhammad Yasir HAYAT MALIK. An innovative group decision-making approach utilizing picture fuzzy hypersoft expert set. *Sigma*, 44(1):210–221, 2026.

[189] Takaaki Fujita. Superhypersoft rough set, superhypersoft expert set, and bipolar superhypersoft set. *Advancing Uncertain Combinatorics through Graphization, Hyperization, and Uncertainization: Fuzzy, Neutrosophic, Soft, Rough, and Beyond*, page 270, 2025.

[190] Ashraf Al-Quran, Nasruddin Hassan, and Emad A. Marei. A novel approach to neutrosophic soft rough set under uncertainty. *Symmetry*, 11:384, 2019.

[191] Xinyi Wang and Qinghai Wang. Uncertainty measurement of variable precision fuzzy soft rough set model. In *CECNet*, 2022.

[192] Tasawar Abbas, Rehan Zafar, Sana Anjum, Ambreen Ayub, and Zamir Hussain. An innovative soft rough dual hesitant fuzzy sets and dual hesitant fuzzy soft rough sets. *VFAST Transactions on Mathematics*, 2023.

[193] Fu Zhang, Weimin Ma, and Hongwei Ma. Dynamic chaotic multi-attribute group decision making under weighted t-spherical fuzzy soft rough sets. *Symmetry*, 15:307, 2023.

[194] Aysun Benek and Taha Yasin Ozturk. A comparative analysis of two different decision-making methods in neutrosophic soft rough set environments. *OPSEARCH*, pages 1–22, 2025.

[195] Jingjing Zhang. Neutrosophic soft rough sets for quality evaluation of interactive music teaching in higher education: A novel approach. *Neutrosophic Sets and Systems*, 90(1):66, 2025.

[196] Siyang Yang. Extending superhypersoft framework: Weighted soft sets for priority-based decision-making in engineering ethics risk analysis based on big data technology. *Neutrosophic Sets and Systems*, 86:119–125, 2025.

[197] K Selvakumari. Solving game problem using weighted soft sets. *Journal of Computer and Mathematical Sciences*, 9(10):1307–1311, 2018.

[198] Holy-Heavy M Balami, Aliyu G Dzarma, and Mohammed A Mohammed. Weighted soft set and its application in parameterized decision making processes. *International Journal of Development Mathematics (IJDM)*, 2(1):131–144, 2025.

[199] Jean Dezert and Florentin Smarandache. An introduction to dsmt. *arXiv preprint arXiv:0903.0279*, 2009.

[200] Florentin Smarandache and Jean Dezert. An introduction to dsmt. *F. Smarandache, J. Dezert (Editors), Applications and Advances of DSmT for Information Fusion, Vol. 3, American Research Press, Rehoboth, 2009*, 2009.

[201] A Tchamova. Multi-target tracking applications of dezert-smarandache theory. *NATO Advanced Study Institute, Albena, Bulgaria*, pages 66–69, 2005.

[202] Hu Lifang, Guan Xin, and He You. Efficient combination rule of dezert-smarandache theory. *Journal of Systems Engineering and Electronics*, 19(6):1139–1144, 2008.

[203] Yilin Dong, Xinde Li, Jean Dezert, Mohammad Omar Khyam, Md Noor-A-Rahim, and Shuzhi Sam Ge. Dezert-smarandache theory-based fusion for human activity recognition in body sensor networks. *IEEE Transactions on Industrial Informatics*, 16(11):7138–7149, 2020.

[204] Takaaki Fujita and Smarandache Florentin. *Survey of Fuzzy, Neutrosophic, and Uncertain Numbers.* Neutrosophic Science International Association (NSIA) Publishing House, 2026.

[205] Takaaki Fujita, Ajoy Kanti Das, Arif Mehmood, and Suman Das. Belief hyperfunctions, belief superhyperfunctions, and dsm-numbers. *Journal of Evidential and Paraconsistent Reasoning*, 2(1), 2027.

[206] Jean Dezert and Florentin Smarandache. *On the generation of hyper-powersets for the DSmT.* Infinite Study, 2003.

[207] Milan Daniel. A generalization of the minc combination to dsm hyper-power sets. In *Proceedings of IPMU*, volume 6, pages 420–427, 2006.

[208] Omer Akguller. Geometric soft sets. *Hittite Journal of Science and Engineering*, 4(2):159–164, 2017.

[209] Abdul Razak Salleh, Shawkat Alkhazaleh, Nasruddin Hassan, and Abd Ghafur Ahmad. Multiparameterized soft set. *Journal of Mathematics and Statistics*, 8(1):92–97, 2012.

[210] Takaaki Fujita and Iqbal M Batiha. Multiparameterized hypersoft set and type-2 hypersoft set. *Neutrosophic Sets and Systems*, 95(1):12, 2026.

[211] Fatma Adam and Nasruddin Hassan. *Interval valued Q-fuzzy multiparameterized soft set and its application.* Infinite Study, 2016.

[212] Young Bae Jun, Seok Zun Song, and G Muhiuddin. Concave soft sets, critical soft points, and union-soft ideals of ordered semigroups. *The Scientific World Journal*, 2014(1):467968, 2014.

[213] Hero M Salih and Pishtiwan O Sabir. Some new properties of convex and concave soft sets. *Diyala Journal for Pure Sciences*, 15(9):84–94, 2018.

[214] Feng Feng, Naim Cagman, Violeta Leoreanu-Fotea, and Muhammad Akram. Emerging trends in soft set theory and related topics. *The Scientific World Journal*, 2015:689457, 2015.

[215] José Carlos R Alcantud and Gustavo Santos-García. Incomplete soft sets: New solutions for decision making problems. In *Decision Economics, In Commemoration of the Birth Centennial of Herbert A. Simon 1916-2016 (Nobel Prize in Economics 1978) Distributed Computing and Artificial Intelligence, 13th International Conference*, pages 9–17. Springer, 2016.

[216] Zhi Kong, Jie Zhao, Lifu Wang, and Junjie Zhang. A new data filling approach based on probability analysis in incomplete soft sets. *Expert Systems with Applications*, 184:115358, 2021.

[217] José Carlos R Alcantud and Gustavo Santos-García. Decision making under incompleteness based on soft set theory. In *International Conference on Information Processing and Management of Uncertainty in Knowledge-Based Systems*, pages 583–595. Springer, 2018.

[218] F Fatimah, D Rosadi, and RBF Hakim. Probabilistic soft sets and dual probabilistic soft sets in decision making with positive and negative parameters. In *journal of physics: conference series*, volume 983, page 012112. IOP Publishing, 2018.

[219] Nemanja Vučićević, Nenad Stojanović, and Aslıhan Sezgin. Algorithms for decision-making based on energies of probabilistic and dual probabilistic soft set. *Scientific Publications of the State University of Novi Pazar Series A Applied Mathematics Informatics and mechanics*, 2026.

[220] Quang-Thinh Bui, Thanh Nha Nguyen, Hung Son Nguyen, and Bay Vo. A novel framework for handling uncertainty: Intuitionistic fuzzy rough soft sets. *Information Sciences*, page 122592, 2025.

[221] Akın Osman Atagün. Novel characterizations of rough soft sets: Equivalent soft sets in pawlak approximation space with applications. *Expert Systems with Applications*, 292:128462, 2025.

[222] Jianming Zhan and Kuanyun Zhu. Reviews on decision making methods based on (fuzzy) soft sets and rough soft sets. *Journal of Intelligent & Fuzzy Systems*, 29(3):1169–1176, 2015.

[223] Jianming Zhan, Qi Liu, and Bijan Davvaz. A new rough set theory: rough soft hemirings. *Journal of Intelligent & Fuzzy Systems*, 28(4):1687–1697, 2015.

[224] Feng Feng, Muhammad Akram, Bijan Davvaz, and Violeta Leoreanu-Fotea. Attribute analysis of information systems based on elementary soft implications. *Knowledge-Based Systems*, 70:281–292, 2014.

[225] Gangqiang Zhang, Zhaowen Li, and Bin Qin. A method for multi-attribute decision making applying soft rough sets. *Journal of Intelligent & Fuzzy Systems*, 30(3):1803–1815, 2016.

[226] Violeta Fotea. Soft sets and soft rough sets. *Information Sciences*, 2011.

[227] Saeid Jafarzadeh Ghoushchi and Bahman Nasiri. Sustainable landfill site selection for hazardous waste using a gis-based mcdm approach with g-number information. *Environment, Development and Sustainability*, 27(9):22049–22080, 2025.

[228] Saeid Jafarzadeh Ghoushchi and Mohammad Khazaeili. G-numbers: importance-necessity concept in uncertain environment. *International Journal of Management and Fuzzy Systems*, 5(1):27–32, 2019.

[229] Rui Cheng, Jianping Fan, and Meiqin Wu. A dynamic multi-attribute group decision-making method with r-numbers based on merec and cocoso method. *Complex & Intelligent Systems*, 9(6):6393–6426, 2023.

[230] Peide Liu, Baoying Zhu, Hamidreza Seiti, and Li Yang. Risk-based decision framework based on r-numbers and best-worst method and its application to research and development project selection. *Information Sciences*, 571:303–322, 2021.

[231] Hamidreza Seiti, Mahdi Fathi, Ashkan Hafezalkotob, Enrique Herrera-Viedma, and Ibrahim A Hameed. Developing the modified r-numbers for risk-based fuzzy information fusion and its application to failure modes, effects, and system resilience analysis (fmesra). *ISA transactions*, 113:9–27, 2021.

[232] Rui Cheng, Jianping Fan, Meiqing Wu, and Hamidreza Seiti. A dual-level multi-attribute group decision-making model considering interaction factors based on ccsd and marcos methods with r-numbers. *Expert Systems with Applications*, 243:122839, 2024.

[233] Qian Zhao, Yanbing Ju, Peiwu Dong, and Ernesto DR Santibanez Gonzalez. A hybrid decision making aided framework for multi-criteria decision making with r-numbers and preference models. *Engineering Applications of Artificial Intelligence*, 111:104777, 2022.

[234] Hamidreza Seiti, Ashkan Hafezalkotob, and Luis Martínez. R-numbers, a new risk modeling associated with fuzzy numbers and its application to decision making. *Information Sciences*, 483:206–231, 2019.

[235] Santosh Kumar Yadav and Rakesh Kumar. A scalable and utility driven profit maximized auction of resources model for cloudlet based mobile edge computing. *Wireless Personal Communications*, 119(1):527–565, 2021.

[236] Peide Liu, Sepehr Hendiani, Morteza Bagherpour, Seyed Farid Ghannadpour, and Amin Mahmoudi. Utility-numbers theory. *IEEE Access*, 7:56994–57008, 2019.

[237] RA Aliev. Approximate arithmetic operations of u-numbers. *Procedia Computer Science*, 102:59–66, 2016.

[238] RA Aliev. Introduction to u-number calculus. *Intelligent Automation & Soft Computing*, pages 1–6, 2017.

[239] Saeid Jafarzadeh Ghoushchi, Abbas Mardani, and Luis Martínez. Trust number: Trust-based modeling for handling decision-making problems. *Knowledge-Based Systems*, 305:112631, 2024.

[240] P. A. Fathima Perveen and Sunil Jacob John. Relations on spherical fuzzy soft sets. *2nd INTERNATIONAL CONFERENCE ON COMPUTATIONAL SCIENCES-MODELLING, COMPUTING AND SOFT COMPUTING (CSMCS 2022)*, 2023.

[241] Sujit Das and Samarjit Kar. Intuitionistic multi fuzzy soft set and its application in decision making. In *Pattern Recognition and Machine Intelligence: 5th International Conference, PReMI 2013, Kolkata, India, December 10-14, 2013. Proceedings 5*, pages 587–592. Springer, 2013.

[242] Muhammad Saeed, Irfan Saif Ud Din, Imtiaz Tariq, and Harish Garg. Refined fuzzy soft sets: Properties, set-theoretic operations and axiomatic results. *Journal of Computational and Cognitive Engineering*, 3(1):24–33, 2024.

[243] Sheikh Zain Majid, Muhammad Saeed, Umar Ishtiaq, and Ioannis K Argyros. The development of a hybrid model for dam site selection using a fuzzy hypersoft set and a plithogenic multipolar fuzzy hypersoft set. *Foundations*, 4(1):32–46, 2024.

[244] Xingsi Xue, Himanshu Dhumras, Garima Thakur, Rakesh Kumar Bajaj, and Varun Shukla. Schweizer-sklar t-norm operators for picture fuzzy hypersoft sets: Advancing suistainable technology in social healthy environments. *Computers, Materials & Continua*, 84(1), 2025.

[245] R Hema, R Sudharani, and M Kavitha. A novel approach on plithogenic interval valued neutrosophic hypersoft sets and its application in decision making. *Indian Journal Of Science And Technology*, 2023.

[246] Takaaki Fujita. Hyperfuzzy hypersoft set and hyperneutrosophic hypersoft set. *Advancing Uncertain Combinatorics through Graphization, Hyperization, and Uncertainization: Fuzzy, Neutrosophic, Soft, Rough, and Beyond*, page 247, 2025.

[247] Francina Shalini. Trigonometric similarity measures of pythagorean neutrosophic hypersoft sets. *Neutrosophic Systems with Applications*, 2023.

[248] Muhammad Saqlain and Xiao Long Xin. *Interval valued, m-polar and m-polar interval valued neutrosophic hypersoft sets.* Infinite Study, 2020.

[249] Yuncheng Jiang, Yong Tang, Qimai Chen, Hai Liu, and Jianchao Tang. Interval-valued intuitionistic fuzzy soft sets and their properties. *Computers & Mathematics with Applications*, 60(3):906–918, 2010.

[250] Harish Garg and Rishu Arora. Bonferroni mean aggregation operators under intuitionistic fuzzy soft set environment and their applications to decision-making. *Journal of the Operational Research Society*, 69:1711 – 1724, 2018.

[251] Harish Garg and Rishu Arora. Topsis method based on correlation coefficient for solving decision-making problems with intuitionistic fuzzy soft set information. In *AIMS mathematics*, 2020.

[252] Harish Garg and Rishu Arora. Generalized maclaurin symmetric mean aggregation operators based on archimedean t-norm of the intuitionistic fuzzy soft set information. *Artificial Intelligence Review*, 54:3173 – 3213, 2020.

[253] Zheng Zhao, Doo Heon Song, and Kwang Baek Kim. Fuzzy attention convolutional neural networks: A novel approach combining intuitionistic fuzzy sets and deep learning. *Computers, Materials & Continua*, 87(2):32, 2026.

[254] Dajiang Chen, Jialiang Xie, and Honghui Wang. Intrusion detection in industrial internet of things based on granular-ball intuitionistic fuzzy sets. *IEEE Transactions on Fuzzy Systems*, 2026.

[255] Akin Osman Atagün, Savcı Rahman Argün, Aslıhan Sezgin, and Hüseyin Bahadır. Comparison models on soft, fuzzy, and intuitionistic fuzzy sets. *International Journal of Systems Science*, 57(7):1769–1789, 2026.

[256] Nguyen Xuan Thao, Tran Duc Quynh, Truong Thi Thuy Duong, and Hai Van Pham. Distance of intuitionistic fuzzy sets based on score function of a decision support system and its application. *Journal of Intelligent & Fuzzy Systems*, 50(4):1111–1130, 2026.

[257] Chunfeng Suo, Le Fu, and Shu Sun. Selection of giant panda habitat based on the overlap degree of interval-valued intuitionistic fuzzy sets. *Expert Systems with Applications*, page 131043, 2026.

[258] P Gajalaxmi, D Jayanthi, and M Priyadharshini. Operators on circular interval-valued intuitionistic fuzzy sets. *Notes on Intuitionistic Fuzzy Sets*, 32(1):51–67, 2026.

[259] Sanghyuk Lee, Eunmi Lee, and Witold Pedrycz. Inclusion and preference analysis in intuitionistic fuzzy sets: Application in decision-making. *International Journal of Fuzzy Systems*, pages 1–12, 2026.

[260] Saalam Ali, Hamza Naveed, Imran Siddique, and Rana Muhammad Zulqarnain. Extension of interaction geometric aggregation operator for material selection using interval-valued intuitionistic fuzzy hypersoft set. *Journal of operations intelligence*, 2(1):14–35, 2024.

[261] Hamza Naveed and Saalam Ali. Addressing decision-making challenges: similarity measures for interval-valued intuitionistic fuzzy hypersoft sets. *Decision Making Advances*, 3(1):175–184, 2025.

[262] Muhammad Naveed Jafar, Muhammad Saeed, Muhammad Haseeb, and Ahtasham Habib. Matrix theory for intuitionistic fuzzy hypersoft sets and its application in multi-attributive decision-making problems. *Theory and Application of Hypersoft Set*, 65, 2021.

[263] Atiqe Ur Rahman, Muhammad Saeed, Hamiden Abd El-Wahed Khalifa, and Walaa Abdullah Afifi. Decision making algorithmic techniques based on aggregation operations and similarity measures of possibility intuitionistic fuzzy hypersoft sets. *Aims Mathematics*, 7(3):3866–3895, 2021.

[264] Krassimir T Atanassov and G Gargov. *Intuitionistic fuzzy logics*. Springer, 2017.

[265] Jose James and Sunil C Mathew. On the categories of neutrosophic sets and neutrosophic topological spaces. *South East Asian Journal of Mathematics and Mathematical Sciences*, 22(2):235–254, 2026.

[266] A Iampan, N Subha, P Maragatha Meenakshi, and N Rajesh. Neutrosophic substructures of lattices: Sublattices, ideals, and filters. *IAENG International Journal of Computer Science*, 53(10):4076, 2026.

[267] Shake Baji et al. Translations, multiplications, and magnification of generalized neutrosophic set on d-algebra: Visualization using matlab algorithm. *IAENG International Journal of Applied Mathematics*, 56(10):4522, 2026.

[268] Nital P Nirmal, Hem Bhatt, Jignesh G Parmar, Bhavik D Upadhyay, and Vishal J Pandya. Ranking and selection of robot using advance correlation coefficient single-valued neutrosophic set multi-attribute decision-making technique. *ITEGAM-JETIA*, 12(61), 2026.

[269] Yakup As, Karahan Kara, and Sait Gül. Assessing the sustainability performance of fishing harbors under uncertainty: A gaussian-neutrosophic framework. *Sustainable Development*, page e71713, 2026.

[270] Samsiah Abdul Razak, Zahari Md Rodzi, Faisal Al-Sharqi, Nazirah Ramli, and Mohammad Shafiq Mohammad Kamari. Ivpnsdwga: An enhanced dombi-based interval-valued pythagorean neutrosophic aggregation model for sdg-driven sustainable investment decision. *Journal of Quality Measurement and Analysis*, 22(3):209–237, 2026.

[271] Yuxu Han, Sukumar Letchmunan, Wulfran Fendzi Mbasso, Narinderjit Singh Sawaran Singh, Zokir Mamadiyarov, Aseel Smerat, and Zhe Liu. Medical waste disposal site selection under uncertainty: An interval-valued neutrosophic critic-marcos decision support framework. *Journal of the Nigerian Society of Physical Sciences*, pages 3644–3644, 2026.

[272] Janak Raj Amit Kumar and I Paulraj Jayasimman. Structured neutrosophic sets with entropy-derived indeterminacy: Injectivity, stability, and aggregation guarantees for variability-sensitive mcdm. *Neutrosophic Sets and Systems*, 100(1):12, 2026.

[273] Xueping Lu, Tong Zhang, and Lilian Shi. Magdm model based linguistic neutrosophic credibility matrix energy for assessment of learning objectives attainment in production internship. *Neutrosophic Sets & Systems*, 102:102, 2027.

[274] Raja Muhammad Hashim, Muhammad Gulistan, and Muhammad Shahzad. Foundations of neutrosophic n-semirings theory and structural properties. *Neutrosophic Sets and Systems*, 100(1):10, 2026.

[275] Shawkat Alkhazaleh. n-valued refined neutrosophic soft set theory. *Journal of Intelligent & Fuzzy Systems*, 32(6):4311–4318, 2017.

[276] Shawkat Alkhazaleh and Ayman A Hazaymeh. N-valued refined neutrosophic soft sets and their applications in decision making problems and medical diagnosis. *Journal of Artificial Intelligence and Soft Computing Research*, 8(1):79–86, 2018.

[277] Muhammad Akram and Sundas Shahzadi. *Representation of graphs using intuitionistic neutrosophic soft sets.* Infinite Study, 2016.

[278] S Broumi and Tomasz Witczak. Heptapartitioned neutrosophic soft set. *International Journal of Neutrosophic Science*, 18(4):270–290, 2022.

[279] Quang-Thinh Bui, My-Phuong Ngo, Vaclav Snasel, Witold Pedrycz, and Bay Vo. The sequence of neutrosophic soft sets and a decision-making problem in medical diagnosis. *International Journal of Fuzzy Systems*, 24:2036 – 2053, 2022.

[280] Hüseyin Kamacı. Linguistic single-valued neutrosophic soft sets with applications in game theory. *International Journal of Intelligent Systems*, 36(8):3917–3960, 2021.

[281] S. Onar. A note on neutrosophic soft set over hyperalgebras. *Symmetry*, 16(10):1288, 2024.

[282] Ayisha Saeed, Kainat Muniba, Sahar Ahmed Idris, Wael Fawzy Mohamed Mobarak, and Zaffar Ahmed Shaikh. Novel aggregation operations of q-neutrosophic hypersoft sets and their applications in sustainable tourism development and cultural preservation. *Journal of Mathematics*, 2026(1):4773712, 2026.

[283] Eman Hussein, Yousef Al-Qudah, HM Jaradat, Faisal Al-Sharqi, Saja Elnajar, and Ali Jaradat. Similarity measure and sine exponential measure of possibility interval-valued neutrosophic hypersoft sets and their applications. *Neutrosophic Sets and Systems*, 81(1):4, 2025.

[284] Takaaki Fujita. Hyperfuzzy hypersoft set and hyperneutrosophic hypersoft set. *Advancing Uncertain Combinatorics through Graphization, Hyperization, and Uncertainization: Fuzzy, Neutrosophic, Soft, Rough, and Beyond*, page 247, 2025.

[285] Faruk Karaaslan. *Neutrosophic soft sets with applications in decision making.* Infinite Study, 2014.

[286] Pabitra Kumar Maji. *Neutrosophic soft set.* Infinite Study, 2013.

[287] Fazeelat Sultana, Muhammad Gulistan, Mumtaz Ali, Naveed Yaqoob, Muhammad Khan, Tabasam Rashid, and Tauseef Ahmed. A study of plithogenic graphs: applications in spreading coronavirus disease (covid-19) globally. *Journal of ambient intelligence and humanized computing*, 14(10):13139–13159, 2023.

[288] Takaaki Fujita and Florentin Smarandache. *Foundations of Upside-Down Logic and Generalized Paradox Theory.* Neutrosophic Science International Association (NSIA) Publishing House, 2026.

[289] A Oyem and SE Ogunfeyitimi. Plithogenic quasigroups with application to decision making. *Neutrosophic Sets & Systems*, 102:31, 2027.

[290] Giorgio Nordo, Florentin Smarandache, et al. Plithogenic topological spaces. *NEUTROSOPHIC SETS AND SYSTEMS*, 103:17–47, 2027.

[291] Maikel Y Leyva-Vázquez and Florentin Smarandache. Declared-loss evaluation of language models as plithogenic neutrosophic structure: Formal embedding, auditor effects, and the limits of verbalised triplets, 2026.

[292] Divya Arora, Devendra K Tayal, and Sumit K Yadav. A context-based approach for word-level sentiment analysis using plithogenic sets. *Intelligent Computing and Communication Techniques*, pages 411–417, 2025.

[293] Nivetha Martin. Introduction to possibility plithogenic soft sets. *Plithogenic Logic and Computation*, 2024.

[294] Shawkat Alkhazaleh. *Plithogenic soft set.* Infinite Study, 2020.

[295] Takaaki Fujita and Florentin Smarandache. A unified framework for $u$-structures and functorial structure: Managing super, hyper, superhyper, tree, and forest uncertain over/under/off models. *Neutrosophic Sets and Systems*, 91:337–380, 2025.

[296] Takaaki Fujita and Florentin Smarandache. *HyperGraph and SuperHyperGraph Theory with Applications (IV): Uncertain Graph Theory*, volume IV of *HyperGraph and SuperHyperGraph Theory with Applications.* Neutrosophic Science International Association (NSIA) Publishing House, 1.0 edition, 2026.

[297] Takaaki Fujita and Florentin Smarandache. *HyperGraph and SuperHyperGraph Theory with Applications.* Neutrosophic Science International Association (NSIA) Publishing House, 2026.

[298] Takaaki Fujita and Florentin Smarandache. *A Dynamic Survey of Fuzzy, Intuitionistic Fuzzy, Neutrosophic, Plithogenic, and Extensional Sets.* Neutrosophic Science International Association (NSIA), 2025.

[299] Takaaki Fujita and Florentin Smarandache. *A Revised, Comprehensive, and Re-Updated Survey of Unified Frameworks for Fuzzy, Neutrosophic, and Uncertain Sets.* Neutrosophic Science International Association (NSIA) Publishing House, 2026.

[300] Takaaki Fujita and Florentin Smarandache. *A Comprehensive and Updated Survey of Fuzzy, Neutrosophic, and Uncertain Sets.* Neutrosophic Science International Association (NSIA) Publishing House, 2026.

[301] John N Mordeson and Premchand S Nair. *Fuzzy graphs and fuzzy hypergraphs*, volume 46. Physica, 2012.

[302] P. R. Kavyasree and B. Surender Reddy. N-cubic sets applied to linear spaces. *Kragujevac Journal of Mathematics*, 2022.

[303] Gundeti Soujanya and B. Surender Reddy. N-cubic picture fuzzy linear spaces. *Indian Journal Of Science And Technology*, 2024.

[304] Young Bae Jun, Seok Zun Song, and Seon Jeong Kim. N-hyper sets. *Mathematics*, 2018.

[305] Hsiang-Chuan Liu and Tung-Sheng Liu. A novel fuzzy measure and its extensional signed fuzzy measure. In *Proceedings of the 10th WSEAS International Conference on Systems Theory and Scientific Computation, ISTASC*, volume 10, pages 107–111, 2010.

[306] Hsiang-Chuan Liu. An extensional signed fuzzy measure of signed rho-fuzzy measure. In *International Conference on Computational Collective Intelligence*, pages 509–519. Springer, 2010.

[307] Witold Pedrycz. Shadowed sets: representing and processing fuzzy sets. *IEEE Transactions on Systems, Man, and Cybernetics, Part B (Cybernetics)*, 28(1):103–109, 1998.

[308] Mingjie Cai, Qingguo Li, and Guangming Lang. Shadowed sets of dynamic fuzzy sets. *Granular Computing*, 2:85 – 94, 2016.

[309] Gianpiero Cattaneo and Davide Ciucci. Shadowed sets and related algebraic structures. *Fundamenta Informaticae*, 55(3-4):255–284, 2003.

[310] Krassimir T Atanassov. *On intuitionistic fuzzy sets theory*, volume 283. Springer, 2012.

[311] RA Borzooei and Hossein Rashmanlou. Cayley interval-valued fuzzy graphs. *UPB Scientific Bulletin, Series A: Applied Mathematics and Physics*, 78(3):83–94, 2016.

[312] Madhumangal Pal, Sovan Samanta, and Hossein Rashmanlou. Some results on interval-valued fuzzy graphs. *International Journal of Computer Science and Electronics Engineering*, 3(3):205–211, 2015.

[313] Xindong Peng. New operations for interval-valued pythagorean fuzzy set. *Scientia Iranica*, 26:1049–1076, 2018.

[314] Weijian Ren and Yuhong Du. Development and application of an interval-valued pythagorean fuzzy set-based grade evaluation method to foreign fiber content. *Journal of Natural Fibers*, 20, 2023.

[315] Manish Kumar Gunjan, Amal Kumar Adak, and Davood Darvishi Salookolaei. Fermatean fuzzy artinian and noetherian rings. *Asian Research Journal of Mathematics*, 21(5):179–190, 2025.

[316] Fei Gao, Meihong Han, Siyang Wang, and Jie Gao. A novel fermatean fuzzy bwm-vikor based multi-criteria decision-making approach for selecting health care waste treatment technology. *Eng. Appl. Artif. Intell.*, 127:107451, 2024.

[317] Palash Dutta, Nidhi Agarwal, and Sadiq Hussain. An integrated multi criteria decision making framework for hyperbolic fuzzy sets based on a robust logarithmic score function for enhanced healthcare decision support. *International Journal of Machine Learning and Cybernetics*, 17(8):385, 2026.

[318] Palash Dutta. Complex hyperbolic fuzzy sets: Theory and construction. *Journal of Computational Intelligence and Decision Analytics*, 1(1):60–83, 2026.

[319] Palash Dutta and Gourangajit Borah. Construction of hyperbolic fuzzy set and its applications in diverse covid-19 associated problems. *New Mathematics and Natural Computation*, 19(01):217–288, 2023.

[320] R Seema, V Surya, G Sudha, P Jamuna Devi, Animesh Biswas, and R Sophia Porchelvi. Pythagorean hyperbolic fuzzy sets: A novel approach to handling uncertainty. *IAENG International Journal of Applied Mathematics*, 55(11):3660–3679, 2025.

[321] Abhilash Kangsha Banik, Alakananda Konwar, and Palash Dutta. A novel multi-criteria decision-making method based on hyperbolic fuzzy sets. *Fuzzy Decision-Making Algorithms: Mathematical Optimization, Metaheuristic Optimization, Supply Chain Management*, page 101, 2026.

[322] Gürkan Işık. Compact fuzzy set: A parameter-free unifying approach for flexible and reliable conflict modeling in group decision making. *Science and Engineering*, 10(2):208–247, 2026.

[323] Ping Wang, Jie Wang, Guiwu Wei, and Cun Wei. Similarity measures of q-rung orthopair fuzzy sets based on cosine function and their applications. *Mathematics*, 2019.

[324] Muhmamet Deveci, Dragan Pamucar, Ilgin Gokasar, Mario Köppen, and Brij Bhooshan Gupta. Personal mobility in metaverse with autonomous vehicles using q-rung orthopair fuzzy sets based opa-rafsi model. *IEEE Transactions on Intelligent Transportation Systems*, 24:15642–15651, 2023.

[325] Sukanta Nayak and Snehashish Chakraverty. Pseudo fuzzy set. *arXiv preprint arXiv:1502.05864*, 2015.

[326] Muhammad Akram, A Nagoor Gani, and A Borumand Saeid. Vague hypergraphs. *Journal of Intelligent & Fuzzy Systems*, 26(2):647–653, 2014.

[327] Humberto Bustince and P Burillo. Vague sets are intuitionistic fuzzy sets. *Fuzzy sets and systems*, 79(3):403–405, 1996.

[328] Juanjuan Lu, Linli Zhu, and Wei Gao. Structured representation of fuzzy data by bipolar fuzzy hypergraphs. In *International Conference on Machine Learning for Cyber Security*, pages 663–676. Springer, 2022.

[329] Wen-Ran Zhang. Bipolar fuzzy sets. 1997.

[330] Wen-Ran Zhang. Bipolar fuzzy sets and relations: a computational framework for cognitive modeling and multiagent decision analysis. *NAFIPS/IFIS/NASA '94. Proceedings of the First International Joint Conference of The North American Fuzzy Information Processing Society Biannual Conference. The Industrial Fuzzy Control and Intellige*, pages 305–309, 1994.

[331] Chen Shouyu and Guo Yu. Variable fuzzy sets and its application in comprehensive risk evaluation for flood-control engineering system. *Fuzzy Optimization and Decision Making*, 5(2):153–162, 2006.

[332] Wen-chuan Wang, Dong-mei Xu, Kwok-wing Chau, and Guan-jun Lei. Assessment of river water quality based on theory of variable fuzzy sets and fuzzy binary comparison method. *Water resources management*, 28(12):4183–4200, 2014.

[333] Yunhai Fang, Xilai Zheng, Hui Peng, Huan Wang, and Jia Xin. A new method of the relative membership degree calculation in variable fuzzy sets for water quality assessment. *Ecological indicators*, 98:515–522, 2019.

[334] Vasile Patrascu. A new penta-valued logic based knowledge representation. *arXiv preprint arXiv:1502.05562*, 2015.

[335] Vasile Patrascu. Similarity, cardinality and entropy for bipolar fuzzy set in the framework of penta-valued representation. *arXiv preprint arXiv:1506.02060*, 2015.

[336] Yong-Uk Cho, Young-Bae Jun, and Seok-Zun Song. Bifuzzy ideals of pseudo mv-algebras. *JOURNAL OF APPLIED MATHEMATICS AND COMPUTING*, 22(1/2):475, 2006.

[337] Tadeusz Gerstenkorn and Jacek Mańko. Bifuzzy probabilistic sets. *Fuzzy sets and systems*, 71(2):207–214, 1995.

[338] Takaaki Fujita and Florentin Smarandache. *A Comprehensive Survey and Generalization of Linguistic, MOD, Interval, Subset, and Higher-Order Structures.* Neutrosophic Science International Association (NSIA) Publishing House, September 2026.

[339] Mohammad Shafiq bin Mohammad Kamari, Zahari Bin Md Rodzi, RH Al-Obaidi, Faisal Al-Sharq, Ashraf Al-Quran, et al. Deciphering the geometric bonferroni mean operator in pythagorean neutrosophic sets framework. *Neutrosophic Sets and Systems*, 75:139–161, 2025.

[340] Raed Hatamleh, Ahmed Salem Heilat, M Palanikumar, and Abdallah Al-Husban. Characterization of interaction aggregating operators setting interval-valued pythagorean neutrosophic set. *Neutrosophic Sets and Systems*, 81:285–305, 2025.

[341] Tapas Kumar Paul and Madhumangal Pal. Sustainable building material supplier assessment in pythagorean neutrosophic setting using itara and macont methods. *Journal of Industrial Information Integration*, page 101000, 2025.

[342] Prasanta Kumar Raut, Surapati Pramanik, and Bhabani S Mohanty. An overview of fermatean neutrosophic graphs. *Neutrosophic Sets and Systems*, 82(1):40, 2025.

[343] Muhammad Saeed, Mehar Un Nisa, Muhammad Haris Saeed, Tmader Alballa, and Hamiden Abd El-Wahed Khalifa. Detecting patterns of infection-induced fertility using fermatean neutrosophic set with similarity analysis. *IEEE Access*, 11:112320–112333, 2023.

[344] S Bhuvaneshwari, C Sweety, Akanksha Singh, Said Broumi, Mohamed Talea, and Prasanta Kumar Raut. A novel and an efficient codas technique to solve real-life magdm problems in fermatean neutrosophic environment. *Neutrosophic Sets and Systems*, 72(1):3, 2024.

[345] Florentin Smarandache. *Indeterminacy Fuzzy Set coincides with the Picture Fuzzy Set, and both are particular cases of the Refined Neutrosophic Set.* Infinite Study, 2026.

[346] Cemal Ardil. Indeterminacy fuzzy cognitive maps: Simplex-constrained causal reasoning with primitive epistemic indeterminacy. *International Journal of Computer and Information Engineering*, 2026.

[347] Chris J Hinde, Robert Steven Patching, and Stephen A McCoy. Semantic transfer and contradictory evidence in intuitionistic fuzzy sets. In *2008 IEEE International Conference on Fuzzy Systems (IEEE World Congress on Computational Intelligence)*, pages 2095–2102. IEEE, 2008.

[348] CJ Hinde, RS Patching, and SA McCoy. Inconsistent intuitionistic fuzzy sets and mass assignment. *Developments in Fuzzy Sets, Intuitionistic Fuzzy Sets, Generalized Nets and Related Topics. Foundations*, pages 133–153, 2007.

[349] Florentin Smarandache and Mumtaz Ali. Neutrosophic triplet group (revisited). *Neutrosophic sets and Systems*, 26(1):2, 2019.

[350] Chao Wang, Minghu Ha, and Xiaowei Liu. A mathematical model of ternary fuzzy set for voting. *Journal of Intelligent & Fuzzy Systems*, 29(6):2381–2386, 2015.

[351] Florentin Smarandache. *The Neutral Set as a Re-Notation of the Single-Valued Neutrosophic Set.* Infinite Study, 2026.

[352] C Ardil. A neutral set approach for applying topsis in maintenance strategy selection. *International Journal of Computer and Systems Engineering*, 18(4):225–234, 2024.

[353] Florentin Smarandache. *Neutral/Neutrality are parts of the generic group called Indeterminacy.* Infinite Study, 2026.

[354] Sankar Das, Ganesh Ghorai, and Madhumangal Pal. Picture fuzzy tolerance graphs with application. *Complex & Intelligent Systems*, 8(1):541–554, 2022.

[355] G Muhiuddin, Sun Shin Ahn, Chang Su Kim, and Young Bae Jun. Stable cubic sets. *J. Comput. Anal. Appl*, 23(5):802–819, 2017.

[356] Madad Khan, Young Bae Jun, Muhammad Gulistan, and Naveed Yaqoob. The generalized version of jun's cubic sets in semigroups. *Journal of Intelligent & Fuzzy Systems*, 28(2):947–960, 2015.

[357] Melvin Fitting. Kleene's three valued logics and their children. *Fundamenta informaticae*, 20(1-3):113–131, 1994.

[358] John T Kearns. The strong completeness of a system for kleene's three-valued logic. *Mathematical Logic Quarterly*, 25(3-6):61–68, 1979.

[359] Muhammad Akram, Danish Saleem, and Talal Al-Hawary. Spherical fuzzy graphs with application to decision-making. *Mathematical and Computational Applications*, 25(1):8, 2020.

[360] Xiaojie Zhao, Yanping Zhou, and Lu Sun. Solution of supply chain management problems using rough t spherical fuzzy set lower and upper approximation spaces. *Scientific Reports*, 2026.

[361] Fatma Kutlu Gündoğdu and Cengiz Kahraman. Spherical fuzzy sets and spherical fuzzy topsis method. *Journal of intelligent & fuzzy systems*, 36(1):337–352, 2019.

[362] Khushbakhat Asif, Muhammad Kamran Jamil, Hanen Karamti, Muhammad Azeem, and Kifayat Ullah. Randić energies for t-spherical fuzzy hamacher graphs and their applications in decision making for business plans. *Computational and Applied Mathematics*, 42:1–32, 2023.

[363] Abhishek Guleria and Rakesh Kumar Bajaj. T-spherical fuzzy graphs: operations and applications in various selection processes. *Arabian Journal for Science and Engineering*, 45(3):2177–2193, 2020.

[364] M Murali Krishna Rao and B Venkateswarlu. Tripolar fuzzy interior ideals of a $\gamma$-semiring. *Asia Pacific Journal of Management*, 5(2):192–207, 2018.

[365] Takaaki Fujita. A mathematical study of tripolar intuitionistic fuzzy graphs. *International Journal of Computer Application (IJCA)*, 16(4):16/1–16/11, 2026.

[366] M Murali Krishna Rao, B Venkateswarlu, and Y Adi Narayana. Tripolar fuzzy soft ideals and tripolar fuzzy soft interior ideals over semiring. *Italian journal of pure and applied Mathematics*, (422019):731743, 2019.

[367] M Murali Krishna Rao. Tripolar fuzzy interior ideals of $\gamma$-semigroup. *Ann. Fuzzy Math. Inform*, 15(2):199–206, 2018.

[368] Muhammad Akram, Ayesha Khan, Uzma Ahmad, José Carlos R Alcantud, and Mohammed M Ali Al-Shamiri. A new group decision-making framework based on 2-tuple linguistic complex q-rung picture fuzzy sets. *Math. Biosci. Eng*, 19:11281–11323, 2022.

[369] Ahmet Sarucan, Evrencan Özcan, and Büşra Güler. Bonferroni mean-based aggregation operators on q-rung picture fuzzy sets for multi-criteria decision making in energy storage systems. *Symmetry*, 18(6):966, 2026.

[370] Shawkat Alkhazaleh. Neutrosophic vague set theory. *Critical Review*, 10:29–39, 2015.

[371] Ashraf Al-Quran, Hazwani Hashim, and Lazim Abdullah. A hybrid approach of interval neutrosophic vague sets and dematel with new linguistic variable. *Symmetry*, 12(2):275, 2020.

[372] Satham Hussain, Jahir Hussain, Isnaini Rosyida, and Said Broumi. Quadripartitioned neutrosophic soft graphs. In *Handbook of Research on Advances and Applications of Fuzzy Sets and Logic*, pages 771–795. IGI Global, 2022.

[373] Mouhamed Bayane Bouraima, Ertugrul Ayyıldız, Melike Erdogan, and Dragan Pamucar. An interval-valued intuitionistic fuzzy group decision model for evaluation of cross-border railway development. *Cognitive Computation*, 18(1):22, 2026.

[374] Ravindar Raj Chinnappan and Dhanasekar Sundaram. Image enhancement framework combining interval-valued intuitionistic fuzzy sets and fractional sobel operator. *Artificial Intelligence Review*, 58(9):291, 2025.

[375] Surendra Kumar. Fuzzy reliability of a marine power plant using interval valued vague sets. *International Journal of Applied Science and Engineering*, 2006.

[376] Qian-sheng ZHANG and Sheng-yi JIANG. Multicriteria fuzzy decision making method based on interval-valued vague set. *Journal of Chinese Computer Systems*, 31(5):942–946, 2010.

[377] Ping Xiong and Huabin Cheng. Evaluation model based on periphery theory and interval-valued vague set. In *5th International Conference on Information Science, Electrical, and Automation Engineering (ISEAE 2023)*, volume 12748, pages 242–246. SPIE, 2023.

[378] Yukun Hu, Hong Wang, Chunhui Li, Ang Xu, Li Yi, Chao Liang, Bingyang Xi, and Fang Dai. An integrated approach for design concept evaluation based on interval-valued pythagorean fuzzy set and consensus model. *Kybernetes*, 55(8):3611–3636, 2026.

[379] Yukun Hu, Hong Wang, Chunhui Li, Ang Xu, Li Yi, Chao Liang, Bingyang Xi, and Fang Dai. An integrated approach for design concept evaluation based on interval-valued pythagorean fuzzy set and consensus model. *Kybernetes*, 2025.

[380] Vahid Mohagheghi, Seyed Meysam Mousavi, Mohammad Mojtahedi, and Sid Newton. Evaluating large, high-technology project portfolios using a novel interval-valued pythagorean fuzzy set framework: An automated crane project case study. *Expert Syst. Appl.*, 162:113007, 2020.

[381] Yoojeong Oh, Hans Pasman, Safyan Akram Khan, and Sunhwa Park. Hydrogen storage selection for saudi arabia: A multi-criteria decision making under interval-valued pythagorean fuzzy environment. *International Journal of Hydrogen Energy*, 105:1281–1293, 2025.

[382] Abdul Wahab Mustafa, Zia Bashir, Jawad Ali, and Muhammed I Syam. A unified framework of information measures for complex linear diophantine fuzzy sets with application to chronic kidney disease. *Scientific Reports*, 2026.

[383] Masooma Raza Hashmi, Muhammad Riaz, Arshid Mahmood, Muhammad Ajmaeen, and Muhammad Aslam. Optimizing performance metrics with new distance and similarity measures using control parameters of linear diophantine fuzzy sets. *Engineering Applications of Artificial Intelligence*, 162:112557, 2025.

[384] Özlem Arslan and Selcuk Cebi. A novel approach for multi-criteria decision making: extending the waspas method using decomposed fuzzy sets. *Computers & Industrial Engineering*, 196:110461, 2024.

[385] Ilanthenral Kandasamy. Double-valued neutrosophic sets, their minimum spanning trees, and clustering algorithm. *Journal of Intelligent systems*, 27(2):163–182, 2018.

[386] Qiuyan Zhao and Wentao Li. Incorporating intelligence in multiple-attribute decision-making using algorithmic framework and double-valued neutrosophic sets: Varied applications to employment quality evaluation for university graduates. *Neutrosophic Sets and Systems*, 76:59–78, 2025.

[387] Qaisar Khan, Peide Liu, and Tahir Mahmood. Some generalized dice measures for double-valued neutrosophic sets and their applications. *Mathematics*, 6(7):121, 2018.

[388] Nguyen Xuan Thao, Florentin Smarandache, and Nguyen Van Dinh. *Support-neutrosophic set: a new concept in soft computing*. Infinite Study, 2017.

[389] Takaaki Fujita and Florentin Smarandache. Local-neutrosophic logic and local-neutrosophic sets: Incorporating locality with applications. *Advancing Uncertain Combinatorics through Graphization, Hyperization, and Uncertainization: Fuzzy, Neutrosophic, Soft, Rough, and Beyond*, page 51, 2025.

[390] Bin Zhu and Zeshui Xu. Some results for dual hesitant fuzzy sets. *Journal of Intelligent & Fuzzy Systems*, 26(4):1657–1668, 2014.

[391] José Carlos R Alcantud, Gustavo Santos-García, Xindong Peng, and Jianming Zhan. Dual extended hesitant fuzzy sets. *Symmetry*, 11(5):714, 2019.

[392] Florentin Smarandache. Ambiguous set is a subclass of the double refined indeterminacy neutrosophic set, and of the refined neutrosophic set in general. *Neutrosophic Sets & Systems*, 58, 2023.

[393] Pritpal Singh et al. Ambiguous set theory: A new approach to deal with unconsciousness and ambiguousness of human perception. *Full Length Article*, 5(1):52–2, 2023.

[394] Florentin Smarandache. Ambiguous set (as) is a particular case of the quadripartitioned neutrosophic set (qns). *nidus idearum*, page 16, 2024.

[395] M Balamurugan, Khalil H Hakami, Moin A Ansari, Anas Al-Masarwah, and K Loganathan. Quadri-polar fuzzy fantastic ideals in bci-algebras: A topsis framework and application. *European Journal of Pure and Applied Mathematics*, 17(4):3129–3155, 2024.

[396] Yuri Kaluzhny and Alexei Yu Muravitsky. A knowledge representation based on the belnap's four-valued logic. *Journal of Applied Non-Classical Logics*, 3(2):189–203, 1993.

[397] Marcelo E Coniglio and Abilio Rodrigues. From belnap-dunn four-valued logic to six-valued logics of evidence and truth. *Studia Logica*, 112(3):561–606, 2024.

[398] Takaaki Fujita and Florentin Smarandache. Antipodal turiyam neutrosophic graphs. *Neutrosophic Optimization and Intelligent Systems*, 5:1–13, 2024.

[399] Takaaki Fujita. Review of rough turiyam neutrosophic directed graphs and rough pentapartitioned neutrosophic directed graphs. *Neutrosophic Optimization and Intelligent Systems*, 5:48–79, 2025.

[400] Takaaki Fujita and Florentin Smarandache. Pythagorean, fermatean, and complex turiyam neutrosophic graphs. *SciNexuses*, 2:39–63, 2025.

[401] Eswaran Poongothai and Florentin Smarandache. On fuzzy neutrosophic turiyam sets in fourth-dimensional data space and its applications. *European Journal of Pure and Applied Mathematics*, 19(2):7675–7675, 2026.

[402] Prem Kumar Singh, Naveen Surathu, Ghattamaneni Surya Prakash, et al. Turiyam based four way unknown profile characterization on social networks. *Full Length Article*, 10(2):27–7, 2024.

[403] Rama Mallick and Surapati Pramanik. Pentapartitioned neutrosophic set and its properties. *Neutrosophic Sets and Systems*, 35:49, 2020.

[404] Radha R. and Stanis Arul Mary. Pentapartitioned neutrosophic pythagorean set. *International Research Journal on Advanced Science Hub*, 2021.

[405] Suman Das, Rakhal Das, and Binod Chandra Tripathy. Topology on rough pentapartitioned neutrosophic set. *Iraqi Journal of Science*, 2022.

[406] Jiarong Jia. Triple-valued neutrosophic off for public digital cultural service level evaluation in craft and art museums. *Neutrosophic Sets and Systems*, 88(1):44, 2025.

[407] Takaaki Fujita. Triple-valued neutrosophic set, quadruple-valued neutrosophic set, quintuple-valued neutrosophic set, and double-valued indetermsoft set. *Neutrosophic Systems with Applications*, 25(5):3, 2025.

[408] Hongxin Wang. Professional identity formation in traditional chinese medicine students: An educational perspective using triple-valued neutrosophic set. *Neutrosophic Sets and Systems*, 88:83–92, 2025.

[409] Takaaki Fujita, Raed Hatamleh, and Ahmed Heilat. Advanced partitioned neutrosophic offsets, oversets, and undersets: Modeling neither agree nor disagree and beyond. *Statistics, Optimization & Information Computing*, 15(6):5239–5283, 2026.

[410] Takaaki Fujita. Advanced partitioned neutrosophic sets: Formalization of hexa-, hepta-, octa-, nona-, and deca-partitioned structures. *preprint*, 2025.

[411] Fang Sun, Mifeng Ren, Yujing Shi, and Xuanbai Feng. Quadruple-valued neutrosophic offset for maintenance platforms design quality evaluation of digital monitoring and remote operation in new energy power generation systems. *Neutrosophic Sets and Systems*, 88:940–953, 2025.

[412] Maha Mohammed Saeed, Sami Ullah Khan, Fatima Suriyya, Arif Mehmood, and Jamil J Hamja. Interval-valued complex neutrosophic sets and complex neutrosophic soft topological spaces. *International Journal of Analysis and Applications*, 23:132–132, 2025.

[413] Ajoy Kanti Das, Nandini Gupta, Rajat Das, Rajib Mallik, and Suman Das. Interval-valued neutrosophic rough soft set based intelligent multi-criteria decision-making framework. *Boletim da Sociedade Paranaense de Matemática*, 2025.

[414] Anjan Mukherjee and Rakhal Das. Generalized interval-valued neutrosophic set and its application. *Trans. Fuzzy Sets Syst.*, 4:1–13, 2025.

[415] Aiyared Aiyared, Murugan Palanikumar, and M. S. Malchijah Raj. Different algebraic structures and their properties setting logarithm operator applied to extended neutrosophic interval-valued set. *International Journal of Neutrosophic Science*, 2025.

[416] Samaneh Salari and Ali Karimi. Introducing an integrated approach for fire safety assessment in healthcare facilities by interval valued neutrosophic-ahp and fuzzy inference system. *Heliyon*, 11(2), 2025.

[417] Sijia Zhu and Zhe Liu. Distance measures of picture fuzzy sets and interval-valued picture fuzzy sets with their applications. *AIMS Math*, 8(12):29817–29848, 2023.

[418] Rizwan Gul and Mehwish Sarfraz. Enhancing artificial intelligence models with interval-valued picture fuzzy sets and sugeno-weber triangular norms. *Spectrum of engineering and management sciences*, 3(1):126–146, 2025.

[419] Irfan Deli, Mumtaz Ali, and Florentin Smarandache. Bipolar neutrosophic sets and their application based on multi-criteria decision making problems. *2015 International Conference on Advanced Mechatronic Systems (ICAMechS)*, pages 249–254, 2015.

[420] Shibo Wu, Changrun Chen, Zhaoyu Wang, and Lin Song. Failure probability assessment method for offshore oil and gas systems based on interval-valued t-spherical fuzzy set and credal networks. *Mathematics*, 14(12):2151, 2026.

[421] Lu Chen. Interval-valued t-spherical fuzzy extended power aggregation operators and their application in multi-criteria decision-making. *Journal of Intelligent Systems*, 33(1):20240039, 2024.

[422] Raed Hatamleh, Nasir Odat, Abdallah Al-Husban, Arif Mehmood Khattak, Alaa M Abd El-latif, Husham M Attaalfadeel, Walid Abdelfattah, and Ahmad M Abdel-Mageed. Mapping love: A hepta-partitioned neutrosophic machine learning study of university students' romantic sensations. *European Journal of Pure and Applied Mathematics*, 18(3):6484–6484, 2025.

[423] Maha Mohammed Saeed, Raed Hatamleh, Hamza Ali Abujabal, Yahya Khan, Abdallah Al-Husban, Amy A Laja, Sulfaisa MM Pangilan, Cris L Armada, Jamil J Hamja, and Arif Mehmood Khattak. Heptapartitioned neutrosophic soft topologies and machine learning techniques for exploring romantic feelings. *European Journal of Pure and Applied Mathematics*, 18(3):6221–6221, 2025.

[424] Wenwen Meng. Quintuple-valued neutrosophic offset for quality evaluation of cross-border e-commerce talent training based on artificial intelligence. *Neutrosophic Sets and Systems*, 88(1):54, 2025.

[425] Yuan Sun. Assessing the performance of basic education informatization in the digital age using quintuple-valued neutrosophic set. *Neutrosophic Sets and Systems*, 88(1):18, 2025.

[426] CV Malavika, Kalaivani Chandran, Sofia Jennifer John, and Florentin Smarandache. Python based tool for neutrosophic cubic sets and their operations. *Journal of Fuzzy Extension and Applications*, 7(2):592–614, 2026.

[427] Berna Joyce and Dr. S. Remigius Mary. Pythagorean neutrosophic cubic set. *Communications on Applied Nonlinear Analysis*, 2025.

[428] Basudev Siddhya. Multi-criteria decision-making using aczel-alsina aggregation operators in a neutrosophic cubic set environment. *International Journal on Science and Technology*, 2025.

[429] Adil Darvesh, Khalid Naeem, and Syed Bilal Hussain Shah. Scrutiny of neutrosophic cubic fuzzy data with different parametric values using bonferroni weighted mean: A robust mcdm approach. *Decision making advances*, 3(1):245–265, 2025.

[430] Takaaki Fujita, Ajoy Kanti Das, Arif Mehmood, and Suman Das. Revisiting tripolar hyperneutrosophic graphs and their applications. *Journal of Evidential and Paraconsistent Reasoning*, 2(2), 2027.

[431] Takaaki Fujita. A mathematical study of tripolar neutrosophic graphs. *International Journal of Research in Engineering & Science (IJRES)*, 16(4):13/1–13/8, 2026.

[432] Shahzaib Ashraf, Saleem Abdullah, and Abbas Qadir. Novel concept of cubic picture fuzzy sets. *Journal of New Theory*, (24):59–72, 2018.

[433] Muhammad Naeem Khan Tanoli, Khadija Rafique, Zafar Mahmood, Nesreen Sirelkhtam Elmki Abdalla, Ioan-Lucian Popa, and Abhinav Kumar. Einstein aggregation operators for multicriteria group decision making in uncertain environments using cubic picture fuzzy sets. *Scientific Reports*, 15(1):36545, 2025.

[434] PR Kavyasree and B Surender Reddy. Cubic picture hesitant fuzzy linear spaces and their applications in multi criteria decision making. In *Real life applications of multiple criteria decision making techniques in fuzzy domain*, pages 533–557. Springer, 2022.

[435] Giovanni Panti. Multi-valued logics. In *Quantified representation of uncertainty and imprecision*, pages 25–74. Springer, 1998.

[436] Sabu Sebastian and TV Ramakrishnan. Multi-fuzzy sets. In *International Mathematical Forum*, volume 5, pages 2471–2476, 2010.

[437] Sabu Sebastian, TV Ramakrishnan, KK Gireesan, and SJ Sangeeth. Multi-fuzzy sets and neural networks: a collaborative tool for artificial intelligence. *International Journal of Information Technology*, 16(6):3419–3423, 2024.

[438] Ryszard Wójcicki. A theorem of the finiteness of the degree of maximality of the n-valued lukasiewicz logic. *Bulletin of the Section of Logic*, 4(1):19–23, 1975.

[439] Takaaki Fujita and Florentin Smarandache. *Polar Concepts in Uncertainty Theory: Fuzzy, Neutrosophic, and Uncertain Set Frameworks.* Neutrosophic Science International Association (NSIA) Publishing House, 2026.

[440] Muhammad Arshad, Muhammad Saeed, and Atiqe Ur Rahman. Convexity cum concavity on refined fuzzy set with some properties. *Uncertainty Discourse and Applications*, 1(1):140–150, 2024.

[441] Deepak Kumar et al. Reliability assessment of refined single-valued neutrosophic data with application in multi-criteria decision-making. *Neutrosophic Sets & Systems*, 94:86, 2025.

[442] Ankita Ankita and Deepak Kumar. Reliability assessment of refined single-valued neutrosophic data with application in multi-criteria decision-making. *Neutrosophic Sets and Systems*, 94(1):6, 2025.

[443] Takaaki Fujita. Iterative refinement in neutrosophic logic: A unified framework for $n$-refined and partitioned neutrosophic models. *Advances in Neutrosophic Mathematics and Informatics*, 1(3), 2026.

[444] ChangXing Fan and Jun Ye. The cosine measure of refined-single valued neutrosophic sets and refined-interval neutrosophic sets for multiple attribute decision-making. *Journal of Intelligent & Fuzzy Systems*, 33(4):2281–2289, 2017.

[445] Vakkas Uluçay. Some concepts on interval-valued refined neutrosophic sets and their applications. *Journal of Ambient Intelligence and Humanized Computing*, 12(7):7857–7872, 2021.

[446] Florentin Smarandache. *Introduction to the multineutrosophic set.* Infinite Study, 2023.

[447] Florentin Smarandache. Neutrosophic set is a generalization of intuitionistic fuzzy set, inconsistent intuitionistic fuzzy set (picture fuzzy set, ternary fuzzy set), pythagorean fuzzy set, spherical fuzzy set, and q-rung orthopair fuzzy set, while neutrosophication is a generalization of regret theory, grey system theory, and three-ways decision (revisited). *Journal of New Theory*, 29:1–31, 2019.

[448] Florentin Smarandache. *Nidus Idearum. Scilogs, XIII: Structure/NeutroStructure/AntiStructure.* Infinite Study, 2024.

[449] Shawkat Alkhazaleh. Shadow soft set theory. *International Journal of Fuzzy Logic and Intelligent Systems*, 22(4):422–432, 2022.

[450] Ayman A Hazaymeh. Time-shadow soft set: Concepts and applications. *International Journal of Fuzzy Logic and Intelligent Systems*, 24(4):387–398, 2024.

[451] Ljubica Djurović, Maja Laković, and Nenad Stojanović. Decision-making algorithm based on the energy of interval-valued fuzzy soft sets. *arXiv preprint arXiv:2405.15801*, 2024.

[452] Murugan Palanikumar and K Arulmozhi. Novel possibility pythagorean interval valued fuzzy soft set method for a decision making. *TWMS Journal of Applied and Engineering Mathematics*, 2023.

[453] Xiuqin Ma, Yanyan Han, Hongwu Qin, and Peng Wang. Knn data filling algorithm for incomplete interval-valued fuzzy soft sets. *International Journal of Computational Intelligence Systems*, 16(1):30, 2023.

[454] Muhammad Saeed, Muhammad Haris Saeed, Rimsha Shafaqat, Salvatore Sessa, Umar Ishtiaq, and Ferdinando Di Martino. A theoretical development of cubic pythagorean fuzzy soft set with its application in multi-attribute decision making. *Symmetry*, 14(12):2639, 2022.

[455] Gulfam Shahzadi, José Carlos R Alcantud, and Rana Talha Ahmad. Multi criteria decision-making model using the circular pythagorean fuzzy soft set model. *Advanced Engineering Informatics*, 70:104134, 2026.

[456] Gulfam Shahzadi, Muhammad Akram, and Bijan Davvaz. Pythagorean fuzzy soft graphs with applications. *Journal of Intelligent & Fuzzy Systems*, 38(4):4977–4991, 2020.

[457] Murat Kirisci. Measures of distance and entropy based on the fermatean fuzzy-type soft sets approach. *Universal Journal of Mathematics and Applications*, 7(1):12–29, 2024.

[458] Ghous Ali and Masfa Nasrullah Ansari. Multiattribute decision-making under fermatean fuzzy bipolar soft framework. *Granular Computing*, 7(2):337–352, 2022.

[459] Raed Hatamleh, Nasir Odat, Hamza Ali Abujabal, Arif Mehmood Khattak, Alaa M Abd El-latif, Husham MM Attaalfadeel, Abdelhalim Hasnaoui, et al. Fermatean double-valued neutrosophic soft topological spaces. *European Journal of Pure and Applied Mathematics*, 18(3):6483–6483, 2025.

[460] Khaleed Alhazaymeh and Nasruddin Hassan. Vague soft set relations and functions. *Journal of Intelligent & Fuzzy Systems*, 28(3):1205–1212, 2015.

[461] Abedallah Al-shboul, Khaleed Alhazaymeh, Kah Lun Wang, and Kok Bin Wong. Fermatean vague soft set and its application in decision making. *Mathematics*, 12(23):3699, 2024.

[462] Esraa Masadeh, Khaleed Alhazaymeh, Abedallah Al-shboul, Waleed Mohammed Abdelfattah, Rana Muhammad Zulqarnain, Imran Siddique, and Hijaz Ahmad. An extended topsis technique in cubic vague soft set relations with its application in renewable energy sources. *European Journal of Pure and Applied Mathematics*, 18(4):7017–7017, 2025.

[463] Orhan Dalkılıç and Naime Demirtaş. Bipolar fuzzy soft set theory applied to medical diagnosis. *Turkish Journal of Mathematics and Computer Science*, 16(2):314–324, 2024.

[464] Ema Carnia, Dwi Susanti, Mohd Zaki Awang Chek, Mugi Lestari, Moch Panji Agung Saputra, et al. A systematic literature review on bipolar fuzzy soft sets in environmental sustainability and the conceptual development of weighted bfss. *Sustainability (2071-1050)*, 18(13):6873, 2026.

[465] Shadya M Mershkhan and Baravan A Asaad. Double-framed bipolar fuzzy soft sets and algorithmic approaches with symmetry for multi-criteria decision-making under uncertainty. *Symmetry*, 18(1):119, 2026.

[466] Devaraj Ajay and P Chellamani. Pythagorean neutrosophic soft sets and their application to decision-making scenario. In *International Conference on Intelligent and Fuzzy Systems*, pages 552–560. Springer, 2021.

[467] M Palanikumar, K Arulmozhi, and T Gunasekar. Possibility pythagorean neutrosophic soft sets and its application of decision making. *J. Int. Math. Virtual Inst*, 12(1):1–16, 2022.

[468] S Anitha and A Francina Shalini. Entropy and distance measures of bipolar pythagorean neutrosophic soft set. In *Engineering, Science, and Sustainability*, pages 182–188. CRC Press, 2023.

[469] Muhammad Saeed, Imrana Shafique, and Hatıra Gunerhan. Fundamentals of fermatean neutrosophic soft set with application in decision making problem. *International Journal of Mathematics, Statistics, and Computer Science*, 3:294–312, 2025.

[470] Najla Althuniyan, Abedallah Al-shboul, Sarah Aljohani, Kah Lun Wang, Kok Bin Wong, Khaleed Alhazaymeh, and Suhad Subhi Aiady. Decision-making with fermatean neutrosophic vague soft sets using a technique for order of preference by similarity to ideal solution. *Axioms*, 14(5):381, 2025.

[471] Ahmed Mostafa Khalil, Sheng-Gang Li, Harish Garg, Hongxia Li, and Shengquan Ma. New operations on interval-valued picture fuzzy set, interval-valued picture fuzzy soft set and their applications. *Ieee Access*, 7:51236–51253, 2019.

[472] Muhammad Kamran, Qingyu Zhang, Shahzaib Ashraf, and Muhammad Nadeem. Enhancing decision-making in environmental management: a complex picture fuzzy soft set approach to sustainability assessment. *Computational and Applied Mathematics*, 44(5):258, 2025.

[473] Samet Memiş. Another view on picture fuzzy soft sets and their product operations with soft decision-making. *Journal of new theory*, (38):1–13, 2022.

[474] Muhammad Jabir Khan, Poom Kumam, Peide Liu, Wiyada Kumam, and Habib ur Rehman. An adjustable weighted soft discernibility matrix based on generalized picture fuzzy soft set and its applications in decision making. *Journal of Intelligent & Fuzzy Systems*, 38(2):2103–2118, 2020.

[475] G Muhiuddin, Feng Feng, and Young Bae Jun. Subalgebras of bck/bci-algebras based on cubic soft sets. *The Scientific World Journal*, 2014(1):458638, 2014.

[476] Rana Muhammad Zulqarnain, Saalam Ali, Hamza Naveed, Mohamed Abubakar Fiidow, Imran Siddique, and Sohaib Abdal. A novel multi-criteria group decision-making technique in traffic flow and safety assessment using interval-valued complex spherical fuzzy soft set. *Scientific Reports*, 2026.

[477] Mehwish Sarfraz, Diaa S Metwally, Hamiden Abd El-Wahed Khalifa, and Muhammad Saeed. Aczel-alsina t-norm and t-conorm of spherical fuzzy soft set and their application traffic flow management. *Journal of Fuzzy Extension and Applications*, page e244909, 2026.

[478] ELIF Güner and HALIS Aygün. Spherical fuzzy soft sets: Theory and aggregation operator with its applications. *Iranian Journal of Fuzzy Systems*, 19(2):83–97, 2022.

[479] Muhammad Akram and Alessio Martino. Multi-attribute group decision making based on t-spherical fuzzy soft rough average aggregation operators. *Granular Computing*, 8(1):171–207, 2023.

[480] Fatih Karamaz, Faruk Karaaslan, and Mustafa Karhan. A new approach to r, s, t-spherical fuzzy soft sets with applications to multi-criteria decision-making. *Big Data and Computing Visions*, 6(1):1–23, 2026.

[481] Shabana Farman, Faiz Muhammad Khan, and Naila Bibi. T-spherical fuzzy soft rough aggregation operators and their applications in multi-criteria group decision-making. *Granular Computing*, 9(1):6, 2024.

[482] Somen Debnath. Fuzzy quadripartitioned neutrosophic soft matrix theory and its decision-making approach. *Journal of Computational and Cognitive Engineering*, 1(2):88–93, 2022.

[483] Somen Debnath. Interval-valued intuitionistic quadripartitioned neutrosophic soft sets with t, f, c, and u as dependent neutrosophic components and their application in decision-making problem. *Journal of New Results in Science*, 11(1):26–47, 2022.

[484] Anjan Mukherjee and Abhik Mukherjee. Interval-valued intuitionistic fuzzy soft rough approximation operators and their applications in decision making problem. *Annals of Data Science*, 9(3):611–625, 2022.

[485] Deepak Kumar and Anita Kumari. Utilization of interval-valued intuitionistic fuzzy soft set in the health sector. *International Journal of Health Sciences*, 6(S6):8437–8444, 2022.

[486] Sonia, Pratiksha Tiwari, and Priti Gupta. Novel distance, similarity and entropy measures for interval valued intuitionistic fuzzy soft set. *Journal of Intelligent & Fuzzy Systems*, 43(3):3067–3086, 2022.

[487] Rana Muhammad Zulqarnain, Hamza Naveed, Hassan Naseer, Rifaqat Ali, Abdullatif Ghallab, Imran Siddique, and Sohaib Abdal. An extended edas model for sustainable supplier selection in food supply chain management using interval-valued pythagorean fuzzy soft set. *Scientific Reports*, 2026.

[488] Rana Muhammad Zulqarnain, Imran Siddique, Muhammad Asif, Hijaz Ahmad, Sameh Askar, and Shahid Hussain Gurmani. Extension of correlation coefficient based topsis technique for interval-valued pythagorean fuzzy soft set: A case study in extract, transform, and load techniques. *Plos one*, 18(10):e0287032, 2023.

[489] Awais Majeed, Khizar Hayat, and Faruk Karaaslan. Distance and similarity measures on generalized interval-valued pythagorean fuzzy soft sets and their decision-making applications in sports. In *Operations Research Forum*, volume 6, page 144. Springer, 2025.

[490] K Ashma Banu, J Vimala, Nasreen Kausar, and Željko Stević. Optimizing road safety: integrated analysis of motorized vehicle using lattice ordered complex linear diophantine fuzzy soft set. *PeerJ Computer Science*, 10:e2165, 2024.

[491] G Punnam Chander and Sujit Das. A linear diophantine fuzzy soft set-based decision-making approach using revised max-min average composition method. In *Fuzzy, Rough and Intuitionistic Fuzzy Set Approaches for Data Handling: Theory and Applications*, pages 165–181. Springer, 2023.

[492] K Ashma Banu, J Vimala, F Karaaslan, and S Nithya Sri. An advanced algorithmic approach based on lattice-ordered complex linear diophantine fuzzy soft set with an associated navigational case study: Ka banu et al. *The Journal of Supercomputing*, 81(5):649, 2025.

[493] Mithun Datta, Kalyani Debnath, and Surapati Pramanik. Pentapartitioned neutrosophic soft set with interval membership. *Neutrosophic Sets and Systems*, 79(1):37, 2025.

[494] Babisha Julit RL et al. Pentapartitioned fermatean neutrosophic soft-rough set and its applications. *South East Asian Journal of Mathematics & Mathematical Sciences*, 21(2):307, 2025.

[495] KALYANI DEBNATH, MITHUN DATTA, and APU KUMAR SAHA. Fusing neutrosophy and soft sets: A pnss based decision making model for healthcare optimization. *Mathematica Montisnigri*, 64:84–98, 2025.

[496] S Kumar, A Mary, and R Radha. Penta partitioned neutrosophic soft topological space. *Fuzzy, Intuitionistic and Neutrosophic Set Theories and their Applications in Decision Analysis*, pages 49–59, 2025.

[497] Irfan Deli. Interval-valued neutrosophic soft sets and its decision making. *International Journal of Machine Learning and Cybernetics*, 8:665–676, 2017.

[498] Said Broumi, Irfan Deli, and Florentin Smarandache. Relations on interval valued neutrosophic soft sets. *Journal of New results in Science*, 3(5), 2014.

[499] Ayman Hazaymeh. Decision-making method based on interval-valued fuzzy soft expert set. *Statistics, Optimization & Information Computing*, 16(5):4837–4869, 2026.

[500] Reefan Mosallam Almozaini and Kholood Mohammad Alsager. Interval-valued picture fuzzy soft rough sets: A new hybrid framework for robust multi-criteria group decision-making. *Symmetry*, 18(3):419, 2026.

[501] S Anita Shanthi and M Gayathri. Accuracy function on interval valued picture fuzzy soft sets. *Materials Today: Proceedings*, 51:2500–2503, 2022.

[502] S Anita Shanthi and M Gayathri. Entropy measure on interval valued picture fuzzy soft set. In *AIP Conference Proceedings*, volume 2764, page 030008. AIP Publishing LLC, 2023.

[503] Nenad Stojanović, Marina Svičević, Tatjana Stojanović, Nemanja Vučićević, and Filip Andrić. A novel decision-making algorithm based on interval-valued bipolar neutrosophic soft sets. *Information*, 17(8):772, 2026.

[504] Khulud Fahad Bin Muhaya and Kholood Mohammad Alsager. Extending neutrosophic set theory: Cubic bipolar neutrosophic soft sets for decision making. *AIMS Mathematics*, 9(10):27739–27769, 2024.

[505] A Priya, P Maragatha Meenakshi, Aiyared Iampan, N Rajesh, and Suganthi Mariyappan. Possibility neutrosophic bipolar fuzzy soft sets and their applications. *International Journal of Neutrosophic Science*, 23(1):176–193, 2023.

[506] Muhammad Gulistan and Nasruddin Hassan. A generalized approach towards soft expert sets via neutrosophic cubic sets with applications in games. *Symmetry*, 11(2):289, 2019.

[507] Surapati Pramanik, S Dalapati, A Alam, and TK Roy. *Some operations and properties of neutrosophic cubic soft set*. Infinite Study, 2017.

[508] Muhammad Gulistan, Ismat Beg, and Naveed Yaqoob. A new approach in decision making problems under the environment of neutrosophic cubic soft matrices. *Journal of Intelligent & Fuzzy Systems*, 36(1):295–307, 2019.

[509] V Chinnadurai and S Madhanraj. Complex cubic picture fuzzy soft matrices. In *International Conference on Trends in Computational and Cognitive Engineering*, pages 59–69. Springer, 2023.

[510] C Antony Crispin Sweety, IR Sumathi, and G Aishwaryapriyadharshini. Priority weighted neutrosophic refined soft set: A multi-criteria decision-making approach. In *Handbook of Research on Advances and Applications of Fuzzy Sets and Logic*, pages 796–832. IGI Global Scientific Publishing, 2022.

[511] Derya BAKBAK and Vakkas ULUÇAY. Hierarchical clustering methods in architecture based on refined q-single-valued neutrosophic sets. *NeutroAlgebra Theory*, 1:122, 2021.

[512] Hamza Naveed and Saalam Ali. Multi-criteria decision-making approach based on correlation coefficient for multi-polar interval-valued neutrosophic soft set. *Neutrosophic Syst. with Appl*, 24:18–33, 2024.

[513] YS Yun. Parametric operations between 3-dimensional triangular fuzzy number and trapezoidal fuzzy set. *Journal of Algebra & Applied Mathematics*, 21(2), 2023.

[514] Tong Shaocheng. Interval number and fuzzy number linear programmings. *Fuzzy sets and systems*, 66(3):301–306, 1994.

[515] Takaaki Fujita and Florentin Smarandache. *Survey of Z-Number, D-Number, and ZD-Number*. Infinite Study, 2026.

[516] Orhan Dalkılıç. Enhancing relationship analysis through topological structures based on virtual fuzzy parameterized soft sets. *Journal of Applied Mathematics and Computing*, 72(1):7, 2026.

[517] Deniz Fidan and SERDAR Enginoğlu. Fuzzy parameterized fuzzy soft measure spaces. *Universal Journal of Mathematics and Applications*, 9(3):203–231, 2026.

[518] Irfan Deli and Naim Çağman. Intuitionistic fuzzy parameterized soft set theory and its decision making. *Applied Soft Computing*, 28:109–113, 2015.

[519] Bhagawati Prasad Joshi, Abhay Kumar, Akhilesh Singh, Pradeep Kumar Bhatt, and Bhupender Kumar Bharti. Intuitionistic fuzzy parameterized fuzzy soft set theory and its application. *Journal of Intelligent & Fuzzy Systems*, 35(5):5217–5223, 2018.

[520] Said Broumi, Irfan Deli, and Florentin Smarandache. Neutrosophic parametrized soft set theory and its decision making. *Collected Papers. Volume X: On Neutrosophics, Plithogenics, Hypersoft Set, Hypergraphs, and other topics*, 40, 2022.

[521] Said Broumi, Irfan Deli, and Florentin Smarandache. Interval valued neutrosophic parameterized soft set theory and its decision making. *Journal of new results in science*, 3(7):58–71, 2014.

[522] Atiqe Ur Rahman, Muhammad Saeed, Mazin Abed Mohammed, Mustafa Musa Jaber, and Begonya Garcia-Zapirain. A novel fuzzy parameterized fuzzy hypersoft set and riesz summability approach based decision support system for diagnosis of heart diseases. *Diagnostics*, 12(7):1546, 2022.

[523] Muhammad Haris Saeed, Fatima Razaq, Muhammad Saeed, Muhammad Salman Habib, and Seung-June Hwang. A decision-theoretic framework for wastewater treatment performance assessment based on a fuzzy parameterized fuzzy hypersoft set approach. *Scientific Reports*, 15(1):20706, 2025.

[524] Naim Çağman, Filiz Çıtak, and Serdar Enginoğlu. Fuzzy parameterized fuzzy soft set theory and its applications. *Turkish Journal of Fuzzy Systems*, 1(1):21–35, 2010.

[525] Fatma Adam and Nasruddin Hassan. Multi q-fuzzy parameterized soft set and its application. *Journal of Intelligent & Fuzzy Systems*, 27(1):419–424, 2014.

[526] Irfan Deli, Yusuf Şubaş, Florentin Smarandache, and Mumtaz Ali. Interval valued bipolar fuzzy weighted neutrosophic sets and their application. In *2016 IEEE international conference on fuzzy systems (FUZZ-IEEE)*, pages 2460–2467. IEEE, 2016.

[527] Ajoy Kanti Das, Nanini Gupta, Rajat Das, Rajib Mallik, and Suman Das. Interval-valued neutrosophic rough soft set based intelligent multi-criteria decision-making framework. *Boletim da Sociedade Paranaense de Matemática (3rd Series)*, 43:1–26, 2025.

[528] Ajoy Kanti Das, Nandini Gupta, Rajat Das, Carlos Granados, and Suman Das. An innovative approach to neutrosophic rough soft set in intelligent multi-criteria decision-making. *OPSEARCH*, pages 1–32, 2025.

[529] Liwen Ma and Meifeng Li. Covering rough set models, fuzzy rough set models and soft rough set models induced by covering similarity. *Information Sciences*, 689:121520, 2025.

[530] Muhammad Abdullah, Khuram Ali Khan, Atiqe Ur Rahman, and Michael Kikomba Kahungu. An intelligent decision-support system for climate change mitigation using similarity measures of hypersoft rough sets. *International Journal of Computational Intelligence Systems*, 19(1):108, 2026.

[531] Jyoti D Thenge, B Surendranath Reddy, and Rupali S Jain. Contribution to soft graph and soft tree. *New Mathematics and Natural Computation*, 15(01):129–143, 2019.

[532] Muhammad Akram and Saira Nawaz. Operations on soft graphs. *Fuzzy information and Engineering*, 7(4):423–449, 2015.

[533] Muhammad Saeed, Muhammad Khubab Siddique, Muhammad Ahsan, Muhammad Rayees Ahmad, and Atiqe Ur Rahman. A novel approach to the rudiments of hypersoft graphs. *Theory and Application of Hypersoft Set, Pons Publication House, Brussel*, pages 203–214, 2021.

[534] Muhammad Saeed, Atiqe Ur Rahman, and Muhammad Arshad. A study on some operations and products of neutrosophic hypersoft graphs. *Journal of Applied Mathematics and Computing*, 68(4):2187–2214, 2022.

[535] Muhammad Saeed, Muhammad Imran Harl, Muhammad Haris Saeed, and Ibrahim Mekawy. Theoretical framework for a decision support system for micro-enterprise supermarket investment risk assessment using novel picture fuzzy hypersoft graph. *Plos one*, 18(3):e0273642, 2023.

[536] R. Jahir Hussain and M. S. Afya Farhana. Fuzzy chromatic number of fuzzy soft cycle and complete fuzzy soft graphs. *AIP Conference Proceedings*, 2023.

[537] Umair Amin, Aliya Fahmi, Yaqoob Naveed, Aqsa Farid, and Muhammad Arshad Shehzad Hassan. Domination in bipolar fuzzy soft graphs. *J. Intell. Fuzzy Syst.*, 46:6369–6382, 2024.

[538] Vakkas Ulucay. Q-neutrosophic soft graphs in operations management and communication network. *Soft Computing*, 25:8441 – 8459, 2021.

[539] S Satham Hussain, R Hussain, and Florentin Smarandache. Domination number in neutrosophic soft graphs. *Neutrosophic Sets and Systems*, 28:228–244, 2019.

[540] Muhammad Akram and Hafiza Saba Nawaz. Implementation of single-valued neutrosophic soft hypergraphs on human nervous system. *Artificial Intelligence Review*, 56(2):1387–1425, 2023.

[541] Bobin George, Jinta Jose, and Rajesh K Thumbakara. Exploring soft hypergraphs through various operations. *New Mathematics and Natural Computation*, 20(02):551–566, 2024.

[542] Takaaki Fujita, Atiqe Ur Rahman, Arkan A Ghaib, Talal Ali Al-Hawary, and Arif Mehmood Khattak. On the properties and illustrative examples of soft superhypergraphs and rough superhypergraphs. *Prospects for Applied Mathematics and Data Analysis*, 5(1):12–31, 2025.

[543] Jinta Jose, Bobin George, and Rajesh K Thumbakara. Advancements in soft directed graph theory: new ideas and properties. *New Mathematics and Natural Computation*, pages 1–17, 2024.

[544] Jinta Jose, Bobin George, and Rajesh K Thumbakara. Soft directed graphs, their vertex degrees, associated matrices and some product operations. *New Mathematics and Natural Computation*, 19(03):651–686, 2023.

[545] Raed Hatamleh, Nasir Odat, Hamza Ali Abujabal, Faria Khan, Arif Mehmood Khattak, Alaa M. Abd El-latif, Husham M. Attaalfadeel, and Abdelhalim Hasnaoui. Fermatean double-valued neutrosophic soft topological spaces. *European Journal of Pure and Applied Mathematics*, 2025.

[546] V Subash and M Angayarkanni. Neutrosophic hypersoft topological spaces via $m$-open sets. *JP Journal of Geometry and Topology*, 31(1):39–54, 2025.

[547] V Subash and M Angayarkanni. Contra m-continuous maps and contra m-irresolute maps in fuzzy hypersoft topological spaces. *International Journal of Environmental Sciences*, 11(6s):431–448, 2025.

[548] Sagvan Younis Musa and Baravan Abdulmuhsen Asaad. Connectedness on bipolar hypersoft topological spaces. *Journal of Intelligent & Fuzzy Systems*, 43(4):4095–4105, 2022.

[549] PG Patil, C Jaya Subba Reddy, Rani Teli, and Vyshakha Elluru. New structures in fuzzy binary soft topological spaces. *International Journal of Mathematics Trends and Technology-IJMTT*, 71, 2025.

[550] Rui Gao and Jianrong Wu. Filter with its applications in fuzzy soft topological spaces. *AIMS Mathematics*, 6(3):2359–2368, 2021.

[551] A Mukherjee and AK Das. Parameterized topological space induced by an intuitionistic fuzzy soft multi topological space. *Ann. Pure and Applied Math*, 7:7–12, 2014.

[552] Francisco Gallego Lupiáñez. On intuitionistic fuzzy topological spaces. *Kybernetes*, 35(5):743–747, 2006.

[553] M Parimala, M Karthika, and Florentin Smarandache. *A review of fuzzy soft topological spaces, intuitionistic fuzzy soft topological spaces and neutrosophic soft topological spaces*. Infinite Study, 2020.

[554] Maha Mohammed Saeed, Raed Hatamleh Hatamleh, Alaa M Abd El-latif, Abdallah Al-Husban, Takaaki Fujita, Cris L Armada, Rabia Andleeb, and Arif Mehmood Khattak. Separation axioms in quadri-partition neutrosophic soft topological spaces. *European Journal of Pure and Applied Mathematics*, 18(3):6324–6324, 2025.

[555] Noori F Al-Mayahi. Soft banach algebra: Theory and applications. *Journal of Iraqi Al-Khawarizmi*, 8(2):44–68, 2024.

[556] Young Bae Jun. Union-soft sets with applications in bck/bci-algebras. *Bulletin of the Korean Mathematical Society*, 50(6):1937–1956, 2013.

[557] Zanyar A Ameen, Tareq M Al-shami, Radwan Abu-Gdairi, and Abdelwaheb Mhemdi. The relationship between ordinary and soft algebras with an application. *Mathematics*, 11(9):2035, 2023.

[558] Nenad Stojanović. Soft sets whose soft measure is zero. *Filomat*, 39(17):5825–5832, 2025.

[559] Vakkas Ulucay, Mehmet Sahin, Necati Olgun, and Adem Klcman. On neutrosophic soft lattices. *Afrika Matematika*, 28:379–388, 2017.

[560] S. Rajareega, J. Felicita Vimala, and D. Preethi. Complex intuitionistic fuzzy soft lattice ordered group and its weighted distance measures. *Mathematics*, 2020.

[561] VD Jobish, KV Babitha, and Sunil Jacob John. On soft lattice operations. *J Adv Res Pure Math*, 5(2):71–86, 2013.

[562] Yingchao Shao and Keyun Qin. Fuzzy soft sets and fuzzy soft lattices. *International Journal of Computational Intelligence Systems*, 5(6):1135–1147, 2012.

[563] Vassilios Petridis and Vassilis G Kaburlasos. Learning in the framework of fuzzy lattices. *IEEE Transactions on Fuzzy Systems*, 7(4):422–440, 2002.

[564] Shio Gai Quek, Ganeshsree Selvachandran, Vimala Jayakumar, Phet Duong, and Le Hoang Son. A new decision making model based on complex intuitionistic fuzzy soft lattice for traffic monitoring in the pandemic scenarios. *Advanced Intelligent Systems*, 6(11):2400145, 2024.

[565] S Rajareega, J Vimala, and D Preethi. Complex intuitionistic fuzzy soft lattice ordered group and its weighted distance measures. *Mathematics*, 8(5):705, 2020.

[566] S Rajareega and J Vimala. Operations on complex intuitionistic fuzzy soft lattice ordered group and cifs-copras method for equipment selection process. *Journal of Intelligent & Fuzzy Systems*, 41(5):5709–5718, 2021.

[567] A Sezgin Sezer and AO Atagün. A new kind of vector space: soft vector space. *Southeast asian bulletin of mathematics*, 40(5):753–770, 2016.

[568] C Gunduz Aras, AYSE Sonmez, and HUSEYIN Cakalli. An approach to soft functions. *J. Math. Anal*, 8(2):129–138, 2017.

[569] Sabir Hussain. On some soft functions. *Mathematical Sciences Letters*, 4(1):55, 2015.

[570] Zanyar A Ameen and Mesfer H Alqahtani. Some classes of soft functions defined by soft open sets modulo soft sets of the first category. *Mathematics*, 11(20):4368, 2023.

[571] Hacı Aktaş and Naim Çağman. Soft sets and soft groups. *Information sciences*, 177(13):2726–2735, 2007.

[572] Ajoy Kanti Das and Carlos Granados. An advanced approach to fuzzy soft group decision-making using weighted average ratings. *SN Computer Science*, 2(6):471, 2021.

[573] Muhammad Saeed, Atiqe Ur Rahman, Muhammad Ahsan, and Florentin Smarandache. An inclusive study on fundamentals of hypersoft set. *Theory and Application of Hypersoft Set*, 1:1–23, 2021.

[574] Abdülkadir Aygünolu and Halis Aygün. Introduction to fuzzy soft groups. *Computers & Mathematics with Applications*, 58(6):1279–1286, 2009.

[575] Majdoleen Abu Qamar and Nasruddin Hassan. *Characterizations of group theory under Q-neutrosophic soft environment.* Infinite Study, 2019.

[576] Yıldıray Celik, Canan Ekiz, and Sultan Yamak. Applications of fuzzy soft sets in ring theory. *Annals Fuzzy Mathematics and Informatics*, 5(3):451–462, 2013.

[577] Jayanta Ghosh, Dhananjoy Mandal, and T Samanta. Soft structures of groups and rings. *International Journal of Scientific World*, 5(2):117–125, 2017.

[578] Ummahan Acar, Fatih Koyuncu, and Bekir Tanay. Soft sets and soft rings. *Computers & Mathematics with Applications*, 59(11):3458–3463, 2010.

[579] Roy Goetschel and William Voxman. Fuzzy matroids. In *Fuzzy sets and systems*, 1988.

[580] Roy Goetschel and William Voxman. Bases of fuzzy matroids. *Fuzzy Sets and Systems*, 31:253–261, 1989.

[581] Ladislav A. Novak. On goetschel and voxman fuzzy matroids. *Fuzzy Sets Syst.*, 117:407–412, 2001.

[582] Kholod M Hassan and Saied A Johnny. Matroidal structure based on soft-sets. In *Journal of Physics: Conference Series*. IOP Publishing, 2020.

[583] Muhammad Akram, Musavarah Sarwar, and Wieslaw A Dudek. Bipolar fuzzy circuits. In *Graphs for the Analysis of Bipolar Fuzzy Information*, pages 281–307. Springer, 2020.

[584] Ahmed B AL-Nafee, Said Broumi, and Florentin Smarandache. *Neutrosophic soft bitopological spaces.* Infinite Study, 2021.

[585] A Kandil, OAE Tantawy, SA El-Sheikh, and Shawqi A Hazza. Pairwise open (closed) soft sets in soft bitopological spaces. *Ann. Fuzzy Math. Inform*, 11(4):571–588, 2016.

[586] Basavaraj M Ittanagi. Soft bitopological spaces. *International Journal of Computer Applications*, 107(7):1–4, 2014.

[587] AF Sayed. Some separation axioms in fuzzy soft bitopological spaces. *J. Math. Comput. Sci.*, 8(1):28–45, 2017.

[588] Taha Yasin Ozturk and Sadi Bayramov. Category of chain complexes of soft modules. *International Mathematical Forum*, 7(20):981–992, 2012.

[589] Mohammed Amare Mohammed, Berehanu Bekele Belayneh, Zelalem Teshome Wale, Gezahagne Mulat Addis, and Mohammed Tesemma. Construction of soft modules over soft abelian groups. *Research in Mathematics*, 13(1):2605729, 2026.

[590] Mikail Bal and Necati Olgun. Soft neutrosophic modules. *Mathematics*, 6(12):323, 2018.

[591] Qiu-Mei Sun, Zi-Long Zhang, and Jing Liu. Soft sets and soft modules. In *International Conference on Rough Sets and Knowledge Technology*, pages 403–409. Springer, 2008.

[592] Sadi Bayramov, Cigdem Gunduz, and M Ibrahim Yazar. Inverse system of fuzzy soft modules. *Annals of Fuzzy Mathematics and Informatics*, 4(2):349–363, 2012.

[593] OA Tantawy and RM Hassan. Soft metric spaces. In *5th International Conference on Mathematics and Information Sciences*, 2016.

[594] İsmet Altıntaş and Peyil Esengul kyzy. Topology of soft partial metric spaces. *Soft Computing*, 29(19):5613–5623, 2025.

[595] Vildan Çetkin, Elif Güner, and Halis Aygün. On 2s-metric spaces. *Soft Computing*, 24(17):12731–12742, 2020.

[596] Sonam, Ramakant Bhardwaj, Josika Mal, Pulak Konar, and Phumin Sumalai. Fixed point results in soft probabilistic metric spaces. *The journal of Analysis*, 33(1):139–166, 2025.

[597] Yuan Zou. Bayesian decision making under soft probabilities. *Journal of Intelligent & Fuzzy Systems*, 44(6):10661–10673, 2023.

[598] DA Molodtsov. Soft probability of large deviations. *Advances in Systems Science and Applications*, 13(1):53–67, 2013.

[599] Jing Qiu, Zhi Xiao, Wei Xu, and Ying Zhou. Soft probability based random forest for financial distress prediction. *Information Sciences*, page 122870, 2025.

[600] Trevor Jack. On the complexity of properties of partial bijection semigroups, 2021.

[601] Rukchart Prasertpong and Aiyared Iampan. Approximation approaches for rough hypersoft sets based on hesitant bipolar-valued fuzzy hypersoft relations on semigroups. *Journal of Mathematics and Computer Science*, 2022.

[602] Young Bae Jun, Kyoung Ja Lee, and Asghar Khan. Soft ordered semigroups. *Mathematical Logic Quarterly*, 56(1):42–50, 2010.

[603] Tahir Mahmood, Muhammad Asif, Ubaid ur Rehman, and Jabbar Ahmmad. T-bipolar soft semigroups and related results. *Spectrum of Mechanical Engineering and Operational Research*, 1(1):258–271, 2024.

[604] Munazza Naz, Muhammad Shabir, and Muhammad Irfan Ali. On fuzzy soft semigroups. *World Applied Sciences Journal (Special Issue of Applied Math)*, 22:62–83, 2013.

[605] Cheng-Fu Yang. Fuzzy soft semigroups and fuzzy soft ideals. *Computers & Mathematics with Applications*, 61(2):255–261, 2011.

[606] M Al Tahan and Bijan Davvaz. Weak chemical hyperstructures associated to electrochemical cells. *Iranian Journal of Mathematical Chemistry*, 9(1):65–75, 2018.

[607] Maria Santilli Ruggero and Thomas Vougiouklis. Hyperstructures in lie-santilli admissibility and isotheories. *Ratio Mathematica*, 33:151, 2017.

[608] Florentin Smarandache. Foundation of superhyperstructure & neutrosophic superhyperstructure. *Neutrosophic Sets and Systems*, 63(1):21, 2024.

[609] Sultan Yamak, Osman Kazancı, and Bijan Davvaz. Soft hyperstructure. *Computers & Mathematics with Applications*, 62(2):797–803, 2011.

[610] Gulay Oguz and Bijan Davvaz. Soft topological hyperstructure. *J. Intell. Fuzzy Syst.*, 40:8755–8764, 2021.

[611] GR Amiri, R Mousarezaei, and S Rahnama. Soft hyperstructures and their applications. *New Mathematics and Natural Computation*, pages 1–19, 2024.

[612] Takaaki Fujita and Florentin Smarandache. *Superhypergraph neural networks and plithogenic graph neural networks: Theoretical foundations*. Infinite Study, 2025.

[613] A Meenakshi, J Shivangi Mishra, Jeong Gon Lee, Antonios Kalampakas, and Sovan Samanta. Advanced risk prediction in healthcare: Neutrosophic graph neural networks for disease transmission. *Complex & Intelligent Systems*, 11(9):413, 2025.

[614] Filip Ekström Kelvinius, Dimitar Georgiev, Artur Toshev, and Johannes Gasteiger. Accelerating molecular graph neural networks via knowledge distillation. *Advances in Neural Information Processing Systems*, 36:25761–25792, 2023.

[615] Daniel Vik, David Pii, Chirag Mudaliar, Mads Nørregaard-Madsen, and Aleksejs Kontijevskis. Performance and robustness of small molecule retention time prediction with molecular graph neural networks in industrial drug discovery campaigns. *Scientific Reports*, 14(1):8733, 2024.

[616] Yingfang Yuan, Wenjun Wang, Xin Li, Kefan Chen, Yonghan Zhang, and Wei Pang. Evolving molecular graph neural networks with hierarchical evaluation strategy. In *Proceedings of the Genetic and Evolutionary Computation Conference*, pages 1417–1425, 2024.

[617] Midhilesh Momidi, Priyanka S Chauhan, Adityaram Komaraneni, Surya Prakash Ghattamaneni, Kamal Upreti, and Nishant Kumar. Uncertainty-aware molecular property prediction using heterogeneous molecular graph neural networks. In *International Conference on Generative Artificial Intelligence, Cryptography, and Predictive Analytics*, pages 243–254. Springer, 2024.

[618] AA Salama, Huda E Khalid, Ahmed K Essa, and Nadheer M Ahmed. A natural language processing environment for rule-based decision making with neutrosophic logic to manage uncertainty and ambiguity. *Neutrosophic Sets and Systems*, 82(1):44, 2025.

[619] Diego Fernando Coka Flores, Ignacio Fernando Barcos Arias, María Elena Infante Miranda, and Omar Mar Cornelio. Applying neutrosophic natural language processing to analyze complex phenomena in interdisciplinary contexts. *Neutrosophic Sets and Systems*, 74:297–305, 2024.

[620] Sultan AlGhozali and Siti Mukminatun. Natural language processing of gemini artificial intelligence powered chatbot. *Balangkas: An International Multidisciplinary Research Journal*, 1(1):41–48, 2024.

[621] Naeemeh Adel. *Fuzzy natural language similarity measures through computing with words.* PhD thesis, Manchester Metropolitan University, 2022.

[622] Yenson Vinicio Mogro Cepeda, Marco Antonio Riofrío Guevara, Emerson Javier Jácome Mogro, and Rachele Piovanelli Tizano. Impact of irrigation water technification on seven directories of the san juanpatoa river using plithogenic n-superhypergraphs based on environmental indicators in the canton of pujilí, 2021. *Neutrosophic Sets and Systems*, 74(1):6, 2024.

[623] Victor Wanjala and Collins Amenya. Feature-augmented drug superhypergraphs: Neural architectures for hierarchical pharmaceutical knowledge representation. *Neutrosophic Knowledge*, 12, 2026.

[624] Mohammad Hamidi, Florentin Smarandache, and Elham Davneshvar. Spectrum of superhypergraphs via flows. *Journal of Mathematics*, 2022(1):9158912, 2022.

[625] Takaaki Fujita. The power graph as a hypergraph and iterated-powerset power graphs as n-superhypergraphs. *International Journal of Fuzzy Mathematical Archive*, 25(1):37–51, 2026.

[626] Fujita Takaaki, Ajoy Kanti Das, Suman Das, et al. Basic theory of hierarchical superhypergraphs: Cross-levelincidence, direction, and bidirection. *Neutrosophic Computing and Machine Learning*, 44:523–550, 2026.

[627] Mohammad Hamidi, Shohre Shirani, and Saeed Mirvakili. Applications of pythagorean fuzzy superhypergraph entropy in uncertainty optimization. *JOURNAL OF MATHEMATICAL EXTENSION*, 20, 2026.

[628] Mohammad Hamidi and Shohre Shirani Nazvani. Pythagorean fuzzy superhypergraphs. *Fuzzy Systems and its Applications*, 9(1):153–179, 2026.

[629] Takaaki Fujita. Modeling hierarchical and higher-order uncertainty in it systems management using strong intuitionistic fuzzy hypergraphs and superhypergraphs. *Journal of Engineering Management and Systems Science*, 1(1):1–25, 2026.

[630] Takaaki Fujita and Florentin Smarandache. Fuzzy and neutrosophic directed superhypergraphs: Definitions, properties, and illustrative real-world applications in decision-making. *IAENG Int. J. Appl. Math*, 56(5), 2026.

[631] Takaaki Fujita. Interval-valued neutrosophic models based on superhypergraphs. *Advances in Neutrosophic Mathematics and Informatics*, 1(1):31–42, 2026.

[632] Takaaki Fujita. Conceptual hypergraph and conceptual superhypergraph. *Neutrosophic Knowledge*, 10:76–90, 2026.

[633] Takaaki Fujita and Florentin Smarandache. *Neutrosophic soft n-super-hypergraphs with real-world applications.* Infinite Study, 2025.

[634] Takaaki Fujita and Florentin Smarandache. Soft directed n-superhypergraphs with some real-world applications. *European Journal of Pure and Applied Mathematics*, 18(4):6643–6643, 2025.

[635] Ajoy Kanti Das, Rajat Das, Suman Das, Bijoy Krishna Debnath, Carlos Granados, Bimal Shil, and Rakhal Das. A comprehensive study of neutrosophic superhyper bci-semigroups and their algebraic significance. *Transactions on Fuzzy Sets and Systems*, 8(2):80, 2025.

[636] Adel Al-Odhari. A brief comparative study on hyperstructure, super hyperstructure, and n-super superhyperstructure. *Neutrosophic Knowledge*, 6:38–49, 2025.

[637] Mohammad Hamidi and Mohadeseh Taghinezhad. *Application of Superhypergraphs-Based Domination Number in Real World.* Infinite Study, 2023.

[638] Mohammad Hamidi, Florentin Smarandache, and Mohadeseh Taghinezhad. *Decision Making Based on Valued Fuzzy Superhypergraphs.* Infinite Study, 2023.

[639] Takaaki Fujita, Atiqe Ur Rahman, Arkan A Ghaib, Talal Ali Al-Hawary, and Arif Mehmood Khattak. On the properties and illustrative examples of soft superhypergraphs and rough superhypergraphs. *Prospects for Applied Mathematics and Data Analysis*, 5(1):12–31, 2025.

[640] Takaaki Fujita. Review of plithogenic directed, mixed, bidirected, and pangene offgraph. *Advancing Uncertain Combinatorics through Graphization, Hyperization, and Uncertainization: Fuzzy, Neutrosophic, Soft, Rough, and Beyond*, page 120, 2024.

[641] Takaaki Fujita. Recursive hypergraphs and recursive superhypergraphs: Exploring more hierarchical and generalized graph concepts. *Neutrosophic Knowledge*, 10:76–90, 2026.

[642] Hans L Bodlaender. Treewidth: Algorithmic techniques and results. In *International Symposium on Mathematical Foundations of Computer Science*, pages 19–36. Springer, 1997.

[643] Hans L Bodlaender, Alexander Grigoriev, and Arie MCA Koster. Treewidth lower bounds with brambles. In *Algorithms–ESA 2005: 13th Annual European Symposium, Palma de Mallorca, Spain, October 3-6, 2005. Proceedings 13*, pages 391–402. Springer, 2005.

[644] Neil Robertson and Paul D Seymour. Graph minors. iv. tree-width and well-quasi-ordering. *Journal of Combinatorial Theory, Series B*, 48(2):227–254, 1990.

[645] Georg Gottlob, Nicola Leone, and Francesco Scarcello. Hypertree decompositions and tractable queries. In *Proceedings of the eighteenth ACM SIGMOD-SIGACT-SIGART symposium on Principles of database systems*, pages 21–32, 1999.

[646] Georg Gottlob, Nicola Leone, and Francesco Scarcello. Hypertree decompositions: A survey. In *Mathematical Foundations of Computer Science 2001: 26th International Symposium, MFCS 2001 Mariánské Lázne, Czech Republic, August 27–31, 2001 Proceedings 26*, pages 37–57. Springer, 2001.

[647] Takaaki Fujita, Iqbal M Batiha, Muhammad Gulistan, Belal Batiha, Essa Lafi Al Smadi, and Iqbal Jebril. Tree decompositions in fuzzy graphs: Foundations and structural insights. *Statistics, Optimization & Information Computing*, 16(1):312–326, 2026.

[648] Takaaki Fujita and Talal Ali Al-Hawary. Short note of superhyperclique-width and local superhypertree-width. *Neutrosophic Sets and Systems*, 86(1):51, 2025.

[649] Ridvan Sahin and Ahmet Kuçuk. Soft filters and their convergence properties. *Ann. Fuzzy Math. Inform*, 6(3):529–543, 2013.

[650] Sujay Goldar, Subhasis Ray, and Amit Sarkar. Interrelations between soft and ordinary sets: an exploration of soft filters and soft nets. *International Journal of Foundations of Computer Science*, pages 1–20, 2025.

[651] Şaziye Yüksel, Naime Tozlu, and Zehra Güzel Ergül. Soft filter. *Mathematical sciences*, 8(1):119, 2014.

[652] A Kandil, OAE Tantawy, SA El-Sheikh, and AM Abd El-latif. $\gamma$-operation and decompositions of some forms of soft continuity of soft topological spaces via soft ideal. *Ann. Fuzzy Math. Inform*, 9(3):385–402, 2015.

[653] G Muhiuddin and Ahsan Mahboob. Int-soft ideals over the soft sets in ordered semigroups. *AIMS Mathematics*, 5(3):2412, 2020.

[654] Rodyna A Hosny. Remarks on soft topological spaces with soft grill. *Far East J. Math. Sci*, 86(1):111–128, 2014.

[655] Rodyna Ahmed Hosny and Alaa Mohamed Abd El-Latif. Pre open soft sets via soft grills. *Journal of new theory*, (8):29–40, 2015.

[656] Alaa Mohamed Abd El-Latif and Rodyna A Hosny. Some soft sets via soft grills. *South Asian J Math*, 8(1):18–30, 2018.

[657] José Carlos R Alcantud. Convex soft geometries. *Journal of Computational and Cognitive Engineering*, 1(1):2–12, 2022.

[658] Feng Feng, Young Bae Jun, and Xianzhong Zhao. Soft semirings. *Computers & mathematics with applications*, 56(10):2621–2628, 2008.

[659] Muhammad Iftikhar and Tahir Mahmood. Some results on lattice ordered double framed soft semirings. *International Journal of Algebra and Statistics*, 7(1-2):123–140, 2018.

[660] Waheed Ahmad Khan and Abdelghani Taouti. Soft sub-nearsemirings, soft ideals and soft-subsemigroups of nearsemirings. *International Journal of Fuzzy Logic and Intelligent Systems*, 22(3):276–286, 2022.

[661] J Ahsan, K Saifullah, and M Farid Khan. Fuzzy semirings. *Fuzzy Sets and systems*, 60(3):309–320, 1993.

[662] Saman Abdurrahman. Ideal fuzzy semiring atas level subset. *Jurnal Fourier*, 11(1):1–6, 2022.

[663] Muhammad Gulistan, Inayatur Rehman, Muhammad Shahzad, Shah Nawaz, and Salma Khan. Generalized neutrosophic semirings. *Neutrosophic Sets and Systems*, 47(1):34, 2021.

[664] Tareq M Al-Shami, Zanyar A Ameen, and Abdelwaheb Mhemdi. The connection between ordinary and soft $\sigma$-algebras with applications to information structures. *Aims Mathematics*, 8(6):14850–14866, 2023.

[665] Shuker Mahmood Khalil, Mayadah Ulrazaq, Samaher Abdul-Ghani, and Abu Firas Al-Musawi. $\sigma$- algebra and $\sigma$- baire in fuzzy soft setting. *Advances in Fuzzy Systems*, 2018(1):5731682, 2018.

[666] Khalid Naeem and Samet Memiş. Picture fuzzy soft $\sigma$-algebra and picture fuzzy soft measure and their applications to multi-criteria decision-making. *Granular computing*, 8(2):397–410, 2023.

[667] ME El-Shafei, M Abo-Elhamayel, and TM Al-Shami. Partial soft separation axioms and soft compact spaces. *Filomat*, 32(13):4755–4771, 2018.

[668] TM Al-Shami and ME El-Shafei. On soft compact and soft lindelöf spaces via soft pre-open sets. *Annals of Fuzzy Mathematics and Informatics*, 17(1):79–100, 2019.

[669] Tareq M Al-shami, S Saleh, Fathea M Osman Birkea, M Arar, and M Jameel. Compact and locally compact spaces via the class of soft $\delta$-open sets. *Fuzzy Information and Engineering*, 18(1):104–120, 2026.

[670] Hawraa Yousif Daher and Jehad R Kider. Some properties of convex fuzzy compact spaces. *Iraqi Journal of Science*, pages 2856–2867, 2025.

[671] Saeid Jafari, Mamadaliev Nodirbek Kamoldinovich, and Said Isaev Usmonovich. Symmetric power functors in the category of fuzzy compact spaces. *Applied General Topology*, 27(2):25270–25270, 2026.

[672] RUNU DHAR, SUMA PAUL, SUPRIYA PAUL, and GOUR PAL. Some neutrosophic compact spaces via grills. *Electronic Journal of Mathematical Analysis and Applications*, 11(1):231–240, 2023.

[673] Raja Mohammad Latif. Neutrosophic $\lambda$-compactness, $\lambda$-connectedness, and $\lambda$-separation axioms in neutrosophic topological spaces. *The Eurasia Proceedings of Science Technology Engineering and Mathematics*, 35:323–345, 2025.

[674] TM Al-Shami. On soft separation axioms and their applications on decision-making problem. *Mathematical Problems in Engineering*, 2021(1):8876978, 2021.

[675] Sabir Hussain and Bashir Ahmad. Soft separation axioms in soft topological spaces. *Hacettepe journal of mathematics and statistics*, 44(3):559–568, 2015.

[676] Murad Arar and Tareq M Al-Shami. A holistic perspective on soft separation axioms. *Filomat*, 39(11):3623–3638, 2025.

[677] Tariq Shah and Salma Shaheen. Soft topological groups and rings. *Ann. fuzzy math. Inform*, 7(5):725–743, 2014.

[678] Nazmiye Alemdar, Hürmet Fulya Akız, and Halim Ayaz. Soft path, soft-path-connectedness, and the category of soft-topological groups. *Mathematics*, 14(5):764, 2026.

[679] Nazmiye Alemdar, Hürmet Fulya Akız, and Halim Ayaz. Soft connectedness, soft path connectedness and the category of soft topological groups. *arXiv preprint arXiv:2511.12724*, 2025.

[680] Essam H Hamouda. On soft topological groups and soft function spaces. *New Mathematics and Natural Computation*, 22(01):163–173, 2026.

[681] Young Bae Jun, Kyoung Ja Lee, and Seok Zun Song. N-ideals of bck/bci-algerbas. *Journal of the Chungcheong mathematical Society*, 22(3):417–417, 2009.

[682] PB Remya and A Francina Shalini. Neutrosophic vague binary bck/bci-algebra. *Neutrosophic Sets & Systems*, 35:45, 2020.

[683] Young Bae Jun and Xiao Long Xin. Complex fuzzy sets with application in bck/bci-algebras. *Bulletin of the Section of Logic*, 48(3):173–185, 2019.

[684] Young Bae Jun. Soft bck/bci-algebras. *Computers & Mathematics with Applications*, 56(5):1408–1413, 2008.

[685] Young Bae Jun and Chul Hwan Park. Applications of soft sets in ideal theory of bck/bci-algebras. *Information sciences*, 178(11):2466–2475, 2008.

[686] Young Bae Jun and Sun Shin Ahn. Double-framed soft sets with applications in bck/bci-algebras. *Journal of Applied Mathematics*, 2012(1):178159, 2012.

[687] G Muhiuddin, Mohamed E Elnair, Ahmed A Khidir, and Mohammed Hassan. Exploring categorical perspectives on soft bck/bci-algebras. *European Journal of Pure and Applied Mathematics*, 18(4):7055–7055, 2025.

[688] Ghulam Muhiuddin and Deena Al-Kadi. Interval valued m-polar fuzzy bck/bci-algebras. *International Journal of Computational Intelligence Systems*, 14(1):1014–1021, 2021.

[689] Youngbae Jun and Ravikumar Bandaru. Subalgebras and ideals in fuzzy normed bck-algebras and bci-algebras. *Universal Journal of Mathematics and Applications*, 9(2):124–137, 2026.

[690] Adesina Abdul Akeem Agboola and Bijan Davvaz. *Introduction to neutrosophic BCI/BCK-algebras.* Infinite Study, 2015.

[691] AAA Agboola and B Davvaz. Research article introduction to neutrosophic bci/bck-algebras, 2015.

[692] Mahlon M Day. Normed linear spaces. In *Normed Linear Spaces*, pages 27–52. Springer, 1973.

[693] Robert C James. Orthogonality and linear functionals in normed linear spaces. *Transactions of the American Mathematical Society*, 61(2):265–292, 1947.

[694] Sujoy Das, Pinaki Majumdar, and SK Samanta. On soft linear spaces and soft normed linear spaces. *arXiv preprint arXiv:1308.1016*, 2013.

[695] H Akewe, EK Osawaru, and OK Adewale. On soft fixed point of picard-mann hybrid iterative sequences in soft normed linear spaces. *Journal of Nonlinear Analysis and Optimization: Theory & Applications (JNAO)*, 8(1):85–94, 2017.

[696] Sujoy Das and SK Samanta. Soft linear functionals in soft normed linear spaces. *Ann. Fuzzy Math. Inform*, 7(4):629–651, 2014.

[697] Sujoy Das and SK Samanta. On soft inner product spaces. *Ann. Fuzzy Math. Inform*, 6(1):151–170, 2013.

[698] Mehmet Şirin Gönci and Hacer Bozkurt. Soft quasilinear inner product spaces. *TWMS Journal of Applied and Engineering Mathematics*, 2024.

[699] Nashat Faried, Mohamed SS Ali, and Hanan H Sakr. Fuzzy soft inner product spaces. *Appl. Math. Inf. Sci*, 14(4):709–720, 2020.

[700] Sujoy Das and SK Samanta. Operators on soft inner product spaces. *Fuzzy Information and Engineering*, 6(4):435–450, 2014.

[701] Lorena Popa and Lavinia Sida. Fuzzy inner product space: literature review and a new approach. *Mathematics*, 9(7):765, 2021.

[702] Raghad I Sabri and Buthainah AA Ahmed. Another type of fuzzy inner product space. *Iraqi Journal of Science*, pages 1853–1861, 2023.

[703] MI Yazar, CG Aras, and S Bayramov. Results on soft hilbert spaces. *TWMS Journal of Applied and Engineering Mathematics*, 9(1):159–164, 2019.

[704] Osmin Ferrer, Arley Sierra, and José Sanabria. Soft frames in soft hilbert spaces. *Mathematics*, 9(18):2249, 2021.

[705] Nashat Faried, Mohamed SS Ali, and Hanan H Sakr. On fuzzy soft linear operators in fuzzy soft hilbert spaces. In *Abstract and Applied Analysis*, volume 2020, page 5804957. Wiley Online Library, 2020.

[706] Sayyed Mehrab Ramezani. Soft g-frames in soft hilbert spaces. *arXiv preprint arXiv:2307.14390*, 2023.

[707] M Goudarzi and SM Vaezpour. On the definition of fuzzy hilbert spaces and its application. *The Journal of Nonlinear Sciences and its Applications*, 2(1):46–59, 2009.

[708] B Daraby, Z Solimani, and A Rahimi. Some properties of fuzzy hilbert spaces. *Complex Analysis and Operator Theory*, 11(1):119–138, 2017.

[709] Kadhim B Tarsh and Hadeel A Hassan. Neutrosophic inner product spaces and the construction of neutrosophic hilbert spaces with application. *Nonlinear Functional Analysis and Applications*, pages 905–924, 2026.

[710] Hai-Long Yang, Xiuwu Liao, and Sheng-Gang Li. On soft continuous mappings and soft connectedness of soft topological spaces. *Hacettepe Journal of Mathematics and Statistics*, 44(2):385–398, 2015.

[711] Tareq M Al-Shami, Sandeep Kaur, Alkan Özkan, M Hosny, and Abdelwaheb Mhemdi. Some characterizations of soft continuous mappings using soft graphs. *Filomat*, 38(22):7823–7830, 2024.

[712] ST Ekram and RN Majeed. Soft continuous mappings in soft closure spaces. *Iraqi Journal of Science*, pages 2676–2684, 2021.

[713] Sabir Hussain. Binary soft connected spaces and an application of binary soft sets in decision making problem. *Fuzzy Information and Engineering*, 11(4):506–521, 2019.

[714] Samer Al Ghour and Zanyar A Ameen. Maximal soft compact and maximal soft connected topologies. *Applied Computational Intelligence and soft computing*, 2022(1):9860015, 2022.

[715] Zhenyuan Wang and George J Klir. *Fuzzy measure theory.* Springer Science & Business Media, 1992.

[716] Aseel Ahmed Shihab Alshabeeb and Reza Saadati. Fuzzy operators and hyers–ulam stability in the context of*-fuzzy measure spaces. *Algorithms*, 19(7):576, 2026.

[717] Jun Li, Masami Yasuda, Qingshan Jiang, Hisakichi Suzuki, Zhenyuan Wang, and George J Klir. Convergence of sequence of measurable functions on fuzzy measure spaces. *Fuzzy Sets and Systems*, 87(3):317–323, 1997.

[718] Mohamed Ismail and Esraa Kamal. Neutrosophic probability and measure for resilient supply chain management under indeterminacy. *Neutrosophic Sets and Systems*, 94(1):22, 2025.

[719] Florentin Smarandache. *Introduction to neutrosophic measure, neutrosophic integral, and neutrosophic probability.* Infinite Study, 2013.

[720] Alan Christian López Castillo, Edward Usaqui Barbaran, Miguel Oscar López Orellana, Aníbal Campos Barreto, Llesica Soria Ramirez, Wilmer Ortega Chávez, Víctor Tedy López Panaifo, and Alex Davis Astohuaman Huaranga. Media pressure and pretrial detention requirements in ucayali, peru: A study based on neutrosophic measures. *Neutrosophic Sets and Systems*, 74(1):47, 2024.

[721] Miguel Ortiz-Barrios, Natalia Jaramillo-Rueda, Andrea Espeleta-Aris, Berk Kucukaltan, and Llanos Cuenca. Integrated fuzzy decision-making methodology with intuitionistic fuzzy numbers: An application for disaster preparedness in clinical laboratories. *Expert Systems with Applications*, 263:125712, 2025.

[722] Hongxing Li and Vincent C Yen. *Fuzzy sets and fuzzy decision-making.* CRC press, 1995.

[723] Dragan Pamucar, Morteza Yazdani, Radojko Obradovic, Anil Kumar, and Mercedes Torres-Jiménez. A novel fuzzy hybrid neutrosophic decision-making approach for the resilient supplier selection problem. *International Journal of Intelligent Systems*, 35(12):1934–1986, 2020.

[724] Arunodaya Raj Mishra, Dragan Pamucar, Pratibha Rani, Rajeev Shrivastava, and Ibrahim M. Hezam. Assessing the sustainable energy storage technologies using single-valued neutrosophic decision-making framework with divergence measure. *Expert Syst. Appl.*, 238:121791, 2023.

[725] G Muhiuddin, Mohamed E Elnair, Satham Hussain, and Durga Nagarajan. Topsis method-based decision-making model for bipolar quadripartitioned neutrosophic environment. *Neutrosophic Sets and Systems*, 85:899–918, 2025.

[726] Muhammet Gul, Suleyman Mete, Faruk Serin, and Erkan Celik. Fine–kinney-based occupational risk assessment using single-valued neutrosophic topsis. In *Fine–Kinney-Based Fuzzy Multi-criteria Occupational Risk Assessment: Approaches, Case Studies and Python Applications*, pages 111–133. Springer, 2020.

[727] Ting-Yu Chen and Chueh-Yung Tsao. The interval-valued fuzzy topsis method and experimental analysis. *Fuzzy sets and systems*, 159(11):1410–1428, 2008.

[728] Manoj Mathew, Ripon Kumar Chakrabortty, and Michael J. Ryan. A novel approach integrating ahp and topsis under spherical fuzzy sets for advanced manufacturing system selection. *Eng. Appl. Artif. Intell.*, 96:103988, 2020.

[729] Ali Azadeh, Morteza Saberi, Nasim Zandi Atashbar, Elizabeth Chang, and Peiman Pazhoheshfar. Z-ahp: A z-number extension of fuzzy analytical hierarchy process. In *2013 7th IEEE International Conference on Digital Ecosystems and Technologies (DEST)*, pages 141–147. IEEE, 2013.

[730] Hamid Reza Pourghasemi, Biswajeet Pradhan, and Candan Gokceoglu. Application of fuzzy logic and analytical hierarchy process (ahp) to landslide susceptibility mapping at haraz watershed, iran. *Natural Hazards*, 63:965–996, 2012.

[731] Mavera Nawaz, Arooj Adeel, and Muhammad Akram. Risk evaluation in failure mode and effect analysis: Ahp-vikor method with picture fuzzy rough number. *Granular Computing*, 9(3):69, 2024.

[732] Xingang Wang, Yushui Geng, Peipei Yao, and Mengjie Yang. Multiple attribute group decision making approach based on extended vikor and linguistic neutrosophic set. *Journal of Intelligent & Fuzzy Systems*, 36(1):149–160, 2019.

[733] Serafim Opricovic and Gwo-Hshiung Tzeng. Compromise solution by mcdm methods: A comparative analysis of vikor and topsis. *Eur. J. Oper. Res.*, 156:445–455, 2004.

[734] Admin Admin, Luis A. Crespo Crespo-Berti, Haro Teran Lilian Fabiola, and Dinara Turaeva. Neutrosophic decision making using saaty's ahp method and vikor. *Journal of Intelligent Systems and Internet of Things*, 2024.

[735] Muhammad Riaz and Syeda Tayyba Tehrim. A robust extension of vikor method for bipolar fuzzy sets using connection numbers of spa theory based metric spaces. *Artificial Intelligence Review*, 54:561 – 591, 2020.

[736] Jean Dezert and Jean-Marc Tacnet. Soft electre tri outranking method based on belief functions. In *2012 15th International Conference on Information Fusion*, pages 607–614. IEEE, 2012.

[737] Muhammad Shazib Hameed, Shahzaib Ashraf, Vladimir Simic, Chiranjibe Jana, Dragan Pamucar, and Nebojsa Bacanin. Assessment of hospital performance using complex spherical fuzzy soft electre and topsis. *European Journal of Pure and Applied Mathematics*, 19(2):7156–7156, 2026.

[738] Muhammad Akram, Maheen Sultan, Arooj Adeel, and Mohammed M Ali Al-Shamiri. Pythagorean fuzzy n-soft promethee approach: a new framework for group decision making. *Aims Mathematics*, 8(8):17354–17380, 2023.

[739] Muhammad Akram, Maheen Sultan, José Carlos R Alcantud, and Mohammed M Ali Al-Shamiri. Extended fuzzy n-soft promethee method and its application in robot butler selection. *Math. Biosci. Eng*, 20:1774–1800, 2023.

[740] Majed Albaity, Tahir Mahmood, and Zeeshan Ali. Impact of machine learning and artificial intelligence in business based on intuitionistic fuzzy soft waspas method. *Mathematics*, 11(6):1453, 2023.

[741] Tahir MAHMOOD and Jabbar AHMMAD. Classifications of artificial intelligence trading systems by using waspas approach based on picture fuzzy soft aggregation operators. *Sigma: Journal of Engineering & Natural Sciences/Mühendislik ve Fen Bilimleri Dergisi*, 44(1):23, 2026.

# A Dynamic Survey of Soft Set Theory and Its Extensions

Takaaki Fujita · Florentin Smarandache

Soft set theory provides a direct framework for parameterized decision modeling by assigning to each attribute (parameter) a subset of a given universe, thereby representing uncertainty in a structured way. Over the past decades, the theory has expanded into numerous variants—including hypersoft sets, superhypersoft sets, TreeSoft sets, bipolar soft sets, and dynamic soft sets—and has been connected to diverse areas such as topology and matroid theory.

In this book, we survey soft sets and their major extensions, emphasizing parameter spaces, value spaces, representative constructions, and the assumptions required for valid mathematical comparisons. We distinguish literature-based definitions from exploratory formulations and discuss limitations of selected representation and reduction claims.

This second edition adds further examples and concepts, corrects identified mathematical and typographical issues, and improves the organization.